\documentclass[]{fairmeta}

\usepackage[utf8]{inputenc}
\usepackage{url}
\usepackage{amsfonts}
\usepackage{nicefrac}
\usepackage{wrapfig}
\usepackage{amsmath}
\usepackage{tikz}
\usepackage{fontawesome5}
\usetikzlibrary{positioning, arrows.meta, fit, backgrounds, calc, shapes.geometric, decorations.pathreplacing}

\newcommand{\neuripsfiguretextbf}[1]{{\bfseries #1}}
\newenvironment{neuripsfigurefonts}{%
  \begingroup
  \let\textbf\neuripsfiguretextbf
  \normalfont
}{%
  \endgroup
}

\crefname{figure}{Figure}{Figures}
\Crefname{figure}{Figure}{Figures}
\crefname{subfigure}{Figure}{Figures}
\Crefname{subfigure}{Figure}{Figures}
\crefname{table}{Table}{Tables}
\Crefname{table}{Table}{Tables}
\crefname{subtable}{Table}{Tables}
\Crefname{subtable}{Table}{Tables}
\crefname{section}{Section}{Sections}
\Crefname{section}{Section}{Sections}
\crefname{subsection}{Section}{Sections}
\Crefname{subsection}{Section}{Sections}
\crefname{subsubsection}{Section}{Sections}
\Crefname{subsubsection}{Section}{Sections}
\crefname{equation}{Equation}{Equations}
\Crefname{equation}{Equation}{Equations}

\titleformat*{\paragraph}{\sffamily\bfseries}

\title{Reinforcement Learning for Code Optimization}

\author[1,2]{Pierre Chambon}
\author[1,3]{Kunhao Zheng}
\author[1,2]{Juliette Decugis}
\author[2]{Benoit Sagot}
\author[1]{Gabriel Synnaeve}

\affiliation[1]{FAIR at Meta}
\affiliation[2]{Inria}
\affiliation[3]{Université Paris Dauphine}

\abstract{RL for code correctness is now established: have the model generate a program, run it against hidden test cases, and reward solutions that pass. Extending this to code optimization seems straightforward: just add execution time to the reward. But in practice, once timing drives the reward, small problems in measurement noise, reward sparsity, or GRPO instability overwhelm the signal and make RL fail: generated solutions are barely faster, and more of them can fail. We make execution time learnable through three stages: (1) how code is tested, by building DMC-Optim with large optimization tests and a calibrated sandbox; (2) how speed is turned into reward, by composing correctness and speed in the RL environment and using an offline simulator to predict the most promising configurations; and (3) how the model learns from that reward, by adapting GRPO and evaluation to the sparser, noisier timed-execution setting. On DMC-Optim, the strongest optimization-aware configurations improve strict top-50\% pass@$1$ from 18.0\% to 31.3\% on Qwen 2.5 7B and from 30.7\% to 50.4\% on CWM 32B. These gains further increase at stricter percentiles such as top-30\%, with 125\% relative improvement for CWM 32B, while preserving pure-correctness scores. When the timing sandbox is degraded, robust optimization RL reaches up to 100\%--200\% improvement over standard RLVR, depending on the evaluation criterion. On LCB, CWM 32B wins up to 83\% of median-sample speed comparisons against standard RLVR. Relative to the fastest correct human submissions per problem, it reaches about half the human rate of complexity-class improvements (13\% vs.\ 22\%).
}
\correspondence{Pierre Chambon at \email{pchambon@meta.com}}

\begin{document}

\maketitle

\section{Introduction}
\label{sec:introduction}

Training for correctness has not made code models reliably fast. On SWE-fficiency~\citep{ma2025swefficiency}, Claude 4.5 Sonnet produces correct patches 81\% of the time but captures only 4.1\% of expert speedup. Repository scale and expensive execution do not fully explain this: the gap also appears in competitive programming, where cheap tests and ample human submissions provide speed baselines. On \textsc{Venus}, o4-mini reaches 89.1\% pass@$1$ but scores only 56.9\% on \textsc{Beyond-T}, a runtime-efficiency percentile against human references~\citep{du2025afterburner}, and Qwen3-32B drops from 70.0\% pure-correctness pass@$1$ to 43.5\% under a best-complexity-class requirement on BigO(Bench)~\citep{chambon2025bigobench}.

The more controlled setting of competitive-programming lets us isolate where optimization RL fails. In standard reinforcement learning with verifiable rewards (RLVR) for code, execution is mostly a binary verifier: a sample is rewarded if it passes the correctness tests. In optimization RL, execution must also measure how fast a correct program runs on the given tests, so timing noise, weak tests, and rewards that give speed credit to fast but wrong code can corrupt the training signal. Evaluation raises the same coupling problem between correctness and speed. For evaluation, we introduce \(p_\tau\): a solution counts only if it is correct and no slower than the \(\tau\)-th-percentile of the human reference leaderboard, with smaller \(\tau\) stricter. This evaluation confirms that the naive optimization training objective is brittle: rewarding lower average runtime on top of a binary correctness reward changes pass@$1$ by only \(+0.6\) points at \(p_{30}\), whatever the reward-range mapping is, and by \(-0.3\) to \(+1.4\) points at \(p_{100}\), which is equivalent to pure correctness. The question is therefore: what makes an execution-time signal learnable?

\begin{figure*}[t!]
\centering
\input{sections/fig_01_overview_v2}
\end{figure*}

Optimization RL fails through a chain of transformations. The timing source must separate solutions (\cref{sec:measurement}); the environment and reward must choose tests, time limits, reference comparisons, and how rewards trade correctness against speed without incentivizing fast-but-wrong code (\cref{sec:reward}); and the optimizer must remain stable under rewards that are sparser and noisier than pass/fail (\cref{sec:training}). If any link fails, generated programs are not faster, and correctness can also degrade.

We make this chain explicit in \cref{fig:overview}. From the DeepMind Code Contests (DMC) corpus~\citep{li2022competition}, we build DMC-Optim with 2,723 cleaned problems, separating correctness tests from larger optimization tests; 1,302 problems have enough duration spread for timing-based rewards. We sweep pre-execution filtering, intra-execution time constraints, post-execution ranking against human references, and rewards from optimization-only/additive blends to multitask, collapsed, and hard-gated variants. An offline simulator screens this space before online GRPO.

This stack gives consistent in-domain gains across three starts: Qwen 2.5 7B/32B base checkpoints~\citep{qwen25technical}, which we SFT on a decontaminated reasoning-only mix, and the released CWM 32B SFT checkpoint~\citep{cwm2025}. On DMC-Optim pass@$1$ at \(p_{50}\), Qwen 2.5 7B/32B and CWM 32B rise from 18.0/21.1/30.7\% to 31.3/39.6/50.4\%; at \(p_{30}\), CWM 32B rises from 13.7\% to 30.9\%, a 125\% relative gain, while pure-correctness scores stay stable. On LCB, where timeout and percentile scores are unreliable, pairwise speed win rates show CWM 32B reaching up to 83.0\% median-sample wins against standard RLVR while retaining 54.4\% pass@$1$.

Our contributions are threefold:
\begin{itemize}
\item We build DMC-Optim and its timing stack: 2,723 cleaned problems from 12,275 raw DMC problems, 1,302 duration-filterable problems, 430,215 new correctness tests, and 352,740 new optimization tests, specifically made to take longer to execute. We explore ways to control the stability of the timing sandbox to make reliable measurements on top of them.
\item We organize optimization constraints used in prior work, along with new formulations, into three families: test filtering before execution, time constraints during execution, and ranking after execution. We then test which formulations make useful RL environments. We show that this depends on the reward placed on top of the environment, compare additive blends, multitask, collapsed, and gated rewards, use an offline simulator to screen promising configurations, and run online RL with a GRPO recipe adapted to sparse timing rewards.
\item We show optimization gains across evaluations: DMC-Optim \(p_{50}\) pass@$1$ improves from 18.0/21.1/30.7\% to 31.3/39.6/50.4\% on Qwen 2.5 7B/32B and CWM 32B; at \(p_{30}\), CWM 32B improves by 125\% relative while preserving pure-correctness scores; CWM 32B reaches an 83.0\% LCB median-sample speed win rate against standard RLVR; and under degraded sandbox state, robust optimization RL improves over RLVR by roughly 100\%--200\%, depending on the evaluation criterion.
\end{itemize}

\section{Related Work}
\label{sec:related_work}

\paragraph{Execution feedback and efficiency measurement.}
RL with verifiable execution feedback has mostly targeted correctness: CodeRL introduced unit-test-based training, and later GRPO-based work showed binary rewards can scale to reasoning and code generation~\citep{le2022coderl,deepseek2025r1}. Efficiency changes the object: rewards must rank already-correct programs, so test construction, runtime noise, and aggregation become part of the signal. Efficiency benchmarks make this gap visible with percentile scores against human references, stress tests, right-censored timeouts, instruction counts, and growth-curve complexity inference~\citep{du2024mercury,huang2024effibench,qing2025effibenchx,liu2024evalperf,qiu2025enamel,peng2025coffe,he2025sweperf,ma2025swefficiency,chambon2025bigobench}. These tools do not directly become online RL rewards: stress execution and curve fitting can be too expensive, instruction counts move away from wall-clock time, and raw timing on small tests is noisy enough that PIE reports 1.91x spurious speedups on identical code~\citep{shypula2024pie}. The missing piece is a training-compatible timing signal.

\paragraph{Learning efficient code.}
Prior efficiency-training work includes performance-feedback fine-tuning, including online RL on lower-level optimization tasks~\citep{nichols2024performancealigned,wei2025supercoder,mikasa2026hpc}, and efficiency-aware supervised fine-tuning on curated efficient-code data~\citep{huang2025efficoder}. PIE learns from offline slow-fast edit pairs~\citep{shypula2024pie}, while other systems search for or iteratively refine faster programs at inference time without updating the policy~\citep{peng2025perfcodegen,gao2025sbllm,ou2025maxcode,romeraparedes2023funsearch,novikov2025alphaevolve}. These approaches are complementary, but easier for optimization: the system starts from an existing program, or receives repeated execution feedback during search. In one-shot generation, the model must choose the algorithm, data structures, and implementation before seeing runtime feedback.

\paragraph{Closest RL setting.}
The closest recent comparison is Afterburner~\citep{du2025afterburner}. It trains with GRPO in an iterative-refinement setting, where the model receives an existing solution plus runtime metrics and additively mixes format, correctness, and efficiency rewards. That recipe improves efficiency, but also increases the fraction of solutions slower than all human references from 0.33\% to 7.33\%. We study the harder one-shot setting, where the LLM must generate from scratch, in a single return and without execution feedback, a correct solution that ranks as high as possible among reference competitive-programmer solutions. We still build on GRPO~\citep{shao2024deepseekmathpushinglimitsmathematical} and recent analyses of clipping, baselines, and degenerate groups~\citep{yu2025dapo,hao2025opo,guo2025abcgrpo,zheng2025spo,liu2025understanding}, but focus on the full timed-execution recipe: a measurable signal, a correctness-preserving reward, and stable training under sparse noisy returns.

We conduct an extensive literature review of how prior work defines optimization constraints, either as benchmark evaluations or as training objectives, and aggregate the classification in \cref{sec:app_related_efficiency_taxonomy}.

\section{How to get a reliable timing measure}
\label{sec:measurement}

Timing-based RL does not work well by simply adding runtime to the reward on the same base data. When rewarding shorter execution time on top of RLVR with base DMC problems, we see gains of at most 5\% under optimization-constrained evaluation (at \(p_{50}\)), with pure-correctness scores moving by only \(-1\%\) to \(+3\%\). The tests accompanying code problems must be improved: their execution must reject wrong code, separate correct solutions by duration, and preserve optimization differences even under timing-measurement noise. Raw DMC does not fulfill these requirements: on the final RL split, original tests average $0.088\,\text{s}$ and reach only $0.145/0.463\,\text{s}$ at $p95/p99$, so tens-of-milliseconds perturbations can dominate decisions (\cref{app:simulation}).

\begin{figure*}[t!]
\centering
\includegraphics[width=\textwidth]{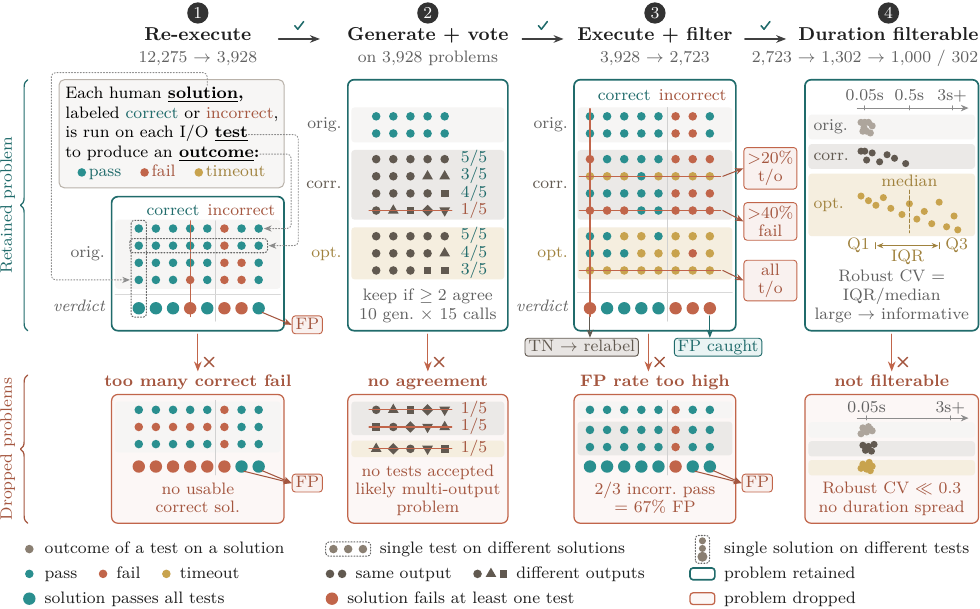}
\caption{\textbf{DMC-Optim construction pipeline.}
From $12{,}275$ raw DMC problems, we re-execute human solutions, add $430{,}215$ correctness tests and $352{,}740$ optimization tests, and filter at the test, solution, and problem levels to obtain a $2{,}723$-problem cleaned corpus.
A final duration-filterability gate retains the $1{,}302$ timing-based RL problems, split into $1{,}000/302$ train/test.
More specifically, for each problem, we augment the original (orig.) tests with generated correctness (corr.) and optimization (opt.) tests, then filter at multiple levels.
Step 1: re-execute human solutions on original tests to confirm labels and catch false positives (FP).
Step 2: generate 10 LLM-written input generators per problem, sample 15 candidate tests from each, and keep only those with output agreement across human solutions.
Step 3: filter at three levels: tests with excessive timeouts or exceptions, contradicted solutions, and problems with few correct solutions or high FP rates.
Step 4: retain problems whose optimization tests span a broad duration range, which we call duration filterable.
Strikethroughs mark removed tests (rows) or relabeled solutions (columns); red boxes are examples of dropped problems.
}
\label{fig:adjudication_pipeline}
\end{figure*}

\paragraph{DMC-Optim builds measurable timing feedback.}
\label{sec:data}
\label{sec:data_sources}
\label{sec:data_test_gen}
\label{sec:data_improving}
We choose to separate the correctness target from optimization in the set of tests used to evaluate the solutions. From 12{,}275 DMC problems, we re-execute human solutions, use verified correct and incorrect submissions as controls, add correctness tests, then generate large-input optimization tests; \cref{fig:adjudication_pipeline} summarizes the corpus. Adjudication is ambiguous: a new failure can mean an invalid test, a mislabeled solution, or an unsuitable problem. We filter at test, solution, and problem levels, trading recall for cleaner labels and enough RL scale. This is detailed in \cref{app:dataset}.

\paragraph{Duration filterability selects the usable subset.}
\label{sec:data_filterability}
Duration filterability is the final gate: averaged verified-correct human runtimes over optimization tests must have robust CV at least $0.3$. Original tests satisfy this for at most 3.8\% of problems; optimization tests reach 48.2\%. On the training split, $p95/p99$ durations are $1.296/3.710\,\text{s}$ versus $0.145/0.463\,\text{s}$ for original tests, enough spread for ranking. Ablations agree: non-filterable training loses strict \(p_{10}\) performance in three of four optimization settings; the remaining character-length setting gains only 3.8\% relative at \(p_{10}\), but remains far below ranked \(p_{30}\) on the duration-filterable pool. For ranked \(p_{30}\), replacing duration-filterable problems by non-filterable ones drops pass@$1$ by 43.5\% relative at \(p_{30}\) and 50.0\% at \(p_{10}\) after 5k training steps.

\paragraph{Local sandbox versus isolated remote execution.}
\label{sec:execution}
\label{sec:ces_noise}
Better tests still fail under the wrong backend. Local execution consists in executing code on the training workers, where sandbox timing contends with inference and rollout orchestration. The other option is to run code executions on I/O tests on a separate dedicated CPU cluster, where each execution is isolated so as to minimize contention; in the rest of the paper, we call this remote sandbox service CES. In standalone reruns of unchanged code on the same tests, local timing moves the mean-percentile ranking of this code against a fixed reference leaderboard of human solutions by 41.2 percentage points on average. When trying to map local durations to our remote sandbox environment, which correctly isolates executions, we only obtain fits with negative cross-validated \(R^2\) on 36{,}660 pairs. We therefore use CES: it isolates timing from worker contention; short-run variability is small; affine service-state drift correction raises stored-vs-fresh Spearman from 0.54 to 0.96, giving \cref{sec:reward} calibrated, correctness-gated timings. \Cref{app:ces_fallback} further explores how to make the execution environment stable enough for timing measurements to be usable by the RL loop.

\section{How to turn it into a learnable reward}
\label{sec:reward}

Reliable timing measurements are still not a usable RL signal. \Cref{sec:measurement} gives us a calibrated execution setup with improved I/O tests that can reject wrong code and accurately compare correct solutions by duration. But while this setup is required for downstream optimization RL, it is not sufficient for obtaining significant gains: naively rewarding direct aggregate duration across tests raises optimization-constrained scores by 15\% (at \(p_{50}\)), while dropping pure-correctness pass@1 by 2--4\%. Designing a proper RL environment and corresponding reward therefore means deciding how the optimization constraint enters the RL loop, so that training improves the policy's ability to generate optimized code without rewarding fast-but-wrong programs. We classify a broad variety of optimization-RL environments under 3 intervention points: timing constraints can enter before execution, during execution, or after execution.

\begin{figure*}[t!]
\centering
\includegraphics[width=\textwidth]{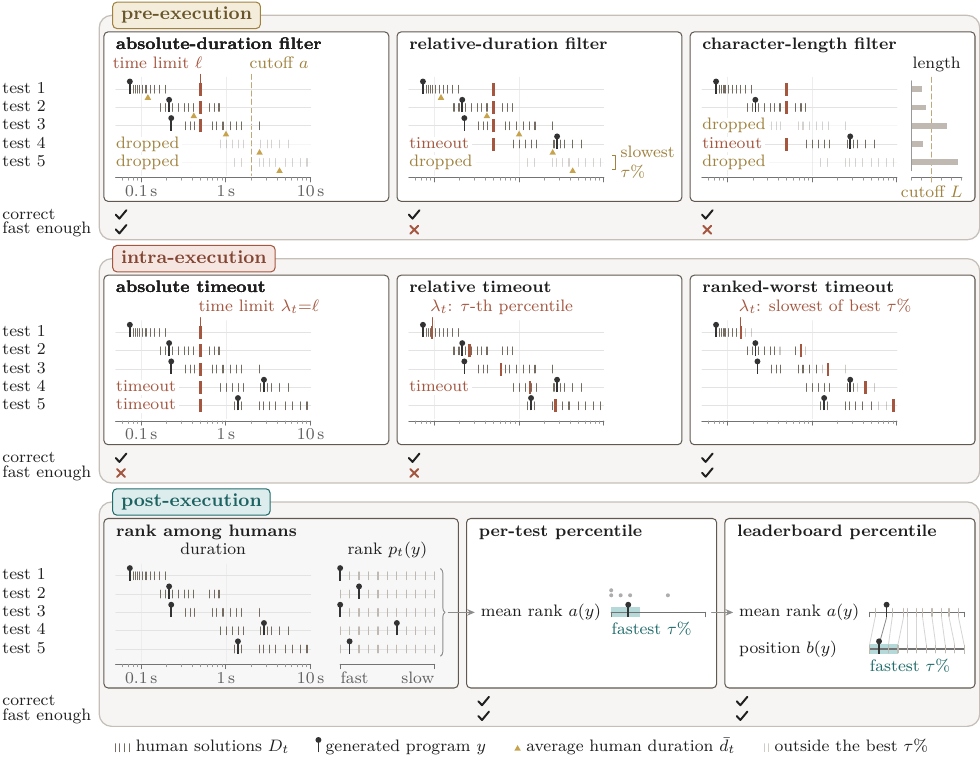}
\caption{\textbf{Different RL environments with different ways of inserting an optimization pressure.}
\emph{Top, pre-execution:} a filter chooses the optimization tests; dropped tests (faded) never run, kept tests run under a corresponding time limit \(\ell\) (red bars).
The absolute-duration filter keeps \(\bar d_t<a\); the relative-duration filter drops the slowest \(\tau\%\) of tests by \(\bar d_t\); the character-length filter keeps input + output \(<L\) characters.
\emph{Middle, intra-execution:} every test runs, and a run with \(\delta_t(y)\geq\lambda_t\) becomes a timeout (red bars).
Absolute: \(\lambda_t=\ell\); relative: \(\lambda_t=\max\{10^{-3},\operatorname{Percentile}_{\tau}(D_t)\}\); ranked-worst: the slowest duration on test \(t\) among the best \(\tau\%\) of human solutions by mean per-test rank.
In both families this comes with a tunable timeout tolerance \(\rho\).
\emph{Bottom, post-execution:} one ranking step, two scalarizations, on top of the naive raw duration approach.
Tests run under no specific timeout limit; on each, the program's normalized percentile \(p_t(y)\) is its rank among the reference solutions on this same test.
The per-test percentile then averages these ranks into an aggregate \(a(y)\); the leaderboard percentile ranks \(a(y)\) among the humans' own mean ranks.
Every environment returns strict correctness \(\tilde c_{\mathrm{cor}}\) (all correctness tests pass; executed optimization tests may time out but not fail), the optimization gate \(g\) and a graded scalar \(q\): \(\phi\) for pre and intra or \(a(y)\) or \(b(y)\).}
\label{fig:env_definition}
\end{figure*}

\begin{figure*}[t!]
\centering
\includegraphics[width=\textwidth]{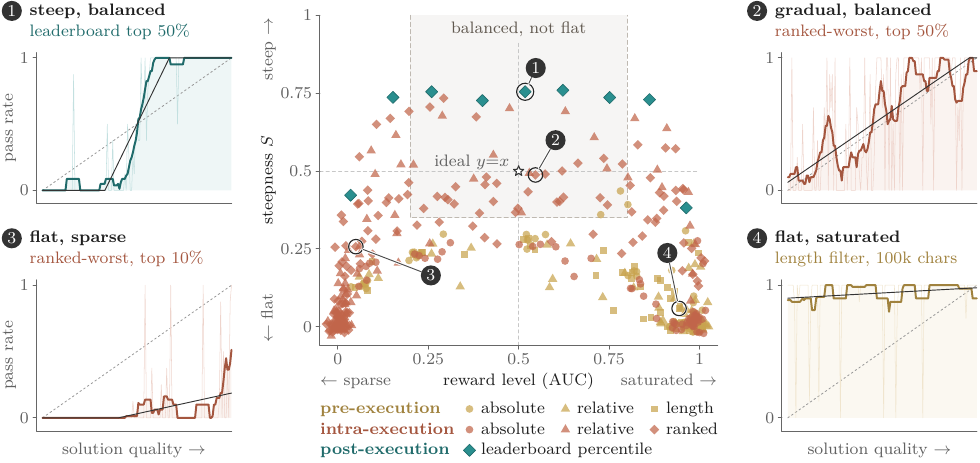}
\caption{\textbf{Offline simulator screening of optimization-aware environments.}
Each point is one environment configuration, positioned by two simulator diagnostics: AUC, the area under the simulated pass-rate curve as sampled solution quality increases, and steepness \(S=a/(1+|a|)\), the bounded slope of a clamped-linear fit to that curve.
The AUC axis separates sparse from saturated settings, while steepness shows whether the transition from weak to strong sampled solutions is flat or correlated.
Color marks the intervention point---pre-execution filtering, intra-execution time limits, or post-execution ranking---and shape marks the subclass; insets show representative curves.
Running thereafter actual RL runs with Qwen 2.5 7B, strict \(p_{30}\) performance correlates most with ordering diagnostics: deviation from \(y=x\) gives \(r_s=-0.832\) (\(p=2\times 10^{-5}\)), raw monotonicity \(r_s=0.787\) (\(p=10^{-4}\)), and steepness \(r_s=0.723\) (\(p=7\times 10^{-4}\)).
}
\label{fig:env_quality_landscape}
\end{figure*}

\paragraph{Where timing enters the environment.}
\label{sec:taxonomy}
We define the optimization-RL environment space by looking at where prior benchmarks and training objectives place the optimization constraint, then also adding new formulations in the same framework. This gives three broad intervention points: pre-execution filtering, intra-execution time constraints, and post-execution ranking. \Cref{sec:app_related_efficiency_taxonomy} shows how previously published work fits into this taxonomy. \emph{Pre-execution} families filter optimization tests before the candidate solution is run, using aggregated absolute duration, within-problem relative duration, or input-output length as a workload proxy; they can remove unstable tails, and are less dependent on reference solutions, but can provide less refined optimization pressure across heterogeneous workloads. \emph{Intra-execution} families impose time constraints while tests run, from global time limits to limits derived from strong human references; they are literal efficiency constraints, but can cut the entire learning signal if not properly tuned. \emph{Post-execution} families leave the execution intact and aggregate durations afterward, typically by inserting the candidate into the calibrated human pool and ranking it across optimization tests. This keeps the most flexibility after execution, since the same measured solution can be reused under different ranking rules or percentile thresholds, but it often requires reference duration distributions or other extra data. \Cref{app:reward} defines the exact environments and families of environments that we explore. See \cref{fig:env_definition}.

\paragraph{The reward-facing interface.}
\label{sec:two_gate}
After execution, all environments reduce to three reward-facing quantities \((c,g,q)\). The correctness gate \(c\in\{0,1\}\) checks the fixed correctness tests; the optimization gate \(g\in\{0,1\}\) checks whether the optimization branch satisfies the timeout tolerance and a potential percentile ranking threshold; the optional graded signal \(q\in[0,1]\) is a continuous equivalent that is not gated. Time-limit environments can expose timeout rate as \(q\). Post-execution ranking environments can return a scalar percentile ranking by comparing each measured duration to calibrated human references, aggregating per-test percentiles, and optionally converting the aggregate into a leaderboard percentile. For this per-test aggregation operation, we screened 17 ranking metrics under repeated noisy measurements and kept the mean-based percentile family because it was the most stable and discriminative: its rerun noise remains small, with scores moving by \(3.0 \pm 1.0\) percentile points under repeated measurements, while its dynamic range remains large, with strong and weak solutions separated by 47 points on average. The full study of ranking functions is in \cref{sec:app_ranking_metrics}.

When using a leaderboard-percentile post-execution environment as an evaluation, this directly defines the \(p_{\tau}\) family of pass@$k$ metrics used throughout the paper in many tables and figures. Let \(q_{\mathrm{lead}}(x,y)\in[0,1]\) be the leaderboard percentile of a generated solution \(y\) after inserting it into the calibrated human reference leaderboard for problem \(x\), with lower values faster. For an evaluation set \(\mathcal X\) and \(n\) sampled solutions \(y_1,\ldots,y_n\) per problem, define
\begin{align}
m_{\tau}(x)
&=
\sum_{i=1}^{n}
\mathbb{I}\!\left\{
\tilde c_{\mathrm{cor}}(x,y_i)=1
\;\wedge\;
q_{\mathrm{lead}}(x,y_i)\leq \frac{\tau}{100}
\right\},\\
\mathrm{pass@}k(p_{\tau})
&=
\frac{1}{|\mathcal X|}
\sum_{x\in\mathcal X}
\left(
1-\frac{\binom{n-m_{\tau}(x)}{k}}{\binom{n}{k}}
\right).
\end{align}
Here \(\tilde c_{\mathrm{cor}}\) is the strict correctness prerequisite: it checks that all base and generated correctness tests succeed, and that no optimization test has a hard failure (but timeouts are not penalized). Thus \(p_{100}\) pass@1 does not have any leaderboard-ranking constraint and is equivalent to a pure-correctness score, while smaller \(\tau\) requires the solution to both satisfy the strict correctness prerequisite and rank in the top \(\tau\%\) of the human reference leaderboard.

\paragraph{Balancing correctness and efficiency.}
Reward composition decides how to allocate the reward range between correctness and efficiency, potentially conditioning one on the other. A reward formulation could be to keep correctness as an outer gate: \(r(x,y)=-1\) when \(c(x,y)=0\), and only correct solutions receive a positive efficiency score such as \(1-q(x,y)\), which can be kept fully continuous, bucketed, or binarized. Such conditioning should help balance learning correctness and learning optimization as policy training proceeds, while avoiding reward-hacking scenarios. We compare various families of rewards, from binary to collapsed, including multitask variants and additive blends.

\paragraph{Pruning the space of candidate RL environments.}
\label{sec:simulator}
Taken together, the many forms of RL environments, including the reward they use, create a large space of RL training possibilities, while each online run uses 8--32 GPU nodes for hours, if not days. We therefore screen with an offline simulator before online GRPO, and try to select the most promising candidates. The simulator does not predict exact learning curves: it keeps the environment and reward computation fixed, replaces model generations by sampled human solutions, replaces live CES calls by precomputed calibrated durations, and asks whether better sampled solutions receive better rewards without saturating, or being overwhelmed by noise. \Cref{fig:env_quality_landscape} uses AUC to expose reward signals that are too sparse or saturated, and steepness to expose reward curves that are too flat; downstream DMC-Optim post-RL test performance correlates the most with diagonal deviation, raw monotonicity, and steepness, while curve noise is not predictive at \(p_{30}\) (\(r_s=-0.002\), \(p=0.994\)). This prunes the environment/reward space before \cref{sec:training} runs GRPO with the sparse, noisy returns tied to live code execution. \Cref{app:rl_simulator} provides more details about the offline RL simulations.

\section{How to train stable RL with it, and getting the results}
\label{sec:training}
\label{sec:experiments}

\Cref{sec:measurement} introduced the RL data and timing-measurement tools; \cref{sec:reward} then defined the RL environments and rewards to be trained on top of them.
The goal is to obtain an RL setup that learns correctness at least as well as standard RLVR, while also making the generated correct code substantially faster.
Because we have many optimization-RL environment families, and many possible parameterizations within each, \cref{sec:reward} also introduced an offline simulator to prune degenerate environments before launching online RL.
This is what we now do in this section:
\begin{enumerate}
    \item \hyperref[sec:setup]{\textbf{Experimental setup.}} We use Qwen and CWM models across two sizes: Qwen 2.5 7B to explore promising optimization-RL setups in detail, and Qwen 2.5 32B/CWM 32B to validate the results at larger scale.
    \item \hyperref[sec:eval_pipeline]{\textbf{Replayed evaluation.}} Since we discussed in detail the noise that can arise when RL training relies on timing measurements, we also adapt the evaluation setup to produce more robust results and comparisons.
    \item \hyperref[sec:baseline_controls]{\textbf{Is all of this work necessary in the first place? Exploring baselines.}} Standard RLVR and naive timing rewards do not solve the task, which is the reason why we explore other RL setups.
    \item \hyperref[sec:grpo]{\textbf{Can GRPO be stable enough for sparse, noisy timing rewards?}} On top of the methods detailed in \cref{sec:measurement,sec:reward}, GRPO itself needs to handle sparser and noisier rewards; otherwise, potential learning could be killed when aggregating the loss and taking gradient steps.
    \item \hyperref[sec:main_results]{\textbf{What optimization RL environments work best?}} Our framework tries to unify the ways previous work benchmarked the notion of code optimization, together with our own formulations; here we test which ones work best when converted into RL-training environments, first at Qwen 2.5 7B scale.
    \item \hyperref[sec:transfer]{\textbf{How do the gains transfer across different optimization evaluation criteria?}} If each optimization constraint can be both an RL training environment and an evaluation setup, we ask whether there is a universally good way to train, and whether one formulation, which can be the same or another, is best at evaluating and discriminating these training runs.
    \item \hyperref[sec:reward_shape]{\textbf{What reward shape maximizes concurrent gains in correctness and optimization?}} We have discussed the construction of the RL environment, but the reward placed on top of it also changes downstream results.
    \item \hyperref[sec:cross_model]{\textbf{If we change the model and/or scale it up, does it keep working? And if evaluated on OOD data?}} Taking the best RL environments and rewards found with Qwen 2.5 7B, we investigate whether they still work at larger scale and on out-of-distribution evaluation data.
    \item \hyperref[sec:compute_fairness]{\textbf{Do we have a fair experiment setup to make these claims, or are we cheating?}} We discuss the overall training setup, the fairness of the comparisons, and how gains could be further scaled up.
    \item \hyperref[sec:calibration_sensitivity]{\textbf{What happens if the timing environment slows down/speeds up?}} As a potential limitation, we explore whether the gains still hold when the evaluation setup, in particular the underlying timing tools, is altered.
    \item \hyperref[sec:uncertainty_estimates]{\textbf{Are these gains just noise?}} Finally, we use a variability study to check the significance of the results.
\end{enumerate}

\paragraph{Experimental setup.}
\label{sec:setup}
\label{sec:sft_setup}
We use two model families across two model sizes: Qwen 2.5~\citep{qwen25technical} at 7B and 32B scale, and CWM 32B~\citep{cwm2025}.
We choose Qwen 2.5 because the released checkpoints are pre-reasoning base models, which lets us control the reasoning trajectories shown to both sizes before optimization RL: we apply one epoch of SFT on a public reasoning-only mix built from OpenCodeReasoning-2 and OpenMathReasoning~\citep{opencodereasoning2,openmathreasoning}, and decontaminate the SFT data against DMC, LCB, and BigO(Bench)~\citep{li2022competition,jain2024livecodebench,chambon2025bigobench}.
This decontamination includes the DMC-Optim test set we build and the training prompts used for RL.
Therefore, during RL, the Qwen models should be reasoning about these problems for the first time (in the sense that they were not SFT on these problems, though examples could still appear somewhere in their broader internet pretraining data) and exploring possible optimizations from the problem statement only, rather than searching inside a prior supervised exposure.

Things are different for CWM 32B, which starts from a released SFT checkpoint already prepared for reasoning RL, so we do not control its SFT data.
The CWM authors report decontamination against LCB, but we cannot make the same claim for DMC-Optim test or for the RL training prompts used here.
Since CWM SFT uses one pass over OpenCodeReasoning, any overlapping problems likely appear with verified R1 generations rather than with the human reference solutions that define our DMC-Optim timing rankings.
Their midtraining data also includes tracing data built on top of DMC problems, using Llama 3.1 70B generations traced with some of the base DMC tests.
CWM 32B therefore probably did not see the ground-truth human solutions used for DMC-Optim evaluation and ranking (in the sense that we do not expect those human references to be part of its SFT or midtraining data, though examples could still appear somewhere in broader internet pretraining data), and did not receive optimization feedback.
This means that RL with CWM 32B instead asks whether a model that may already know some problem statements and was exposed to sampled generated solutions, though unordered and without optimization-signal supervision, can learn to search this space with an optimization criterion, and potentially discover new optimization tricks not present in the exposed data.

For the Qwen SFT stage, all GPUs are trainers: the 7B run uses 16 H100 nodes (128 GPUs, tensor parallelism 4) and the 32B run uses 32 H100 nodes (256 GPUs, tensor parallelism 8), with data parallelism 32 in both cases.
Both use sequence length 32{,}768, per-data-parallel-rank batch size 2, no gradient accumulation, and learning rate \(8.6\times10^{-6}\) after a 1k-step warmup, giving \(32\times2\times32{,}768=2{,}097{,}152\) tokens per optimizer step.
Over 24{,}829 steps, this is 52.1B packed training tokens, corresponding to 43.5B raw unpadded tokens from the reasoning-only mix.

Main RL runs use 10{,}000 optimizer steps, temperature 1.0 sampling with no top-\(p\) truncation, learning rates \(1\times10^{-7}\) for 7B and \(1.4\times10^{-7}\) for 32B, and a 50/50 split between rollout and training nodes.
A 7B RL run uses 8 NVIDIA H100 nodes for about one day, while a 32B RL run uses 32 NVIDIA H100 nodes for about a day and a half.
Although the two model sizes use different parallelization schemes, both end up with data parallelism 16, local batches capped at 32{,}000 tokens, and no trajectory splitting across batches.
With 16 samples per prompt and a typical average of 10{,}000 tokens per rollout, this gives about 480{,}000 tokens per global batch, or roughly three problems with their 16 trajectories per optimizer step.
Since DMC-Optim train contains about 1{,}000 problems, 10{,}000 training steps correspond to roughly 30 passes over the full training set.
More setup details are shared in \cref{app:training_support}.

\paragraph{Replayed evaluation.}
\label{sec:eval_pipeline}
We evaluate optimization RL in two regimes and keep them separate in the discussion.
DMC-Optim test is the in-domain benchmark for measuring whether a model produces correct code that also meets strict human-speed percentile thresholds.
LiveCodeBench (LCB)~\citep{jain2024livecodebench} is the out-of-distribution transfer check, where we report both pass@$1$ and pairwise speed win rates against the standard-RLVR baseline.
As \cref{sec:app_empirical,sec:app_simulation} show, LCB tests are too small and fast to support reliable per-test timeout sweeps: changing timeout thresholds moves little signal, while win rate remains more robust because it compares accumulated durations across all tests.
In order to spare compute usage, and try to mitigate potential noise across re-runs, for each evaluation step, we first run LLM generation and sample 20 solutions per problem.
We then run timing in shared CES runs: tests from all problems and all model dumps are shuffled and executed concurrently, and human-reference solutions are included to calibrate durations across evaluation runs.
Only after this timing pass do we replay the stored per-test verdicts and durations through different percentile thresholds, time budgets, affine calibrations, and LCB win-rate comparisons without re-running either the model or the timing sandbox.

\paragraph{Is all of this work necessary in the first place? Exploring baselines.}
\label{sec:baseline_controls}
Before comparing different types of optimization RL environments, we first check whether standard RLVR or direct timing rewards already achieve good code-optimization performance; if so, the work on refining the dataset, sandbox, and rewards would be unnecessary.
As seen in \cref{tab:qwen7b_baseline_duration_rewards}, the first reference is standard RLVR on Qwen 2.5 7B: it uses the base tests only, rewards correctness, and after 10k RL steps reaches 43.5\% pass@$1$ at \(p_{100}\), which corresponds to correctness with an infinite-timeout evaluation.
This then degrades sharply: performance drops by almost 60\% at \(p_{50}\), where the solution must rank in the top half of the human-solution leaderboard for the problem.
Adding our generated correctness and optimization tests, which reduce false positives and provide longer executions, improves pure pass@$1$ to 46.9\% at \(p_{100}\), but the stricter-threshold performance does not move substantially: \(p_{50}\) sees only a 14\% relative gain and \(p_{30}\) a 21\% relative gain over standard RLVR.

We then explored a direct timing reward: among correct solutions, faster executions receive higher rewards.
For each solution, we measure the average runtime across tests and assign a proportional reward only when the solution is correct.
This does not resolve the problem and leads to disappointing results: on base tests, stricter-percentile performance barely improves, reaching 18.9\% and 18.0\% pass@$1$ at \(p_{50}\) under the two reward shapings, compared to 18.0\% for standard RLVR.
We believe that, beyond the poorly refined reward shape, this is mainly because the base tests are poorly suited to any form of optimization feedback.

Using the optimization tests gives a more satisfying signal: relative to standard RLVR on base tests, direct timing rewards improve pass@$1$ by up to 21\% at \(p_{50}\) and 44\% at \(p_{30}\).
The picture is more nuanced if we isolate the reward change by comparing to the RLVR reference that already uses the generated correctness and optimization tests: in that comparison, the best \(p_{50}\) gain is only 5\%, and it comes with an 8\% relative degradation at \(p_{100}\).
We therefore trade some pure-correctness performance for stricter-percentile performance, and lose some of the total aggregated performance in the translation.
This trade-off is further explored by \citet{zheng2026extrapolativeweightaveragingreveals}, who show that code RL methods expose correctness-efficiency frontiers rather than a single monotonic improvement direction.
The added optimization tests do provide optimization signal, but more work is needed to make RL push correctness and efficiency together.
At this point, the naive raw-duration reward either fails when the data is not prepared for optimization feedback, or works only by trading off correctness for efficiency.
Unfortunately, the strongest baseline remains standard RLVR with improved underlying data.

\begin{table}[t!]
\refstepcounter{table}
\label{tab:qwen7b_baseline_duration_rewards}
\normalsize
\noindent\textbf{Table~\thetable: Qwen 2.5 7B baseline RLVR and naive-duration rewards on DMC-Optim test.}
Rows compare standard correctness-only RLVR with base tests, with our enriched correctness tests and optimization tests, and raw-duration rewards on either base tests or optimization tests as well.
Columns report pass@$1$ under the leaderboard-percentile post-execution environment at different percentile levels.
Here, \(p_{100}\) is equivalent to correctness-only pass@$1$, whereas \(p_{50}\) requires solutions to be both correct and rank in the top 50\% of the human reference solutions for the same problem.
\par\vspace{0.35em}

\centering
\small
\setlength{\tabcolsep}{5.0pt}
\begin{tabular*}{0.78\textwidth}{@{\extracolsep{\fill}}l rrrr@{}}
\toprule
& \multicolumn{4}{c}{\bfseries pass@$1$} \\
\cmidrule(lr){2-5}
\bfseries RL training environment & \bfseries $p_{100}$ & \bfseries $p_{50}$ & \bfseries $p_{30}$ & \bfseries $p_{10}$ \\
\midrule
\multicolumn{5}{@{}l}{\textsc{RLVR references}} \\[1pt]
\quad Standard RLVR & 43.5 & 18.0 & 7.7 & 1.9 \\
\quad Our enriched correctness tests + optimization tests & 46.9 & 20.6 & 9.3 & 2.7 \\
\midrule
\multicolumn{5}{@{}l}{\textsc{Naive raw-duration reward}} \\[1pt]
\quad Base tests, linear & 44.9 & 18.9 & 8.3 & 2.3 \\
\quad Base tests, log & 43.2 & 18.0 & 8.3 & 2.3 \\
\quad Optimization tests, linear & 43.1 & 21.7 & 10.9 & 3.6 \\
\quad Optimization tests, log & 42.5 & 21.2 & 11.1 & 3.5 \\
\bottomrule
\end{tabular*}
\end{table}

\paragraph{Can GRPO be stable enough for sparse, noisy timing rewards?}
\label{sec:training_pipeline}
\label{sec:grpo}
Standard RLVR on base tests shows the limited optimization capability of post-trained LLMs: correctness improves, but the model still loses almost 60\% from \(p_{100}\) to \(p_{50}\).
We showed that improving the underlying RL data already helps optimization capabilities with RLVR.
The goal of our work is to find the environment and reward formulation that allows us to squeeze more gains from that improved data, especially as the direct timing reward above did not bring the expected results.
Before launching RL runs and exploring our candidate environments, we took a look at GRPO itself, the update rule that ultimately turns any of these scalar rewards into model-weight changes: we need to make sure that potentially better optimization RL environments do not get shadowed by bad loss aggregation and inherently noisy optimizer steps.

The offline simulator in \cref{sec:simulator} shows that even after the sandbox and timing tool improvements, our environment candidates remain tied to noisy timing measurements and give sparse rewards.
Compared with correctness-only RLVR, correctness-plus-optimization rewards have fewer useful rollouts and more zero-advantage groups (when all rollouts get the same reward, either all \(1\) when the RL problem is too easy, or more frequently in our case all \(-1\) when not a single rollout manages to solve it, at least partially); on a Qwen 2.5 7B run, for example, the zero-advantage-context rate at the beginning of training can be around 40--50\% for binary optimization rewards like the ones we evaluate.
The sandbox state can also vary under concurrent CES load, and therefore the ranges of the timing measurements that are performed can shift, so two rewards collected at different times are not perfectly comparable even if the model policy has not changed: we want to avoid giving more reward to a rollout that is in theory less optimized than another one, simply because the sandbox state was more favorable for the first rollout and made it run faster.

For these reasons, all optimization RL runs below use the same stabilized training recipe.
We keep a standard async-GRPO structure (that we detail in \cref{app:training_support}), but improve the following points:
\begin{enumerate}
    \item We first increase the number of same-prompt rollouts on the worker side: this helps with zero-advantage problems and makes runs reach useful updates faster while dropping fewer batches of data; it also improves the quality of the Monte Carlo estimates used in the advantage computation, which is helpful when individual rewards can be noisy.
    This reduces the number of problems seen per global trainer batch, but decreases the wall-clock duration of the RL runs.
    \item On the trainer side, we also increase the training batch size, improving pure-correctness performance in RL runs by 10--25\% across different optimization RL environments, optimization performance at \(p_{50}\) by up to 35\%, and optimization performance at \(p_{30}\) by up to 60\%.
    This is consistent with the fact that batches that are too small become prone to unstable gradient estimates under timing noise, and enlarging these batches helps smooth it out.
    \item Following the Dr. GRPO analysis that standard-deviation normalization can introduce a difficulty bias~\citep{liu2025understanding}, we center returns but do not divide by the group standard deviation.
    This matters more for timing-based optimization rewards: we do not want to overweight prompts where the model does not find optimization tricks, or prompts where a single timing fluctuation creates an isolated reward difference, relative to prompts where rollouts expose a variety of optimization tricks and therefore carry a more useful teaching signal.
    \item At the problem level, we use a token-weighted prompt mean for the advantage baseline and normalize the loss by a fixed token horizon \(N=32768\), rather than by each rollout length.
    The token-weighted mean keeps the global batch gradient neutral, so total positive and negative gradients cancel.
    The fixed horizon corrects the bias of per-rollout length normalization, which would not penalize long incorrect trajectories enough and would under-reward successful deep-reasoning trajectories.
    This is especially important in our setting, where rewards can be very sparse and the model can get lost in indefinite incorrect reasoning.
    With this setting, by the end of training, models tend to generate around 9k tokens on easy problems, while hard problems average around 16k.
    We do not observe a particular difference between RLVR training and optimization training in reasoning length.
    \item Finally, to limit stale timing comparisons, we first tune the worker/trainer ratio so generated batches are consumed promptly: we prefer mild trainer-side staleness over worker-side staleness, although this slightly hurts GPU utilization, because queued worker-side batches can become stale and be mixed with batches collected under different sandbox states.
    On top of this, we enforce an additional hard filtering of contexts getting too old, discarding contexts older than \(S_{\max}=30\) optimizer steps, which gives a 5\% score gain.
    Finally, we avoid a replay buffer that would reuse rewards from older sandbox states.
\end{enumerate}
More details are shared in \cref{app:training_support}.

\paragraph{What optimization RL environments work best?}
\label{sec:main_results}
Taking the full pipeline built so far, the goal is to find an optimization RL setup, first on Qwen 2.5 7B, that improves over the strongest standard-RLVR data baseline: using our enriched correctness and optimization tests, this baseline reaches 46.9 pass@$1$ at \(p_{100}\) and 20.6 at \(p_{50}\).
\Cref{tab:qwen7b_environments} reports the optimization-aware RL environments we tried, sorted by the environment families introduced in \cref{sec:taxonomy,sec:app_env_qwen7b_definitions}.
The details and mathematical formulations of each environment are in \cref{sec:app_env_qwen7b_definitions}; overall, the three main families of environments are:
\begin{enumerate}
    \item Pre-execution environments, where the optimization constraint is introduced by selecting which optimization tests evaluate the generated solution; one parameterization, for example, filters tests by the mean runtime of human reference solutions.
    \item Intra-execution environments, which introduce the optimization constraint through the timeout itself.
    \item Post-execution environments, which leave execution intact, then manipulate the recorded durations by ranking generated solutions against human references.
\end{enumerate}
We report pass@$1$ and pass@$10$ under several percentile thresholds: \(p_{100}\) corresponds to pure-correctness evaluation, while \(p_{10}\) requires solutions to be both correct and in the top 10\% of the human reference solutions for the same problem.

Across these three families of environments, we explore three pre-execution subfamilies, three intra-execution subfamilies, and two post-execution subfamilies, each parameterized by thresholds or percentile cutoffs.
We use the offline simulator to remove configurations that are too degenerate: too sparse, too flat because weak and strong solutions receive similar rewards, or too noisy.
We also skip parameterizations that produce nearly identical environment behavior and run one representative instead.
This gives the set of parameterized RL environments explored in \cref{tab:qwen7b_environments}.
Based on this sweep, what works best?

Looking first at pass@$1$, the best \(p_{100}\) values remain around 47, so the optimization-aware environments do not substantially improve pure correctness over the RLVR baselines, but they also do not degrade it.
This is already progress over the naive duration rewards from \cref{tab:qwen7b_baseline_duration_rewards}, which traded away correctness performance.
The best strict scores are around 31 at \(p_{50}\), 19 at \(p_{30}\), and 6 at \(p_{10}\).
These are noticeable gains over standard RLVR on base tests: that baseline drops by roughly 60\% from 43.5 at \(p_{100}\) to 18.0 at \(p_{50}\), and by roughly 80\% to 7.7 at \(p_{30}\).
By contrast, the strongest rows, in particular the top-30\% per-test-percentile post-execution configuration, reach 31.3 at \(p_{50}\) and 19.1 at \(p_{30}\), which is about a 75\% relative gain at \(p_{50}\) and a 150\% relative gain at \(p_{30}\) over standard RLVR.

The gains remain substantial against RLVR with optimization tests as well: the same row improves \(p_{50}\) by roughly 50\% and \(p_{30}\) by roughly 100\%, while staying in the same pure-correctness range.
The mild \(p_{100}\) improvement over base-test RLVR should mostly be read as an improved-data effect: the standard-RLVR baseline already moves from 43.5 to 46.9 when trained with our enriched correctness and optimization tests.

The strict optimization gains are also not only a pass@$10$-to-pass@$1$ transfer effect, where RL would make the first sample closer to the best sample among ten while leaving the sampled solution distribution unchanged.
Pass@$10$ itself improves substantially: standard RLVR on base tests reaches 41.1 at \(p_{50}\) and 7.4 at \(p_{10}\), while the top-30\% per-test-percentile post-execution configuration reaches 56.1 and 18.6.
This suggests that the model changes the sampled solution distribution and learns new optimization behavior, rather than only better aligning pass@$1$ with optimization knowledge already induced by standard RLVR.

\begin{table}[t!]
\refstepcounter{table}
\label{tab:qwen7b_environments}
\normalsize
\noindent\textbf{Table~\thetable: Qwen 2.5 7B optimization environment comparison on DMC-Optim test.}
Rows are RL training configurations from the taxonomy of \cref{sec:taxonomy}, grouped by families of rewards according to where the optimization constraint is inserted relative to code execution.
Columns first report pass@$1$ and then pass@$10$, each at \(p_{100}\), \(p_{50}\), \(p_{30}\), and \(p_{10}\).
The first block isolates standard RLVR; the next adds optimization and more-correctness tests from our data pipeline; the remaining blocks add rewards before, during, or after execution.
The \(p_{100}\) columns check correctness preservation, while \(p_{50}\), \(p_{30}\), and \(p_{10}\) progressively emphasize stricter speed.
Unless marked otherwise, all models are trained with the same async-GRPO setup and evaluated after 10k steps.
Rows marked \({}^{*}\) are timeout-relative ablations stopped at 5k because they produced too many zero-advantage groups and training became too slow; gray rows are 5k references from nearby configurations, not additional ablations.
Bold marks the best model result in each column.
We used group-averages when possible, and further study variability at the end of this section.
\par\vspace{0.35em}

\centering
\small
\setlength{\tabcolsep}{2.0pt}
\newcommand{\refcell}[1]{\textcolor{black!50}{#1}}
\begin{tabular*}{\textwidth}{@{\extracolsep{\fill}}l rrrr rrrr@{}}
\toprule
& \multicolumn{4}{c}{\bfseries pass@$1$} & \multicolumn{4}{c}{\bfseries pass@$10$} \\
\cmidrule(lr){2-5} \cmidrule(lr){6-9}
\bfseries Configuration & \bfseries $p_{100}$ & \bfseries $p_{50}$ & \bfseries $p_{30}$ & \bfseries $p_{10}$ & \bfseries $p_{100}$ & \bfseries $p_{50}$ & \bfseries $p_{30}$ & \bfseries $p_{10}$ \\
\midrule
\multicolumn{9}{@{}l}{\textsc{Standard RLVR}} \\[1pt]
\quad Standard RLVR                 & 43.5 & 18.0 & 7.7 & 1.9 & 65.1 & 41.1 & 22.9 & 7.4 \\
\midrule
\multicolumn{9}{@{}l}{\textsc{+ Optimization/More Correctness Tests}} \\[1pt]
\quad More Correctness Tests (MC)   & 46.7 & 19.8 & 8.7 & 2.4 & 67.0 & 44.7 & 26.0 & 8.3 \\
\quad MC + Large 10s Timeout        & 46.4 & 19.8 & 8.8 & 2.3 & 68.2 & 46.8 & 27.9 & 8.9 \\
\quad MC + Optimization tests       & 46.9 & 20.6 & 9.3 & 2.7 & 67.6 & 47.2 & 28.2 & 9.6 \\
\midrule
\multicolumn{9}{@{}l}{\textsc{+ Reward: Pre-execution} (test filtering)} \\[1pt]
\quad \textit{Absolute duration filter} & & & & & & & & \\
\quad\quad $\tau=0.1$\,s            & 46.7 & 20.2 & 9.2 & 2.5 & 67.6 & 47.0 & 28.7 & 9.2 \\
\quad\quad $\tau=0.5$\,s            & 46.5 & 25.3 & 13.3 & 4.3 & 67.2 & 51.9 & 34.9 & 13.8 \\
\quad\quad $\tau=2$\,s              & 46.9 & 29.2 & 17.4 & 5.7 & 67.5 & 54.7 & 39.0 & 17.3 \\[1pt]
\quad \textit{Character-length filter} & & & & & & & & \\
\quad\quad $\geq 1$k chars          & 46.7 & 20.0 & 8.8 & 2.0 & 67.6 & 46.2 & 27.7 & 7.3 \\
\quad\quad $\geq 100$k chars        & 46.4 & 21.2 & 10.3 & 3.1 & 67.7 & 48.1 & 28.8 & 11.4 \\
\quad\quad $\geq 10$M chars         & 45.6 & 28.6 & 16.8 & 5.5 & 65.2 & 54.0 & 37.9 & 16.4 \\[1pt]
\quad \textit{Relative duration filter} & & & & & & & & \\
\quad\quad top 30\%                 & \textbf{47.4} & 22.8 & 11.7 & 3.7 & \textbf{68.4} & 48.9 & 32.4 & 12.4 \\
\quad\quad top 50\%                 & 47.0 & 21.5 & 10.2 & 3.0 & 67.6 & 46.9 & 27.7 & 10.1 \\
\quad\quad top 80\%                 & 47.2 & 21.3 & 9.7 & 2.7 & 67.0 & 46.8 & 28.0 & 10.3 \\
\midrule
\multicolumn{9}{@{}l}{\textsc{+ Reward: Intra-execution} (timeout-based)} \\[1pt]
\quad \textit{Absolute timeout} & & & & & & & & \\
\quad\quad $T=0.5$\,s               & 47.3 & 29.4 & 17.5 & 5.5 & 67.3 & 55.9 & 41.0 & 16.9 \\
\quad\quad $T=2$\,s                 & 46.6 & 22.1 & 10.6 & 3.4 & 66.8 & 48.7 & 30.4 & 11.8 \\[1pt]
\quad \textit{Ranked-worst timeout} & & & & & & & & \\
\quad\quad from $p_{20}$ ranked     & 39.2 & 26.6 & 16.4 & 5.6 & 61.4 & 49.8 & 37.9 & 16.6 \\
\quad\quad from $p_{50}$ ranked     & 45.7 & 23.6 & 12.2 & 3.7 & 66.6 & 48.7 & 32.8 & 11.3 \\
\quad\quad from $p_{80}$ ranked     & 45.0 & 29.3 & 17.7 & \textbf{6.0} & 65.8 & 54.3 & 39.8 & 17.2 \\[1pt]
\quad \textit{Relative timeout ablation} & & & & & & & & \\
\quad\quad \refcell{ref.\ abs filter 2s @5k} & \refcell{42.9} & \refcell{21.1} & \refcell{10.4} & \refcell{3.4} & \refcell{65.4} & \refcell{48.3} & \refcell{31.1} & \refcell{11.5} \\
\quad\quad \refcell{ref.\ lead.\ per.\ 30\% @5k} & \refcell{40.9} & \refcell{26.8} & \refcell{16.1} & \refcell{5.2} & \refcell{65.0} & \refcell{53.7} & \refcell{40.5} & \refcell{15.2} \\
\quad\quad top 30\%\({}^{*}\)       & 31.0 & 19.4 & 10.8 & 3.9 & 56.6 & 44.4 & 31.2 & 12.5 \\
\quad\quad top 50\%\({}^{*}\)       & 35.5 & 21.5 & 12.6 & 4.4 & 60.0 & 47.1 & 35.1 & 15.6 \\
\quad\quad top 80\%\({}^{*}\)       & 35.1 & 21.4 & 12.2 & 4.7 & 59.2 & 48.2 & 33.4 & 14.8 \\
\midrule
\multicolumn{9}{@{}l}{\textsc{+ Reward: Post-execution} (ranking-based)} \\[1pt]
\quad \textit{Leaderboard percentile} & & & & & & & & \\
\quad\quad Two-gate bucketed        & 47.1 & 27.5 & 15.5 & 5.1 & 66.8 & 53.1 & 37.4 & 15.7 \\
\quad \textit{Per-test percentile} & & & & & & & & \\
\quad\quad top 80\%                 & 46.2 & 22.9 & 11.7 & 3.5 & 68.0 & 49.7 & 31.7 & 11.9 \\
\quad\quad top 50\%                 & 46.2 & 30.5 & 18.4 & 5.7 & 65.8 & \textbf{56.2} & \textbf{43.3} & 17.9 \\
\quad\quad top 30\%                 & 46.2 & \textbf{31.3} & \textbf{19.1} & \textbf{6.0} & 67.8 & 56.1 & 42.7 & \textbf{18.6} \\
\quad\quad Two-gate bucketed        & 46.5 & 25.4 & 13.8 & 4.4 & 67.5 & 52.7 & 36.2 & 13.8 \\
\bottomrule
\end{tabular*}
\end{table}

Among the families that could have been promising, one fails noticeably: relative-timeout intra-execution variants improve stricter percentiles, but their \(p_{100}\) pure-correctness scores drop by roughly 17--27\%.
They therefore do not improve optimization capability without hurting correctness.
This is the kind of optimization RL environment that moves along the correctness-efficiency Pareto frontier rather than breaking it.
Other environments land on the opposite side of the frontier: the \(\tau=0.1\,\mathrm{s}\) pre-execution absolute-duration filter preserves correctness, but gives almost the same strict-percentile profile as standard RLVR with optimization tests.
Although the environment is formulated to expose an optimization constraint, it does not make that constraint useful enough for training.

Most environment configurations in \cref{tab:qwen7b_environments} use the binary collapsed reward to include both the correctness and optimization constraints.
We discuss the form of the rewards themselves later in this section, beyond the root environment formulation.

Based on this sweep, we retain four environment settings for the follow-up experiments: the \(2\,\mathrm{s}\) absolute pre-execution filter, the ranked-worst timeout from \(p_{80}\), and the \(p_{50}\) and \(p_{30}\) per-test-percentile post-execution rewards.
On Qwen 2.5 7B, these settings reach broadly similar improvements across strict percentiles, with different tradeoffs in pure correctness.
We therefore next ask whether they remain good RL setups across other models and sizes, and whether their gains transfer when evaluated on other benchmarks.

\begin{figure}[t!]
\refstepcounter{figure}
\label{fig:improvement_profile}
\noindent\makebox[\linewidth][c]{\includegraphics[width=\textwidth]{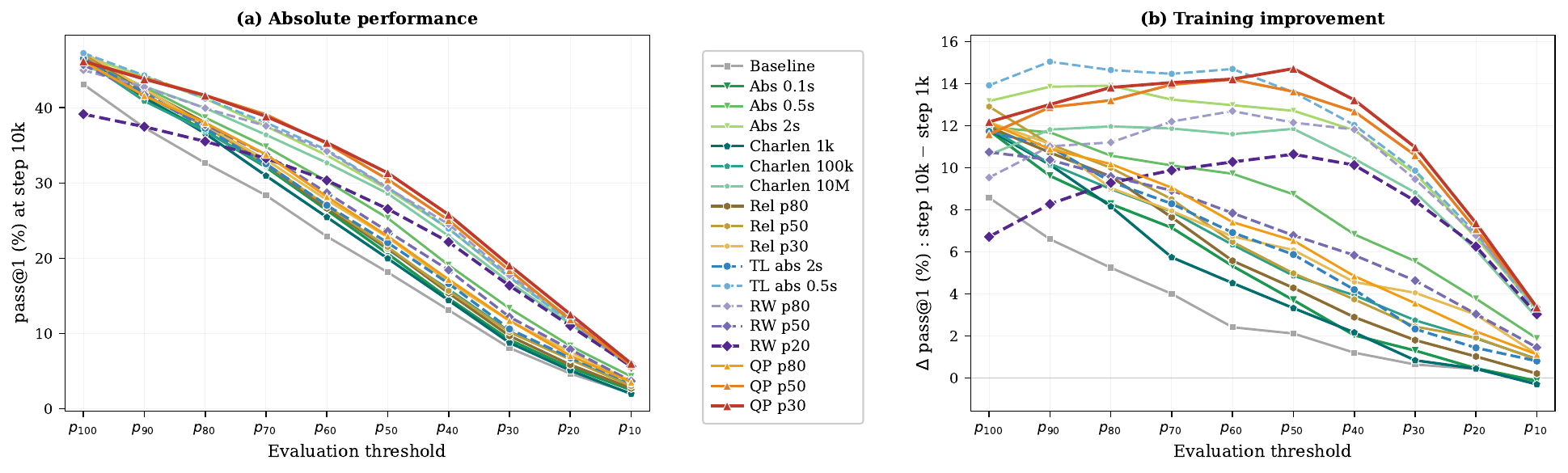}}
\par\vspace{0.2em}

\small
\noindent\textbf{Figure~\thefigure: Absolute and training-gain profiles by evaluation threshold.}
\emph{Left:} pass@$1$ at step 10k as the evaluation threshold tightens from \(p_{100}\) to \(p_{10}\).
\emph{Right:} \(\Delta\) pass@$1$ from step 1k to step 10k over the same thresholds.
Both panels separate mostly-correctness gains, which fade along the diagonal as the decile becomes stricter, from more balanced optimization gains that stay large deeper into the percentile sweep.
\end{figure}

\paragraph{How do the gains transfer across different optimization evaluation criteria?}
\label{sec:transfer}
In \cref{tab:qwen7b_environments}, the different training environments insert the optimization constraint in different ways, but are evaluated under a single evaluation family: leaderboard-percentile post-execution ranking, at \(p_{100}\), \(p_{50}\), \(p_{30}\), and \(p_{10}\).
\Cref{fig:improvement_profile} first generalizes this view across all deciles, comparing both absolute performance after 10k RL steps and relative improvement from 1k to 10k.
We use 1k rather than the shared SFT model as the reference point because the first 1{,}000 RL steps mostly account for learning the reasoning format, after which task-level exploration and exploitation become more visible.
Standard RLVR is clearly under-performing across deciles.
\par\vspace{0.35em}
The relative-gain panel separates two families of runs.
Some environments produce good gains in pure correctness and only partial transfer to stricter deciles, with transfer magnitude decreasing as the evaluation criterion tightens; these appear as diagonal curves in the right panel.
Other environments produce good gains in pure correctness and maintain, or even improve, these gains as the decile decreases, until the ranking criterion eventually becomes too strict and absolute performance falls.
These appear as more concave curves.
Among them, some trade pure correctness for optimization performance, visible as lower \(p_{100}\) starting points such as RW \(p_{20}\).
Others, such as QP \(p_{30}\)/\(p_{50}\) (per-test-percentile post-execution ranking) and Abs 2s (pre-execution absolute filtering), improve pure correctness as much as the correctness-oriented runs and keep that level, or improve it, through roughly the first half of decile values before the curve starts to diminish.
The \(p_{50}\) decile looks like an inflection point where most runs begin to suffer from the harder ranking criterion and transfer less.
We believe these curves provide a useful view of how balanced the learning is: in the ideal case, the absolute-performance curve would stay flat near the top as the criterion tightens.
Abs 0.5s appears to be in a transition regime, between runs that mostly learn correctness and runs with more balanced learning across correctness and optimization.
\par\vspace{0.35em}
\begin{figure}[!ht]
\refstepcounter{figure}
\label{fig:train_eval_heatmap}
\noindent\makebox[\linewidth][c]{\includegraphics[width=\textwidth]{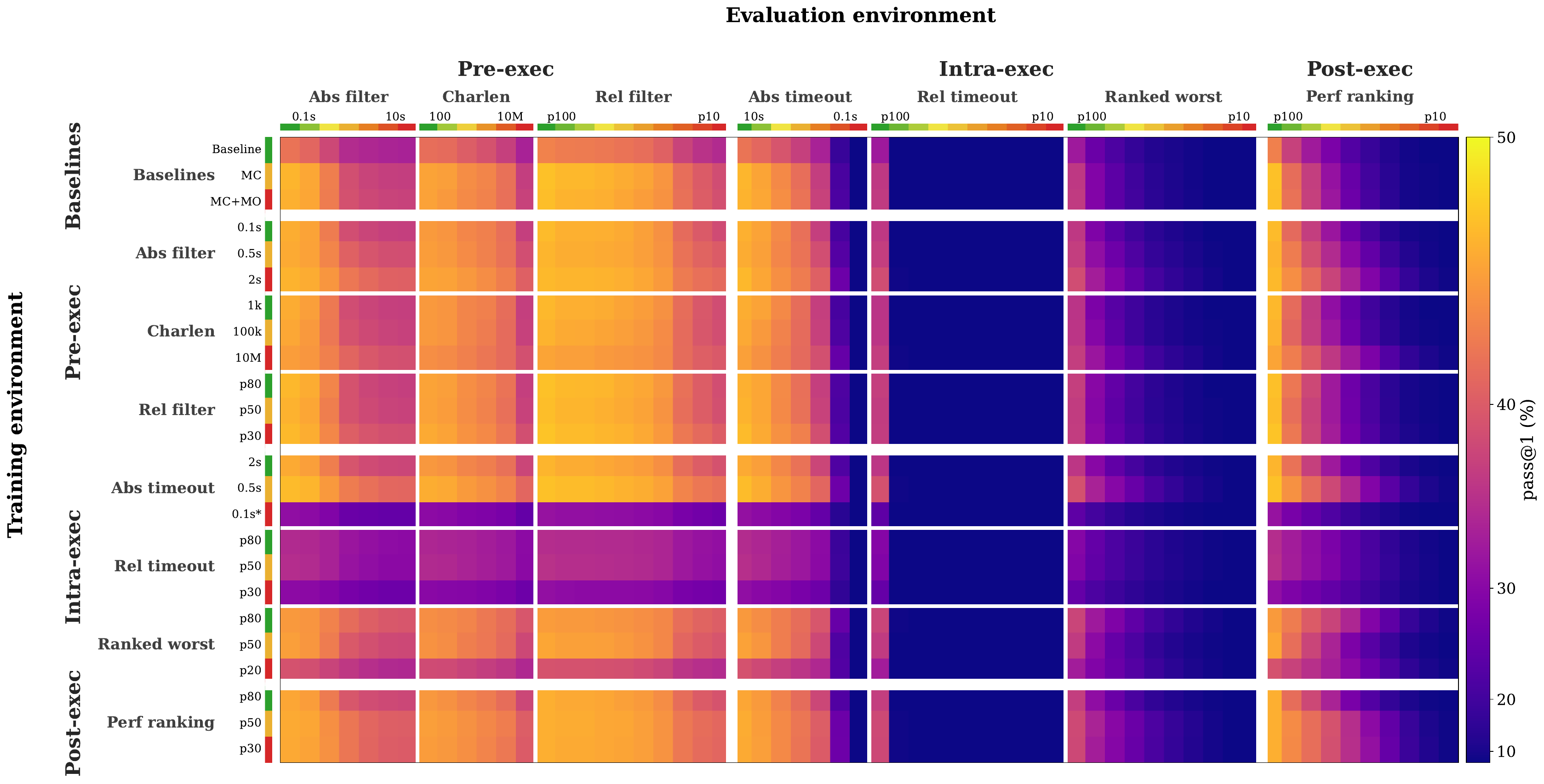}}
\par\vspace{0.2em}

\small
\noindent\textbf{Figure~\thefigure: Train-by-eval cross-evaluation for Qwen 2.5 7B optimization RL.}
Rows are training environments and columns are evaluation environments, grouped by pre-execution, intra-execution, and post-execution families; each reports pass@$1$ of a trained environment under an evaluation environment.
The row and column colored side bars mark the within-family strictness ordering from easier to harder settings.
Post-execution performance-ranking provides both the strongest discrimination across training environments at fixed parameterization, visible as large color variation within a column, and a broad range of difficulty as the parameterization changes, visible as color variation across each row.
\end{figure}
\Cref{fig:train_eval_heatmap} further generalizes this cross-evaluation across many evaluation environments and parameterizations, taking each training family from \cref{tab:qwen7b_environments} and covering a wider set of thresholds: all deciles for percentile-based environments, absolute timeouts from \(10\,\mathrm{s}\) to \(0.1\,\mathrm{s}\), and multiple pre-execution filters.
Because this produces a large number of scores, we summarize the full grid as a train-by-eval heatmap.
Across pre-execution, intra-execution, and post-execution evaluation criteria, a stable top group remains near the best score: QP \(p_{50}\) is within one pass@$1$ point of the best row in 58/60 evaluation columns, QP \(p_{30}\) in 52/60, Abs 2s in 52/60, and TL abs 0.5s in 54/60.
By contrast, RW \(p_{80}\), which looked competitive in \cref{tab:qwen7b_environments}, is more evaluator-dependent: it is within one point in 16/60 columns and within two points in 52/60.

The choice of evaluation criterion matters most for the middle of the leaderboard.
Strong configurations stay strong almost everywhere, while weaker configurations often look better under criteria close to their home training environment.
For example, RW \(p_{20}\) and relative-filter \(p_{30}\) exchange many leaderboard positions depending on whether we evaluate under ranked-worst intra-execution timeouts or relative pre-execution filtering.
Looking now at the evaluation criteria, the heatmap helps identify the ones that behave similarly: ranked-worst and post-execution ranking induce almost the same leaderboard over training configurations (\(\rho \simeq 0.995\)), explaining why RW \(p_{80}\) transfers reasonably well to the post-execution ranking used in \cref{tab:qwen7b_environments}; char-length and relative pre-execution filters behave similarly as well (\(\rho \simeq 0.975\)).

This cross-evaluation therefore helps identify what makes a useful optimization evaluator.
We want an evaluator that both discriminates training runs at a fixed parameter value and spans a broad range of difficulty as the parameter changes.
By this diagnostic, leaderboard-percentile post-execution ranking is the cleanest choice: its family-average spread across training rows is 13.5 pass@$1$ points, and its parameter sweep spans about 40 points from \(p_{100}\) to \(p_{10}\), justifying a posteriori why we use this evaluation criterion in \cref{tab:qwen7b_environments}.
Absolute, character-length, and relative pre-execution filters discriminate runs well, with spreads around 15.5--15.8 points, but their parameter sweeps are narrower, around 6--7 points.
Relative-timeout evaluation has the opposite failure mode: it spans 35.1 points as the threshold tightens, but only separates training rows by 2.3 points on average, so it mostly changes global difficulty without clearly ranking trained models.
Looking more closely, this is even worse, as the apparent difficulty spread mostly comes from a \(p_{100}\) setting that removes the optimization constraint and still evaluates correctness correctly; once the relative-timeout constraint is active, scores become flat and low across all parameterizations.
The same family is weak as a training environment: its three trained policies fall in the bottom five under every family-averaged evaluator, and across individual evaluation columns they are bottom-5 in 39/60, 40/60, and 50/60 columns for TL rel \(p_{80}\), TL rel \(p_{50}\), and TL rel \(p_{30}\), respectively.
Another example is TL abs 0.1s, which is a degenerate parameterization, but belongs to a family where other parameterizations perform well, among which TL abs 0.5s belongs to the stable top group.
Among that group, however, the cleanest leaderboard-percentile post-execution ranking criterion still favors QP \(p_{30}\), QP \(p_{50}\), and Abs 2s at stricter percentiles, while TL abs 0.5s falls slightly behind on this specific evaluation axis, elected as the most trustworthy.
For this reason, we mainly further explore the differences between the first three.

\paragraph{What reward shape maximizes concurrent gains in correctness and optimization?}
\label{sec:reward_shape}
\Cref{tab:qwen7b_reward_composition_main} takes the most promising environment configurations---Abs 2s pre-execution filtering and per-test-percentile \(p_{30}/p_{50}\) post-execution ranking---and explores, as introduced in \cref{sec:reward,sec:app_reward_composition}, several reward forms to balance the three shared signals produced by the environments: correctness, the optimization gate, and the graded optimization quality.
As shown in the table results, we observe that putting only an optimization objective leads to complete policy collapse, while the counterpart standard RLVR on the right tests keeps good correctness and already shows signs of improvement on optimization metrics.
Training on these two objectives jointly as a multitask RL run does not work much better, and fails to break the Pareto frontier of correctness--optimization.
Additive blending, an approach used in RL for other tasks, remains disappointing: with this reward shape, snippets of code that are fast but wrong can still receive partial reward during training, which trades a large amount of pure-correctness performance for decent but limited optimization gains.
Following \citet{zheng2026extrapolativeweightaveragingreveals}, one can argue that, in weight space, training for correctness alone and training for optimization alone can be linearly combined to obtain a policy close to one trained directly on the combined task.
Which in turn ends up navigating a correctness--optimization Pareto front, which is what we observe in practice for multitask and blending rewards.
With the nuance that the blended reward still includes both objectives in each advantage computation, which we believe helps explain why the resulting models sit slightly above the multitask variant, but we do not observe a major Pareto break for either of them.

\providecommand{\dash}{\textcolor{black!35}{--}}
\begin{table}[t!]
\refstepcounter{table}
\label{tab:qwen7b_reward_composition_main}
\normalsize
\noindent\textbf{Table~\thetable: Reward-composition sweep on Qwen 2.5 7B.}
DMC-Optim test pass@$1$ after 10k of RL training.
Columns compare training with the 2\,s pre-execution absolute filter with the post-execution per-test percentile environment, each evaluated with a post-execution ranking environment.
For the pre-execution environment, the binary optimization signal uses a 10\% timeout-tolerance gate, while non-binary variants use the proportion of timeouts.
For the post-execution environment, the binary optimization signal uses the \(p_{30}\) percentile gate, while non-binary variants use the percentile ranking itself.
The continuous 50/50, 66/33, and 75/25 rows are two-gate ranked variants that set the reward range of correct solutions to respectively \([0,1]\), \([0.33,1]\), and \([0.5,1]\), while keeping incorrect-solution reward at \(-1\).
\par\vspace{0.5em}

\centering
\small
\renewcommand{\arraystretch}{0.92}
\setlength{\tabcolsep}{2.2pt}
\begin{tabular*}{\textwidth}{@{\extracolsep{\fill}}l rrrr rrrr@{}}
\toprule
\bfseries Training config. & \multicolumn{4}{c}{\bfseries Pre-exec abs. filter 2\,s} & \multicolumn{4}{c}{\bfseries Post-exec per-test percentile} \\
\cmidrule(lr){2-5} \cmidrule(lr){6-9}
\bfseries Eval. percentile & \bfseries $p_{100}$ & \bfseries $p_{50}$ & \bfseries $p_{30}$ & \bfseries $p_{10}$ & \bfseries $p_{100}$ & \bfseries $p_{50}$ & \bfseries $p_{30}$ & \bfseries $p_{10}$ \\
\midrule
\multicolumn{9}{@{}l}{\textsc{Collapsed reward}} \\[1pt]
\quad Binary & 46.9 & 29.2 & 17.4 & 5.7 & 46.2 & 31.3 & 19.1 & 6.0 \\
\quad Bucketed & 46.2 & 24.8 & 13.6 & 4.3 & 45.6 & 30.1 & 18.6 & 5.7 \\
\midrule
\multicolumn{9}{@{}l}{\textsc{Two-gate reward}} \\[1pt]
\quad Binary & 45.7 & 26.1 & 14.9 & 4.9 & 47.6 & 29.8 & 17.8 & 5.8 \\
\quad Bucketed & 46.6 & 22.1 & 10.6 & 3.4 & 47.1 & 27.5 & 15.5 & 4.9 \\
\quad Continuous, 50/50 split & \dash & \dash & \dash & \dash & 47.6 & 24.3 & 12.7 & 4.2 \\
\quad Continuous, 66/33 split & \dash & \dash & \dash & \dash & 47.1 & 21.9 & 10.7 & 3.3 \\
\quad Continuous, 75/25 split & \dash & \dash & \dash & \dash & 46.4 & 21.4 & 10.4 & 3.2 \\
\midrule
\multicolumn{9}{@{}l}{\textsc{Additive blend}} \\[1pt]
\quad Binary & 39.1 & 24.4 & 13.7 & 5.2 & 35.8 & 25.4 & 15.1 & 5.2 \\
\quad Bucketed & 42.2 & 21.9 & 11.3 & 3.6 & 39.5 & 25.4 & 15.0 & 5.0 \\
\midrule
\multicolumn{9}{@{}l}{\textsc{Optimization only}} \\[1pt]
\quad Binary & 0.0 & 0.0 & 0.0 & 0.0 & 0.0 & 0.0 & 0.0 & 0.0 \\
\quad Bucketed & 0.0 & 0.0 & 0.0 & 0.0 & 0.0 & 0.0 & 0.0 & 0.0 \\
\midrule
\multicolumn{9}{@{}l}{\textsc{Multitask}} \\[1pt]
\quad Binary & 38.0 & 22.7 & 12.5 & 4.4 & 30.3 & 21.8 & 13.6 & 4.8 \\
\quad Bucketed & 41.2 & 20.5 & 10.2 & 3.4 & 39.8 & 26.7 & 16.0 & 5.5 \\
\bottomrule
\end{tabular*}
\end{table}

Our experiments underline two reward formulations that can break this front: either a gating objective, or a single collapsed objective.
For two-gate rewards, we explored, beyond the binary and bucketed formulations, some continuous equivalents, where the binary correctness signal assigns a forced \(-1\) reward to incorrect code and, when the code is correct, allows a continuous optimization signal in the range 0--1.
The continuous signal hurts optimization performance relative to bucketed rewards, which we believe is because bucketing filters out part of the timing-measurement noise before it reaches the gradient steps.
Reducing the continuous optimization-based credit range to higher values does not further boost correctness either, and instead hurts lower-percentile scores.

Finally, between bucketed and binary rewards, across two-gate and collapsed formulations and across the different optimization RL environments, the binary approach consistently improves all percentiles (exception to be made for the \(p_{100}\) bucketed pre-execution Abs 2s cell, but it performs substantially worse at stricter percentiles).
This joins the classic DeepSeek-R1 observation that favoring simple binary rewards can help avoid hidden reward hacking~\citep{Guo_2025}.
It also justifies a posteriori the choice made in \cref{tab:qwen7b_environments} to mainly report binary collapsed reward variants of each RL environment.
We keep these rewards as the main variants below, while tracking two-gate and bucketed variants when their scores remain close to the binary ones.

\paragraph{If we change the model and/or scale it up, does it keep working? And if evaluated on OOD data?}
\label{sec:cross_model}
\Cref{tab:cross_model} reports cross-model DMC-Optim transfer, using the post-execution leaderboard-percentile evaluation environment, and \cref{tab:lcb_transfer} reports the same models on LiveCodeBench (LCB).
As introduced above and further explored in \cref{sec:app_empirical,sec:app_simulation}, LCB tests are too short for stable timeout-style optimization evaluation, so win rates against each model-family standard-RLVR baseline are better suited to measure optimization gains.

\providecommand{\dash}{\textcolor{black!35}{--}}
\begin{table}[t!]
\refstepcounter{table}
\label{tab:cross_model}
\normalsize
\noindent\textbf{Table~\thetable: Cross-model DMC-Optim Scores.}
DMC-Optim test pass@$1$ under various evaluation percentiles.
Columns report \(p_{100}\), \(p_{50}\), \(p_{30}\), and \(p_{10}\) for each model, Qwen~2.5~7B, CWM~32B, and Qwen~2.5~32B; \dash{} indicates that a configuration was not run for that model.
Bold marks the best displayed value within each model and evaluation threshold.
We used group-averages when possible, and further study variability at the end of this section.
\par\vspace{0.5em}

\centering
\small
\setlength{\tabcolsep}{1.35pt}
\begin{tabular*}{\textwidth}{@{\extracolsep{\fill}}l rrrr rrrr rrrr@{}}
\toprule
& \multicolumn{4}{c}{\bfseries Qwen 2.5 7B} & \multicolumn{4}{c}{\bfseries Qwen 2.5 32B} & \multicolumn{4}{c}{\bfseries CWM 32B} \\
\cmidrule(lr){2-5} \cmidrule(lr){6-9} \cmidrule(lr){10-13}
\bfseries Configuration & \bfseries $p_{100}$ & \bfseries $p_{50}$ & \bfseries $p_{30}$ & \bfseries $p_{10}$ & \bfseries $p_{100}$ & \bfseries $p_{50}$ & \bfseries $p_{30}$ & \bfseries $p_{10}$ & \bfseries $p_{100}$ & \bfseries $p_{50}$ & \bfseries $p_{30}$ & \bfseries $p_{10}$ \\
\midrule
\multicolumn{13}{@{}l}{\textsc{Standard RLVR}} \\[1pt]
\quad Standard RLVR & 43.5 & 18.0 & 7.7 & 1.9 & 54.8 & 21.1 & 9.4 & 2.4 & 69.8 & 30.7 & 13.7 & 3.3 \\
\midrule
\multicolumn{13}{@{}l}{\textsc{+ Optimization/More Correctness Tests}} \\[1pt]
\quad MC & 46.7 & 19.8 & 8.7 & 2.4 & 56.4 & 24.6 & 11.3 & 3.0 & \textbf{71.5} & 35.5 & 17.6 & 5.0 \\
\quad MC + 10s timeout & 46.4 & 19.8 & 8.8 & 2.3 & 59.4 & 26.5 & 12.7 & 3.6 & 71.1 & 34.2 & 15.7 & 4.1 \\
\quad MC + opt. tests & 46.9 & 20.6 & 9.3 & 2.7 & \textbf{60.0} & 28.7 & 14.2 & 4.4 & \textbf{71.5} & 35.6 & 17.1 & 4.7 \\
\midrule
\multicolumn{13}{@{}l}{\textsc{+ Reward: Pre-execution} (test filtering)} \\[1pt]
\quad \textit{Absolute duration filter} & & & & & & & & & & & & \\
\quad\quad $\tau=0.5$\,s & 46.5 & 25.3 & 13.3 & 4.3 & \dash & \dash & \dash & \dash & 70.9 & 48.3 & 28.6 & 8.7 \\
\quad\quad $\tau=2$\,s & 46.9 & 29.2 & 17.4 & 5.7 & 48.2 & 31.2 & 19.2 & 6.3 & 70.5 & 49.1 & 30.1 & 9.5 \\
\midrule
\multicolumn{13}{@{}l}{\textsc{+ Reward: Intra-execution} (timeout-based)} \\[1pt]
\quad \textit{Ranked-worst timeout} & & & & & & & & & & & & \\
\quad\quad from $p_{80}$ ranked & 45.0 & 29.3 & 17.7 & \textbf{6.0} & 54.4 & 35.4 & 21.3 & 6.9 & 69.9 & 49.7 & 30.6 & \textbf{9.8} \\
\midrule
\multicolumn{13}{@{}l}{\textsc{+ Reward: Post-execution} (ranking-based)} \\[1pt]
\quad \textit{Leaderboard percentile} & & & & & & & & & & & & \\
\quad\quad Two-gate bucketed & \textbf{47.1} & 27.5 & 15.5 & 5.1 & 59.3 & 37.6 & 22.8 & 7.5 & 71.2 & 49.0 & 29.9 & 9.3 \\
\quad \textit{Per-test percentile} & & & & & & & & & & & & \\
\quad\quad top 50\% & 46.2 & 30.5 & 18.4 & 5.7 & 55.5 & 39.2 & \textbf{24.2} & 8.3 & 71.4 & 48.9 & 29.4 & 9.4 \\
\quad\quad top 30\% & 46.2 & \textbf{31.3} & \textbf{19.1} & \textbf{6.0} & 55.7 & \textbf{39.6} & \textbf{24.2} & \textbf{8.5} & 70.2 & \textbf{50.4} & \textbf{30.9} & \textbf{9.8} \\
\quad\quad Two-gate bucketed & 46.5 & 25.4 & 13.8 & 4.4 & 59.6 & 36.3 & 21.8 & 6.8 & 70.1 & 46.6 & 28.0 & 8.6 \\
\bottomrule
\end{tabular*}
\end{table}

\noindent Trying Qwen 2.5 32B and CWM 32B with the optimization RL environments and rewards that worked best on Qwen 2.5 7B preserves most of the gains, showing that the enhanced optimization-RL training scales beyond the original Qwen 2.5 7B setting.
On DMC-Optim, the larger models show the same pattern as Qwen 2.5 7B: with top-30\% per-test-percentile ranking, Qwen 2.5 32B has a small pure-correctness improvement and much larger gains at stricter thresholds, with \(p_{50}\) rising by 88\%, \(p_{30}\) by 157\%, and \(p_{10}\) by 254\% over standard RLVR (though it is worth mentioning that the \(p_{10}\) relative gain is almost unfair as the baseline RLVR can simply not solve this criterion).
CWM 32B with the same environment family also keeps pure-correctness stable and improves \(p_{50}\) by 64\% and \(p_{30}\) by 126\%.
The ranking of environment families is coherent across scales: we do not observe a configuration that works at Qwen 2.5 7B and collapses at larger scale, or a weak small-scale configuration that suddenly becomes the best larger-scale run.
We also recover the Qwen 2.5 7B data-improvement results: better data quality and the added tests improve performance, but an optimization RL environment with the right reward extracts significant additional strict-percentile gains on top.

\providecommand{\dash}{\textcolor{black!35}{--}}
\begin{table}[t!]
\refstepcounter{table}
\label{tab:lcb_transfer}
\normalsize
\noindent\textbf{Table~\thetable: Cross-model LCB transfer.}
LCB pass@$1$ and speed win rates versus each model-family standard RLVR baseline.
WR$_b$/WR$_m$ use the fastest/median passing sample and exclude draws.
Rows mirror the DMC-Optim best configurations in \cref{tab:cross_model}, where available; \dash{} indicates that a configuration was not run for that model.
Bold marks the best displayed value within each model and metric.
\par\vspace{0.5em}

\centering
\small
\renewcommand{\arraystretch}{0.92}
\setlength{\tabcolsep}{1.7pt}
\begin{tabular*}{\textwidth}{@{\extracolsep{\fill}}l rrr rrr rrr@{}}
\toprule
& \multicolumn{3}{c}{\bfseries Qwen 2.5 7B} & \multicolumn{3}{c}{\bfseries Qwen 2.5 32B} & \multicolumn{3}{c}{\bfseries CWM 32B} \\
\cmidrule(lr){2-4} \cmidrule(lr){5-7} \cmidrule(lr){8-10}
\bfseries Config. & \bfseries p@1 & \bfseries WR$_b$ & \bfseries WR$_m$ & \bfseries p@1 & \bfseries WR$_b$ & \bfseries WR$_m$ & \bfseries p@1 & \bfseries WR$_b$ & \bfseries WR$_m$ \\
\midrule
\multicolumn{10}{@{}l}{\textsc{Standard RLVR}} \\[1pt]
\quad Standard RLVR & \textbf{44.2} & \dash & \dash & \textbf{52.7} & \dash & \dash & \textbf{56.4} & \dash & \dash \\
\midrule
\multicolumn{10}{@{}l}{\textsc{+ Optimization/More Correctness Tests}} \\[1pt]
\quad MC & 42.6 & 51.2 & 57.7 & 50.7 & 51.5 & 68.6 & 55.3 & 62.1 & 73.0 \\
\quad MC + 10s timeout & 43.1 & 56.6 & 57.3 & 51.5 & 53.7 & 71.2 & 54.7 & 59.3 & 68.4 \\
\quad MC + opt. tests & 42.1 & 55.4 & 62.1 & 51.9 & 52.3 & 72.5 & 55.2 & 60.4 & 69.3 \\
\midrule
\multicolumn{10}{@{}l}{\textsc{+ Reward: Pre-execution} (test filtering)} \\[1pt]
\quad $\tau=0.5$\,s & 41.7 & 62.2 & 65.3 & \dash & \dash & \dash & 54.4 & 64.0 & \textbf{83.0} \\
\quad $\tau=2$\,s & 41.1 & \textbf{63.2} & 69.5 & 45.6 & 57.7 & 70.9 & 54.1 & 63.0 & 78.0 \\
\midrule
\multicolumn{10}{@{}l}{\textsc{+ Reward: Intra-execution} (timeout-based)} \\[1pt]
\quad from $p_{80}$ ranked & 39.7 & 59.5 & 69.5 & 49.4 & 57.4 & 76.7 & 53.2 & 63.7 & 78.5 \\
\midrule
\multicolumn{10}{@{}l}{\textsc{+ Reward: Post-execution} (ranking-based)} \\[1pt]
\multicolumn{10}{@{}l}{\quad \textit{Leaderboard percentile}} \\[1pt]
\quad\quad Two-gate bucketed & 41.9 & 62.7 & 70.7 & 52.3 & 61.4 & 74.6 & 54.0 & 61.7 & 79.3 \\
\multicolumn{10}{@{}l}{\quad \textit{Per-test percentile}} \\[1pt]
\quad\quad top 50\% & 41.1 & 61.4 & \textbf{73.0} & 48.7 & 52.5 & 73.4 & 54.2 & \textbf{68.1} & 80.7 \\
\quad\quad top 30\% & 40.6 & 60.3 & 68.5 & 49.1 & 61.9 & 72.9 & 53.9 & 66.3 & 82.9 \\
\quad\quad Two-gate bucketed & 41.6 & 59.8 & 65.0 & 52.4 & \textbf{66.7} & \textbf{77.8} & 53.8 & 63.5 & 76.6 \\
\bottomrule
\end{tabular*}
\end{table}

\Cref{tab:lcb_transfer} gives the out-of-distribution LCB results.
On this benchmark, per-test-percentile post-execution training, especially \(p_{30}\) and \(p_{50}\), gives large median-sample speed win rates across models, from 68.5\% for Qwen 2.5 7B with \(p_{30}\) training up to 82.9\% for CWM 32B with \(p_{30}\) training as well.
Under our WR$_m$ definition, this means that if we sample 20 solutions from the standard-RLVR model and 20 from the optimization-trained model, keep the median passing sample by speed for each side, and compare them on the same problem, the optimization-trained CWM 32B sample is faster in 82.9\% of comparisons.
WR$_b$ instead asks whether the best passing sample among 20 is faster, which gives the standard-RLVR baseline an oracle over its own batch and captures cases where it occasionally samples a fast solution (the same oracle is given to the optimization-trained model, but we expect it to benefit the base model more, since even with less consistency over its generations it can randomly sample a good optimization trick from time to time).
Even under this stricter best-sample comparison, CWM 32B reaches 66.3--68.1\% against standard RLVR for the \(p_{30}/p_{50}\) post-execution rows.
This suggests that inference-time search over a standard-RLVR model can recover some speed, but it does not fully recover the faster solution modes and new optimization tricks reached by optimization RL; and the so-trained models have more direct control over the efficiency of each of their generations.

\providecommand{\dash}{\textcolor{black!35}{--}}
\begin{table}[t!]
\refstepcounter{table}
\label{tab:cwm32b_reward_composition_main}
\normalsize
\noindent\textbf{Table~\thetable: Reward-composition sweep on CWM 32B.}
DMC-Optim test pass@$1$ at step 10k.
Columns compare the same 2\,s pre-execution absolute filter and post-execution per-test percentile environments as in \cref{tab:qwen7b_reward_composition_main}.
The table includes the reward-composition cells that were also run for Qwen 2.5 7B.
\par\vspace{0.5em}

\centering
\small
\renewcommand{\arraystretch}{0.92}
\setlength{\tabcolsep}{2.2pt}
\begin{tabular*}{\textwidth}{@{\extracolsep{\fill}}l rrrr rrrr@{}}
\toprule
& \multicolumn{4}{c}{\bfseries Pre-exec abs. filter 2\,s} & \multicolumn{4}{c}{\bfseries Post-exec per-test percentile} \\
\cmidrule(lr){2-5} \cmidrule(lr){6-9}
\bfseries Reward variant & \bfseries $p_{100}$ & \bfseries $p_{50}$ & \bfseries $p_{30}$ & \bfseries $p_{10}$ & \bfseries $p_{100}$ & \bfseries $p_{50}$ & \bfseries $p_{30}$ & \bfseries $p_{10}$ \\
\midrule
\multicolumn{9}{@{}l}{\textsc{Collapsed reward}} \\[1pt]
\quad Binary & 70.5 & 49.1 & 30.1 & 9.5 & 70.2 & 50.4 & 30.9 & 9.8 \\
\quad Bucketed & 71.3 & 46.7 & 27.8 & 8.5 & 70.2 & 49.3 & 29.8 & 9.7 \\
\midrule
\multicolumn{9}{@{}l}{\textsc{Two-gate reward}} \\[1pt]
\quad Binary & 70.5 & 47.2 & 28.7 & 9.0 & 70.4 & 48.2 & 29.1 & 9.1 \\
\quad Bucketed & 70.8 & 40.0 & 21.7 & 6.8 & 70.1 & 46.6 & 28.0 & 8.6 \\
\midrule
\multicolumn{9}{@{}l}{\textsc{Additive blend}} \\[1pt]
\quad Binary & 68.7 & 48.5 & 29.5 & 9.2 & 68.3 & 49.9 & 31.0 & 10.0 \\
\quad Bucketed & 69.2 & 45.3 & 27.4 & 8.8 & 69.4 & 48.1 & 29.0 & 8.8 \\
\bottomrule
\end{tabular*}
\end{table}

Concerning the LCB pure-correctness scores, for the \(p_{30}\) post-execution row, pass@$1$ decreases by 3.6 points on Qwen 2.5 7B, or about 8\%, which reduces to only a 4\% decrease as we scale it up to CWM 32B.
We believe this degradation is partly due to LCB hard problems, which are much harder than DMC-Optim, and so the task of solving them can conflict with producing optimized code at the same time.
It also reflects a generous evaluation setting much in favor of poorly optimized but correct solutions: for the LCB correctness numbers, we remove timeouts on a dataset whose tests are too short anyway to measure optimization differences, so pass@$k$ does not reward any optimization trick whatsoever as long as the solution remains correct.
We did not rebuild LCB tests as we did for DMC-Optim, and a version with time-limit constraints closer to online programming contests would be substantially more favorable to optimization-trained models.
The corresponding CWM 32B pass@$10$ scores also support that pure-correctness capabilities of optimization-trained models are not really degraded: standard RLVR reaches 66.3, while the top-30\% and top-50\% post-execution rows reach 65.6 and 66.3.
The model's pure solving capacity is therefore pretty much equivalent to standard RLVR at pass@$10$, even though some of it is not converted to pass@$1$ yet; further RL training can help mitigate this by translating more of that capacity back into single-sample correctness.

As a side note, the experimental setup at the beginning of this section mentioned that CWM 32B has seen DMC problem statements during SFT, unlike the Qwen 2.5 variants that were decontaminated for their SFT against the DMC-Optim train and test prompts (though the CWM 32B exposure was through generated solutions rather than the human references used to assign evaluation percentile scores).
The standard-RLVR rows show that the model did not particularly benefit from this possible exposure: without a correctly defined RL environment with an optimization constraint, the model does not seem to learn to search its knowledge space inherited from SFT for the most optimized existing solution.
Finally, \cref{tab:cwm32b_reward_composition_main} also reports CWM 32B reward-composition variants on the most successful environments.
It confirms the Qwen 2.5 7B finding that collapsed binary rewards are the best balanced choice, although the CWM 32B differences are smaller and additive binary rewards become competitive in the strictest cells, while still giving up more pure-correctness score.

\paragraph{Do we have a fair experiment setup to make these claims, or are we cheating?}
\label{sec:compute_fairness}
All runs above, whether standard RLVR or one of the optimization-aware RL environments, train on the same underlying DMC-Optim seed set of roughly 1{,}000 prompts and use the same 10{,}000 optimizer-step budget.
This is a small prompt budget compared with larger code-RL runs: the CWM 32B report uses 81{,}000 code-problem prompts for RL~\citep{cwm2025}.
We therefore read these experiments as a controlled small-scale comparison rather than as a saturated training recipe.
Given the effect sizes observed here, we expect larger prompt pools to help optimization RL further: the main claims are not 5\% gains that might need scale to become visible, but roughly 100\% relative improvements, or more, at stricter percentiles already in this limited setup.

The training dynamics point in the same direction.
For Qwen 2.5 32B, the standard-RLVR baseline starts after 1k steps at 49.2\% \(p_{100}\) pass@$1$ on DMC-Optim test and finishes at 54.8\%, whereas the top-30\% per-test-percentile run starts lower, at 46.1\%, but catches up within 10k steps and reaches 55.7\%.
At stricter optimization thresholds, standard RLVR simply does not learn optimization capabilities: \(p_{50}\) pass@$1$ moves from 22.2\% after 1k steps to 21.1\% after 10k, and \(p_{30}\) from 10.7\% to 9.4\%.
This suggests that pure correctness RL may even forget some occasional efficient samples inherited from the more diverse post-SFT model.
By contrast, the top-30\% optimization run moves from 23.4\% to 39.6\% at \(p_{50}\), with the same direction at stricter percentiles.
We observe the same findings with the other models.

The equal-step comparison controls the trainer-side GPU budget: runs perform the same number of optimizer updates on similarly sized batches.
The worker side differs more.
Standard RLVR on base tests makes no optimization-test CES calls; RLVR with the generated optimization tests adds those calls and makes it harder to get a positive reward; the top-30\% optimization environment uses the same CES calls, adds a small amount of CPU-side reward computation, and makes the reward again a bit sparser.
On Qwen 2.5 7B, standard RLVR skips 25\% of batches because they have zero advantage; adding optimization tests raises this by 3 points, and the top-30\% environment adds another 4 points.
On Qwen 2.5 32B, the skipped-batch rate increases by only 9\% relative, and on CWM 32B it decreases by 10\%.
The latter happens because larger models, which are more performant in the first place, end up making many baseline prompts all-success groups; optimization RL can turn some of those otherwise too-easy prompts back into useful training signal, reducing wasted generated rollouts.

In rollout generation time, the added CES calls account for about 15\% overhead relative to standard RLVR on base tests, but the top-30\% optimization environment is on par with RLVR on the same optimization tests: the remaining optimization RL-specific operations are CPU-side scalar computations that are negligible in wall-clock time.

Putting workers and trainers together and looking at the full RL runs, moving from base-test RLVR to optimization-test RLVR costs about 25\% average wall-clock training time, while the difference between top-30\% optimization RL and optimization-test RLVR stays within 5\%.
If we instead force equal wall-clock by stopping optimization RL earlier, the conclusion remains: stopping the Qwen 2.5 7B top-30\% run at 7k steps and comparing it with standard RLVR at 10k keeps \(p_{100}\) pass@$1$ on par, and still improves \(p_{30}\) pass@$1$ from 7.7\% to 17.6\%, about a 130\% relative gain (and similar observations are made with the other models).

\begin{figure*}[t!]
\refstepcounter{figure}
\label{fig:alphabeta_sweep}
\begin{minipage}[t]{0.49\textwidth}
\centering
\includegraphics[width=0.88\linewidth]{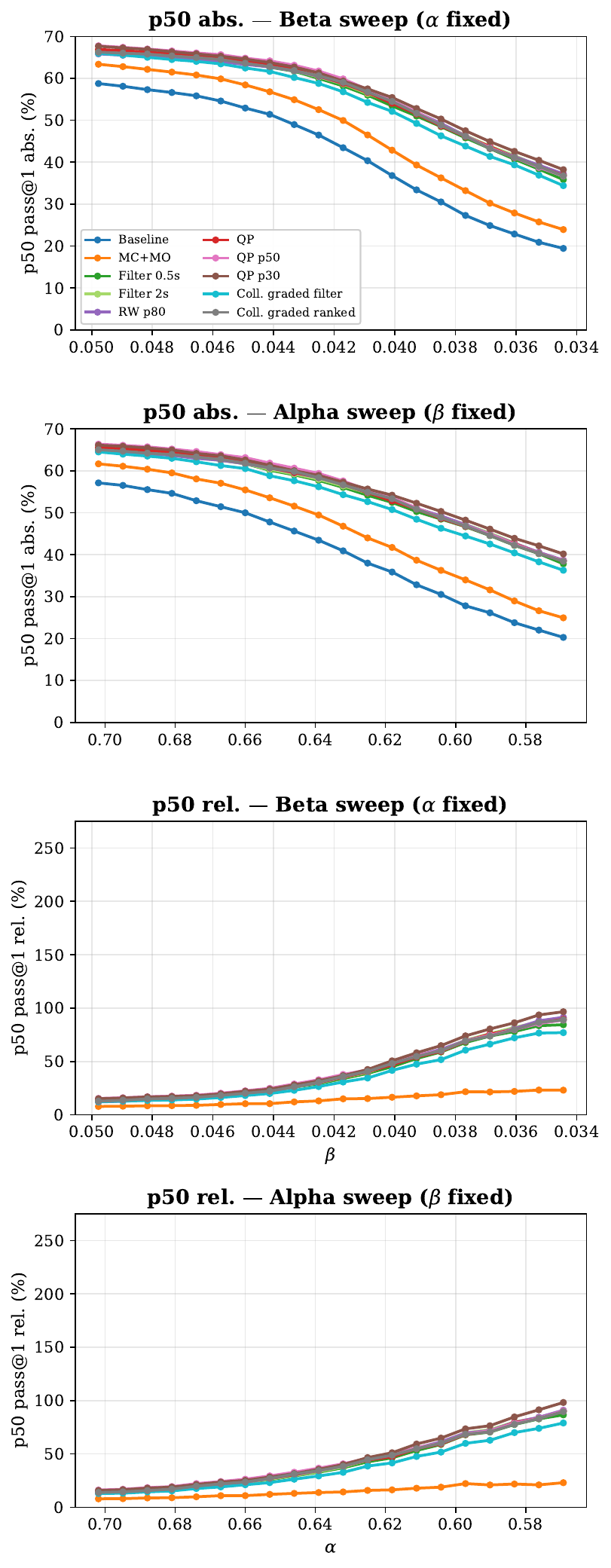}
\end{minipage}\hfill
\begin{minipage}[t]{0.49\textwidth}
\centering
\includegraphics[width=0.88\linewidth]{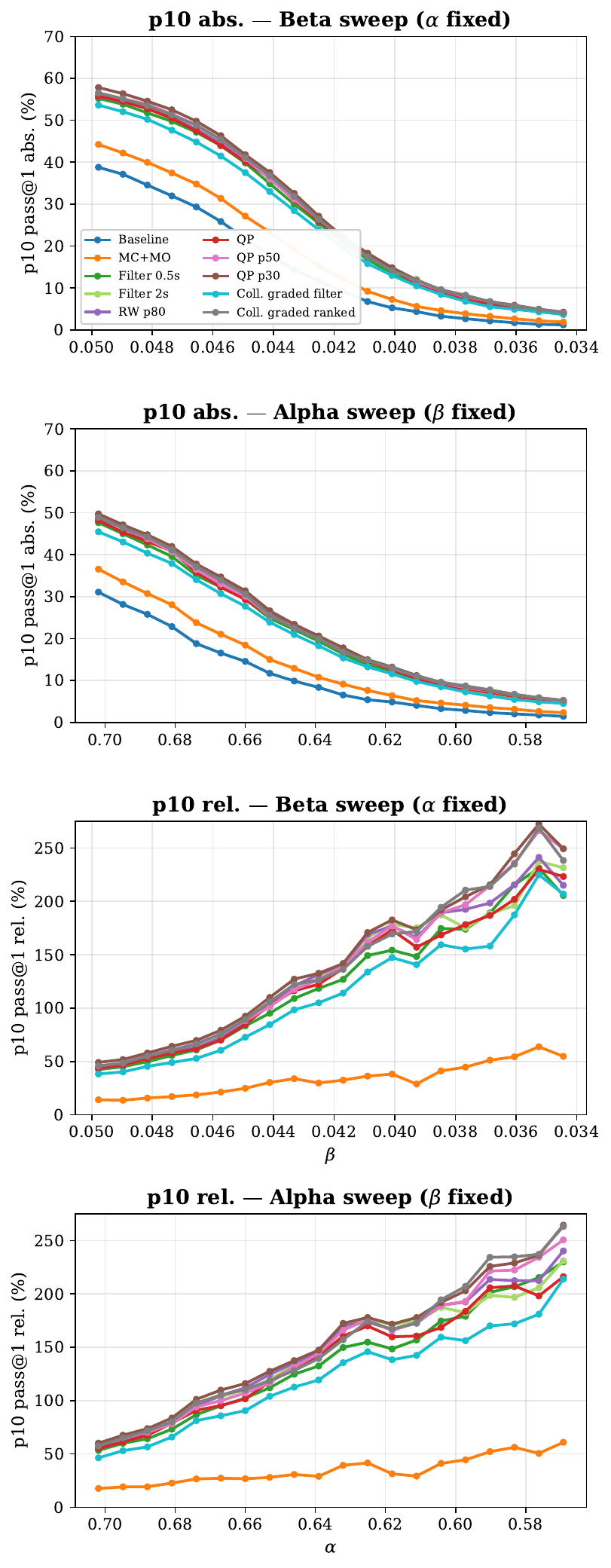}
\end{minipage}
\par\vspace{0.2em}

\small
\noindent\textbf{Figure~\thefigure: CWM 32B trained models under different evaluation-time sandbox states, from ``cheater'' fast sandbox to degraded slow sandbox.}
\emph{Left:} pass@$1$ and relative improvement at \(p_{50}\) under affine calibration shifts applied to recorded durations.
\emph{Right:} the same sweep at \(p_{10}\).
The left side of each x-axis corresponds to a faster timing environment, while the right side corresponds to a slower one, with the human-reference leaderboard kept fixed.
Across this timing sweep, base RLVR remains below optimization-test RLVR, and both stay below the top-performing optimization RL runs.
Optimization-RL gains remain positive even when the sandbox is made substantially faster, and are especially significant at stricter percentiles like \(p_{10}\).
When the sandbox slows down, gains increase significantly.
In practice, the sweep applies \(d'_{\mathrm{ref}}=\alpha d_{\mathrm{ref}}+\beta\) to stored reference durations while keeping generated-solution recorded durations fixed, so larger \(\alpha\) and/or \(\beta\) make submitted LLM solutions look faster relative to the leaderboard.
\end{figure*}

\paragraph{What happens if the timing environment slows down/speeds up?}
\label{sec:calibration_sensitivity}
\Cref{fig:alphabeta_sweep} varies the affine calibration applied to sandbox-recorded durations, simulating what happens when execution timing shifts at evaluation time.
For a fixed leaderboard of reference solutions into which an LLM candidate solution is inserted, changing the calibrated durations changes the output score.
At the beginning of this section, we introduced an evaluation setup that uses human reference solutions to calibrate recorded durations at evaluation time; here we ask what happens if we let go of that calibration, and whether the conclusions change.
Looking at \cref{fig:alphabeta_sweep}, the ordering of runs is mostly preserved across the sweep: base RLVR consistently performs worse than RLVR augmented with optimization tests, which itself remains below the top-performing optimization RL environments selected above.

On all subplots, values on the left side of the x-axis correspond to faster timing environments, and values on the right correspond to slower environments.
As the evaluation-time sandbox degrades and measures slower durations, the relative advantage of optimization RL over base RLVR grows faster than linearly, whereas optimization-test RLVR improves only at a slower, roughly linear pace.
Conversely, when the sandbox is made much faster, many solutions from base RLVR are classified as top submissions on the reference leaderboards even without a clear optimization trick.
Even then, the curves reach an asymptotic regime where the top-30\% trained model maintains about 7\% improvement over optimization-test RLVR and 15\% over base RLVR at \(p_{50}\), and roughly 30--35\% and 50--60\% respectively at \(p_{10}\).
This confirms that even when we cheat the sandbox by making execution substantially faster without updating the leaderboard, optimization RL keeps added value.
In the tables reported above, our evaluation setup tries to estimate the ``true'' timing values, so the reported numbers reflect the gains one could expect on average by submitting the generated solutions to an online code competition platform for instance.

\paragraph{Are these gains just noise?}
\label{sec:uncertainty_estimates}
The previous results show clear gains for the best optimization RL runs over standard RLVR.
Nevertheless, even with our data, sandbox, RL-environment, GRPO, and evaluation contributions we described above, these trainings and evaluation tasks remain particularly prone to noise.
We therefore try to estimate how much variability can appear during training and during evaluation.
\Cref{tab:variability_intervals} reports training-time and evaluation-time 95\% confidence-interval half-widths, to give an order of magnitude of the noise one can expect around the reported scores.

During training, even when we try to fix random seeds, train on the same prompt set, and preserve the same prompt ordering, the async RL setup remains strongly stochastic: rollout sampling, zero-advantage batch dropping, CES timing measures, among else, can all differ across launches.
On the evaluation side, variability comes from temperature sampling, which is partly controlled by sampling many candidates and using the unbiased pass@$k$ estimator of \citet{chen2021evaluatinglargelanguagemodels}, and more importantly from the measured durations used to assign optimization scores and from the finite DMC-Optim test set itself.
Most previous tables report averaged scores over multiple training runs and evaluations when available; here we report CIs for single training-run reported scores, so averaged scores would have less variance than these CIs, but this gives a good worst-case estimate of having a single run.

Training-time rows in \cref{tab:variability_intervals} pool the within-recipe deviations of repeated launches, combining them across all training recipes to compute a single, more robust variance estimate and shared CI.
Test-time rows use a classic bootstrapping approach of the 302 DMC-Optim problems: we resample problems with replacement and recompute the aggregated scores.
These quantities can be used differently depending on the comparison we are trying to make.
If we compare two runs on a fixed test set and fixed evaluation round, the training-time absolute half-width from \cref{tab:variability_intervals} multiplied by \(\sqrt{2}\) gives the half-width of the difference CI; therefore, if the two models' score difference exceeds that value, it is significant at 95\%.
If the checkpoints are fixed and we want to ask how they compare on a general distribution of DMC-like problems, to be able to say ``checkpoint A is better than checkpoint B on general DMC-like problems'', the evaluation-time interval is the relevant scale.
If we want to know whether one training method is better than another in general, a conservative approach is to combine both CIs.
Our goal remains to simply give an order of magnitude of the variations while staying pessimistic: we do not compute paired bootstrap intervals, nor one-sided tests that would assume in advance which run should be better.

\par\vspace{0.2em}
\begin{table}[t!]
\refstepcounter{table}
\label{tab:variability_intervals}
\normalsize
\noindent\textbf{Table~\thetable: Training-time and test-time uncertainty for DMC-Optim evaluation.}
Values are 95\% confidence-interval half-widths in pass@$1$ points for a single training run's score.
Training-time rows estimate how much a single run's score may vary, by pooling deviations across repeated launches and building a shared robust CI from the variance of deviations across all trainings.
Test-time rows use a classic bootstrapping approach of the 302 DMC-Optim problems.
For pairwise comparisons between two single-run configurations on the same test set, to know whether their difference is statistically meaningful with 95\% confidence, compare the difference with the corresponding training-time CI half-width multiplied by \(\sqrt{2}\) (though in practice paired differences tend to remove the \(\sqrt{2}\) penalty according to \citet{wang2026measuringnoisesllmevals}); test-time CIs give the additional uncertainty when the evaluation set is not fixed.
\par\vspace{0.2em}

\centering
\small
\renewcommand{\arraystretch}{0.82}
\setlength{\tabcolsep}{2.0pt}
\begin{tabular*}{\textwidth}{@{\extracolsep{\fill}}l rrrr rrrr@{}}
\toprule
& \multicolumn{4}{c}{\bfseries pass@$1$} & \multicolumn{4}{c}{\bfseries pass@$10$} \\
\cmidrule(lr){2-5} \cmidrule(lr){6-9}
\bfseries Source and step & \bfseries \(p_{100}\) & \bfseries \(p_{50}\) & \bfseries \(p_{30}\) & \bfseries \(p_{10}\) & \bfseries \(p_{100}\) & \bfseries \(p_{50}\) & \bfseries \(p_{30}\) & \bfseries \(p_{10}\) \\
\midrule
\multicolumn{9}{@{}l}{\textsc{Training-time}} \\[1pt]
\multicolumn{9}{@{}l}{\quad \textit{Qwen 2.5 7B}} \\[1pt]
\quad\quad After 5k RL steps & $\pm$1.9 & $\pm$1.7 & $\pm$1.3 & $\pm$0.8 & $\pm$1.9 & $\pm$1.7 & $\pm$3.1 & $\pm$1.7 \\
\quad\quad After 10k RL steps & $\pm$1.4 & $\pm$1.4 & $\pm$1.1 & $\pm$0.5 & $\pm$1.5 & $\pm$2.4 & $\pm$2.8 & $\pm$1.7 \\
\multicolumn{9}{@{}l}{\quad \textit{CWM 32B}} \\[1pt]
\quad\quad After 5k RL steps & $\pm$1.1 & $\pm$4.4 & $\pm$4.2 & $\pm$1.4 & $\pm$1.6 & $\pm$2.5 & $\pm$5.5 & $\pm$3.4 \\
\quad\quad After 10k RL steps & $\pm$1.1 & $\pm$1.4 & $\pm$1.5 & $\pm$0.8 & $\pm$1.4 & $\pm$1.3 & $\pm$1.9 & $\pm$2.5 \\
\midrule
\multicolumn{9}{@{}l}{\textsc{Test-time}} \\[1pt]
\multicolumn{9}{@{}l}{\quad \textit{CWM 32B}} \\[1pt]
\quad\quad After 5k RL steps & $\pm$4.4 & $\pm$4.2 & $\pm$3.5 & $\pm$2.1 & $\pm$4.0 & $\pm$4.6 & $\pm$5.0 & $\pm$4.3 \\
\quad\quad After 10k RL steps & $\pm$4.3 & $\pm$4.3 & $\pm$3.8 & $\pm$2.3 & $\pm$3.9 & $\pm$4.5 & $\pm$5.1 & $\pm$4.5 \\
\bottomrule
\end{tabular*}
\end{table}

Under this reading, the main optimization RL gains remain well above the estimated variability.
At \(p_{50}\), the top-30\% post-execution trained model improves over standard RLVR by 19.7 points and over optimization-test RLVR by 14.8 points, which is \(9.9\times\) and \(7.4\times\) the training-time half-width of differences, respectively.
Even under a conservative combination of training-time and test-time CIs (by simply adding the variances), the half-width is 4.7 points at \(p_{50}\), so these gaps remain \(4.2\times\) and \(3.1\times\) above it.
At \(p_{30}\), the corresponding gaps are 17.2 and 13.8 points, which is again well over significance levels.
The smaller optimization-test RLVR gain over standard RLVR is more borderline under test-set uncertainty, but if we fix the test set, gains remain above training-time variability.

We also include uncertainties after only 5k RL steps.
They are larger than the final 10k intervals, especially for CWM 32B at strict thresholds, suggesting that performance of RL reruns varies more significantly at intermediate training steps, though they converge toward a more consistent performance level as training proceeds.
We also note the test-time variance at \(p_{100}\): since LCB and DMC-Optim have similar test-set sizes, this variance should be of similar order on LCB, and this also mitigates the actual significance of the 4\% pure-correctness decrease observed for some optimization RL configurations.

\section{What Does the Model Learn?}
\label{sec:analysis}

\begin{figure}[t!]
\refstepcounter{figure}
\label{fig:radar_difficulty}
\noindent\begin{minipage}[t]{0.49\textwidth}
\centering
\includegraphics[width=0.86\linewidth]{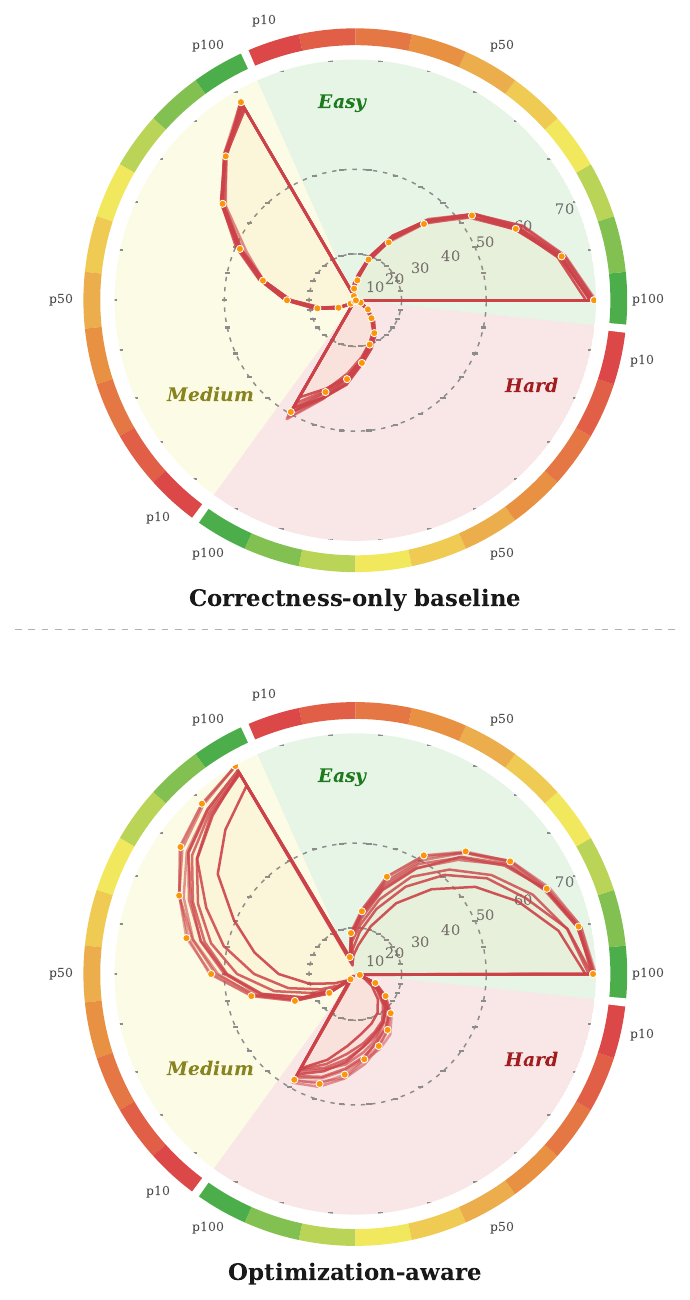}
\par\vspace{0.1em}
{\small\textbf{(a)} Absolute pass@$1$ at each training step}
\end{minipage}\hfill
\begin{minipage}[t]{0.49\textwidth}
\centering
\includegraphics[width=0.86\linewidth]{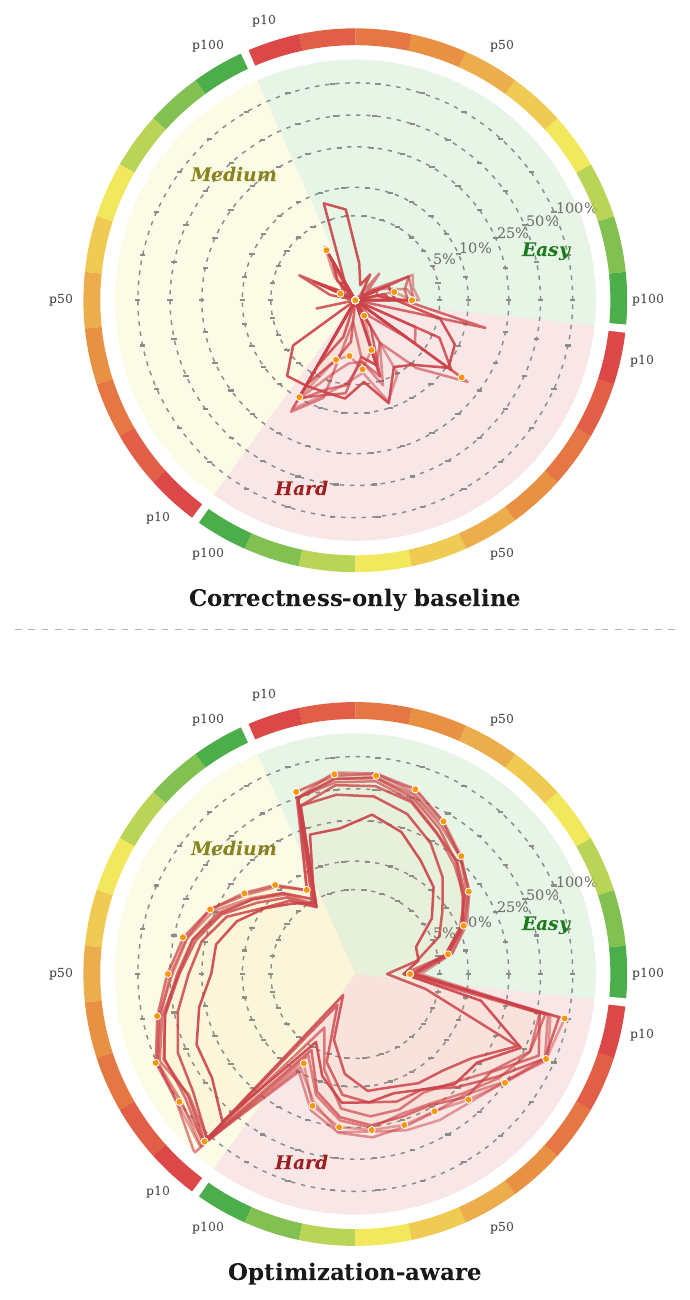}
\par\vspace{0.1em}
{\small\textbf{(b)} Relative pass@$1$ of each training step compared to step 1k}
\end{minipage}
\par\vspace{0.2em}

\small
\noindent\textbf{Figure~\thefigure: Per-difficulty CWM 32B pass@$1$ and training gains.}
\emph{Left:} absolute pass@$1$ profiles across difficulty, evaluation threshold, and training step for the correctness-only baseline and top-30\% RL training.
The top plot corresponds to the standard RLVR training, and the bottom plot to an optimization-RL training.
Each ring corresponds to the evaluation of a single training step; both trainings are 10k steps long, with an evaluation every 1k steps.
Along a ring, we distinguish the evaluation on easy, medium, and hard problems, and within each category we sweep the optimization constraint of the evaluation, from \(p_{100}\), which is pure correctness, all the way to \(p_{10}\), which is a hard optimization constraint.
\emph{Right:} relative profiles over the same difficulty-threshold grid, where gains are computed relative to the same model's performance at 1k training steps, emphasizing that optimization RL provides structured improvement as training proceeds, whereas RLVR gains are uncontrolled and inconsistent: sometimes lower on some difficulty levels or percentile thresholds, sometimes improving, but never as much.
During training, optimization RL brings gains across all percentiles, and the amplitude of the gains is greater as the evaluation constraint gets stricter.
\end{figure}

The results discussed in the previous section show large gains on optimization tasks when LLMs are RL-trained with an objective that correctly combines correctness and optimization constraints.
But higher performance could come from simple implementation hacks, such as faster input parsing, or from more interesting algorithmic discoveries, and could hide discrepancies across difficulty levels.

\paragraph{Do model optimization capabilities depend on the underlying problem difficulty?}
\Cref{fig:radar_difficulty} compares standard-RLVR-trained CWM 32B with a post-execution optimization-RL-trained variant by slicing DMC-Optim test scores by problem difficulty, evaluation percentile threshold, and training step.
At 10k steps, optimization RL keeps \(p_{100}\) pass@$1$ slightly above standard RLVR on all three splits, moving from 74.6 to 76.4 on Easy, 72.5 to 74.9 on Medium, and 49.5 to 50.6 on Hard.
The stricter thresholds move much more: at \(p_{50}\), pass@$1$ rises from 31.9 to 51.3 on Easy, 32.4 to 52.8 on Medium, and 21.9 to 30.6 on Hard; at \(p_{30}\), gains are in the order of 125\% on Easy/Medium and 70\% on Hard.
Optimization RL therefore brings gains across all levels of difficulty, without trading off one difficulty level for another, and in particular without sacrificing hard-problem pure correctness (at least on DMC-Optim) to improve lower-difficulty optimization performance.
We also observe that gains grow across all levels as evaluation percentiles get stricter.
They are nevertheless smaller on Hard problems than on Easy/Medium, which naturally follows from the fact that these problems are harder to solve in the first place and mobilize more model capacity before any optimization can be considered.
\begin{table}[t!]
\refstepcounter{table}
\label{tab:cwm_lcb_difficulty}
\normalsize
\noindent\textbf{Table~\thetable: CWM 32B LCB transfer by difficulty.}
Rows use the same CWM 32B configurations as \cref{tab:lcb_transfer}.
The overall column reports pass@$1$ on all LCB tasks.
Per-difficulty columns report pass@$1$ and WR$_{\mathrm{med}}$ within the easy, medium, and hard subsets; WR$_{\mathrm{med}}$ uses the median passing samples of an optimization RL configuration and the baseline RLVR, and compares them to determine which one is faster.
Bold marks the best displayed value within each column.
\par\vspace{0.5em}

\centering
\small
\renewcommand{\arraystretch}{0.92}
\setlength{\tabcolsep}{1.6pt}
\begin{tabular*}{\textwidth}{@{\extracolsep{\fill}}l r rr rr rr@{}}
\toprule
& \multicolumn{1}{c}{\bfseries Overall} & \multicolumn{6}{c}{\bfseries Per difficulty} \\
\cmidrule(lr){2-2} \cmidrule(lr){3-8}
& & \multicolumn{2}{c}{\bfseries Easy} & \multicolumn{2}{c}{\bfseries Medium} & \multicolumn{2}{c}{\bfseries Hard} \\
\cmidrule(lr){3-4} \cmidrule(lr){5-6} \cmidrule(lr){7-8}
\bfseries Config. & \bfseries p@1 & \bfseries p@1 & \bfseries WR$_{\mathrm{med}}$ & \bfseries p@1 & \bfseries WR$_{\mathrm{med}}$ & \bfseries p@1 & \bfseries WR$_{\mathrm{med}}$ \\
\midrule
\multicolumn{8}{@{}l}{\textsc{Standard RLVR}} \\[1pt]
\quad Standard RLVR & \textbf{56.4} & 98.3 & \dash & 67.7 & \dash & \textbf{29.8} & \dash \\
\midrule
\multicolumn{8}{@{}l}{\textsc{+ Optimization/More Correctness Tests}} \\[1pt]
\quad MC & 55.3 & 98.1 & 50.0 & 68.1 & 82.5 & 27.3 & 69.2 \\
\quad MC + 10s timeout & 54.7 & \textbf{98.5} & 66.7 & 66.6 & 67.6 & 26.7 & 69.2 \\
\quad MC + opt. tests & 55.2 & \textbf{98.5} & 36.5 & \textbf{68.2} & 74.4 & 26.9 & 72.5 \\
\midrule
\multicolumn{8}{@{}l}{\textsc{+ Reward: Pre-execution} (test filtering)} \\[1pt]
\quad $\tau=0.5$\,s & 54.4 & 98.2 & 63.6 & 66.2 & \textbf{93.4} & 26.4 & 79.4 \\
\quad $\tau=2$\,s & 54.1 & 98.2 & 58.3 & 65.7 & 88.8 & 26.0 & 75.4 \\
\midrule
\multicolumn{8}{@{}l}{\textsc{+ Reward: Intra-execution} (timeout-based)} \\[1pt]
\quad from $p_{80}$ ranked & 53.2 & 98.3 & 63.4 & 65.5 & 86.3 & 24.6 & 76.0 \\
\midrule
\multicolumn{8}{@{}l}{\textsc{+ Reward: Post-execution} (ranking-based)} \\[1pt]
\multicolumn{8}{@{}l}{\quad \textit{Leaderboard percentile}} \\[1pt]
\quad\quad Two-gate bucketed & 54.0 & 98.3 & 54.7 & 66.8 & 91.8 & 25.4 & 75.0 \\
\multicolumn{8}{@{}l}{\quad \textit{Per-test percentile}} \\[1pt]
\quad\quad top 50\% & 54.2 & 98.3 & 56.2 & 66.9 & 82.6 & 25.7 & \textbf{86.0} \\
\quad\quad top 30\% & 53.9 & 98.2 & \textbf{70.2} & 65.8 & 88.7 & 25.8 & 81.4 \\
\quad\quad Two-gate bucketed & 53.8 & 98.1 & 51.1 & 66.6 & 88.7 & 25.1 & 73.9 \\
\bottomrule
\end{tabular*}
\end{table}

\Cref{tab:cwm_lcb_difficulty} gives the out-of-distribution version of the same question on LCB.
The Easy split is essentially saturated in pure-correctness pass@$1$.
This matches what we observed in the previous section: for CWM 32B, standard RLVR probably throws away many batches because the corresponding prompts are too easy and produce zero-advantage all-success groups, whereas optimization RL can turn part of those otherwise useless samples into useful training signal.
The win rates on Easy problems are therefore not as large as on other difficulty levels.
Since the problems are easy, standard RLVR has more chances to sample a good algorithm already; the best median-sample win rate is 70.2\% for the top-30\% post-execution run, while most other optimization environments are closer to 60\%.

On Easy and Medium, we do not diagnose large correctness differences.
The top-50\% post-execution run has the same Easy pass@$1$ as standard RLVR, and is within one point on Medium.
Only on Hard problems do we observe a real solve-rate difference: top-50\% and top-30\% post-execution training reduce pass@$1$ from 29.8\% to about 25.7--25.8\%, a roughly 13\% relative drop.
The optimization levels are nevertheless very large, reaching 86.0\% WR$_{\mathrm{med}}$ on Hard problems.
We already discussed why LCB can show lower pure-correctness numbers for this specific setting: the benchmark is out-of-distribution, the hard split is more difficult than DMC-Optim, and the pass@$1$ on this specific dataset only uses small tests with infinite timeout, which simply gives no competitive advantage to any form of optimization.
Looking instead at hard-problem pass@$10$ after 10k RL steps, the top-50\% run keeps the same 44.7 score as standard RLVR, while the top-30\% run decreases by only about 3\%.
Maybe this is too optimistic, but we believe that this shows that further RL training should help translate more of this preserved pass@$10$ capacity back into pass@$1$, and the fact that the underlying multi-sample solving capacity is moving as much as it does under standard RLVR is a good signal (something that inference-time search methods can also exploit).

\begin{wrapfigure}[35]{r}{0.50\linewidth}
\vspace{-0.9em}
\refstepcounter{figure}
\label{fig:judge_category_comparison}
\noindent\makebox[\linewidth][c]{\includegraphics[width=\linewidth]{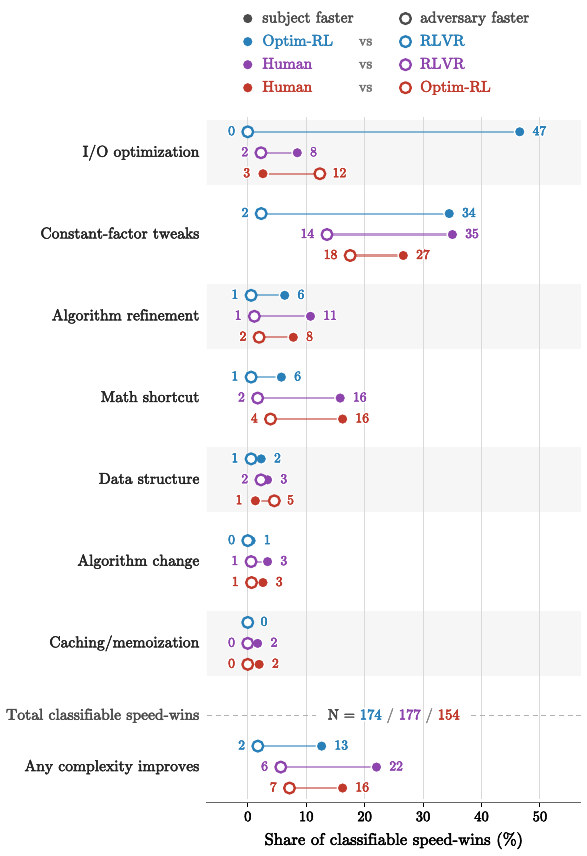}}
\vspace{-0.7em}

\small
\noindent\textbf{Figure~\thefigure: Breakdown of speed-wins on DMC-Optim test per category of code improvement, on judge-classifiable pairs of solutions.}
Numbers are proportions of the given total number of pairs that were classifiable, where a pair is made of a subject best solution and adversary best solution on a unique problem.
\vspace{-1.0em}
\end{wrapfigure}

\paragraph{What are the code optimization tricks that optimization RL trained models learned to generate?}
We compare solutions from three sources on the 302 DMC-Optim test problems: a post-execution optimization-trained CWM 32B, the CWM 32B RLVR baseline, and the fastest correct human Codeforces solution available for each problem.
For the model solutions, we select the best-of-20 passing solution per problem by optimization-test duration; for the human source, we use the fastest correct human solution.
We then discard pairs where the relative speed difference is below 1\%, and send the remaining blinded code pairs to GPT-OSS~120B~\citep{openai2025gptoss120bgptoss20bmodel}.
Each pair receives 10 randomized judge rollouts with the two solutions randomly assigned to ``Solution 1'' and ``Solution 2''.
The judge must first identify which solution is faster, which we know from execution on optimization tests, as a filtering step to capture whether the judgment that follows has any chance to be correct.
It then emits a structured classification of the differences between the solution identified as fastest and the one identified as slowest: implementation-level versus algorithmic, the sub-category of improvement (among I/O optimization, math shortcut, data-structure change, algorithm change, etc.; see \cref{fig:judge_category_comparison}), and whether the asymptotic complexity appears to improve.

For the trained optimization-vs-RLVR model comparison, 78 of the 302 tasks are filtered before judging or classification: 39 because both systems fail to produce a passing sample, 12 because only one system does, and 27 because the best passing samples are speed ties.
This leaves 224 pairs where one of the two models win by speed, with the trained model faster in 200 of them.
After 2{,}240 judge rollouts, pairs with fewer than three speed-correct rollouts, meaning rollouts where the judge identifies the true fastest solution, are marked unclassifiable, leaving 174 classified entries (we also note that GPT-OSS on the task of finding the fastest sample has 68\% accuracy).
\Cref{fig:judge_category_comparison} summarizes the resulting classification results.
On the classifiable pairs of the optimization-RL versus RLVR comparison, 47\% are speed wins in favor of the optimization-RL model due to better I/O optimization.
More interesting cases are the 6\% of pairs where it finds an algorithmic improvement and therefore beats RLVR, 6\% with a math shortcut, 2\% with a better data structure, and 1\% with an actual full algorithm change.
We also notice the 34\% of cases that correspond to constant-factor tweaks, which can still be non-negligible changes that make a difference in code competitions: avoiding intermediate allocations and building things on the fly, tightening loop bounds, or identifying early-termination cases.

\paragraph{How close are we to the best human solutions?}
Optimization-RL dominates standard RLVR, yet humans still beat the trained model in 153 of 227 speed-win pairs (67\%).
It is also useful to notice that optimization-RL beats the best humans in 33\% of cases.
I/O optimization is something the model has learned well: optimization-RL dominates humans on this type of speed win, and wins 12\% of matches thanks to such tricks.
We also note that 4\% of speed wins are obtained by the model over humans thanks to a math shortcut.
In general, however, humans remain stronger at finding such improvements: compared to the baseline, humans win 16\% of cases thanks to math shortcuts, while optimization-RL only beats RLVR this way in 6\% of cases, and the same holds for algorithmic refinements, with 11\% versus 6\%.
Finally, looking at the bottom row of \cref{fig:judge_category_comparison}, we record that in 13\% of classified pairs, according to GPT-OSS, optimization-RL wins over RLVR with a complexity improvement.
Humans still find more such improvements, with 22\% of cases against RLVR, but optimization-RL finds more than half as many complexity improvements over RLVR.
Directly comparing humans to optimization-RL still favors humans, which win in 16\% of cases with a complexity improvement, but the converse also happens in 7\% of the classified pairs.

Looking at the per-difficulty breakdown, humans and optimization-RL are roughly on par on easy problems in terms of direct speed wins, even though humans find a few more complexity improvements: 10\% of cases are human wins with a complexity improvement, versus 7.7\% for optimization-RL.
We also note that on easy problems, compared to RLVR, optimization-RL actually gets more speed wins than humans.
As difficulty increases, the human advantage becomes clearer: on hard problems, humans beat optimization-RL in 70\% of cases, and 23\% are won by humans with a complexity improvement.
Compared to RLVR, however, optimization-RL still shines on hard problems, finding a complexity improvement over RLVR in 25\% of cases.
Taken together, these results show that humans still overcome our best optimization-RL configuration, and that this RL training could benefit from favoring some types of improvements, maybe with a stronger focus on learning real algorithmic refinements or changes over I/O optimizations.
But even in its current formulation, optimization-RL learns several optimization tricks that beat the best human solutions on some problems, with non-naive code improvements.

These findings remain conditioned on the validity of the labels assigned by GPT-OSS, although we tried to control for this with a ground-truth speed task and multi-sampling.
In particular, only a subset of the problems where one model or the other wins by speed are classified.
The judge speed-identification accuracy is 68\%, and drops to 59\% on human-vs-optimization-RL pairs where the recorded-duration differences are tighter.
To ground the analysis above, we manually reviewed the optimization-RL-vs-RLVR labels and found the labels on entries deemed classifiable to be accurate: complexity categorization, meaning whether the faster solution is a complexity improvement, was correctly solved by GPT-OSS in 94.8\% of cases.
The main limitations of this judge study are that 22\% of the speed wins between optimization-RL and RLVR were not classifiable and are therefore not reported in the numbers of \cref{fig:judge_category_comparison}.
Among these are probably interesting cases, since GPT-OSS could not accurately judge them; we believe that an I/O optimization trick is easier to spot, so the skipped optimization tricks are more likely to contain other forms of improvements.
The judge was also forced to classify each pair into a single category, which can also similarly hide more interesting concurrent improvements.

\paragraph{Limitations.}
DMC-Optim is still a narrow setting for software optimization.
The tasks are single-file Python competitive-programming problems, some rewards compare against a fixed pool of human reference timings, and the timed signal depends on noisy sandbox executions.
This makes the benchmark useful for finding and comparing different shape of rewards, but it does not cover repository-scale profiling, memory objectives, multi-language systems code, or long-horizon edit loops.
The full pipeline is also expensive: it requires stronger tests, many sandbox executions, and large compute resources for such timing rewards to be learned through RL.
The simulator helps reduce this cost before online RL, but end-to-end it remains non-negligible.
Recent work points toward these settings through repository workloads~\citep{he2025sweperf,ma2025swefficiency}.
Those directions need rewards that are not too tied to static human pools.
For post-execution ranking rewards, one route is to replace the fixed reference distribution with generated or learned distributions, or even to refine it adversarially~\citep{romeraparedes2023funsearch,novikov2025alphaevolve}.
The current pre-execution filtering RL family could also be a useful candidate, although generalizing how to filter the test pool and choose the corresponding timeout would require additional work.
As full SWE tasks are definitively harder, we would also probably need to revisit how to balance correctness and optimization objectives.

The LCB benchmarking also exposed an unexpected emerging behavior of optimization RL-trained models.
They tend to remove overhead that is useful for a benchmark harness but slower at runtime, such as class wrappers, method dispatch, or unused interface structure.
This is not necessarily a general unavoidable limitation of our RL methods; it is a consequence of rewarding fast executable code on DMC-Optim train, which accepts a stripped I/O version, as well as of removing the instruction-following split of the SFT, at least for Qwen 2.5 7B and 32B.
We therefore evaluate on LCB with I/O evaluation rather than relying on call-based wrappers, but a model trained for downstream usage would likely need instruction-following rehearsal or interface-preserving constraints.

Finally, the judge analysis suggests a potential angle for improving the rewards, maybe with value models.
The reward observes that a solution is faster, but it does not say whether the gain came from I/O handling, data structures, math, algorithm change, or complexity improvements.
A significant part of the learning is therefore spent on I/O and constant-factor improvements, but the human comparison suggests that closing the remaining gap, as well as making the model produce more interesting outputs for algorithm discovery, may require feedback that is algorithm-aware.

\section{Conclusion}
RL for code optimization becomes trainable when the underlying data, measurement tools, RL environments and rewards, and GRPO are composed together to propagate a clear timing-feedback signal: larger optimization tests, calibrated execution, pre-/intra-/post-execution optimization constraints, collapsed binary rewards, and stable GRPO updates.
On DMC-Optim, this improves strict optimization across models and difficulties while preserving \(p_{100}\) pure-correctness: top-50\% pass@$1$ rises from 18.0\% to 31.3\% on Qwen 2.5 7B and from 30.7\% to 50.4\% on CWM 32B, with respective 150\% and 125\% relative gains at \(p_{30}\).
On LCB, we observe similar optimization gains with pure-correctness mostly preserved, and only contained decreases when they appear.
The model learns I/O optimization and constant-factor improvements, but also some algorithmic refinements: on a comparison set of problems, optimization-RL beats standard RLVR in 89\% of speed-win pairs and wins with complexity improvements in 13\% of cases, while the best human submissions beat the same baseline with complexity improvements in 22\% of cases.
Optimization-RL also beats human submissions themselves with a complexity improvement in 7\% of cases.
We view this as a path toward RL training for algorithm discovery: algorithm-aware feedback, potentially through value models, should further boost these capabilities.

The same correctness-efficiency gap we originally measured on competitive programming---where standard RLVR drops from 43.5\% \(p_{100}\) pass@$1$ to 7.7\% at \(p_{30}\) on Qwen 2.5 7B, and from 69.8\% to 13.7\% on CWM 32B---also appears on real-world software-engineering tasks, where it is, if anything, even more significant.
On SWE-fficiency, at the time it was released, Claude 4.5 Sonnet produced correct patches 81\% of the time but captured only 4.1\% of expert speedup~\citep{ma2025swefficiency}.
The gap between writing correct code and writing efficient code is therefore a problem that translates from competitive programming to SWE tasks as well.

RL methods to improve code generation correctness have already transferred from ``playground'' settings to real-world software engineering.
The methods that enabled SWE-RL~\citep{wei2025swerladvancingllmreasoning} to apply RL to real-world software engineering were developed first in less compute-intensive settings: GRPO was introduced on math problems through DeepSeekMath~\citep{shao2024deepseekmathpushinglimitsmathematical}, and rule-based verifiable rewards and the \texttt{\textless think\textgreater} reasoning template were validated on math and competitive coding through DeepSeek-R1~\citep{deepseek2025r1}.
SWE-RL adapted the reward and scaffolding for the real-world software-engineering domain, swapping execution-based rewards for text similarity because running code in real repositories was too expensive.

Efficiency has not been correctly mastered on software-engineering tasks yet.
Not only are software-engineering tasks harder and far more expensive to execute, slowing down RL iterations, but the challenges that make competitive-programming efficiency hard---measurement reliability, reward noise, and training instability---are all exacerbated on real-world software-engineering tasks.
We see this work as a step toward a similar transfer for efficiency: it develops some methods in the competitive-programming ``playground'', where the feedback loop is more easily handled, and may provide a basis for later adaptation to real-world software optimization.

\section*{Acknowledgments}
We thank the authors of the CWM paper whose RL codebase made this work possible, in particular Jonas Gehring, Jade Copet, Quentin Carbonneaux, David Zhang, Badr Youbi Idrissi, Vegard Mella, and Taco Cohen.
We also thank Zacharias Fisches, Sida Wang, Mathurin Videau, and João Maria Janeiro for helpful discussions.
We thank Andrew Hamiel, Don Landrum, and Eslam Elnikety for their support.

\bibliographystyle{assets/plainnat}
\bibliography{paper}

\clearpage
\beginappendix

\section{Simulation study of code optimization success metrics under measurement noise}
\label{app:simulation}

This appendix isolates the measurement question behind \cref{sec:measurement}: if two correct solutions differ in runtime, when can a noisy execution system rank them reliably enough to support either an RL reward or an evaluation metric?
We deliberately remove the RL loop and study a smaller problem, namely how per-execution timing perturbations interact with the test suite and the metric used to aggregate durations.
The noise considered here is random variation around a fixed execution regime: two executions of the same solution on the same test can differ because of dispatch latency, scheduling jitter, or other local timing effects.
More systematic shifts in CES, our remote sandbox execution service, such as changing load patterns, sandbox upgrades, or redeploying the sandbox service to different infrastructure, affect measurements in a more consistent way across many executions and are treated separately in \cref{sec:app_duration_correction,sec:app_temporal_stability}, together with stored-duration calibration.
The test generation step that creates the timing measurements studied here, and more broadly the dataset cleaning and duration-filterability pipeline, are treated in \cref{app:dataset}.
The point of this appendix is therefore not to prove that a particular reward trains, but to explain why the paper first changes the test regime before comparing reward designs, and why the code-optimization success metric should depend on the timing regime of the test set being used.
\suppressfloats[t]

Two design choices matter in this simplified setting.
First, the test suite must make algorithmic differences large relative to fixed dispatch and scheduling overheads; otherwise a tens-of-milliseconds perturbation can be a large fraction of the signal.
Second, the metric should not depend on one fragile boundary when a relative comparison can aggregate many per-test duration comparisons.
We study these choices by contrasting the test-suite properties of LiveCodeBench (LCB), a dataset with short base tests and fewer tests per problem, against DMC-Optim, a dataset with larger optimization tests, longer runtimes, and more tests per problem; we then simulate a thresholded timeout metric against a threshold-free win-rate metric.
We discuss the original DMC tests in prose only: as shown in \cref{sec:data_test_gen}, they remain in the same fast-test regime as LCB, so we use LCB as the representative benchmark for short-test regimes in the quantitative comparison.

\subsection{Empirical timing regimes}
\label{sec:app_empirical}

\paragraph{Fast-test and slow-test regimes.}
The empirical difference between the two regimes is visible before any simulation, but the evidence comes from two complementary sources.
\Cref{sec:data_test_gen} reports human-reference statistics showing the construction effect inside DMC: original DMC tests reach only $0.145\,\text{s}$ at $p95$ and $0.463\,\text{s}$ at $p99$, whereas generated optimization tests reach $1.296\,\text{s}$ and $3.710\,\text{s}$.\footnote{These quantiles are computed on the final RL training split. Unless stated otherwise, the remaining empirical statistics and figures in Appendix~A use held-out executions from LCB and from the DMC-Optim test split.}
The training-split quantiles and the held-out comparisons below support the same qualitative point: original DMC and LCB are fast-test regimes, while DMC-Optim moves much more mass into durations where timing differences can be measured.

\begin{figure}[b!]
\centering
\includegraphics[width=\textwidth]{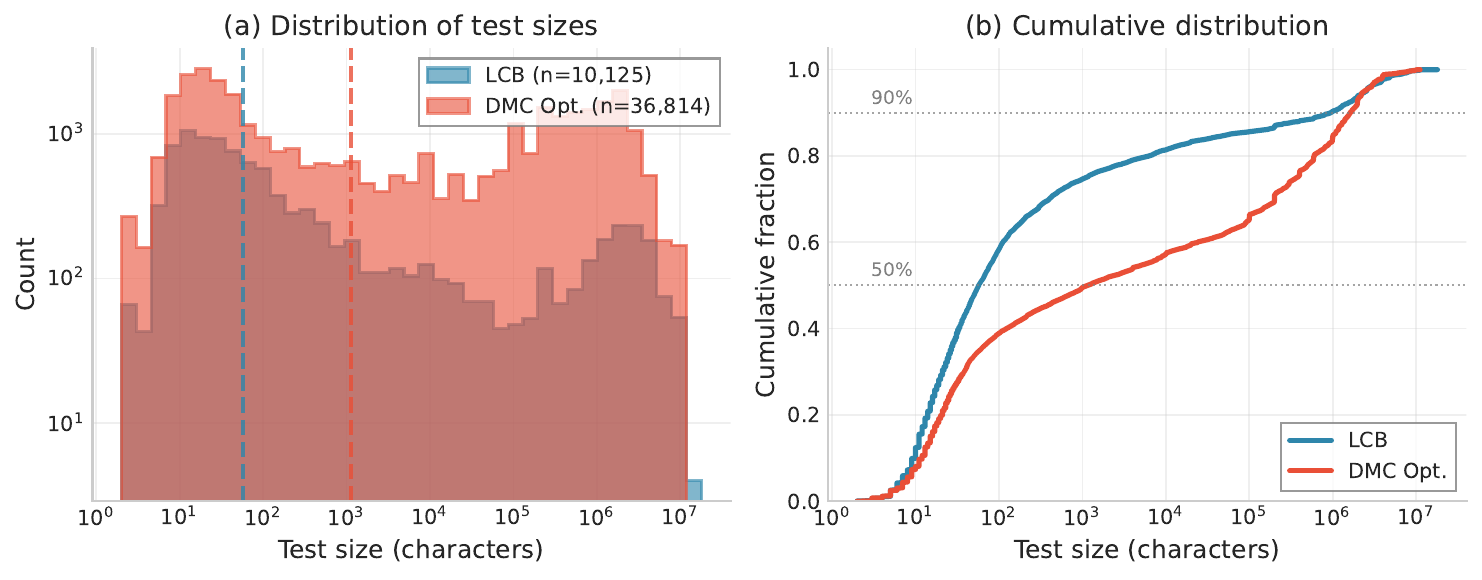}
\caption{\textbf{Test-size regimes for LCB and DMC-Optim.} \emph{Left:} Histograms of per-test input-plus-output character counts on log--log axes, with dashed vertical lines at the medians. LCB is concentrated on small public/private tests, while DMC-Optim optimization tests are shifted toward larger inputs and retain a much heavier upper tail. \emph{Right:} Cumulative distributions on a log-scaled x-axis. The separation is already visible around the median and widens further in the upper tail, which is where larger inputs can induce measurable runtime differences between solutions.}
\label{fig:app_test_sizes}
\end{figure}

\Cref{fig:app_test_sizes} shows that DMC-Optim uses both larger optimization tests and more tests per problem than LCB.
LCB aggregates 10{,}125 public/private tests over 287 problems, or 35.3 tests per problem on average, while DMC-Optim contributes 36{,}814 optimization tests over 302 problems, or 121.9 per problem.
The typical LCB test remains in the tens of characters, whereas the typical DMC-Optim test is around $10^3$ characters, with a much heavier upper tail.

\begin{figure}[t!]
\centering
\includegraphics[width=\textwidth]{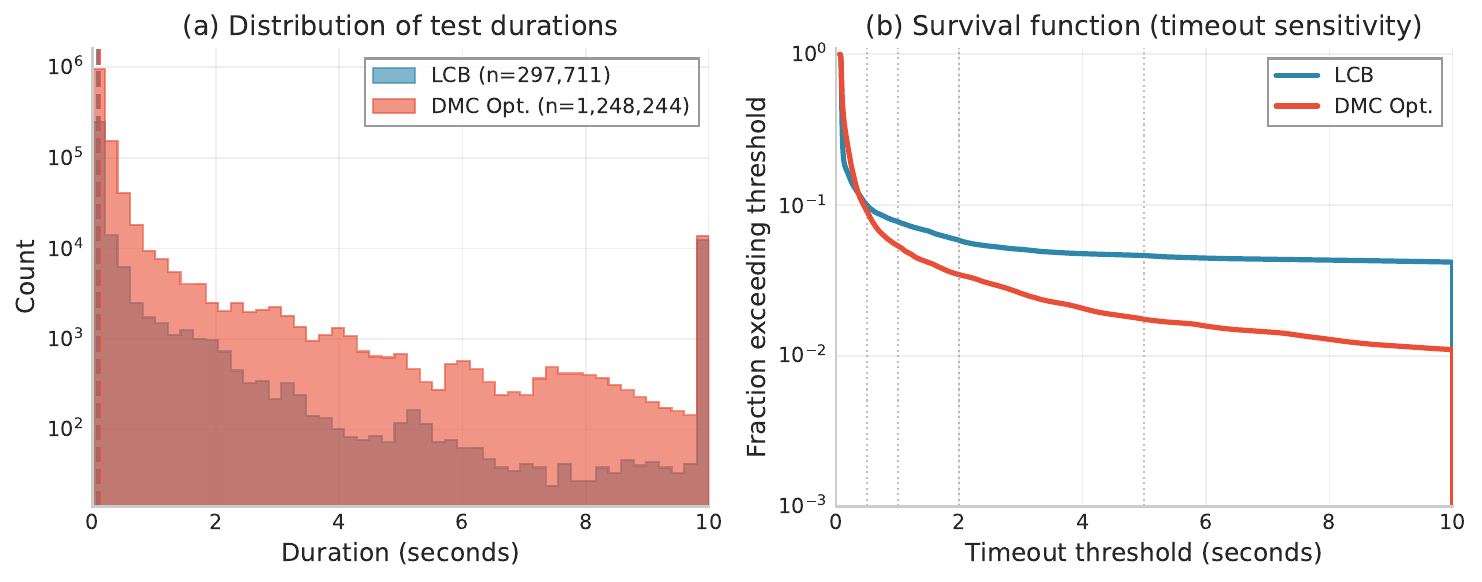}
\caption{\textbf{Observed duration distributions for LCB and DMC-Optim.} \emph{Left:} Histograms of per-test execution durations on a log-scaled y-axis for 297{,}711 LCB executions and 1{,}248{,}244 DMC-Optim executions, aggregated over two model evaluations on the DMC-Optim test split. LCB places more mass near very fast executions and exhibits a visible timeout-induced tail, while DMC-Optim has a broader body and a smoother upper tail; the spike at 10\,s is the timeout bin, corresponding to executions that were too slow to finish under the service limit. \emph{Right:} Survival functions showing the fraction of tests above each threshold, again on a log-scaled y-axis. LCB becomes nearly flat between 2 and 10\,s, so threshold changes in that region move little useful signal; DMC-Optim decays more steadily, making threshold choice materially more informative.}
\label{fig:app_durations}
\end{figure}

The same pattern appears in runtime.
LCB is not the original DMC benchmark, but it occupies the same fast-test regime: \cref{fig:app_durations} shows a sharp concentration on very fast executions and a broad plateau in the survival curve between 2 and 10\,s, so changing the timeout across that range moves little signal.
DMC-Optim instead shows a steadier decay, which means the same change in threshold can still move a non-trivial fraction of tests across the boundary.

\subsection{A compact noise hypothesis}

\paragraph{Additive and multiplicative terms.}
Let $d_{S,i}^{\text{true}}$ be the execution time of solution $S$ on test $i$ in a fixed execution environment.
We model the observed duration as
\begin{equation}
    d_{S,i}^{\text{obs}} = d_{S,i}^{\text{true}} \left(1 + \epsilon_{S,i}^{\text{mult}}\right) + \epsilon_{S,i}^{\text{add}}.
    \label{eq:app_noise_model}
\end{equation}
The multiplicative term captures proportional slowdowns that scale with the workload.
The additive term captures fixed overheads such as scheduling jitter, dispatch latency, or context-switch costs.
These effects are not observed separately and are hard to model mechanistically, so we treat their residual contribution as random additive noise.
The quantity that matters more on fast tests is the relative additive burden
\begin{equation}
    r_{S,i}^{\text{add}} = \frac{\left|\epsilon_{S,i}^{\text{add}}\right|}{d_{S,i}^{\text{true}}}.
    \label{eq:app_relative_burden}
\end{equation}
As $d_{S,i}^{\text{true}}$ becomes small, the same absolute perturbation occupies a larger fraction of the measurement.

This decomposition is not only synthetic.
The stored-to-fresh CES calibration in \cref{sec:app_duration_correction} fits an affine map with a $53\,\text{ms}$ intercept, which is consistent with a non-negligible fixed-cost term in the execution stack.
Because this intercept corresponds to a systematic shift between sandbox states, we remove it with a calibration model rather than treating it as random per-call noise.
It nevertheless gives the right scale for the additive component studied here.
As a corroborative check on the test executions used for the benchmark comparison, a hypothetical $53\,\text{ms}$ perturbation would amount to at least half of the observed duration for 75.9\% of LCB executions and 52.4\% of DMC-Optim executions; at $100\,\text{ms}$, it would exceed the full observed duration for 68.8\% and 46.3\%, respectively.
These auxiliary numbers are not our main duration statistics, but they point in the same direction as \cref{sec:data_test_gen,fig:app_durations}: a fixed tens-of-milliseconds term is much harder to absorb in the fast-test regime.

\subsection{Simulation study: timeout versus win-rate}
\label{sec:app_simulation}

\paragraph{Why compare these two metrics.}
Every metric in this section is built from the same noisy per-test durations, but the aggregation determines where noise can flip the decision.
We therefore compare two minimal and natural families: a thresholded absolute metric, represented by timeout, and a threshold-free relative metric, represented by win-rate.
Timeout is the simplest way to ask whether a solution satisfies a fixed latency budget, while win-rate is the simplest way to ask whether one correct solution is faster than another without choosing an absolute threshold.
These are not meant to specify the final reward design.
Each can instead be read as a simple code-optimization success metric: during RL, it can be converted into a reward that pushes generation toward faster correct solutions; at evaluation time, it can be used as a score for comparing models on a test set.
The simulation keeps only the thresholded-versus-relative distinction, which is the part that can be studied without the full RL environment.

\paragraph{Two simple metric families.}
For a fixed timeout $\tau$, the timeout score of solution $S$ is
\begin{equation}
    m_{\tau}(S) = 1\!\left[\max_i d_{S,i}^{\text{obs}} \le \tau \right].
    \label{eq:app_timeout_metric}
\end{equation}
In a pairwise comparison between solutions $A$ and $B$, the timeout metric declares $B$ the winner only if $m_{\tau}(B) > m_{\tau}(A)$.
This makes the decision hinge on a single threshold and on the slowest observed test for each solution.
Even if $B$ is faster than $A$ on every test, a fixed timeout $\tau$ gives $B$ no credit whenever one test of $B$ remains above $\tau$: both solutions fail if $A$ is also above $\tau$, and the speed gap is invisible to this metric.
Only an oracle threshold satisfying
\(
\max_i d_{B,i}^{\text{true}} < \tau < \max_i d_{A,i}^{\text{true}}
\)
would separate the pair by timeout.
Thus, under aggregate notions of speed such as average duration or per-test win rate, timeout can misrank or fail to distinguish solutions because its decision boundary is not a simple comparison of which solution is usually faster.
Noise can then make the same threshold comparison less stable, because each per-test measurement is another chance for an execution near the boundary to cross it.

For win-rate, we compare the two observed durations test by test:
\begin{equation}
    w(B;A) = \sum_i 1\!\left[d_{B,i}^{\text{obs}} < d_{A,i}^{\text{obs}}\right].
    \label{eq:app_winrate_metric}
\end{equation}
The win-rate metric declares $B$ the winner if $w(B;A) > w(A;B)$.
It is still noisy, though it does not require tuning an external threshold and it averages the comparison over all tests rather than over a single boundary event, making it easier for per-test duration noise to cancel out.

\begin{figure}[t!]
\centering
\includegraphics[width=\textwidth]{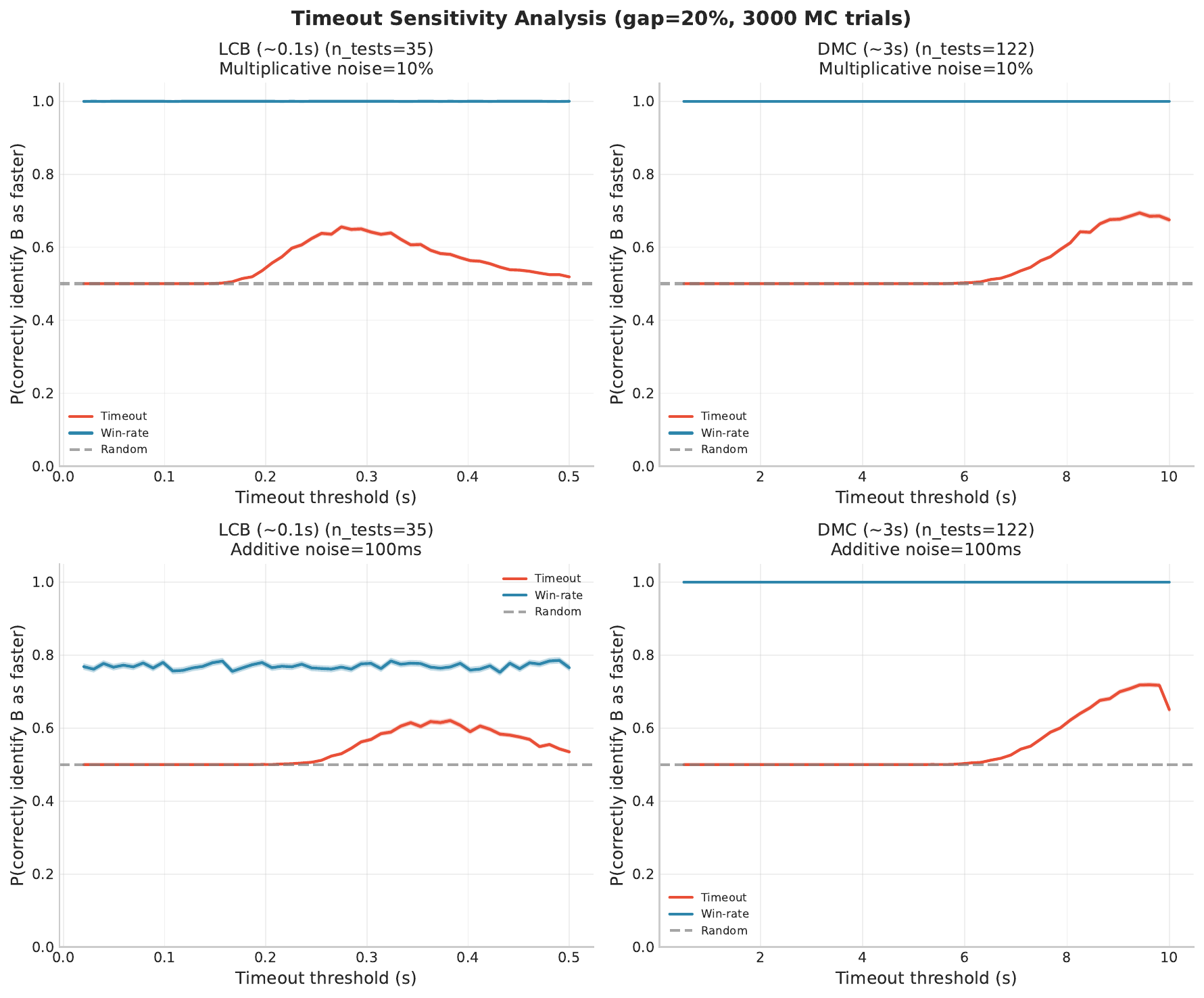}
\caption{\textbf{Threshold sensitivity of timeout versus win-rate.} Accuracy is plotted against the timeout threshold for LCB-like problems (left) and DMC-Optim-like problems (right), under multiplicative noise (top) and additive noise (bottom). Timeout is the red curve and win-rate the blue horizontal line. Timeout is only useful in a narrow interval between the two solutions' extremes; outside that interval both solutions either pass or fail together, producing ties. On LCB-like problems with additive noise, the useful interval largely disappears and timeout remains close to random guessing, while win-rate stays meaningfully higher because it compares durations directly rather than through a single boundary.}
\label{fig:app_timeout_sweep}
\end{figure}

\begin{figure}[t!]
\centering
\includegraphics[width=\textwidth]{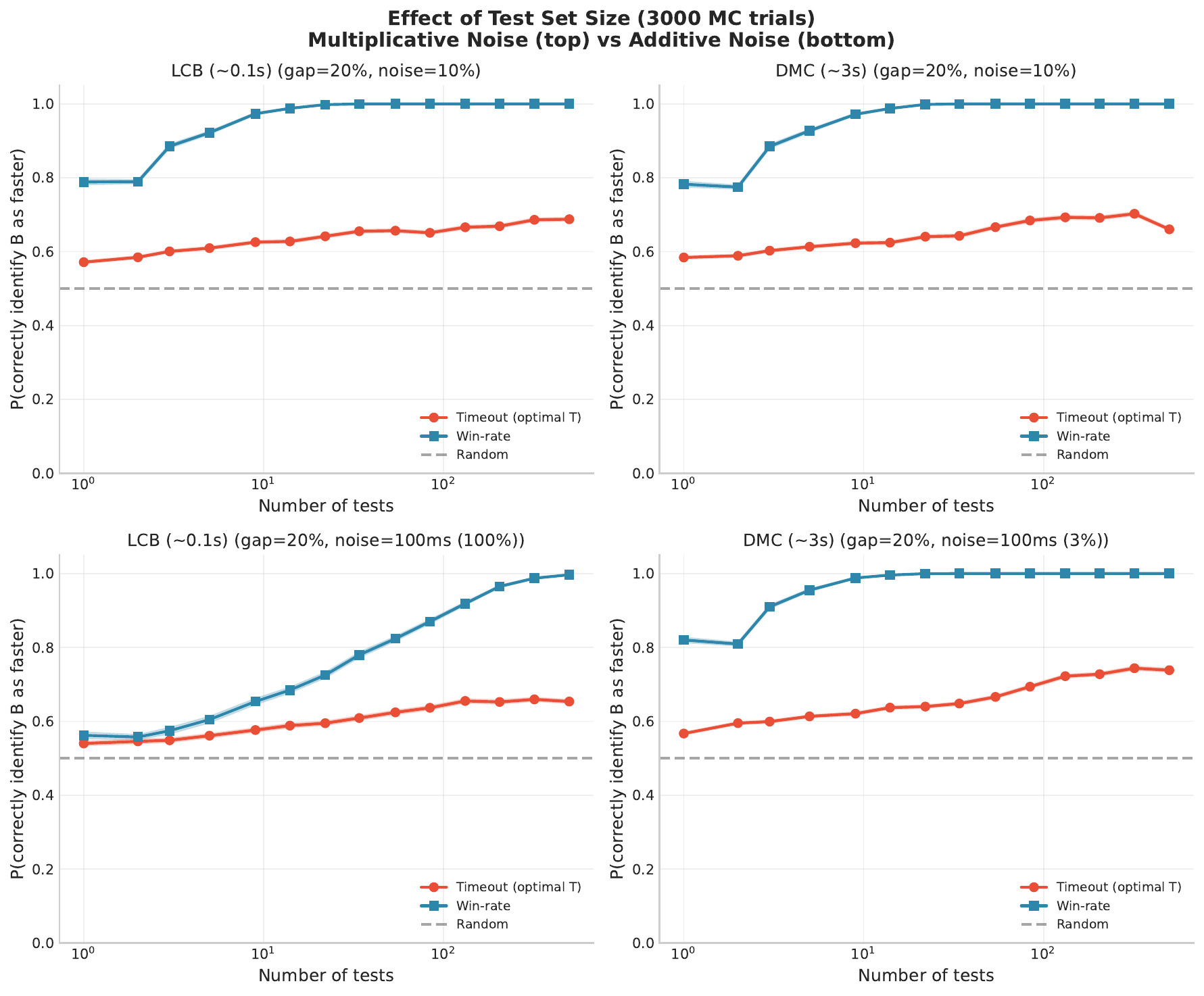}
\caption{\textbf{Effect of the number of tests per problem.} Accuracy is plotted against the number of tests for LCB-like problems (left) and DMC-Optim-like problems (right), under multiplicative noise (top) and additive noise (bottom). On fast problems with additive noise, win-rate continues to gain from additional tests because it averages pairwise comparisons across the suite, while timeout improves at a much slower rate because every decision still hinges on a single threshold. On slower problems, both metrics benefit more from having more tests, which is one reason DMC-Optim is easier to measure reliably than LCB.}
\label{fig:app_ntests}
\end{figure}

\begin{figure}[t!]
\centering
\includegraphics[width=\textwidth]{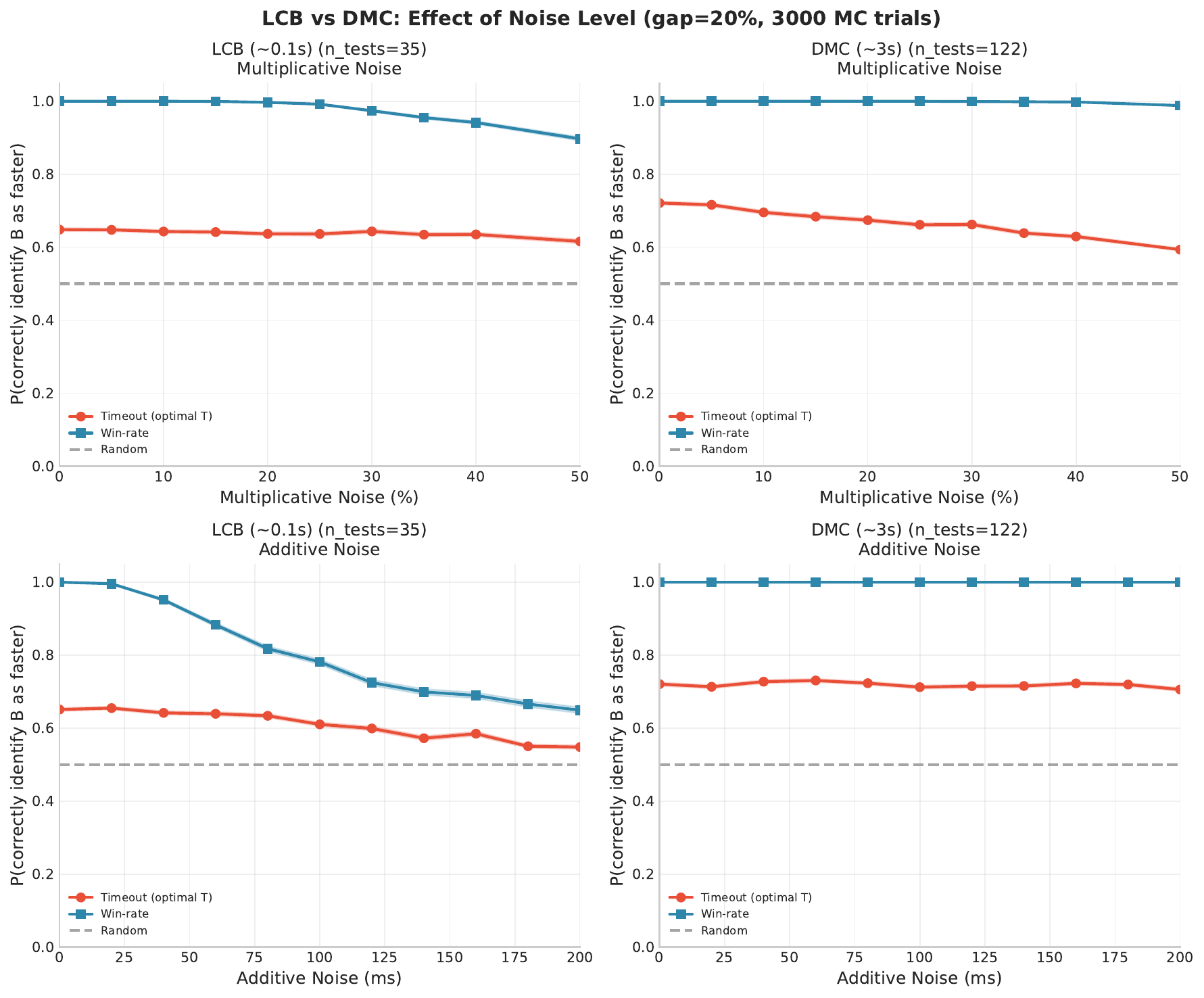}
\caption{\textbf{Noise sensitivity across the LCB-like and DMC-Optim-like regimes.} Each panel plots accuracy against noise magnitude for LCB-like problems (left) and DMC-Optim-like problems (right), under multiplicative noise (top) and additive noise (bottom). Multiplicative noise degrades both families gradually. Additive noise instead separates the regimes: on LCB-like problems the timeout metric approaches random guessing while win-rate remains meaningfully above random guessing, whereas on DMC-Optim-like problems both metrics remain usable over a much wider range. Longer and more numerous tests therefore help both metric families, but the thresholded one fails earlier.}
\label{fig:app_lcb_vs_dmc}
\end{figure}

\paragraph{Synthetic problem families.}
The simulation uses empirically anchored synthetic LCB-like and DMC-Optim-like problems that isolate the qualitative regime difference; the next paragraph gives the exact parameterization.
The parameters are anchored to the empirical test counts and duration scales in \cref{fig:app_test_sizes,fig:app_durations}: an LCB-like family with 35 tests per problem and characteristic sub-second durations, and a DMC-Optim-like family with 122 tests per problem and substantially slower tests.
In each trial, solution $B$ is the true winner and solution $A$ is 20\% slower on average, with additional per-test variability so that some individual tests can still reverse the ordering.
We inject either multiplicative noise or additive noise and measure how often each metric identifies the correct winner.
Except for the explicit threshold sweep in \cref{fig:app_timeout_sweep}, timeout results use an oracle threshold chosen after the fact, so the reported timeout curves should be read as optimistic.
The three figures below answer complementary questions: \cref{fig:app_timeout_sweep} shows whether threshold tuning can rescue timeout at all, \cref{fig:app_ntests} shows how much more tests help each metric family, and \cref{fig:app_lcb_vs_dmc} shows how the two dataset regimes separate as noise grows.

\paragraph{Synthetic playground specification.}
For regime $r \in \{\text{LCB-like}, \text{DMC-Optim-like}\}$, we first draw base test durations \(b_i \sim \mathrm{LogNormal}(\mu_r, \sigma_r)\), with \((\mu_r, \sigma_r) = (-2.3, 0.5)\) for LCB-like problems and \((1.0, 0.5)\) for DMC-Optim-like problems.
We use $N_r = 35$ and $N_r = 122$ tests, respectively, to match the empirical test-count gap.
We then generate the true durations of the two competing solutions as
\begin{equation}
    d_{A,i}^{\text{true}} = b_i \left(1 + \frac{\Delta}{2} + \eta_{A,i}\right),
    \qquad
    d_{B,i}^{\text{true}} = b_i \left(1 - \frac{\Delta}{2} + \eta_{B,i}\right),
\end{equation}
with mean gap \(\Delta = 0.2\) and per-test variability \(\eta_{A,i}, \eta_{B,i} \sim \mathcal{N}(0, 0.15^2)\), before applying the observation model from \cref{eq:app_noise_model}.
These values are empirical anchors chosen to preserve the main ingredients that matter for the argument: positive and strongly right-skewed durations, the order-of-magnitude separation between fast and slow tests, and the empirical difference in tests per problem.
Each Monte Carlo trial resamples the base durations, the test-specific gap factors, and the measurement noise, so the reported curves average over both execution noise and problem-to-problem variability rather than over a single frozen synthetic instance.

\paragraph{Threshold fragility.}
\Cref{fig:app_timeout_sweep} shows the first difference between the two metric families.
Win-rate is flat because it has no threshold to tune.
Timeout is only informative in a narrow band between the two solutions' extremes, so its accuracy varies sharply with the chosen threshold even under multiplicative noise.
Under additive noise on LCB-like problems, that useful band largely disappears and timeout remains close to random guessing across the full sweep, whereas win-rate stays clearly above random guessing.

\paragraph{More tests help, but not equally.}
\Cref{fig:app_ntests} shows that increasing the number of tests improves both metric families, but the gain is not symmetric.
This matters because DMC-Optim provides about 3.5 times more tests per problem than LCB.
On LCB-like problems under additive noise, win-rate keeps improving as more tests are added, whereas timeout improves at a much slower rate.
On DMC-Optim-like problems, especially under additive noise, the gains from additional tests are much stronger for both metrics, which is exactly the behavior we want from a slower and denser test suite.

\paragraph{The regime split appears under additive noise.}
\Cref{fig:app_lcb_vs_dmc} summarizes the comparison by sweeping the noise magnitude directly.
Under multiplicative noise, both datasets degrade gradually and the gap between timeout and win-rate is modest.
The qualitative split appears under additive noise.
On LCB-like problems, timeout approaches random guessing while win-rate remains meaningfully above random guessing.
On DMC-Optim-like problems, both metrics remain usable over a much wider range, although win-rate stays higher throughout.

\paragraph{Practical implication for experiment design.}
These figures suggest a simple order of operations.
First change the data regime: larger inputs, longer runtimes, and more tests per problem improve both metric families at once.
Only then does it become worthwhile to decide whether a hard timeout should remain part of the optimization objective.
If the suite stays in the fast-test regime, tighter threshold tuning does not fix the underlying problem; it only searches for a narrow operating window in a measurement setting where thresholded decisions are intrinsically brittle.
For training and DMC-Optim evaluation, this is why our experiments first move from the original fast-test regime to DMC-Optim, and only then study how different efficiency metrics behave on top of that dataset.
\Cref{tab:qwen7b_environments} shows that strict DMC-Optim thresholds can separate Qwen 2.5 7B training configurations once the test regime has been changed.
For LCB, the same argument leads to a different evaluation choice: because LCB remains in the fast-test regime, we use win-rate-style speed scores in \cref{tab:lcb_transfer} rather than absolute timeout sweeps.
\Cref{tab:app_lcb_timeout_sweep} gives the corresponding empirical check.
Across Qwen 2.5 7B checkpoints, LCB timeout pass@$1$ moves mostly as a shared function of the threshold and provides little configuration separation; at the tightest threshold, the displayed rows span only 1.0 pass@$1$ point, from 29.6\% to 30.6\%, so the sweep compresses differences between training configurations rather than giving a useful optimization ranking.

\begin{table}[t!]
\centering
\caption{\textbf{LCB timeout pass@$1$ is weakly discriminative across Qwen 2.5 7B training configurations.}
Rows are the Qwen 2.5 7B checkpoints from the environment sweep in \cref{tab:qwen7b_environments,tab:lcb_transfer}; naming follows the LCB transfer table.
Each column reports LCB pass@$1$ after applying an absolute timeout threshold to the same re-executed samples.
The threshold sweep mostly moves all configurations together, and the strictest threshold compresses the displayed rows to a 1.0-point band, which is why \cref{tab:lcb_transfer} reports LCB speed transfer with win-rate-style scores instead.}
\label{tab:app_lcb_timeout_sweep}
\footnotesize
\setlength{\tabcolsep}{3.2pt}
\begin{tabular}{@{}lrrrrrrrr@{}}
\toprule
\bfseries Configuration & \bfseries 10\,s & \bfseries 5\,s & \bfseries 2\,s & \bfseries 1\,s & \bfseries 0.5\,s & \bfseries 0.3\,s & \bfseries 0.2\,s & \bfseries 0.1\,s \\
\midrule
\multicolumn{9}{@{}l}{\textsc{Standard RLVR}} \\[1pt]
\quad Standard RLVR & 44.2 & 43.6 & 42.0 & 40.1 & 37.2 & 35.8 & 34.4 & 30.6 \\
\midrule
\multicolumn{9}{@{}l}{\textsc{+ Optimization/More Correctness Tests}} \\[1pt]
\quad MC & 42.6 & 42.3 & 41.2 & 39.5 & 37.0 & 35.6 & 34.2 & 29.9 \\
\quad MC + 10s timeout & 43.1 & 42.8 & 41.6 & 40.0 & 37.2 & 35.9 & 34.5 & 30.1 \\
\quad MC + opt. tests & 42.1 & 41.9 & 40.8 & 39.1 & 36.6 & 35.1 & 33.8 & 29.6 \\
\midrule
\multicolumn{9}{@{}l}{\textsc{+ Reward: Pre-execution} (test filtering)} \\[1pt]
\quad $\tau=2$\,s & 41.1 & 40.9 & 40.5 & 38.7 & 36.8 & 35.1 & 34.0 & 29.7 \\
\midrule
\multicolumn{9}{@{}l}{\textsc{+ Reward: Post-execution} (ranking-based)} \\[1pt]
\multicolumn{9}{@{}l}{\quad \textit{Leaderboard percentile}} \\[1pt]
\quad\quad Two-gate bucketed & 41.9 & 41.6 & 40.9 & 39.2 & 37.0 & 35.3 & 34.2 & 29.6 \\
\multicolumn{9}{@{}l}{\quad \textit{Per-test percentile}} \\[1pt]
\quad\quad top 50\% & 41.1 & 41.0 & 40.7 & 38.6 & 37.0 & 35.2 & 34.3 & 30.2 \\
\quad\quad top 30\% & 40.6 & 40.6 & 40.2 & 38.5 & 36.6 & 34.9 & 34.1 & 29.6 \\
\bottomrule
\end{tabular}
\end{table}

\subsection{Takeaways}
\label{sec:app_implications}

This appendix supports the order of operations used in the paper.
First, a mixed noise model with an additive term is the right abstraction for fast test suites: when durations are close to dispatch and scheduling overheads, the same absolute perturbation can dominate the measurement.
Second, larger inputs, longer runtimes, and more tests per problem help both thresholded and relative metrics by making the timing signal larger and by averaging over test-specific reversals.
Original DMC and LCB remain on the fragile side of this divide, while DMC-Optim moves enough mass into the slower regime to make timing-based decisions more informative.
Third, relative aggregation survives longer than a hard timeout in the fast-test regime because it compares durations across many tests instead of hinging on one boundary.
A downstream consequence is that LCB transfer is evaluated with win-rate-style speed scores rather than timeout sweeps: LCB is a short-test benchmark, and the simulations above together with \cref{tab:app_lcb_timeout_sweep} show that thresholded comparisons in that regime do not support stable model comparisons.

These conclusions do not remove the need for the rest of the measurement stack.
DMC-Optim still needs duration filterability to select problems where correct solutions can be ranked by time (\cref{app:dataset}), and CES calibration is still needed so that measured durations remain comparable across service states (\cref{app:ces_fallback}).
The simulation explains the first step: before choosing a reward, the benchmark has to leave the regime where fixed 10--100\,ms perturbations are comparable to the duration signal.
Once the test suite provides larger inputs, longer durations, and more trials per problem, \cref{sec:reward} can ask how to turn those measurements into a correctness-gated RL reward.

\clearpage
\section{Building DMC-Optim: stronger correctness signal and new optimization feedback}
\label{app:dataset}
This appendix expands the data half of \cref{sec:measurement}. The main paper states the causal chain in compressed form: raw DMC is too weak as both a correctness gate and a timing source; DMC-Optim rebuilds the test regime; duration filterability then selects the subset where timing can rank correct solutions. Here we spell out the construction details of DMC-Optim. The main pipeline is \cref{fig:adjudication_pipeline}; this appendix records the thresholds, counts, side branches, and diagnostic figures that support it.

\paragraph{Roadmap.}
The appendix has four parts. First, \cref{sec:app_dataset_claim} states what an optimization-RL dataset must prove, and \cref{sec:app_source_data} explains why raw DMC fails both as a correctness gate and as a timing source. Second, \cref{sec:app_construction_stages} gives the full reduction from raw DMC to the $2{,}723$-problem DMC-Optim corpus and the $1{,}302$-problem duration-filterable RL pool, \cref{sec:app_test_generation} describes how the new tests are generated, \cref{sec:app_filtering_generated_tests} explains how generated tests are filtered before they are trusted, and \cref{sec:app_before_duration_selection} reports the corpus that remains before duration selection. Third, \cref{sec:app_filterability} defines duration filterability and \cref{sec:app_duration_stats} describes the final duration-filterable RL pool, including the stored human-reference durations used by the reward and evaluation pipeline. Finally, \cref{sec:app_filterability_training_controls} trains models to test whether DMC-Optim helps with optimization RL, and whether dataset construction alone is enough without problem-relative rewards and environment design. Timing-backend reliability and calibration of stored durations are handled in \cref{app:ces_fallback}.

\subsection{What the dataset must establish}
\label{sec:app_dataset_claim}

Not every coding dataset can serve as seed data for optimization RL. Even a correctness-only objective demands careful data curation, and adding execution time makes the requirements stricter. The data must satisfy three conditions at once. First, the problems must lie in a learnable regime: if they are too easy, RL saturates quickly; if they are too hard, nearly every rollout fails and the optimization objective has no chance to appear. Second, the tests must span enough computational load that execution time reflects code behavior rather than only scheduler noise, interpreter overhead, or a fixed sandbox cost. Third, if such tests are generated, the generation process needs trusted positive and negative controls. Verified correct human solutions are used both to validate candidate tests and, later, as timing references. Verified incorrect solutions are used as negative controls for checking whether an augmented suite still lets wrong code pass.

This is why DMC-Optim separates two roles that are often collapsed in programming benchmarks. Correctness tests tighten the pass/fail gate: they should reduce false positives without introducing false negatives. Optimization tests create duration spread: after correctness has already been enforced, they should separate correct solutions by runtime. The final dataset therefore stores not only test inputs and expected outputs, but also measured reference durations for the verified human solutions, because several downstream reward and evaluation variants compare model rollouts against this stored human timing distribution.

\subsection{Source data, limitations, and decontamination}
\label{sec:app_source_data}

\paragraph{DeepMind Code Contests.}
The base dataset is the DeepMind Code Contests (DMC) dataset~\citep{li2022competition}, a collection of approximately $12{,}275$ competitive-programming problems sourced from platforms such as Codeforces. Each problem contains a natural-language description, input/output specifications, constraints, example input/output pairs, and three original test categories: public tests visible during competition, private tests used for final grading, and generated tests produced by the DMC authors with input generators and brute-force verifiers. Human-submitted solutions are provided in multiple programming languages and labeled as correct or incorrect.

We retain Python solutions. C++ is the dominant language in competitive programming, but it is also the most controlled setting for low-level speed: compiler optimizations, memory layout, and hand-tuned implementation details can dominate the methodology. Python is noisier and more overhead-heavy, which makes it the more demanding setting for a timing methodology that must operate inside an RL loop. If optimization RL can learn useful algorithmic changes in this noisier setting, we expect the same ideas to transfer to lower-overhead languages such as C++, although language-specific calibration would still be needed.

\paragraph{Original tests are too small for timing.}
The original DMC tests were designed for functional correctness, not for measuring efficiency. Even after cleaning, the original categories remain extremely small: on the final RL training split, median input-plus-output sizes are 18, 24, and 31 characters for public, private, and DMC-generated tests. Across the original categories, the pooled mean execution duration over correct human solutions is only $0.088\,\text{s}$, with $p95=0.145\,\text{s}$ and $p99=0.463\,\text{s}$ (\cref{tab:dataset_stats}). At that scale, additive perturbations of 10--100\,ms can match or exceed the signal. The simulation study in \cref{app:simulation} shows why this regime is fragile: with additive noise around 100\,ms, timeout-based comparisons on such fast tests can fall to roughly $55\%$ accuracy, close to chance. Relative runtime comparisons are more stable because they aggregate many per-test comparisons instead of depending on one timeout boundary, but they remain weak while the tests are this short; they become reliable only when larger inputs, longer runtimes, and more tests make the timing signal larger than the execution-noise floor.

\paragraph{Original tests are also too weak as a correctness gate.}
The original DMC tests are also weak for evaluating correctness itself. This matters because the paper compares optimization-RL recipes that can trade off generating faster programs against preserving correctness. If the seed training data already lets incorrect solutions pass, later changes in model pass rate are hard to interpret: they could come from uncontrolled label noise rather than from the reward or environment being tested. We therefore need a cleaner correctness gate before asking whether an RL recipe improves speed, degrades correctness, or changes both. CodeContests+~\citep{codecontestsplus} reports that only $67.1\%$ of the mutation-generated tests in the original dataset pass input validation, and that more than $4{,}000$ problems misclassify nearly all correct submissions as incorrect. Our own augmentation analysis found the same failure mode: among the 257 problems in a random 300-problem sample from the $3{,}928$ execution-validated pool for which generated correctness tests were available, 81 already had false-positive rates above $5\%$ on the original tests alone. Thus raw DMC is limited both as a source of timing signal and as a reliable correctness gate.

\paragraph{Deduplication and quality filtering.}
To reduce leakage between training and evaluation, we deduplicate at the problem level against evaluation benchmarks such as LiveCodeBench~\citep{jain2024livecodebench}. Matching is performed on problem descriptions. We also remove malformed, missing, or ambiguous problem statements identified by an instruction-tuned model. The deduplication stage and quality filtering stage reduce the corpus from $12{,}275$ to $11{,}468$ problems.

\subsection{Overview of DMC-Optim Construction Stages}
\label{sec:app_construction_stages}

The DMC-Optim pipeline produces two related endpoints. The first is a $2{,}723$-problem cleaned corpus with validated solutions, generated correctness tests, and generated optimization tests. The second is the RL pool used in the main paper: after selecting problems whose optimization tests are duration-filterable, we obtain $1{,}302$ problems and resplit them into $1{,}000$ training and 302 test problems. \Cref{tab:pipeline_overview} gives the high-level reduction. The table should be read as a sequence of claims rather than only as attrition: stages 1--3 establish trustworthy solution controls, stages 4--6 add and filter the generated tests, and stages 7--8 impose the additional measurement requirement needed by optimization RL.

\begin{table}[h!]
\centering
\small
\caption{\textbf{Dataset construction pipeline.}
Each row shows a processing stage and the number of problems remaining afterward.
The pipeline yields two related artifacts: a $2{,}723$-problem intermediate cleaned dataset, and the $1{,}302$-problem duration-filterable pool used for optimization RL.
The experiments in this paper resplit that pool into $1{,}000$ training and 302 test problems.
}
\label{tab:pipeline_overview}
\begin{tabular}{clr}
\toprule
{\bfseries Stage} & {\bfseries Operation} & {\bfseries Problems} \\
\midrule
0 & Raw DMC data & 12{,}275 \\
1 & Deduplication against evaluation benchmarks & 11{,}468 \\
2 & Require both correct \& incorrect solutions & 6{,}706 \\
3 & Execution-based solution validation & 3{,}928 \\
4 & Test augmentation \& quality filtering & 3{,}061 \\
5 & Keep problems retained by all generated-test filtering passes & 2{,}978 \\
6 & Require non-empty correctness and optimization tests & 2{,}723 \\
7 & Select duration-filterable problems & 1{,}302 \\
8 & Resplit into RL train / test & 1{,}000 / 302 \\
\bottomrule
\end{tabular}
\end{table}

\paragraph{Solution-type filtering.}
We require every retained problem to have both correct and incorrect Python solutions. Problems with only correct solutions have no negative controls, so the test suite cannot be checked for false positives. Problems with only incorrect solutions have no positive controls, so generated tests cannot be validated by executing known-correct code. Problems with no solutions cannot be used for construction or training. This step removes $2{,}210$ correct-only problems, $1{,}045$ incorrect-only problems, and $1{,}507$ problems with no usable solutions, leaving $6{,}706$ problems.

\paragraph{Execution-based validation.}
We re-execute retained solutions against the original public, private, and DMC-generated tests. This prevents stale labels from being amplified during test generation. Correct solutions that fail re-execution are removed from the verified-correct pool; incorrect solutions are retained as negative controls. After this stage, $3{,}928$ problems remain with validated solutions.

\paragraph{From validated problems to the intermediate cleaned corpus.}
The generated-test stages in \cref{sec:app_test_generation,sec:app_filtering_generated_tests} leave $3{,}061$ problems after test augmentation and quality filtering. Restricting to problems that survive every generated-test filtering pass leaves $2{,}978$ problems. We then keep only problems where the generated correctness tests and generated optimization tests are both present, and remove 255 problems where one of these generated test sets is empty after filtering. This leaves a $2{,}723$-problem cleaned corpus with both generated correctness tests and generated optimization tests for every retained problem. Having optimization tests is not the same as being usable for timing-based RL: many retained cleaned-corpus problems are later rejected as not duration-filterable, even though their optimization tests are present. This cleaned corpus also keeps the original tests and remains useful for future work that does not require duration filterability.

\paragraph{From the cleaned corpus to the RL pool.}
For optimization RL, duration filterability is applied to the full $2{,}723$-problem cleaned corpus. It keeps problems where the optimization tests span enough workload for timing to be informative, yielding $1{,}302$ duration-filterable problems and a complementary pool of $1{,}421$ non-duration-filterable problems. We then resplit the $1{,}302$ duration-filterable problems into the final DMC-Optim RL train/test split with $1{,}000$ training problems and 302 test problems. Across this final duration-filterable train/test pool, DMC-Optim stores $176{,}249$ verified correct human reference solutions together with duration annotations on $607{,}849$ tests; the final train and test sets contain $127{,}841$ and $48{,}408$ reference solutions after incorrect solutions are removed from the training artifact.
The $1{,}169$ duration-filterable number reported later in \cref{tab:filterability} refers only to a seed-42 pre-resplit $2{,}423$-problem training partition used for intermediate diagnostics. The final construction instead applies duration filterability to the full $2{,}723$-problem corpus, yielding the $1{,}302$-problem pool that is resplit into the final $1{,}000/302$ DMC-Optim RL split.

This duration-filterability selection separates dataset cleanliness from optimization readiness. Both the $1{,}302$ duration-filterable problems and the $1{,}421$ non-duration-filterable complement come from the same $2{,}723$-problem cleaned corpus; the difference is not whether tests exist, but whether the optimization tests create enough duration spread to support a timing reward.

\subsection{Test generation}
\label{sec:app_test_generation}

\paragraph{Why generate tests at all.}
The cleaned corpus still inherits the main DMC limitation: original tests are small and often fail to expose incorrect submissions. Test generation is therefore not an optional scale-up step; it is the mechanism that turns verified human solutions into a stronger pass/fail gate and, separately, into larger workloads for timing. The construction deliberately asks the model only for input generators, not for expected outputs, so that generated inputs can be checked against trusted human code rather than against another model completion.

\paragraph{Generation setup.}
We generate additional tests with a supervised fine-tuned \textsc{CWM} 32B checkpoint. For each problem, we sample 10 independent input-generator candidates. The prompt contains the full problem statement and one randomly selected verified human solution. The model is not asked to produce expected outputs directly. It emits a Python \texttt{InputGenerator} class, which the environment executes to produce candidate inputs. This design matters: one model-generated \texttt{InputGenerator} class can expand into many candidate inputs, while the output side remains grounded in trusted reference code.

The supervised checkpoint is also a conservative choice for diversity. The generator's job is not to consistently solve the task in a one-shot fashion; it is to produce varied, constraint-respecting input programs under repeated sampling. Using the SFT checkpoint reduces the risk that the generator has already collapsed toward a narrow reward-driven pattern on the task, which would undercut the purpose of sampling multiple candidate test generators.

This design is close in spirit to CodeContests+~\citep{codecontestsplus}, which also uses LLM-written generators to strengthen DMC-style test suites, but the target is different. A binary time-limit-exceeded gate is not enough for our reward design: optimization RL needs tests that create a range of durations so correct solutions can be ranked by efficiency, not only classified as passing or timing out. We therefore store reproducible input/output pairs and later remove ambiguous problems rather than rely on custom checkers.

\paragraph{InputGenerator task environment.}
The environment implements two campaign variants. The correctness campaign repeatedly samples \texttt{edge}, \texttt{small}, and \texttt{sample} cases. The optimization campaign swaps \texttt{small} for \texttt{large}. Each attempt executes 15 method calls in total, so one problem can yield up to 150 candidate tests across the 10 sampled generators. \Cref{tab:iotestgen_pseudocode} summarizes the acceptance loop before dataset-level filtering.

\begin{table}[h!]
\centering
\small
\caption{\textbf{Pseudo-code of the input-generator environment.}
A sampled test set is kept only if the generator executes successfully and at least two verified human solutions agree on all produced outputs.
The later dataset-level filtering stage enforces stricter agreement and removes tests, solutions, or whole problems when the generated suite is ambiguous.
}
\label{tab:iotestgen_pseudocode}
\begin{tabular}{p{0.97\linewidth}}
\toprule
{\bfseries Input:} problem statement $P$, prompt solution $s_{\mathrm{prompt}}$, verification pool $\mathcal{S}$ with up to 10 verified human solutions, campaign mode $m$. \\
\midrule
\textbf{1.} Prompt the model with $P$ and $s_{\mathrm{prompt}}$ to generate a Python \texttt{InputGenerator} class. \\
\textbf{2.} Choose the call schedule: correctness uses $(\texttt{edge}, \texttt{small}, \texttt{sample}) \times 5$; optimization uses $(\texttt{edge}, \texttt{large}, \texttt{sample}) \times 5$. \\
\textbf{3.} Execute the generated class and collect 15 candidate input strings. If the generator crashes or cannot be parsed, reject the attempt. \\
\textbf{4.} Run each $s \in \mathcal{S}$ on the 15 candidate inputs and record the produced outputs. \\
\textbf{5.} Reject the attempt unless at least two solutions produce the exact same 15-output list. \\
\textbf{6.} Pair each accepted input with the consensus output, re-execute one agreeing solution on the resulting input/output pairs, and keep the test set only if this final verification succeeds. \\
\bottomrule
\end{tabular}
\end{table}

\paragraph{Generation campaigns.}
We run two campaigns. The first produces additional correctness tests on the execution-validated pool. After these tests are inserted and the dataset is filtered once, the optimization campaign runs on the cleaner $3{,}061$-problem pool and targets large instances near the problem constraints. \Cref{tab:test_gen_stats} reports both campaigns. The optimization acceptance rate should not be interpreted as large-input generation being intrinsically easier; this campaign runs after the first filtering pass has already removed many problematic problems and mislabeled solutions.

\begin{table}[h!]
\centering
\small
\caption{\textbf{Test generation statistics by campaign.}
Each problem receives 10 independent \texttt{InputGenerator} samples, and each accepted sample contributes 15 validated input/output pairs.
With 10 attempts, the fraction of problems receiving at least one accepted test set is pass@10.
}
\label{tab:test_gen_stats}
\resizebox{\linewidth}{!}{
\begin{tabular}{lcccccc}
\toprule
{\bfseries Campaign} & {\bfseries Pool} & {\bfseries Problems} & {\bfseries Problems with} $\geq 1$ {\bfseries accepted set} & {\bfseries Pass@1} & {\bfseries Total accepted tests} & {\bfseries Mean tests / problem} \\
\midrule
\textbf{Correctness} & execution-validated & 3{,}929 & 3{,}547 (90.30\%) & 73.00\% & 430{,}215 & 109.50 \\
\textbf{Optimization} & post-filtered & 3{,}061 & 2{,}854 (93.24\%) & 76.82\% & 352{,}740 & 115.24 \\
\bottomrule
\end{tabular}
}
\end{table}

The accepted test counts translate into a substantial workload shift in the retained RL pool. In the final $1{,}000$-problem training split, optimization tests have median input-plus-output size 928 characters and mean human-reference runtime $0.334\,\text{s}$, compared with 36 characters and $0.137\,\text{s}$ for retained correctness tests. The separation is not only that optimization tests are more numerous; they occupy a different part of the feasible workload range.

At this stage the pipeline has produced more tests, not yet a final dataset. Accepted generator attempts can still contain inputs that are too slow for correctness checking, violate hidden constraints, or reveal that a supposedly correct solution was mislabeled. The next subsection is therefore the adjudication step that decides which tests, solutions, and problems survive.

\paragraph{Generation prompt.}
\Cref{fig:iotestgen_prompt} shows a representative prompt skeleton. The real prompt contains the full problem statement and one randomly selected verified human solution, then asks the model to emit only imports and the \texttt{InputGenerator} class. Each method call returns one randomized input string, so repeatedly calling the same generated class yields a small test campaign rather than one static example.

\begin{figure}[h!]
\centering
\fbox{
\begin{minipage}{0.96\linewidth}
\scriptsize
\textbf{Instruction.} Write a Python \texttt{InputGenerator} class with methods \texttt{generate\_edge\_cases}, \texttt{generate\_large\_cases}, \texttt{generate\_small\_cases}, and \texttt{generate\_sample\_cases}. Each call must return one randomized input string that respects the problem constraints. \\
\textbf{Problem statement.} [Full competitive-programming description inserted here.] \\
\textbf{Reference human solution.} [One verified Python solution inserted here.] \\
\textbf{Output format.} Return only the necessary imports and the \texttt{InputGenerator} class inside a Python code block; do not include example usage or auxiliary text.
\end{minipage}
}
\caption{\textbf{Representative prompt used to synthesize an \texttt{InputGenerator} class.}
The real prompt contains the entire problem statement and one randomly selected verified human solution.
The model only proposes inputs; outputs are inferred later from agreement across human solutions.
}
\label{fig:iotestgen_prompt}
\end{figure}

The high-level placement of test generation is already shown in \cref{fig:adjudication_pipeline}, while \cref{tab:pipeline_overview} gives the corresponding problem-count reductions. The details that are specific to generation are therefore kept in the prompt skeleton and in \cref{tab:iotestgen_pseudocode,tab:test_gen_stats}: the model proposes input generators, human solutions define the outputs, and the two campaigns separately target correctness coverage and optimization workloads.

\subsection{Filtering generated tests}
\label{sec:app_filtering_generated_tests}

The generators produce candidate tests, not final evaluation suites. When a newly generated test is executed against verified correct solutions and some of them fail, the failure can mean three different things. The test may be invalid because it violates constraints or triggers unrelated runtime errors. The test may be valid and the failing solution may be a newly discovered false positive. Or the problem may admit multiple valid outputs for the same input, making strict input/output comparison unsuitable without a custom checker. The pipeline has to navigate this ambiguity across tests, solutions, and whole problems.

This is the central data-quality tension in DMC-Optim. Aggressive filtering produces cleaner labels but can shrink the pool below the scale needed for RL. It can also bias the retained pool toward easier problems: if every problem where generated tests expose many false positives or many positive-control failures is removed, the remaining dataset may mostly contain problems whose original tests were already adequate, with few edge cases where solutions sit near the correctness boundary, leaving little useful signal for RL. Conversely, weak filtering preserves scale but leaves the reward exposed to mislabeled solutions or ambiguous tests. We therefore filter at three levels and tune thresholds through manual case studies on small problem subsets, while preserving enough duration-filterable problems for the final RL split.

The post-generation filtering path corresponds to the test-augmentation and quality-filtering stage in \cref{fig:adjudication_pipeline} and to stages 4--6 in \cref{tab:pipeline_overview}. For the correctness-test branch, distributed re-execution outputs are joined back to the augmented problem records by problem identifier, then the pipeline removes pathological generated tests, removes solutions whose labels are contradicted by the filtered suite, and finally removes whole problems whose remaining labels are still too weak. In parallel, raw pass-rate diagnostics and manual cases guide threshold selection. Those diagnostics are useful for understanding what the generated tests expose, but the final training set uses the filtered tests and filtered solution labels.

\paragraph{Filtering parameters and thresholds.}
\Cref{tab:filtering_params} lists the thresholds as a compact reference. The following paragraphs define each rate and cap in the context where it is used: first the test-level filters, then solution-level classification, and finally problem-level filtering. The correctness-cleaning round is stricter because it is responsible for relabeling solutions, removing ambiguous problems, and reducing false positives. The later optimization round relaxes timeout handling because large-input timeouts are part of the efficiency signal rather than necessarily evidence of an invalid test.

\begin{table}[h!]
\centering
\small
\caption{\textbf{Post-processing filtering parameters.}
The table is a compact index of the thresholds; the paragraphs that follow define the denominators, caps, and decision rules in context.
The correctness round is the main correctness-cleaning stage.
The optimization round reuses the same exception filtering and correct-solution drop cap, but relaxes timeout handling because large-input timeouts are part of the efficiency signal rather than necessarily an error.
}
\label{tab:filtering_params}
\begin{tabular}{lcc}
\toprule
{\bfseries Parameter} & {\bfseries Correctness} & {\bfseries Optimization} \\
\midrule
Min correct solutions & 5 & 1 \\
Min incorrect solutions & 1 & 1 (all retained) \\
Max correct-solution drop rate & 25\% & 25\% \\
Max false-positive rate & 5\% & 100\% \\
Timeout fraction threshold & 20\% & 90\% \\
Timeout treated as success & no & yes \\
Timeout removal cap & 25\% & 25\% \\
Exception fraction threshold & 40\% & 40\% \\
Exception/failure removal cap & 30\% & 30\% \\
\bottomrule
\end{tabular}
\end{table}

\paragraph{Implementation invariants.}
The filters never remove original public, private, or DMC-generated tests; they only remove columns from the newly generated test suite. When more generated tests exceed a removal threshold than the cap allows, the tests are sorted by the offending fraction and the worst tests, meaning those that create the most positive-control failures, are removed first. The exception/failure filter excludes syntax-error outcomes, so it targets generated inputs that make verified-correct solutions fail at runtime or produce wrong outputs rather than solution-parser failures. Within each test-removal stage, the filtered pass criterion is monotone: removing generated tests can only keep or increase the set of solutions that pass the filtered suite. The problem-level gates then decide whether any remaining false-positive or correct-solution attrition is acceptable.

\paragraph{Test-level filters.}
For correctness tests, a timeout is evidence that the generated input is too aggressive to serve as a reliable pass/fail check. We remove tests on which more than 20\% of correct solutions exceed the time limit, and tests on which more than 40\% raise runtime exceptions or wrong-output failures. For optimization tests, large inputs are intentional. A timeout can be exactly the signal that a test probes a slower part of the valid input space. We therefore tolerate timeout rates up to 90\% on an optimization test. The test becomes uninformative only when almost all correct solutions time out, because then it no longer separates faster from slower implementations. The timeout filter can remove at most 25\% of a problem's generated tests, while the later exception/failure filter can remove at most 30\% of the remaining generated tests. These caps prevent a small number of bad samples from erasing an otherwise useful suite, while still removing the tests most likely to be invalid.

\paragraph{Solution-level classification.}
After test-level filtering, the filtered augmented suite is used to update the positive and negative solution pools. A verified-correct solution remains a positive control only if it passes the filtered suite; otherwise it is removed from the retained correct-solution pool. An incorrect solution remains a negative control only if it still fails the filtered suite; if it passes the stronger suite, it is no longer a reliable negative example and is removed from the retained incorrect-solution pool. This is the step that turns generated-test evidence into the solution-count changes reported in \cref{sec:app_before_duration_selection}. If generation produces no usable tests for a problem, this step cannot repair the original test suite; high-false-positive problems in that category can only be removed later by the problem-level gate.

\paragraph{Problem-level filters.}
In the correctness-cleaning round, a retained problem must keep at least 5 verified correct solutions, lose at most 25\% of its original correct solutions, and have false-positive rate below 5\% after augmentation. The false-positive rate is computed over the original incorrect-solution pool: if more than 5\% of those negative controls pass the augmented suite, the problem is removed rather than made to look clean by deleting most of its hard negative examples. This removes ambiguous multi-output problems, mislabeled correct solutions, and weak original suites. Problems that make too many correct solutions fail across many generated tests are likely unsuitable for strict input/output evaluation. Problems that still let too many incorrect solutions pass cannot provide reliable reward signal. In the later optimization round, the problem has already passed the stricter correctness-cleaning stage, so optimization tests are not used as the primary false-positive filter. We still cap correct-solution attrition at 25\% and require at least one surviving correct solution, but we let timeouts function as duration outcomes rather than as immediate correctness verdicts.

\begin{figure}[h!]
\centering
\includegraphics[width=\textwidth]{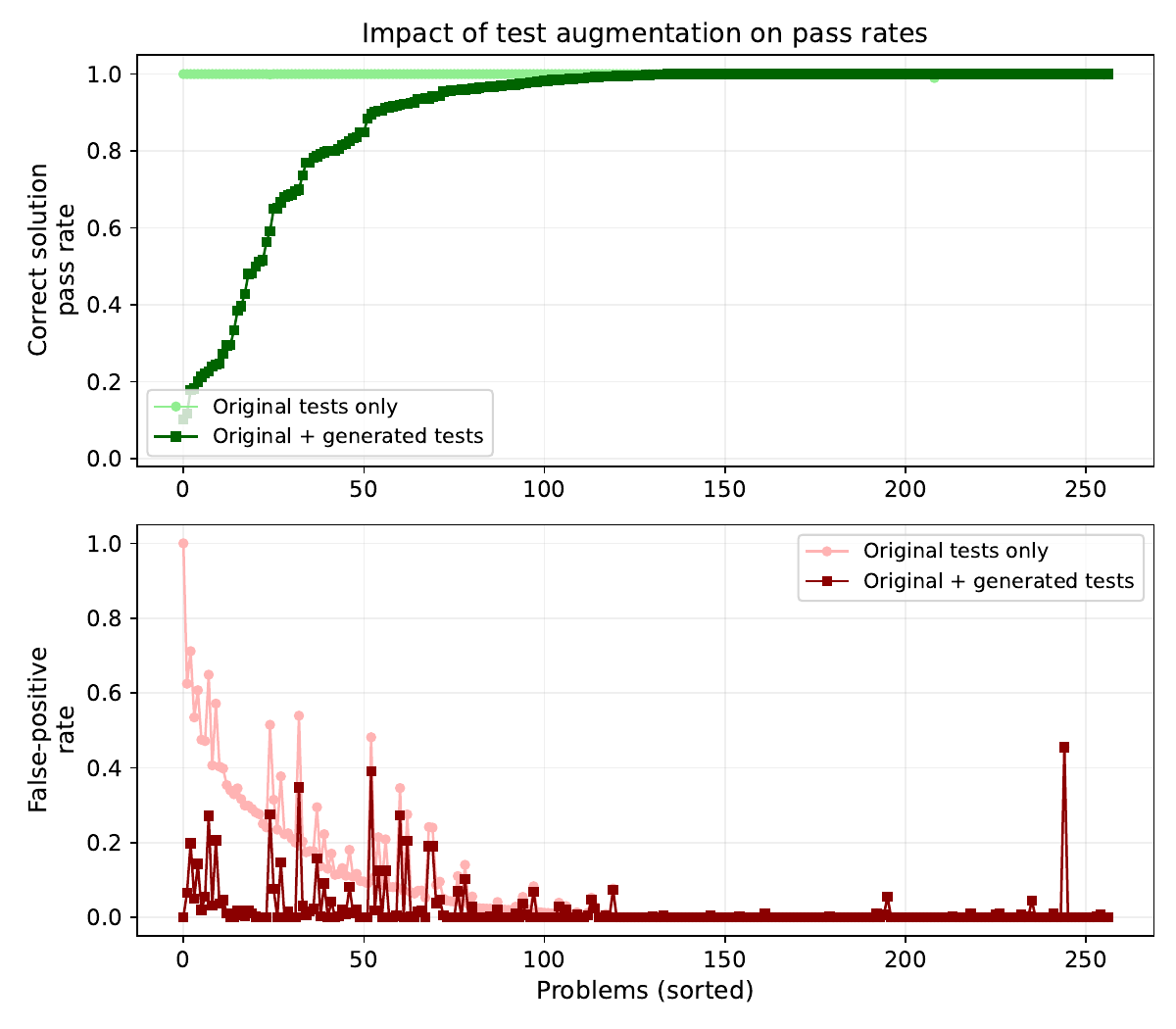}
\caption{\textbf{Generated correctness tests reduce false positives on a random 300-problem sample from the execution-validated pool.}
\emph{Top:} correct-solution pass rates before and after adding generated tests, for the 257 problems where generated tests were available.
The post-augmentation curve remains near one for many problems but drops on cases where newly generated tests expose bugs or invalid generated tests before filtering.
\emph{Bottom:} false-positive rates, defined as the fraction of incorrect solutions that pass all tests.
Mean false-positive rate drops from 8.27\% to 2.11\%; the number of problems above the 5\% threshold drops from 81 to 27; and no problem has a higher false-positive rate after adding tests.
These curves use raw generated-test outcomes, not the final training set.
}
\label{fig:test_augmentation_impact}
\end{figure}

\begin{figure*}[t!]
\centering
\begin{minipage}{0.43\textwidth}
\centering
\includegraphics[width=\linewidth]{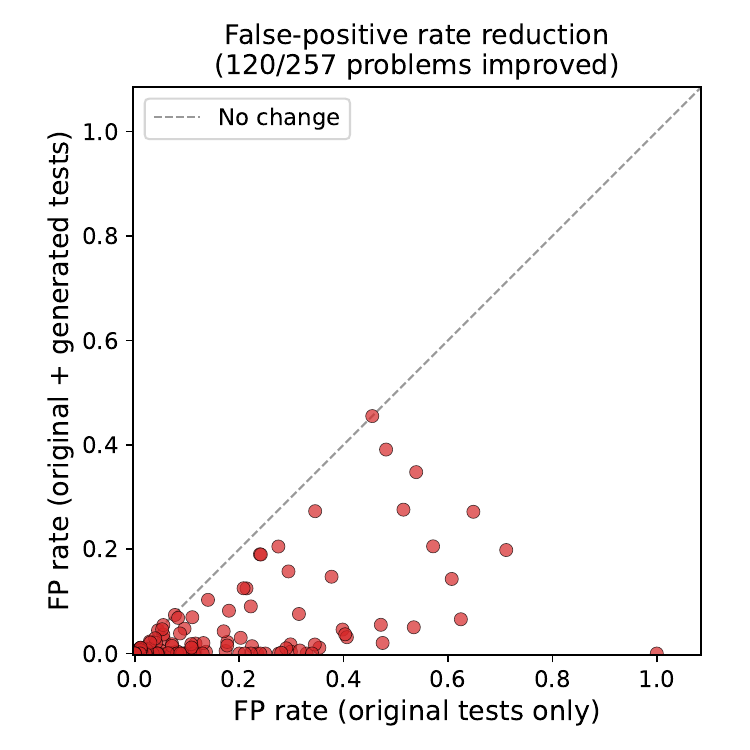}
\end{minipage}
\hfill
\begin{minipage}{0.55\textwidth}
\centering
\includegraphics[width=\linewidth]{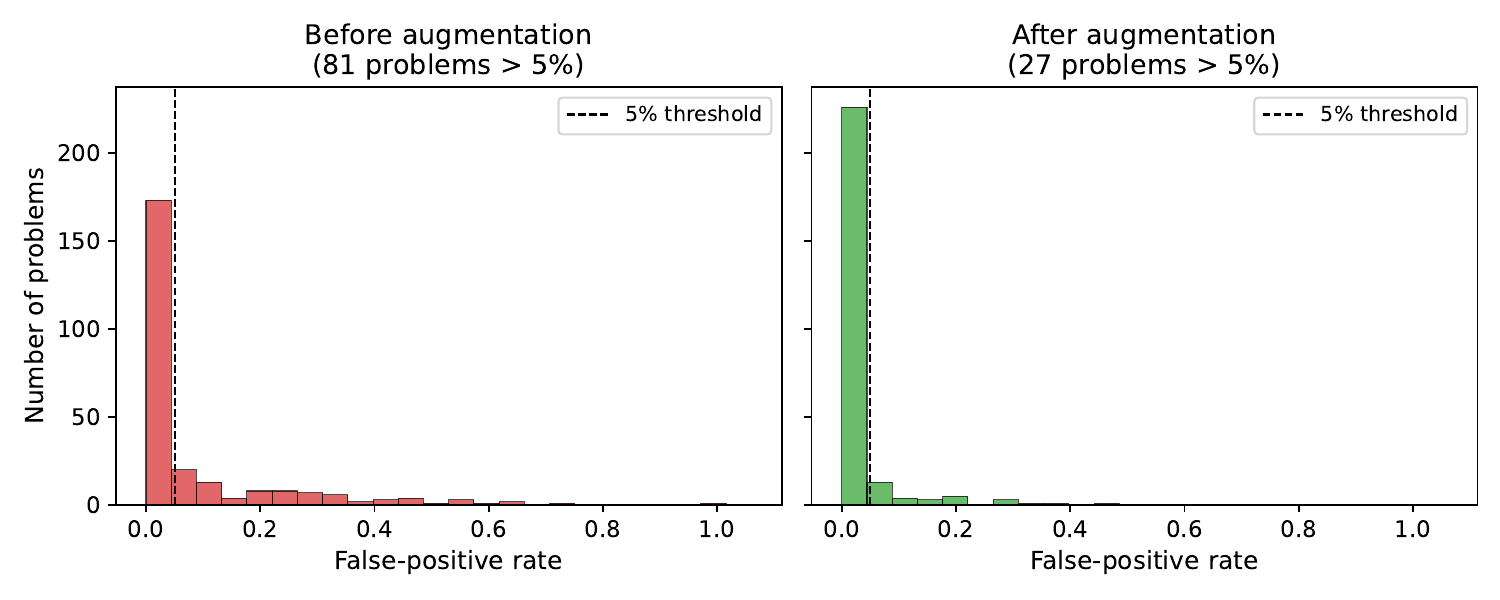}
\end{minipage}
\caption{\textbf{False-positive reductions are broad rather than driven by one outlier.}
\emph{Left:} Each point compares one problem's false-positive rate on original tests and on original plus generated tests; points below the diagonal improve, and 120 of 257 generated-test problems improve.
\emph{Right:} The false-positive distribution shifts left after augmentation, with the number of problems above the 5\% removal threshold falling from 81 to 27.
The remaining high-false-positive problems are handled by the problem-level filter rather than kept in the final training artifact.
}
\label{fig:test_augmentation_fp_300}
\end{figure*}

\begin{figure}[t!]
\centering
\includegraphics[width=0.5\textwidth]{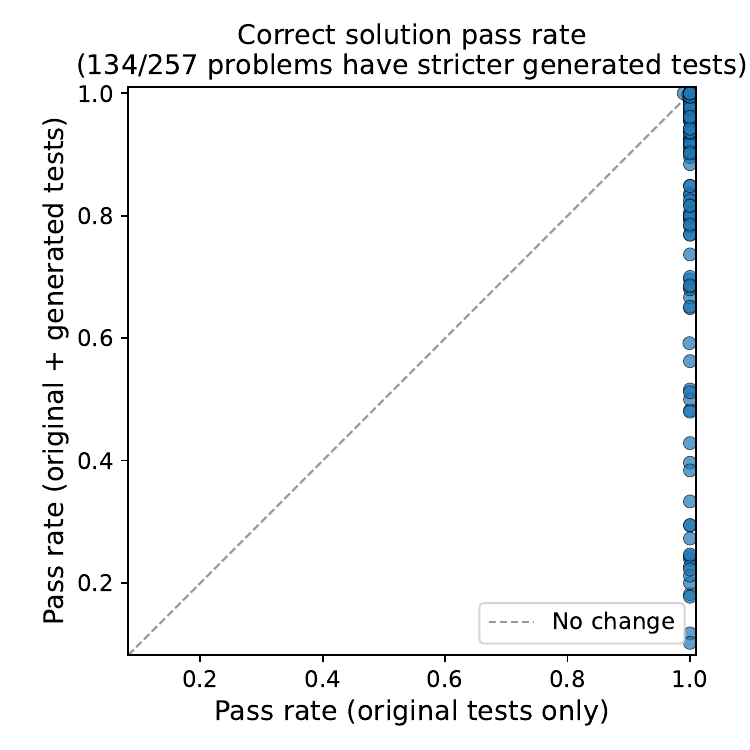}
\caption{\textbf{Correct-solution attrition from generated correctness tests on a random 300-problem sample from the execution-validated pool.}
Each point compares one generated-test problem before and after augmentation.
The x-axis is concentrated near one because the preceding execution-validation stage already checked that retained positive controls pass the original public, private, and DMC-generated tests.
Most of the visible variation is therefore vertical: it measures the additional attrition introduced by generated correctness tests.
Most problems remain near one, but 134 of 257 problems have stricter generated tests that reduce the positive-control pass rate.
This scatter is computed before the timeout, exception/failure, and problem-level filters, so it shows why the filtering caps are needed rather than the final retained training artifact.
}
\label{fig:test_augmentation_correct_pr_300}
\end{figure}

\paragraph{300-problem sample: false-positive reduction.}
\Cref{fig:test_augmentation_impact,fig:test_augmentation_fp_300,fig:test_augmentation_correct_pr_300} show the effect of generated correctness tests on a random 300-problem sample from the $3{,}928$ execution-validated pool. Generated tests were available for 257 of these 300 problems. On those 257 problems, mean false-positive rate drops from 8.27\% to 2.11\%, the median false-positive rate moves from 0.60\% to 0.00\%, and 49 problems go from a nonzero false-positive rate to no observed false positives. The number of problems above the 5\% false-positive threshold drops from 81 to 27; the counts above 10\% and 20\% drop from 60 to 17 and from 43 to 8. False-positive rate is reduced or unchanged on 120 and 137 problems respectively, with no increases. The cost is stricter positive control: mean correct-solution pass rate drops from 99.99\% to 89.98\%. These figures use raw generated-test pass/fail outcomes, before the timeout and exception/failure filters above, so they measure what the generated tests expose. The final training set is the more conservative artifact after test-level, solution-level, and problem-level filtering.

\paragraph{300-problem sample: correct-solution pass-rate cost.}
The false-positive reduction is not free. On the same 257 generated-test problems, correct-solution pass rate drops on 134 problems, is unchanged on 121, and increases on 2 because of timeout-related re-execution stochasticity. Among the dropped problems, the mean drop is 19.21 percentage points, the median drop is 6.40 points, and the largest drop is 89.77 points. The tail motivates the 25\% problem-level correct-solution drop cap: 53 problems drop by more than 10 points, 34 by more than 25 points, and 20 by more than 50 points before problem-level filtering.

\begin{table}[h!]
\centering
\small
\caption{\textbf{Correct-solution attrition before filtering on a random 300-problem sample from the execution-validated pool.}
The table uses raw original-plus-generated test outcomes on the 257 problems with generated correctness tests.
It is a diagnostic for threshold selection, not the final filtered training artifact.
}
\label{tab:correct_pr_attrition_300}
\begin{tabular}{lr}
\toprule
{\bfseries Statistic} & {\bfseries Value} \\
\midrule
Generated-test coverage & 257 / 300 problems \\
Mean correct pass rate & 99.99\% \(\to\) 89.98\% \\
Problems with lower / unchanged / higher correct pass rate & 134 / 121 / 2 \\
Mean / median drop among dropped problems & 19.21pp / 6.40pp \\
Largest drop & 89.77pp \\
Problems dropping by \(>10\)pp / \(>25\)pp / \(>50\)pp & 53 / 34 / 20 \\
\bottomrule
\end{tabular}
\end{table}

\begin{table*}[t!]
\centering
\small
\caption{\textbf{Largest false-positive reductions in a random 300-problem sample from the execution-validated pool.}
These examples show that the aggregate false-positive reduction is not driven only by small-solution-count cases.
Some problems improve with no loss of correct-solution pass rate, while others are later removed because the generated tests are too strict for positive controls.
}
\label{tab:largest_fp_reductions_300}
\resizebox{\textwidth}{!}{
\begin{tabular}{lrrrrr}
\toprule
{\bfseries Problem} & {\bfseries FP before} & {\bfseries FP after} & {\bfseries Reduction} & {\bfseries Correct PR before \(\to\) after} & {\bfseries Correct / incorrect sol.} \\
\midrule
train/532 & 100.00\% & 0.00\% & 100.00pp & 100.0\% \(\to\) 92.3\% & 13 / 1 \\
train/814 & 62.46\% & 6.56\% & 55.90pp & 100.0\% \(\to\) 95.5\% & 179 / 1{,}753 \\
train/395 & 71.16\% & 19.82\% & 51.33pp & 100.0\% \(\to\) 89.6\% & 173 / 7{,}627 \\
train/293 & 53.45\% & 5.03\% & 48.42pp & 100.0\% \(\to\) 100.0\% & 192 / 1{,}233 \\
train/609 & 60.71\% & 14.29\% & 46.43pp & 100.0\% \(\to\) 42.9\% & 14 / 56 \\
train/782 & 47.47\% & 2.02\% & 45.45pp & 100.0\% \(\to\) 100.0\% & 11 / 99 \\
train/709 & 47.13\% & 5.50\% & 41.64pp & 100.0\% \(\to\) 96.0\% & 126 / 3{,}019 \\
train/921 & 64.86\% & 27.15\% & 37.71pp & 100.0\% \(\to\) 100.0\% & 180 / 3{,}381 \\
train/733 & 40.62\% & 3.12\% & 37.50pp & 100.0\% \(\to\) 24.7\% & 73 / 32 \\
train/862 & 57.14\% & 20.54\% & 36.61pp & 100.0\% \(\to\) 100.0\% & 65 / 112 \\
\bottomrule
\end{tabular}
}
\end{table*}

\begin{figure*}[t!]
\centering
\includegraphics[width=\textwidth]{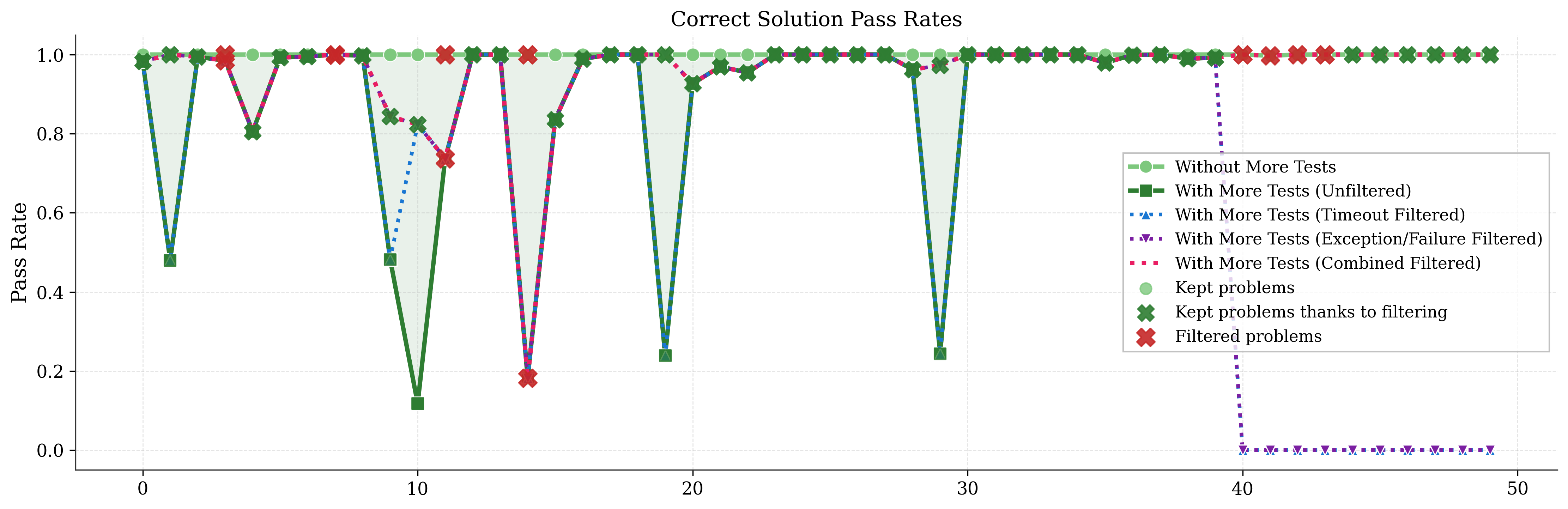}
\vspace{0.4em}
\includegraphics[width=\textwidth]{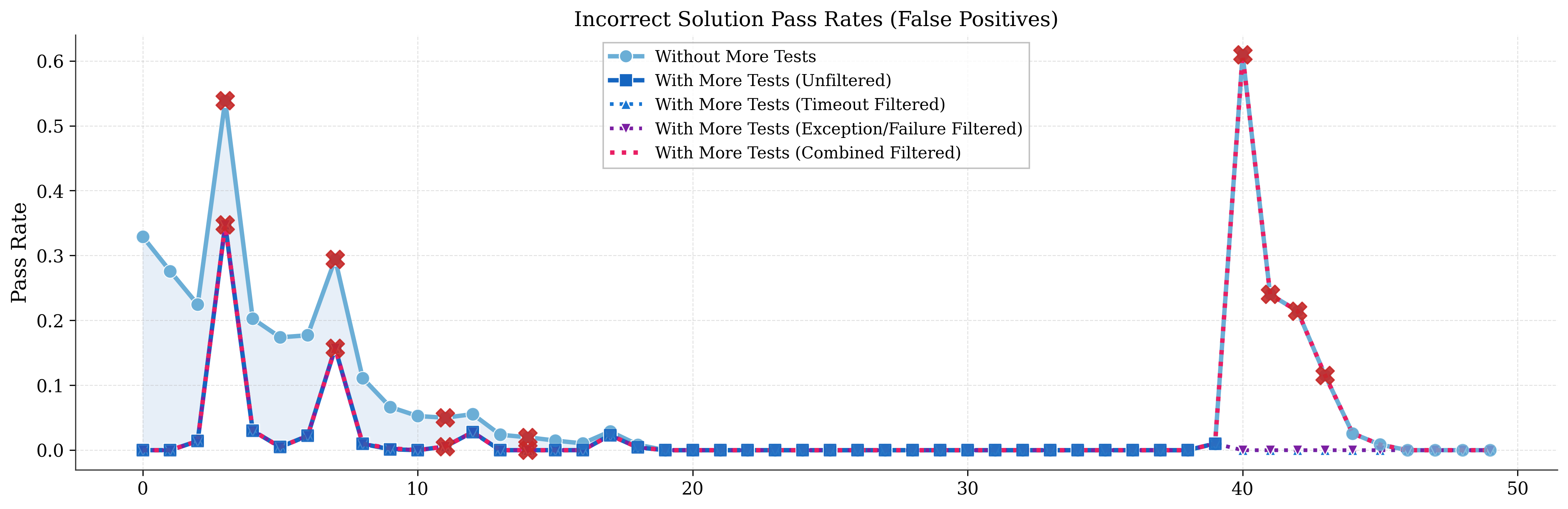}
\caption{\textbf{Fifty-problem mechanism diagnostics for positive-control and negative-control pass rates.}
Problems appear in the same order in both panels, sorted by false-positive reduction.
\emph{Top:} The light-green curve is the pass rate of verified-correct solutions on the original tests.
The dark-green curve adds the raw generated correctness tests.
The blue, purple, and magenta curves then apply timeout filtering, exception/failure filtering, and the final combined filtered suite.
The final combined filtered curve is the retained positive-control pass rate after test-level filtering, before solution-level reclassification and problem-level filtering.
When filtered curves separate upward from the raw generated-test curve, the filters recovered positive controls by removing generated tests that failed too many of them; when the curves overlap, filtering did not change the aggregate verdict for that problem.
Green crosses mark problems kept thanks to filtering, while red crosses mark problems removed by the later problem-level gates.
\emph{Bottom:} The same stages are shown for originally incorrect solutions, so lower values mean fewer false positives.
The shaded area is the false-positive mass removed by raw generated tests.
When filtered curves move upward relative to the raw generated-test curve, recovering agreement with positive controls also allowed some incorrect solutions to pass again; when all generated-test curves overlap near zero, the generated tests reject the negative controls without needing test removal.
}
\label{fig:test_augmentation_diagnostics_50}
\end{figure*}

\begin{figure*}[p]
\centering
\includegraphics[width=0.80\textwidth]{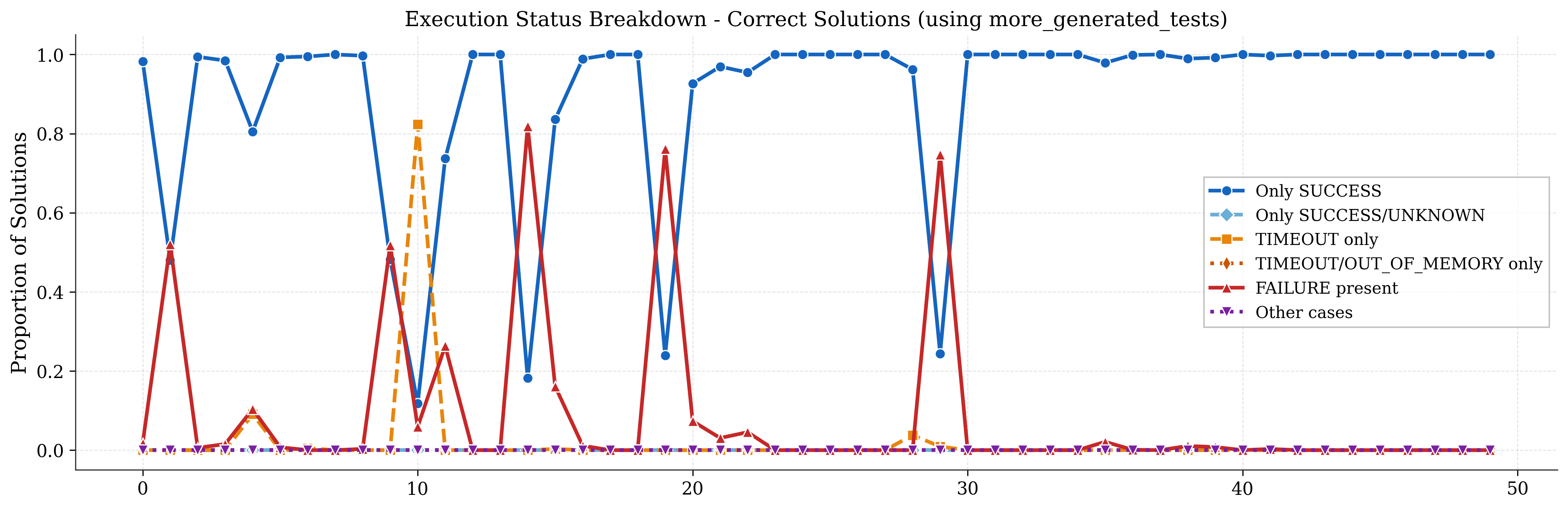}
\vspace{-0.2em}

\includegraphics[width=0.80\textwidth]{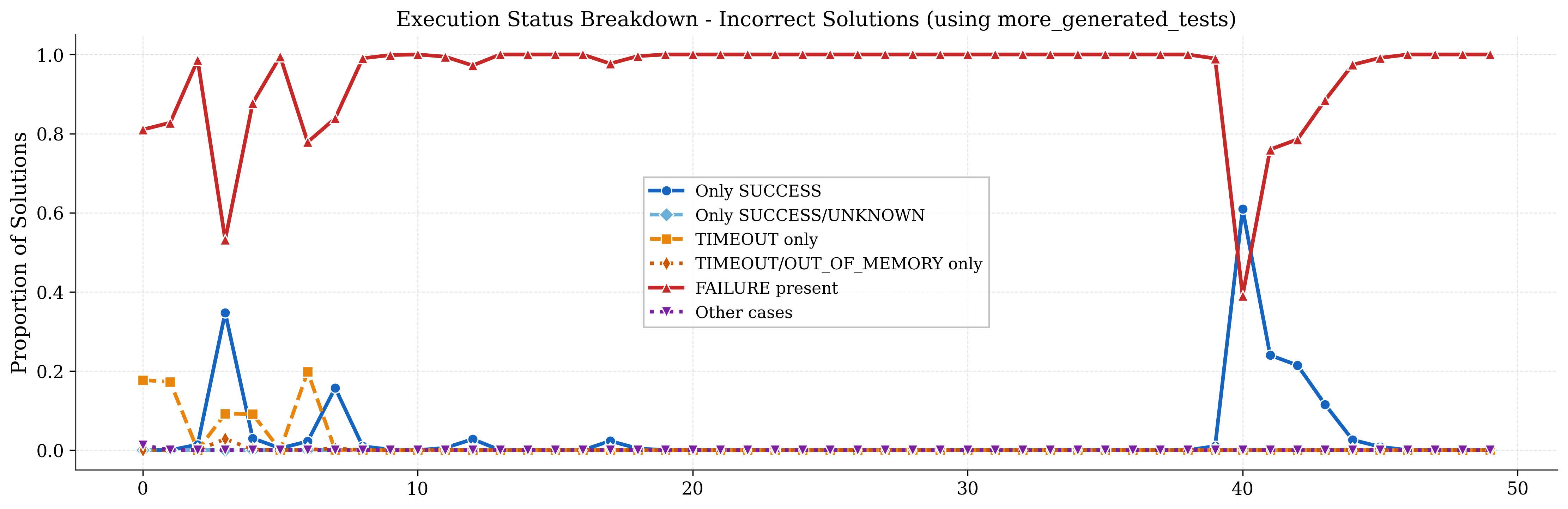}
\vspace{-0.2em}

\includegraphics[width=0.80\textwidth]{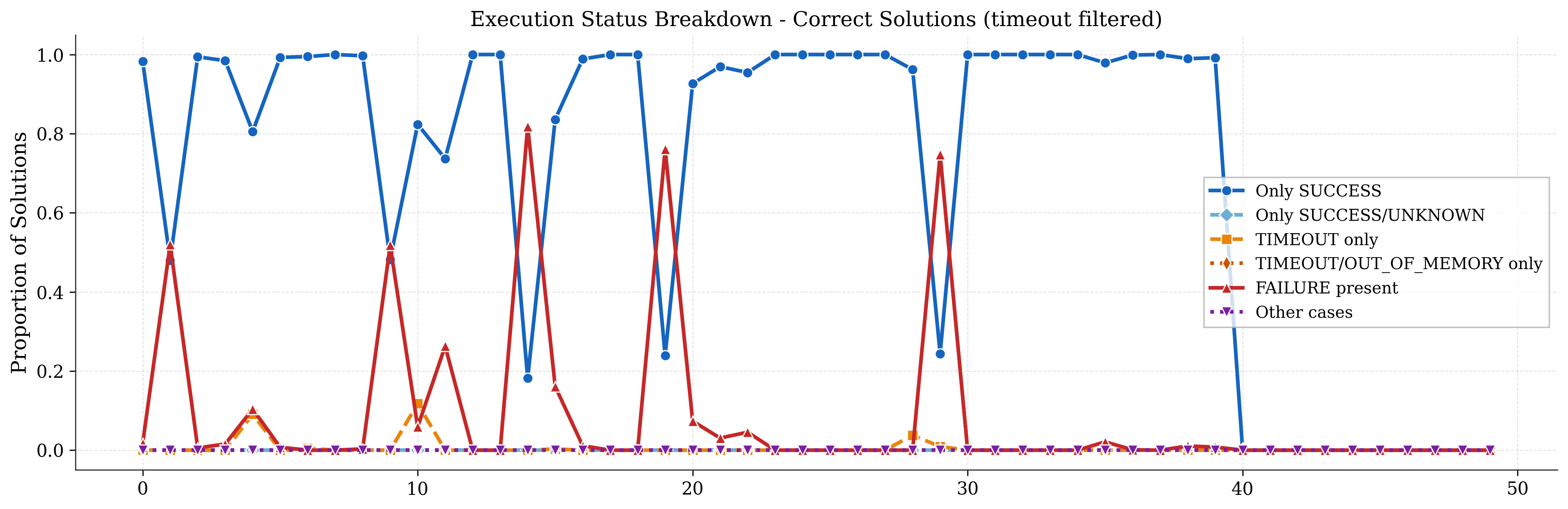}
\vspace{-0.2em}

\includegraphics[width=0.80\textwidth]{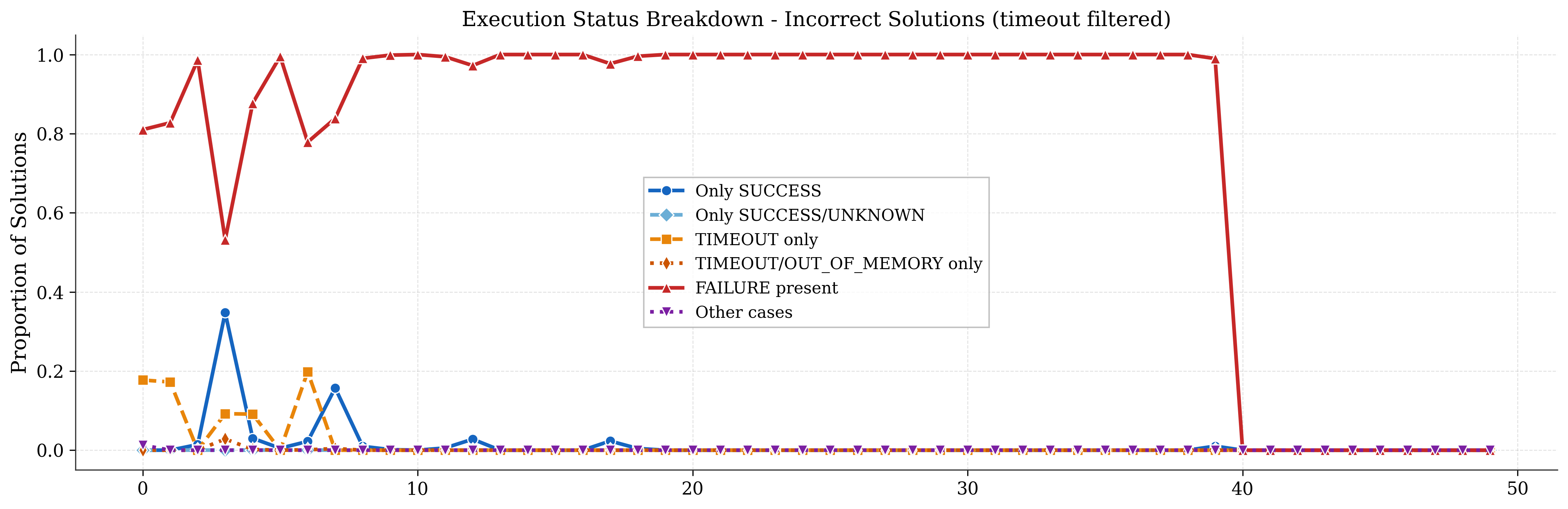}
\caption{\textbf{Execution-status diagnostics before and after timeout filtering.}
All panels use the same problem order as \cref{fig:test_augmentation_diagnostics_50}.
From top to bottom: \emph{(a)} raw generated-test executions on verified-correct solutions; \emph{(b)} raw generated-test executions on originally incorrect solutions; \emph{(c)} executions on verified-correct solutions after removing timeout-heavy generated tests; and \emph{(d)} executions on originally incorrect solutions after the same timeout filtering.
Each curve is the fraction of solutions in a problem whose generated-test executions fall into one status pattern: all successes, success/unknown only, timeout only, timeout/out-of-memory only, at least one failure, syntax error, or exception, or another mixed case.
Comparing \emph{(a)} and \emph{(c)} shows whether timeout filtering recovers positive controls by increasing the all-success curve and reducing timeout mass.
Comparing \emph{(b)} and \emph{(d)} shows the cost of that repair on negative controls: if the failure-present curve falls or the all-success curve rises for originally incorrect solutions, timeout filtering has weakened false-positive rejection.
The subsequent exception/failure filtering stage is shown in \cref{fig:test_augmentation_exception_failure_stages_50}.
}
\label{fig:test_augmentation_filtering_stages_50}
\end{figure*}

\begin{figure*}[t!]
\centering
\includegraphics[width=0.95\textwidth]{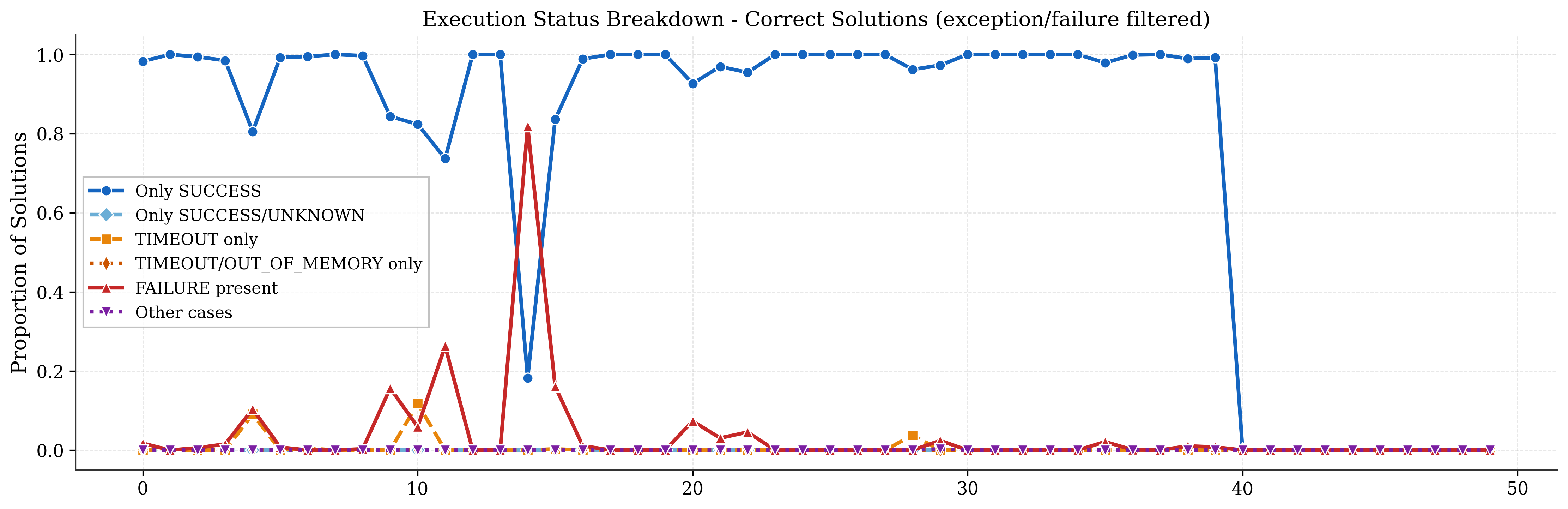}
\vspace{-0.2em}

\includegraphics[width=0.95\textwidth]{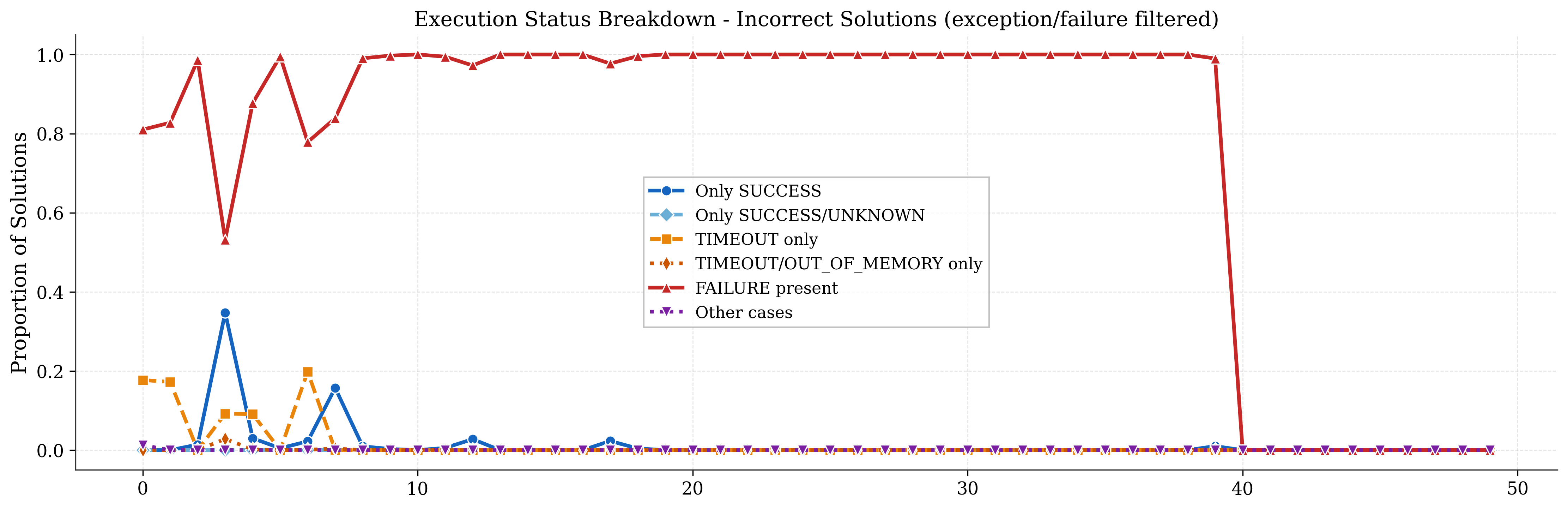}
\caption{\textbf{Execution-status diagnostics after exception/failure filtering.}
This figure continues \cref{fig:test_augmentation_filtering_stages_50} using the same problem order.
From top to bottom: \emph{(e)} executions on verified-correct solutions after the subsequent exception/failure filtering; and \emph{(f)} executions on originally incorrect solutions after both filtering stages.
Comparing \emph{(e)} with \emph{(a)} and \emph{(c)} in \cref{fig:test_augmentation_filtering_stages_50} shows whether the second filtering stage further recovers positive controls by removing generated tests that create failures or exceptions.
Comparing \emph{(f)} with \emph{(b)} and \emph{(d)} shows whether that recovery weakens false-positive rejection.
Large separations between raw and filtered panels mark problems where filtering changed the generated-test evidence, while overlapping curves indicate that the stage had little effect on aggregate execution statuses.
This is why test-level filtering is followed by solution-level and problem-level gates rather than keeping every recovered problem.
}
\label{fig:test_augmentation_exception_failure_stages_50}
\end{figure*}

\paragraph{50-problem diagnostics: why filtering is needed and examples.}
The 300-problem displays in \cref{fig:test_augmentation_impact,fig:test_augmentation_fp_300,tab:correct_pr_attrition_300,fig:test_augmentation_correct_pr_300,tab:largest_fp_reductions_300} give the headline statistics; the 50-problem slice in \cref{fig:test_augmentation_diagnostics_50,fig:test_augmentation_filtering_stages_50,fig:test_augmentation_exception_failure_stages_50,tab:manual_augmentation_cases,tab:no_generated_tests_cases,tab:largest_correct_pr_drops} explains the mechanisms behind the filters. Out of 50 problems, 40 received generated correctness tests and 10 did not. On the generated-test subset, the mean false-positive rate falls from 6.67\% to 1.64\%, the median false-positive rate falls from 0.43\% to 0.00\%, and no problem has a higher false-positive rate after augmentation. At the same time, mean correct-solution pass rate falls from 100.00\% to 87.07\%. The useful question is therefore not whether generated tests are simply ``good'' or ``bad'', but which drops in correct-solution pass rate are evidence of stale labels, which are caused by invalid or over-large generated tests, and which problems remain too ambiguous to keep. When the correct-solution pass rate drops by only ${\sim}1\%$, did the new tests find a small family of buggy solutions, or only measurement noise? When the drop is 50\% or more, did the tests discover many stale labels, or did they introduce timeout or constraint artifacts? When false positives disappear without hurting positive controls, can the problem be kept as a clean win? When no usable generated tests are produced, should a high-false-positive problem be removed rather than repaired?

\Cref{fig:test_augmentation_diagnostics_50,fig:test_augmentation_filtering_stages_50,fig:test_augmentation_exception_failure_stages_50} should be read with that distinction in mind. The desired pattern is visible on many problems: the false-positive curve moves down while the correct-solution curve stays near one. Those are the cases where generated tests add coverage without disturbing positive controls. The asymmetry is mechanical: false positives can only decrease when tests are added, but correct-solution pass rate can decrease for several reasons. The left tail of the correct-solution plot is therefore the reason filtering cannot be skipped, and it is also where ambiguous cases surface. A raw generated suite can make many verified-correct solutions fail because a test found a genuine edge case, because it is too slow for a correctness suite, because it violates a hidden constraint and therefore unfairly penalizes solutions that respect the problem boundaries, or because it exposes a problem whose strict input/output semantics are ambiguous.

The manual cases in \cref{tab:manual_augmentation_cases} make this ambiguity concrete. Train/103 has a small drop that manual inspection confirms as label repair: generated tests remove 26 false positives while only three previously verified-correct solutions fail. By contrast, train/67 and train/88 have drops that are too large to treat as harmless evidence. The pipeline therefore treats correct-solution failures as evidence to explain, not as proof that every failing solution has an independent bug. The pattern across solutions matters: a few idiosyncratic solutions failing a valid edge case differs from a large family of similar solutions failing together, which may reveal a shared bug but may also mean that the generated test has shifted the problem boundary. The pattern across generated tests matters as well: several diverse tests rejecting the same solution family suggests a missing property in those solutions, while one isolated generated test rejecting many positive controls is more likely to be too slow, out-of-spec, or ambiguous. The filtering rules use these patterns to decide whether to remove tests, update the solution pools, or remove the problem.

The execution-status plots in \cref{fig:test_augmentation_filtering_stages_50,fig:test_augmentation_exception_failure_stages_50} explain why the same pass-rate drop requires different responses. A timeout-heavy pattern on verified-correct solutions means the generated input is too expensive for a correctness check, even if it catches an incorrect solution. A failure- or exception-heavy pattern can indicate an out-of-spec input, a missing checker, or a multi-output ambiguity. In those cases, removing the worst generated tests first is not weakening the benchmark; it prevents one invalid or over-large test from discarding an otherwise useful problem. The removal caps define when this repair is still acceptable. If a problem can be recovered by removing only a small fraction of generated tests that fail many positive controls, it remains valuable. If it requires removing too much of the generated suite, the evidence is ambiguous: either the positive controls remain the source of truth and many generated tests are wrong, or the generated tests have revealed a shared bug pattern in a large family of solutions labeled correct. We remove such problems rather than decide this with a coarse threshold; a finer adjudication procedure would need to analyze clusters of solutions and clusters of generated tests jointly.

The example tables in \cref{tab:manual_augmentation_cases,tab:no_generated_tests_cases,tab:largest_correct_pr_drops} explain why problems that look similar in the aggregate curves can require different decisions. Some generated suites are clean wins, removing false positives while preserving positive controls. Some are clean no-ops, where the original tests were already sufficient. Some expose real bugs in solutions previously labeled correct, so a modest correct-solution drop is desirable. Others fail because generation produced tests that are too slow, out-of-spec, or too aggressive even after capped removal. This is why the filtering rules are stated before the performance plots: the purpose is not to maximize false-positive reduction at any cost, but to reduce false positives under an explicit positive-control budget while keeping the borderline problems that can still teach the RL system.

\begin{table*}[t!]
\centering
\small
\caption{\textbf{Manual validation cases used to interpret the generated-test filters.}
The cases come from the 50-problem diagnostic slice and cover the main outcomes: ideal false-positive removal, no-op augmentation, timeout-heavy tests, constraint violations, large-scale false-positive reduction, legitimate edge cases, and problems removed because the generated suite remains too aggressive after filtering.
}
\label{tab:manual_augmentation_cases}
\resizebox{\textwidth}{!}{
\begin{tabular}{llccp{6.5cm}}
\toprule
{\bfseries Problem} & {\bfseries Observed pattern} & {\bfseries FP before \(\to\) after} & {\bfseries Correct PR before \(\to\) after} & {\bfseries Interpretation} \\
\midrule
train/103 & Edge-case discovery & 32.91\% \(\to\) 0.00\% & 100.00\% \(\to\) 98.28\% & Ideal outcome: generated tests eliminate 26 false positives while rejecting only three buggy positive-control solutions. \\
train/64 & Clean no-op & 0.00\% \(\to\) 0.00\% & 100.00\% \(\to\) 100.00\% & Original tests were already sufficient; generated tests remain consistent with the original verdict. \\
train/67 & Timeout storm & 5.26\% \(\to\) 0.00\% & 100.00\% \(\to\) 11.76\% & Generated inputs are too expensive for many correct solutions; this motivates the 20\% timeout filter. \\
train/97 & Constraint violation & 27.59\% \(\to\) 0.00\% & 100.00\% \(\to\) 48.00\% & A generated input violates the specification; this motivates exception/failure filtering. \\
train/86 & Subtle constraint violations & 6.66\% \(\to\) 0.14\% & 100.00\% \(\to\) 48.19\% & Generated tests expose many false positives but include invalid or ambiguous cases that must be filtered or removed. \\
train/77 & Large-scale FP reduction & 20.28\% \(\to\) 3.00\% & 100.00\% \(\to\) 80.52\% & The problem passes final gates: the false-positive rate falls below 5\% while correct-solution drop stays below 25\%. \\
train/20 & Legitimate strictness & 0.00\% \(\to\) 0.00\% & 100.00\% \(\to\) 92.67\% & Generated tests find genuine bugs in solutions that the original suite labeled correct; the drop is small enough to retain the problem. \\
train/88 & Too aggressive after filtering & 1.99\% \(\to\) 0.00\% & 100.00\% \(\to\) 18.18\% & The suite rejects too many positive controls even after capped test removal, so the problem is removed. \\
\bottomrule
\end{tabular}
}
\end{table*}

The eight examples in \cref{tab:manual_augmentation_cases} are deliberately heterogeneous. Train/103 is the ideal case: the generated tests remove a large false-positive pocket with almost no positive-control loss. Train/64 is also useful, even though nothing changes, because it shows that augmentation is allowed to be conservative when the original suite is already adequate. Train/20 is the case that prevents a simplistic interpretation of correct-solution attrition: a 7.33pp drop is not treated as noise because manual inspection shows the generated test is valid and reveals genuine bugs in solutions that DMC had labeled correct. Train/77 shows why the problem-level cap should not be too strict. It has a large false-positive reduction, from 20.28\% to 3.00\%, and a nontrivial but acceptable correct-solution drop, from 100.00\% to 80.52\%; manual inspection suggests this is indeed a hard problem where many subtleties were not controlled by the original test suite. Removing such cases would bias the dataset away from the hard boundary region where stronger tests matter.

The rejected cases in \cref{tab:manual_augmentation_cases} explain the other side of the threshold. Train/67 catches a false positive, but it does so through timeout-heavy generated inputs that make most correct solutions fail. Train/97 and train/86 show constraint-violation patterns, where generated tests expose incorrect solutions but also ask verified-correct code to handle inputs outside the intended specification. Train/88 shows the residual failure mode after capped filtering: the generated suite consistently rejects most positive controls, and no sufficiently large subset of generated tests agrees with a sufficiently large subset of positive controls. The problem is therefore removed rather than forced into the dataset.

\begin{table}[t!]
\centering
\small
\caption{\textbf{High-false-positive problems without generated tests in the 50-problem slice.}
When generation fails, augmentation cannot repair the original suite; such problems can only be removed by the problem-level false-positive gate.
}
\label{tab:no_generated_tests_cases}
\begin{tabular}{lrrr}
\toprule
{\bfseries Problem} & {\bfseries Correct sol.} & {\bfseries Incorrect sol.} & {\bfseries FP before} \\
\midrule
train/24 & 38 & 41 & 60.98\% \\
train/26 & 337 & 229 & 24.02\% \\
train/131 & 33 & 154 & 21.43\% \\
train/135 & 100 & 104 & 11.54\% \\
\bottomrule
\end{tabular}
\end{table}

The no-generated-test cases in \cref{tab:no_generated_tests_cases} teach a different lesson. They are not merely missing a source of extra coverage; they may be exactly the cases where extra coverage was needed. In the 50-problem slice, train/24, train/26, train/131, and train/135 all have false-positive rates above 5\% before augmentation, but generation produced no usable tests for them. The solution-level step has no new evidence to relabel these problems, so the only conservative option is removal by the problem-level false-positive gate. This is another reason the final pool should not be understood as the result of adding tests everywhere. It is the subset where generation, filtering, and the original controls jointly leave a usable correctness signal. A useful follow-up would be to study why input-generator synthesis fails on these problems even when the model sees both the problem statement and a correct solution: they may require harder program reasoning from the example solution, have unusual input structure, or depend on constraints that are unstated, hidden in the problem text, and hard to infer even from solution code.

\begin{table*}[t!]
\centering
\small
\caption{\textbf{Largest correct-solution pass-rate drops in the 50-problem diagnostic slice.}
Large drops are not automatically treated as beneficial evidence.
Some correspond to genuine edge cases, while others indicate generated tests that are too slow, invalid, or too aggressive; the 25\% correct-solution drop cap removes these cases from the final dataset.
}
\label{tab:largest_correct_pr_drops}
\begin{tabular}{lrrrr}
\toprule
{\bfseries Problem} & {\bfseries Correct PR before \(\to\) after} & {\bfseries Change} & {\bfseries Correct sol.} & {\bfseries FP before \(\to\) after} \\
\midrule
train/67 & 100.00\% \(\to\) 11.76\% & -88.24pp & 17 & 5.26\% \(\to\) 0.00\% \\
train/88 & 100.00\% \(\to\) 18.18\% & -81.82pp & 11 & 1.99\% \(\to\) 0.00\% \\
train/13 & 100.00\% \(\to\) 23.91\% & -76.09pp & 46 & 0.00\% \(\to\) 0.00\% \\
train/49 & 100.00\% \(\to\) 24.40\% & -75.60pp & 332 & 0.00\% \(\to\) 0.00\% \\
train/97 & 100.00\% \(\to\) 48.00\% & -52.00pp & 25 & 27.59\% \(\to\) 0.00\% \\
train/86 & 100.00\% \(\to\) 48.19\% & -51.81pp & 83 & 6.66\% \(\to\) 0.14\% \\
train/120 & 100.00\% \(\to\) 73.68\% & -26.32pp & 133 & 5.00\% \(\to\) 0.56\% \\
train/77 & 100.00\% \(\to\) 80.52\% & -19.48pp & 154 & 20.28\% \(\to\) 3.00\% \\
\bottomrule
\end{tabular}
\end{table*}

The largest-drop table in \cref{tab:largest_correct_pr_drops} is included to avoid misleading success stories. Large correct-solution drops are not automatically beneficial, even when false positives fall to zero. Train/67 and train/88 show that an apparent false-positive improvement can come from tests that overwhelm positive controls. Train/13 and train/49 show a different warning sign: large correct-solution drops with no false-positive benefit, suggesting that the added tests shift the pass/fail boundary against positive controls rather than toward the false positives missed by the original suite. By contrast, train/77 remains below the 25\% correct-drop cap while reducing false positives below 5\%, so it represents the kind of hard-but-usable case the pipeline should retain. The cap removes problems where generated tests reject too many positive controls, while still keeping hard problems where generated tests expose subtleties missed by the original suite.

The second generation phase, which adds optimization tests after the correctness tests, adds one more layer to this logic. For correctness tests, a timeout on many positive controls is usually a warning that the generated input is too large for pass/fail adjudication. For optimization tests, we deliberately push toward larger inputs, so some correct human solutions timing out is not at all a defect: it can help create the slope that separates faster and slower correct implementations. The useful signal is not only per-test hardness, but diversity across the optimization-test pool. We want different tests to stress different pressure points of the solutions, exposing which implementations fail because of input size, value ranges, graph shape, combinatorial structure, or other workload properties. If all generated optimization tests time out the same subset of solutions, the suite has limited downstream use: it behaves like one repeated timeout gate rather than a ranking signal. This pool-level requirement also constrains each individual test: neither extreme is useful. If nearly every correct solution times out, the test becomes only a timeout gate and no longer ranks speed. If no correct solution is stressed, the test adds little optimization signal. The useful regime is the middle, where optimization tests create duration spread and partial timeouts across diverse solution and input patterns while still leaving enough successful executions to compare correct solutions.

Taken together, the 300-problem pool studied in \cref{fig:test_augmentation_impact,fig:test_augmentation_fp_300,tab:correct_pr_attrition_300,fig:test_augmentation_correct_pr_300,tab:largest_fp_reductions_300} and the fine-grained manual review of the 50-problem pool in \cref{fig:test_augmentation_diagnostics_50,fig:test_augmentation_filtering_stages_50,fig:test_augmentation_exception_failure_stages_50,tab:manual_augmentation_cases,tab:no_generated_tests_cases,tab:largest_correct_pr_drops} support the same conservative interpretation. Generated correctness tests expose many false positives missed by original DMC, but the raw augmented suite is not treated as ground truth because raw augmentation mixes label repair, valid edge-case discovery, over-large workloads, constraint violations, generation failures, and problems whose strict input/output semantics are too brittle for this pipeline. The retained corpus keeps only the subset that survives test-level, solution-level, and problem-level checks, while avoiding a filter that would remove every difficult boundary case.

\subsection{What remains before duration selection}
\label{sec:app_before_duration_selection}

\paragraph{Problem and solution counts.}
\Cref{tab:pipeline_detailed} reports the production-scale counterpart of the diagnostic analysis in \cref{sec:app_filtering_generated_tests}: the counts after generated-test filtering and before applying duration filterability. This is the point where the corpus has cleaned labels plus both generated correctness and generated optimization tests, but it has not yet been restricted to problems with enough timing spread. Incorrect solutions are needed during construction because they measure false positives, but they are not used as training rollouts in the final RL artifact.

\begin{table}[h!]
\centering
\small
\caption{\textbf{Problem and solution counts before duration selection.}
Problem and solution counts through generated-test filtering.
Solution counts are shown where available from the pipeline logs.
}
\label{tab:pipeline_detailed}
\begin{tabular}{lrrr}
\toprule
{\bfseries Stage} & {\bfseries Problems} & {\bfseries Correct Sol.} & {\bfseries Incorrect Sol.} \\
\midrule
Raw DMC data & 12{,}275 & --- & --- \\
After deduplication & 11{,}468 & --- & --- \\
After solution-type filtering & 6{,}706 & --- & --- \\
After execution validation & 3{,}928 & 938{,}227 & 1{,}490{,}355 \\
After test augmentation filtering & 3{,}061 & 797{,}896 & 1{,}032{,}816 \\
After intersection & 2{,}978 & --- & --- \\
Before duration selection & 2{,}723 & ${\sim}$800{,}000 & ${\sim}$1{,}030{,}000 \\
\bottomrule
\end{tabular}
\end{table}

The largest production attrition happens at the test-augmentation filtering stage: the execution-validated pool drops from $3{,}928$ to $3{,}061$ problems, removing 867 problems, or 22.1\% of the pool. At the solution level, this retains 85.1\% of the verified-correct solutions and 69.1\% of the incorrect-solution pool. That asymmetry is expected: generated correctness tests intentionally remove false positives and unreliable negative controls from the incorrect pool, while the correct-solution drop is capped to avoid keeping problems where generated tests appear too noisy or underspecified.

This completes the correctness half of the construction. The remaining question is different: among problems with cleaned labels and both generated correctness and optimization tests, which ones have optimization tests whose durations vary enough to define a useful timing objective?

\paragraph{Corpus variants used in our experiments.}
At this stage we maintain two corpus variants. The full variant is approximately 117\,GB and contains all correct and incorrect solutions, all test categories, and per-test execution-duration metadata. The lightweight variant is approximately 35\,GB, keeps one correct and one incorrect solution per problem, and excludes optimization tests for faster development iteration. Both variants share the same split.

\subsection{Duration filterability}
\label{sec:app_filterability}

Duration filterability is the measurement gate, not another label-cleaning filter. Duration-based rewards need within-problem timing spread: if all correct solutions run in nearly the same time on all tests, a reward can only amplify noise or constant-factor artifacts. For each test category of each problem, we aggregate runtimes over verified correct human solutions to obtain one representative duration per test: \(\bar{d}_j\) is the average runtime of test \(j\) across those solutions. We then define the robust coefficient of variation
\begin{equation}
\label{eq:robust_cv}
\text{Robust CV} =
\frac{\text{IQR}(\bar{d}_1, \dots, \bar{d}_N)}
{\text{median}(\bar{d}_1, \dots, \bar{d}_N)}.
\end{equation}
Using IQR rather than standard deviation keeps the metric insensitive to a small number of outlier tests, while dividing by the median makes it comparable across problems with different absolute runtimes. We call a problem duration-filterable when this value is at least 0.3, meaning the interquartile spread is at least 30\% of the median duration. When this condition holds, filtering or ranking by duration can expose meaningful differences between fast and slow solutions. When it does not, the available tests mostly exercise the same workload, and duration-based rewards collapse toward timing noise.

We deliberately do not define filterability from the spread of runtimes across correct human solutions on a fixed test. That alternative would bias problem selection toward the incidental composition of the available human-solution pool: low spread may simply mean that the retained solutions are homogeneous, while high spread may be driven by a few constant-factor outliers or pathological but still correct implementations. It can also make a problem look informative even when all tests remain sub-second and therefore exposed to additive timing noise. We instead use variation across tests, which is more directly a property of the test suite.

Some environment variants use test length before execution, so we also track length filterability through the Pearson correlation between test length and duration, with \(r \ge 0.9\) as the proxy threshold. This property answers a different question: not whether timing is informative, but whether serialized length can stand in for timing without executing the tests. We still require the duration-spread criterion for the final optimization dataset because some problems have clear duration spread even when runtime depends more on input values or structure than on character count.

\Cref{tab:filterability} reports these filterability counts on the pre-resplit $2{,}423$-problem training partition used for intermediate diagnostics.

\begin{table}[h!]
\centering
\small
\caption{\textbf{Filterability of test suites across the pre-resplit $2{,}423$-problem DMC-Optim training partition.}
A problem is duration filterable (D) if $\text{Robust CV} \geq 0.3$ and length filterable (L) if Pearson $r \geq 0.9$ between test length and duration.
Columns D and L report marginal filterability rates; ``Both'' and ``Neither'' report the corresponding intersections.
Original tests rarely span enough workload to support timing-based optimization.
}
\label{tab:filterability}
\begin{tabular}{lrrrr}
\toprule
{\bfseries Test category} & {\bfseries Duration (D)} & {\bfseries Length (L)} & {\bfseries Both} & {\bfseries Neither} \\
\midrule
Public tests        & 24 (1.0\%)       & 181 (7.5\%)      & 18 (0.7\%)       & 2{,}236 (92.3\%) \\
Private tests       & 92 (3.8\%)       & 131 (5.4\%)      & 13 (0.5\%)       & 2{,}213 (91.3\%) \\
Generated tests     & 88 (3.6\%)       & 110 (4.5\%)      & 12 (0.5\%)       & 2{,}237 (92.3\%) \\
Correctness tests   & 368 (15.2\%)     & 882 (36.4\%)     & 209 (8.6\%)      & 1{,}382 (57.0\%) \\
Optimization tests  & 1{,}169 (48.2\%) & 1{,}031 (42.6\%) & 863 (35.6\%)     & 1{,}086 (44.8\%) \\
\bottomrule
\end{tabular}
\end{table}

\Cref{tab:filterability} shows that optimization tests are the only category that remains informative at scale. On the pre-resplit training partition, the original public, private, and generated tests are almost never duration-filterable. Correctness tests help, but mostly by adding isolated hard cases. Optimization tests behave differently: 1{,}169 problems (48.2\%) are duration-filterable, and 863 (35.6\%) satisfy both duration and length filterability. Among duration-filterable optimization problems, 863 / 1{,}169 (73.8\%) are also length-filterable, so requiring both criteria would discard 306 problems whose runtimes are already informative even when length is only a weak proxy. Changing timeout thresholds on raw competition tests can change reward strictness, but it cannot create workload diversity that is absent from the tests themselves. This makes the complementary non-duration-filterable pool a natural negative control. If duration filterability is selecting problems where timing rewards can actually distinguish faster from slower correct solutions, then replacing duration-filterable problems with non-filterable ones should mostly hurt timing-based training. It should matter much less for correctness-oriented training, where the reward does not rely on within-problem duration spread.

\begin{figure*}[p]
\centering
\includegraphics[width=0.72\textwidth]{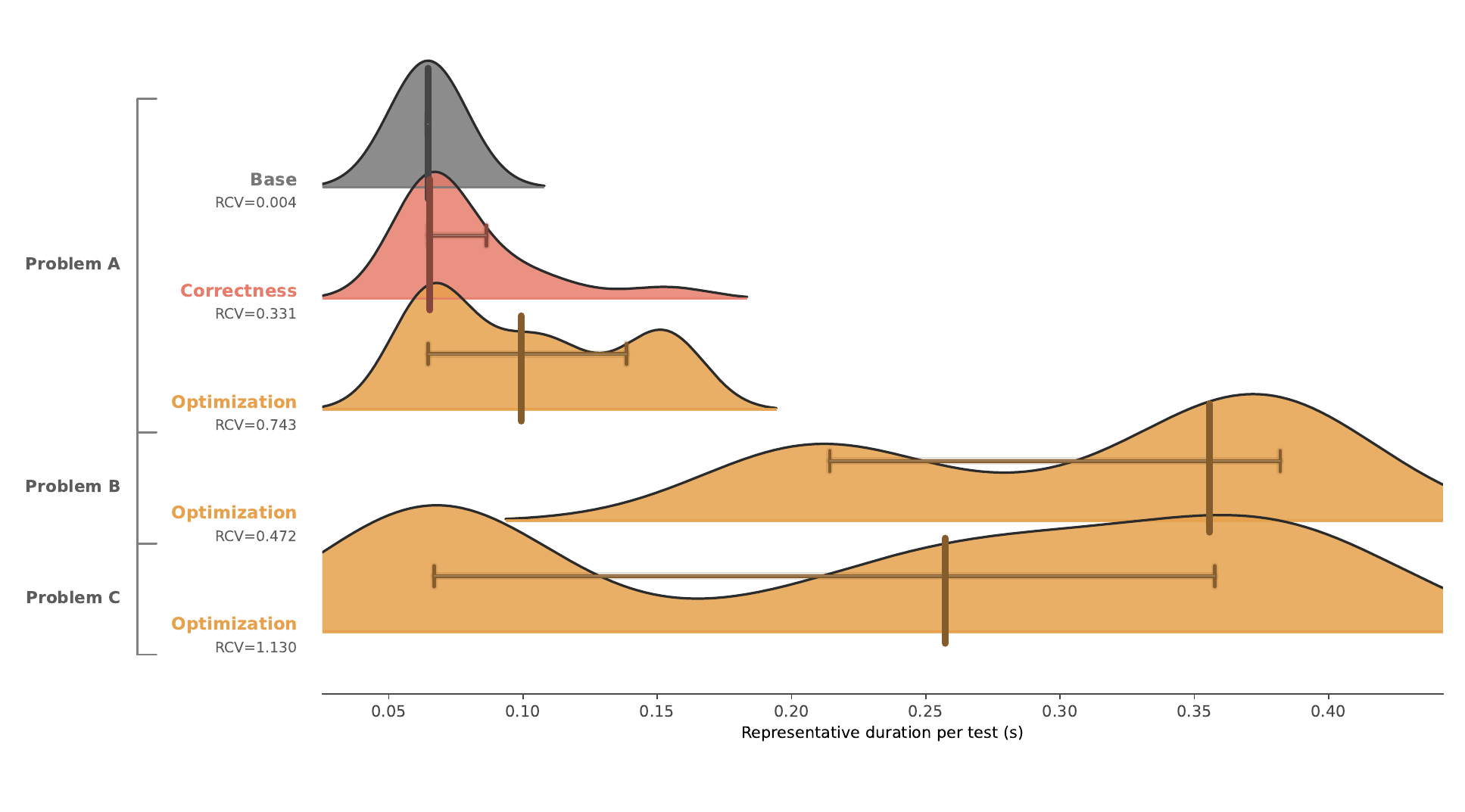}
\par\vspace{0.15em}
{\small\textbf{(a) Duration-filterable examples}}\par
\vspace{0.45em}

\includegraphics[width=0.72\textwidth]{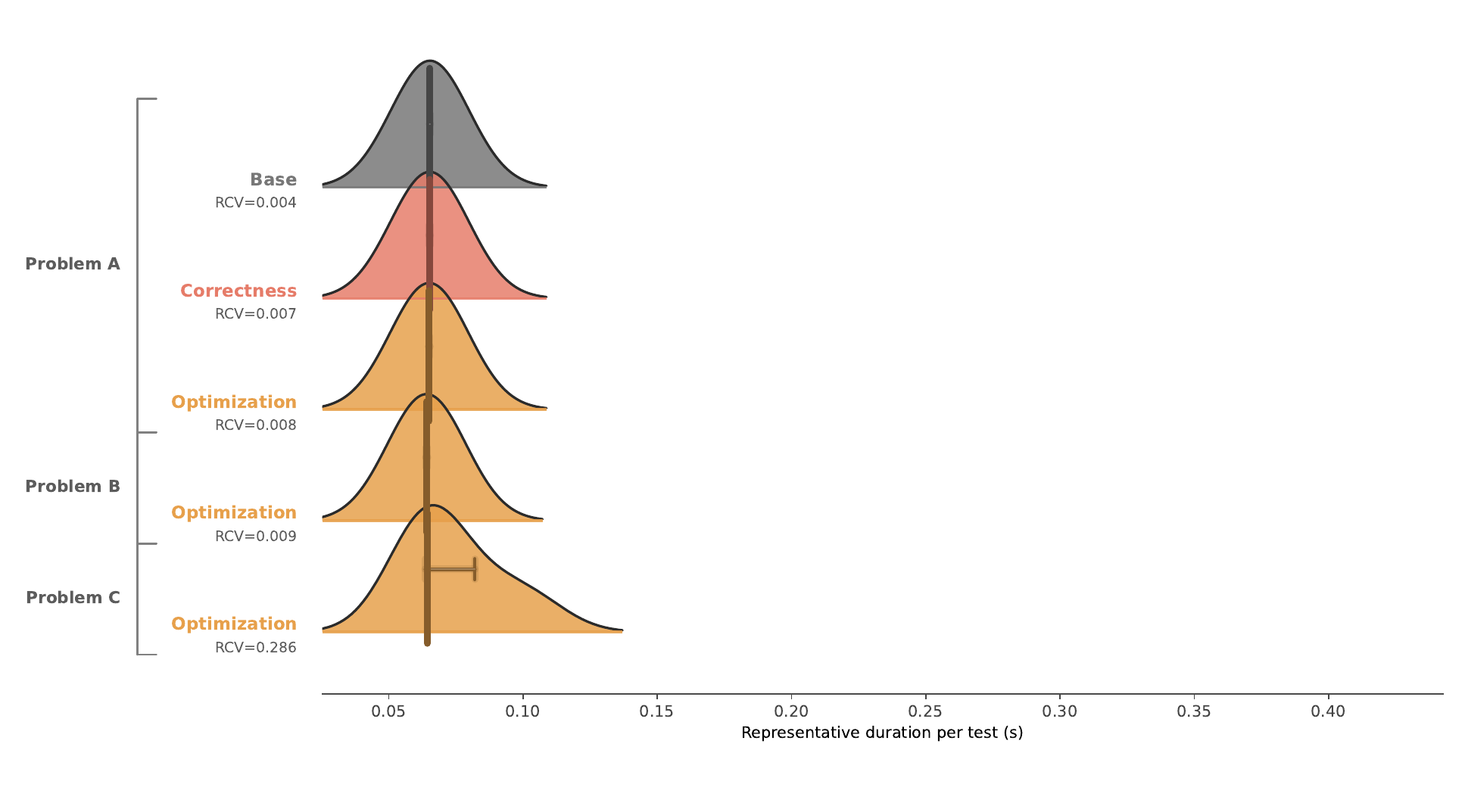}
\par\vspace{0.15em}
{\small\textbf{(b) Non-duration-filterable examples}}\par
\caption{\textbf{Duration-filterable versus non-duration-filterable optimization profiles under a shared x-axis scale.}
\emph{Panel a} shows a retained duration-filterable cluster.
For Problem A, the original base tests remain concentrated near $0.065\,\text{s}$, while generated correctness tests widen the profile and generated optimization tests create a broader multi-modal distribution.
Problems B and C show optimization-only examples on the same retained side of the threshold; their optimization robust CVs are 0.47 and 1.13, respectively, compared with 0.74 for Problem A.
\emph{Panel b} shows the matched rejected cluster on the same x-axis scale.
All three profiles remain concentrated around the original sub-second runtime regime, with optimization robust CVs 0.01, 0.01, and 0.29 for Problems A, B, and C.
The robust CV is computed as the interquartile range of the representative per-test durations divided by their median.
For the retained Problem A, a value of 0.74 means that the middle half of the optimization tests spans 74\% of the median duration; for the rejected Problem A, a value of 0.01 means that the middle half is almost degenerate around the median.
Using the interquartile range rather than the maximum makes the criterion depend on a persistent spread across the suite rather than on one isolated slow test, while dividing by the median makes the threshold comparable across problems with different absolute runtimes.
The comparison illustrates why duration filterability is problem-relative: useful optimization tests need not be slow in absolute terms, but they must create enough within-problem spread to rank correct solutions by speed.
}
\label{fig:filterability_case_studies_main}
\end{figure*}

\begin{figure*}[t!]
\centering
\includegraphics[width=\textwidth]{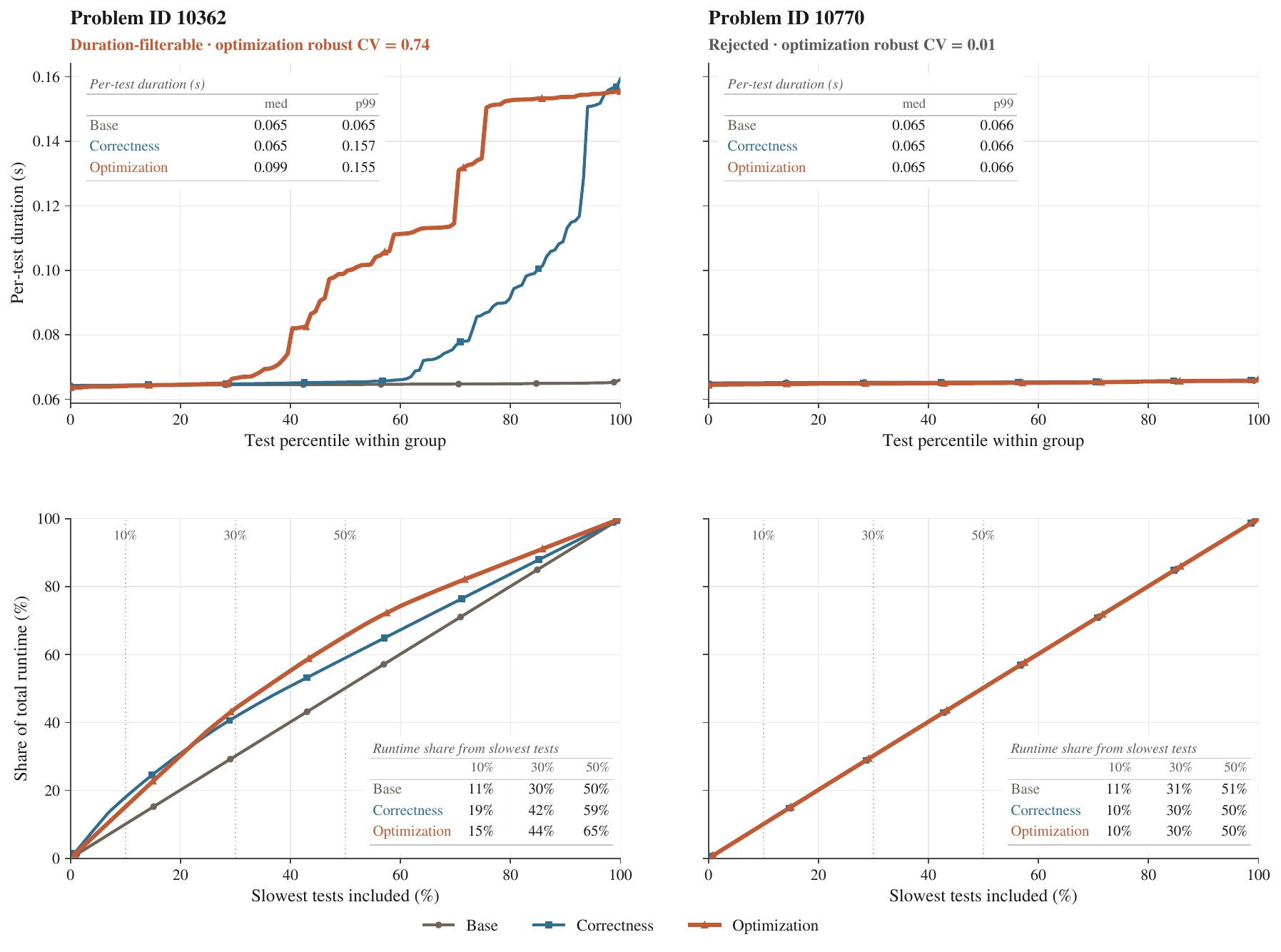}
\caption{\textbf{Retained versus rejected optimization-test profiles.}
The figure reads both rows through the same duration-profile shapes.
In the top row, tests are sorted by representative duration.
A flat curve means that the whole suite has nearly one runtime, so there is no timing signal.
A corner means that most tests remain fast and only a small tail is slow; this creates duration spread, but concentrates the useful workload variation in a small subset.
A diagonal-like ramp is the desired shape: runtimes increase across much of the suite, giving the reward a broad range of workloads on which to compare correct solutions.
The rejected right problem is the flat case, with base, correctness, and optimization tests all compressed near 0.065\,s.
The retained left problem shows the difference between the two generated branches.
Generated correctness tests add slower cases, but the curve is still closer to a corner: many tests stay near the original runtime and the increase is concentrated near the slow end.
Generated optimization tests are closer to a ramp, with robust CV 0.74 and median duration 0.099\,s, indicating a broader workload range.
The bottom row shows the same distinction through cumulative runtime share carried by the slowest tests.
If all tests have the same duration, this curve is diagonal: the slowest 30\% of tests carry 30\% of runtime, as in the rejected problem.
If 90\% of tests are very fast and 10\% are very slow, the curve makes a sharp corner: the slowest 10\% carry most of the runtime, and adding the next 40\% adds little.
The useful regime lies between these extremes, with a bowed curve above the diagonal but not collapsed into a corner.
In the retained left problem, the optimization tests have this pattern: the slowest 10\%, 30\%, and 50\% account for 15\%, 44\%, and 65\% of total runtime.
}
\label{fig:filterability_case_studies}
\end{figure*}

\Cref{tab:filterability} gives the count-level result; the next two figures show what those selected and rejected profiles look like.
\Cref{fig:filterability_case_studies_main} uses the same duration scale for retained and rejected examples.
Retained optimization suites spread across runtimes, while rejected suites stay clustered near the base runtime.
\Cref{fig:filterability_case_studies} zooms in on one retained/rejected pair and separates three visual cases: flat, corner, and ramp.
Flat profiles have no timing signal; corner profiles rely on a small slow tail; ramp-like profiles distribute workload variation across the suite.
This is the pattern that the robust-CV threshold is meant to capture.

\FloatBarrier

\subsection{The final duration-filterable RL pool}
\label{sec:app_duration_stats}

\Cref{tab:dataset_stats} reports pooled timing statistics on the final $1{,}000$-problem RL training split after removing non-duration-filterable problems. The original DMC tests reach only $0.145\,\text{s}$ at $p95$ and $0.463\,\text{s}$ at $p99$; correctness tests reach $0.407\,\text{s}$ and $1.378\,\text{s}$; optimization tests reach $1.296\,\text{s}$ and $3.710\,\text{s}$. The share of tests above $1\,\text{s}$ likewise rises from 0.42\% on the original suite to 1.54\% on correctness tests and 6.87\% on optimization tests. The $1\,\text{s}$ cutoff is a common reference point rather than a requirement that every problem cross it: some valid problem domains cannot produce second-scale tests, so the main signal is the relative enrichment, about 16$\times$ the original-test rate.

\begin{table}[h!]
\centering
\small
\caption{\textbf{Pooled timing statistics on the final $1{,}000$-problem RL training split.}
Original DMC tests pool public, private, and DMC-generated tests.
Statistics are computed over all test-duration records within each test category, without averaging per problem; per-problem statistics are reported in \cref{tab:duration_tail_summary}.
}
\label{tab:dataset_stats}
\begin{tabular}{lrrrrrr}
\toprule
{\bfseries Test group} & {\bfseries Pooled tests} & {\bfseries Mean (s)} & {\bfseries $p90$ (s)} & {\bfseries $p95$ (s)} & {\bfseries $p99$ (s)} & {\bfseries Tests $> 1$\,s} \\
\midrule
Original DMC tests & 106{,}974 & 0.088 & 0.090 & 0.145 & 0.463 & 0.42\% \\
Correctness tests & 118{,}551 & 0.137 & 0.209 & 0.407 & 1.378 & 1.54\% \\
Optimization tests & 119{,}420 & 0.334 & 0.736 & 1.296 & 3.710 & 6.87\% \\
\bottomrule
\end{tabular}
\end{table}

For each retained problem, DMC-Optim stores measured durations for verified human reference solutions on the retained tests. These durations anchor the problem-relative rewards and timing-sensitive evaluation metrics used later: in some optimization-RL environments, a fresh model rollout can be compared with the stored human timing distribution for the same problem and test suite, rather than with one global runtime threshold. Because those reference durations and the fresh rollout durations may be collected under different execution-service states, \cref{app:ces_fallback} describes the calibration step that maps stored durations into the current timing scale before they are used for reward or evaluation.

\Cref{tab:duration_tail_summary} checks that the tail in \cref{tab:dataset_stats} is not only a pooled artifact, but appears across individual problems. Their pooled $p99$ duration reaches $3.710\,\text{s}$, versus $1.378\,\text{s}$ for correctness tests and $0.463\,\text{s}$ for the original DMC tests. At the problem level, 18.9\% of problems still have at least 10\% of optimization tests above $1\,\text{s}$, versus 5.2\% for correctness tests and 1.2\% for the original tests.

\begin{table}[h!]
\centering
\small
\caption{\textbf{Duration-tail summary on the final $1{,}000$-problem RL training split.}
``Pooled'' statistics are computed over all tests in a category.
Problem-level statistics measure how often slow tests persist within individual problems.
Optimization tests are the only category that simultaneously reaches a multi-second pooled tail and leaves a substantial fraction of problems with many slow tests.
}
\label{tab:duration_tail_summary}
\resizebox{\linewidth}{!}{
\begin{tabular}{lrrrrrr}
\toprule
{\bfseries Test category} & {\bfseries Pooled median (s)} & {\bfseries Pooled $p99$ (s)} & {\bfseries Median $p99 / \text{median}$} & {\bfseries Tests $> 1$\,s} & {\bfseries Problems with} $\geq 10\%$ {\bfseries tests} $> 1$\,s & {\bfseries Problems with any test} $> 3$\,s \\
\midrule
Original DMC tests & 0.065 & 0.463 & 1.02$\times$ & 0.4\% & 1.2\% & 0.8\% \\
Correctness tests  & 0.066 & 1.378 & 4.02$\times$ & 1.5\% & 5.2\% & 3.9\% \\
Optimization tests & 0.099 & 3.710 & 3.07$\times$ & 6.9\% & 18.9\% & 6.7\% \\
\bottomrule
\end{tabular}
}
\end{table}

\begin{figure*}[t!]
\centering
\includegraphics[width=\textwidth]{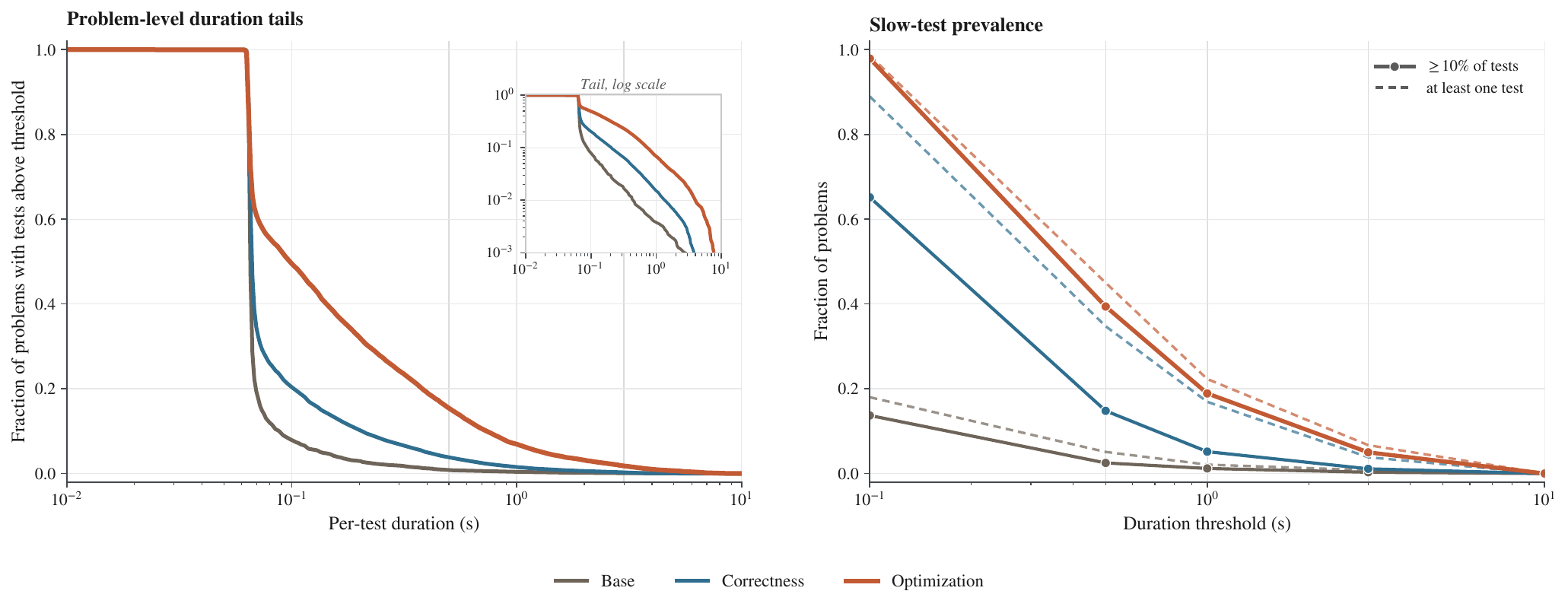}
\caption{\textbf{Aggregate duration-tail view on the final $1{,}000$-problem RL training split.}
The left panel plots an equal-problem-weight duration tail: for each problem and threshold, the fraction of tests above threshold, averaged across problems (tail repeated on a log scale, inset). Optimization tests dominate across the full sweep and stay separated in the multi-second regime.
The right panel summarizes slow-test prevalence at each threshold, distinguishing problems with ``at least one slow test'' from those with ``at least 10\% slow tests.''
At a 0.5\,s threshold, 39.4\% of problems still have at least 10\% of optimization tests above threshold, versus 14.8\% (correctness) and 2.5\% (original); at 1\,s: 18.9\%, 5.2\%, 1.2\%.
}
\label{fig:duration_tail_overview}
\end{figure*}

\begin{figure*}[p]
\centering
\begin{minipage}{0.98\textwidth}
\centering
\includegraphics[width=\linewidth]{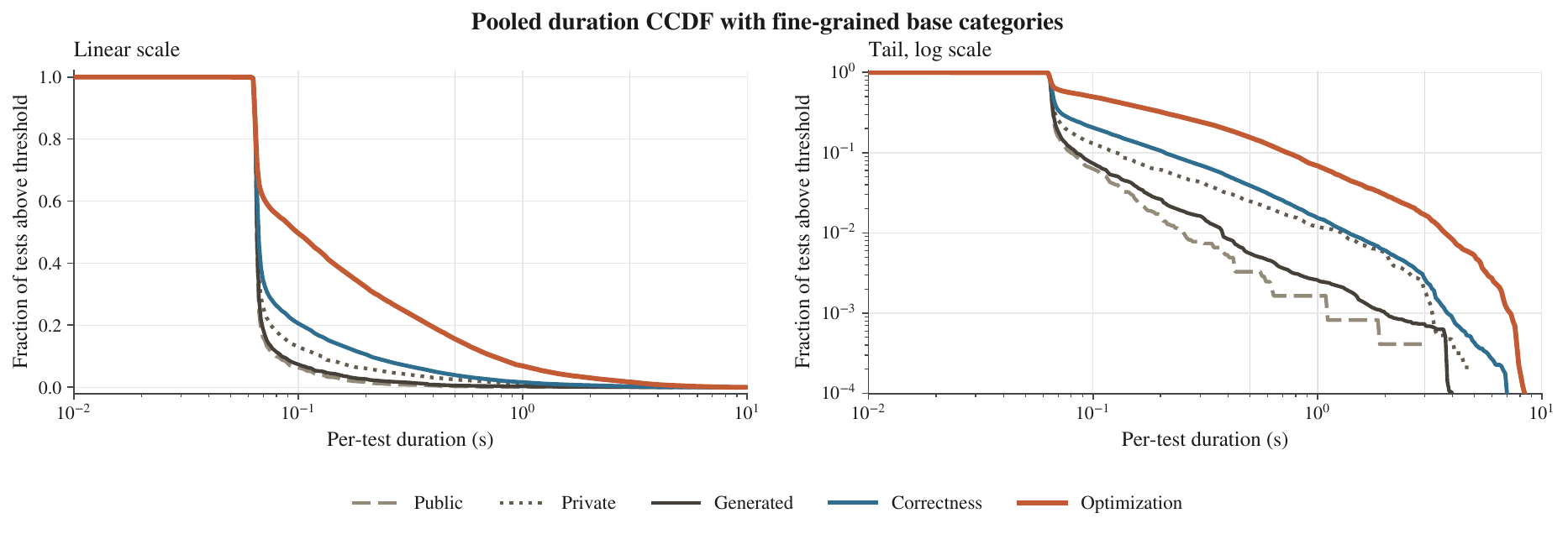}
\par\vspace{0.15em}
{\small\textbf{(a) Fine-grained pooled duration tails}}\par
\end{minipage}
\vspace{0.55em}
\begin{minipage}[t]{0.49\textwidth}
\centering
\includegraphics[width=\linewidth]{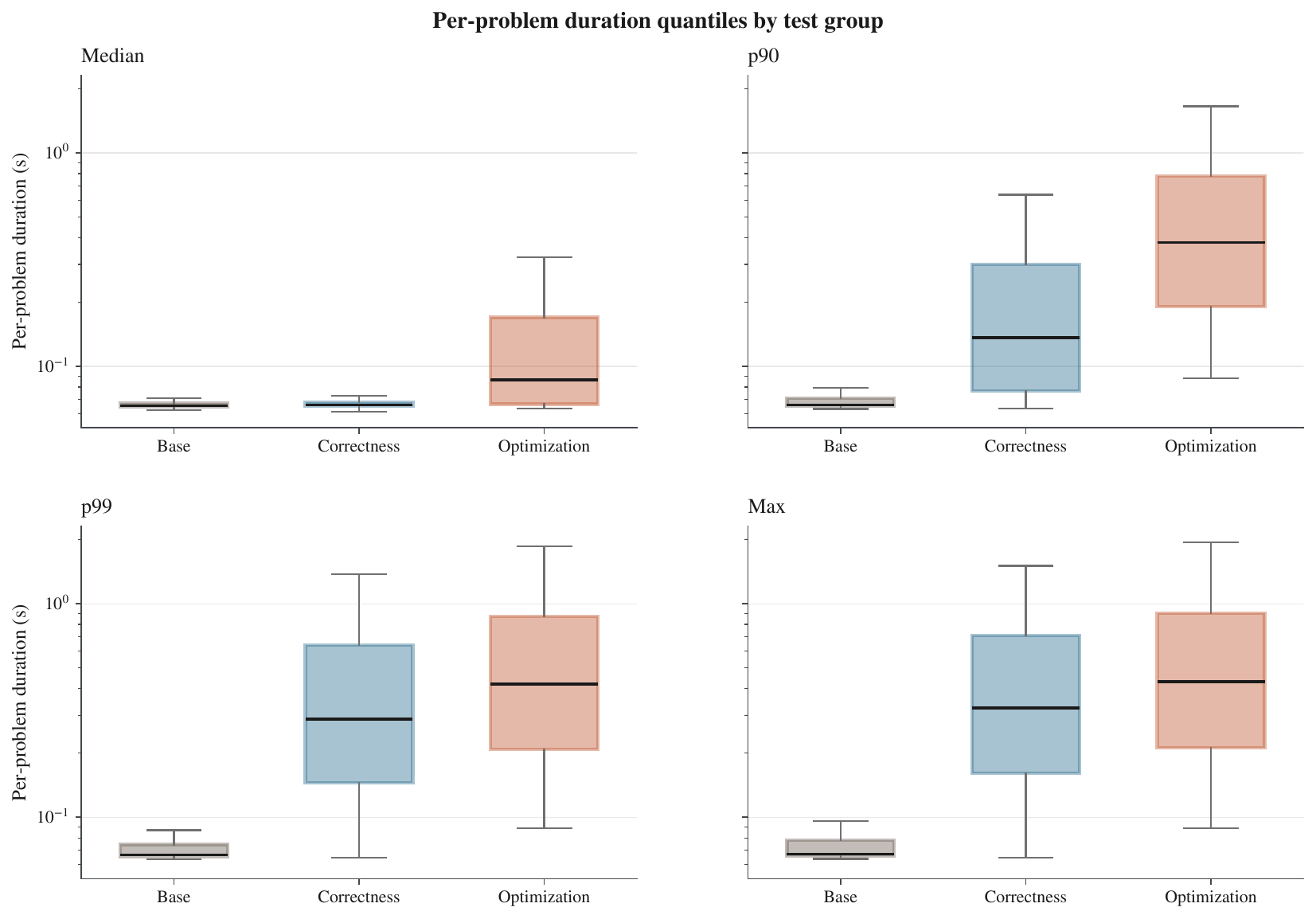}
\par\vspace{0.15em}
{\small\textbf{(b) Per-problem duration quantiles}}\par
\end{minipage}
\hfill
\begin{minipage}[t]{0.49\textwidth}
\centering
\includegraphics[width=\linewidth]{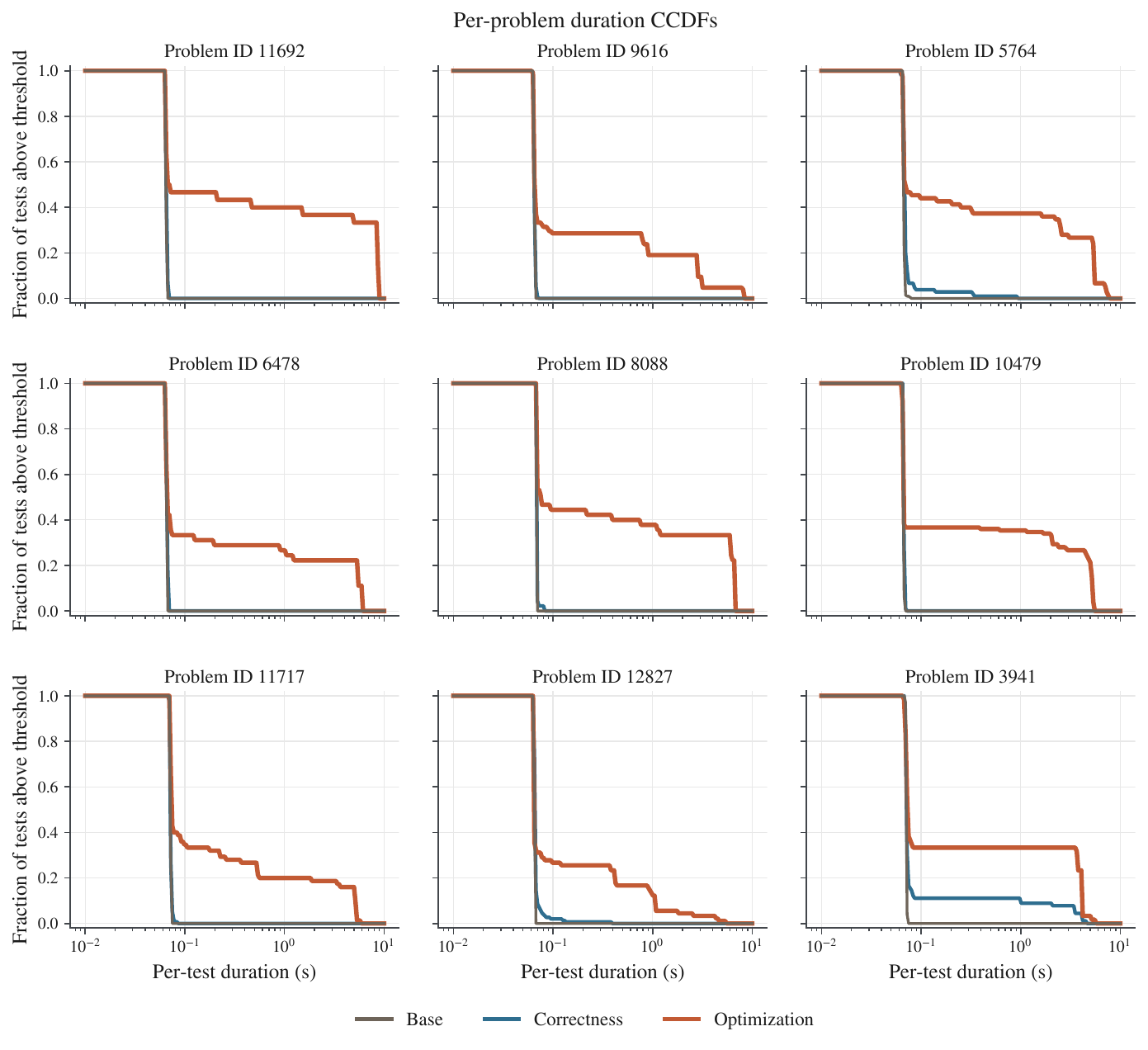}
\par\vspace{0.15em}
{\small\textbf{(c) Example problem-level tails}}\par
\end{minipage}
\caption{\textbf{Additional duration-tail diagnostics on the final $1{,}000$-problem RL training split.}
\emph{Panel a} separates the original public, private, and DMC-generated tests from the generated correctness and optimization tests.
The left subplot shows the pooled complementary cumulative distribution function (CCDF) with a linear y-axis, while the right subplot shows the same tail on a log y-axis.
All original categories leave the multi-second regime quickly; correctness tests add a longer tail, but optimization tests keep the largest fraction of tests above every threshold in the slow regime.
\emph{Panel b} reports per-problem median, $p90$, $p99$, and maximum durations on log-scaled y-axes.
The median panel shows that many problems remain near the same sub-second center, while the higher-quantile panels show that the extra optimization signal mainly appears in the tail rather than through a uniform shift of every test.
\emph{Panel c} shows representative problem-level CCDFs selected because optimization tests create a heavier tail than the base suite.
Across these examples, the base and correctness curves often collapse shortly after the sub-second threshold, whereas optimization tests keep nonzero mass into the second-scale range.
Together with \cref{fig:duration_tail_overview}, these plots show that DMC-Optim does not merely add more tests; it adds tests that broaden the within-problem timing range available to the reward.
}
\label{fig:duration_tail_additional_diagnostics}
\end{figure*}

The goal when creating DMC-Optim was not to push most tests beyond a fixed absolute threshold. For many problems, the valid input domain and intrinsic algorithmic structure do not support such a regime; forcing it would either violate constraints or collapse the suite onto only the hardest corner. The goal is to cover as much of each problem's feasible duration range as possible, so that enough tests become slow within that valid range for timing to remain informative. \Cref{fig:duration_tail_overview} visualizes this at the aggregate level, and \cref{fig:duration_tail_additional_diagnostics} breaks the same pattern into pooled tails, per-problem quantiles, and representative problem examples.

These duration-tail measurements establish the measurement property of the retained pool, but they do not by themselves prove that duration-filterable selection matters for learning, nor that data construction alone is sufficient for optimization RL. We therefore run two training ablations: first changing the training pool while keeping the reward, evaluation set, and training budget fixed, and then asking whether a naive raw-duration reward can replace the problem-relative reward used in the main experiments.

\clearpage
\subsection{Training ablations: does DMC-Optim help, and is it enough to solve optimization RL?}
\label{sec:app_filterability_training_controls}

The preceding diagnostics are dataset-only measurements. We therefore train models to ask whether the data choices behind those measurements change learning behavior. The first ablation isolates problem selection: we compare three equal-sized training pools of 1{,}000 problems under the same reward and environment. The pools are the duration-filterable pool used in the main runs; a mixed-proportion pool with 478 duration-filterable and 522 non-duration-filterable problems, matching the full-corpus ratio $1{,}302/2{,}723$; and a non-filterable pool sampled from the $1{,}421$-problem complement whose optimization tests fail the duration-filterability criterion. The non-filterable companion is constructed from the runtime-merged cleaned corpus before duration-filterable selection, not by editing the already filtered training split: it preserves the same transformed training schema, removes incorrect solutions, and keeps aggregated duration metadata and duration distributions. The reward, RL environment, added-test family, CES execution backend, evaluation set, evaluation cadence, and training budget are kept fixed within each row. We report pass@$1$ at evaluation percentile \(p_{30}\) as an absolute score and, in parentheses, its relative change from the duration-filterable baseline,
\[
\Delta_{\mathrm{rel}} = 100 \times \frac{s_{\mathrm{variant}} - s_{\mathrm{filterable}}}{s_{\mathrm{filterable}}},
\]
so negative values are relative drops.

\begin{table}[h!]
\centering
\small
\caption{\textbf{Training-pool composition controls at fixed dataset size.}
Entries are pass@$1$ at evaluation percentile \(p_{30}\) after 5k RL steps; values in parentheses are relative changes against the same environment trained on the duration-filterable pool.
The comparison isolates the problem-selection step: the added-test family, reward definition, RL environment, evaluation set, and number of training problems are held fixed within each row.
The Charlen 100k row filters optimization tests above 100k serialized input-output characters and applies a 0.5\,s optimization gate with 10\% timeout tolerance.
}
\label{tab:filterability_training_controls}
\begin{tabular}{lrrr}
\toprule
{\bfseries Training environment} & {\bfseries Filterable pool} & {\bfseries Mixed pool} & {\bfseries Non-filterable pool} \\
\midrule
MC+MO correctness gate & 9.3 & 8.9 (-4.3\%) & 8.3 (-10.8\%) \\
Filter 2s timing gate & 10.4 & 9.6 (-7.7\%) & 9.3 (-10.6\%) \\
Charlen 100k + 0.5\,s gate & 9.5 & 9.4 (-1.1\%) & 9.4 (-1.1\%) \\
Ranked \(p_{30}\) timing gate & 16.1 & 13.3 (-17.4\%) & 9.1 (-43.5\%) \\
\bottomrule
\end{tabular}
\end{table}

\paragraph{Duration filterability boosts optimization RL.}
\Cref{tab:filterability_training_controls} separates two questions: what each reward does on the same training pool, and how the same reward changes when duration-filterable problems are replaced by mixed or non-filterable problems.
The MC+MO row uses the retained correctness and optimization tests as a pass/fail gate, without adding a problem-relative speed ranking.
Its \(p_{30}\) score changes only moderately across pools (9.3, 8.9, 8.3), which shows that the non-filterable complement is not uninteresting problems, nor problems that would be broken for standard pure-correctness checks.
What we suspect is that any optimization-based reward would break on these problems and tests.

The character-length row explains an important caveat, as it suffers less than the MC+MO baseline when changing the pool of problems.
On the duration-filterable pool, large serialized tests often overlap with the measured duration tail; removing tests above 100k input-output characters can therefore remove some of the optimization pressure that made the pool useful.
This is why the filterable-pool charlen score, 9.5\%, is below the filterable-pool Filter 2s score of 10.4\% and well below the ranked \(p_{30}\) score of 16.1\%.
As the pool is replaced by mixed and then non-filterable problems, the same large-character tests are less aligned with a useful duration profile: by definition, these problems do not have enough within-problem duration spread for the optimization tests to support a problem-relative timing reward.
Dropping very large tests is then less costly: those tests were not creating a useful timing ladder anyway.
On the remaining smaller tests, the stricter 0.5\,s optimization gate with 10\% timeout tolerance can be a better choice than the 10\,s optimization timeout used by the MC+MO baseline, because it applies optimization pressure better matched to the retained test sizes.
This explains why charlen stays at stable performance on the mixed and non-filterable pools, and why it can beat MC+MO at the same non-filterable pool (9.4\% versus 8.3\% at \(p_{30}\)).
On the duration-filterable pool, on the contrary, the removed large tests are part of the tail that distinguishes important optimization differences between correct solutions.
Charlen then has to rely on the smaller retained tests after discarding a meaningful part of the timing signal.
On the non-filterable pool, by contrast, the retained small tests provide weaker and flatter timing pressure, but at least the discarded large tests were not carrying a separate branch of useful optimization pressure.
It does not mean that the non-filterable pool is as good an optimization-RL dataset: its best charlen score on non-filterable data still remains below ranked \(p_{30}\) on the duration-filterable pool (9.4\% versus 16.1\%).

The decisive comparison is therefore ranked \(p_{30}\).
This reward uses the human-reference timing distribution to rank correct solutions by speed, so it needs a real within-problem duration profile.
When duration-filterable problems are replaced by mixed or purely non-filterable pools, ranked \(p_{30}\) falls from 16.1\% to 13.3\% and then 9.1\%, ending only 0.8 points above MC+MO on the same non-filterable pool.
This is a bad regime for a ranking reward: if the optimization-test duration profile is nearly flat, the reward still forces a speed ordering, but that ordering is weakly supported by the tests and mostly adds noise to the reward.
The conclusion is not that non-filterable problems are useless, but that duration filterability predicts where problem-relative timing rewards have a usable training signal.

\paragraph{Good optimization tests are not enough for good optimization RL.}
After constructing DMC-Optim, a natural question is whether the dataset work is enough by itself: once the tests are stronger and include larger optimization workloads, can we simply reward shorter raw execution time, or do the downstream RL environment and reward still add substantial signal?
We test this by running raw-duration rewards on two branches. The base-test variant measures absolute seconds on the original correctness-style tests; the optimization-test variant measures absolute seconds on the generated optimization tests. Each is tested with a linear and a logarithmic map from seconds to reward. \Cref{tab:naive_duration_training_controls} reports relative \(p_{30}\) changes against two references: MC+MO, which asks whether raw timing improves over correctness-only RLVR on the same added tests, and ranked \(p_{30}\), which asks whether raw timing can match a problem-relative optimization-RL environment and reward design.

\begin{table}[h!]
\centering
\small
\caption{\textbf{Naive raw-duration reward controls.}
Entries are relative changes in pass@$1$ at evaluation percentile \(p_{30}\) after 10k RL steps.
The first column compares each raw-duration reward to MC+MO; the second compares it to the ranked \(p_{30}\) timing reward.
Raw duration on optimization tests is better than raw duration on base tests, but it remains well below the problem-relative timing reward.
}
\label{tab:naive_duration_training_controls}
\begin{tabular}{lrr}
\toprule
{\bfseries Raw-duration reward} & {\bfseries vs. MC+MO} & {\bfseries vs. ranked \(p_{30}\)} \\
\midrule
Base tests, linear map & -10.8\% & -56.5\% \\
Base tests, log map & -10.8\% & -56.5\% \\
Optimization tests, linear map & +17.2\% & -42.9\% \\
Optimization tests, log map & +19.4\% & -41.9\% \\
\bottomrule
\end{tabular}
\end{table}

The base-test raw-duration rows show that adding absolute timing to the original test regime is not enough for optimization RL: both variants drop by 10.8\% relative to MC+MO at \(p_{30}\).
The log-map rows are a useful diagnostic.
Given the very short base-test durations in \cref{tab:dataset_stats}, one might expect a logarithmic map to help by expanding reward differences among small runtimes.
This is not what we observe, but the log map also does not provide the missing signal: on base tests, it is indistinguishable from the linear map (-10.8\% versus -10.8\% relative to MC+MO), and on optimization tests it only slightly improves the linear variant (+19.4\% versus +17.2\%).
This suggests that extra resolution at very small runtimes mostly turns timing noise into reward noise, while the larger duration gaps that carry more useful optimization signal still need a problem-relative reward.
Moving the same raw-duration idea onto optimization tests helps more, improving over MC+MO by 17.2--19.4\%, which confirms that better tests expose some timing signal.
However, these same rows remain 41.9--42.9\% below the ranked \(p_{30}\) reward, so raw duration still fails to recover the problem-relative timing signal.

Together with \cref{tab:filterability_training_controls}, these controls give a two-sided conclusion.
Improving the dataset and then doing raw optimization RL is not enough: even on generated optimization tests, the best raw-duration control remains far below the more fine-grained ranked \(p_{30}\) environment.
Conversely, using an explicitly designed timing reward on the wrong data is also not enough: ranked \(p_{30}\) falls from 16.1\% on the duration-filterable pool to 9.1\% on the non-filterable pool.
DMC-Optim is therefore useful only together with downstream reward and environment design: the dataset must create measurable within-problem duration structure, and the RL recipe must turn that structure into a stable training signal.

\clearpage
\section{Making runtime measurements trustworthy: code execution backend}
\label{app:ces_fallback}
This appendix expands the second half of \cref{sec:measurement}. \Cref{app:dataset} first showed why DMC-Optim needs optimization tests with useful duration spread: without such spread, a timing reward has little signal. That is still not enough. If the backend cannot measure these durations reliably, the apparent spread will not be usable for training or evaluation, no matter how carefully the tests were generated. In correctness-only RLVR, the sandbox mostly decides whether a solution passes before the timeout; small runtime perturbations are secondary. In optimization RL, the runtime itself enters the reward, so worker load, retries, fallback behavior, and backend drift can change the learning signal. Here we record how our initial local-sandbox setup was discarded and replaced by a more controlled remote code execution service (CES), how CES is used during construction, training, and evaluation, how infrastructure failures are handled, and how stored human-reference durations are calibrated against fresh executions.

\paragraph{Reading map.}
\Cref{sec:app_execution_claim} states what a code execution backend must provide for stable optimization RL.
\Cref{sec:app_local_vs_ces} then starts from the initial local-sandbox setup, which is our base setup for standard RLVR, and shows why it fails for optimization RL: it changes tight-timeout rankings, creates unstable within-problem orderings, inflates duration filterability, fails calibration or correction attempts, and breaks down inside the RL loop.
\Cref{sec:app_exec_architecture,sec:app_conservative_timing,sec:app_fallback_impact,sec:app_fallback_tradeoffs} describe CES as the more controlled replacement: the execution path, fallback policy, empirical failure rates, and alternative fallback designs we use or considered.
\Cref{sec:app_duration_correction,sec:app_temporal_stability,sec:app_calibration_sensitivity,sec:app_training_eval_calibration_sensitivity} analyze the remaining variability of that CES setup, separating short-run noise during live execution from long-term drift between stored human-reference durations and fresh CES measurements, then testing whether affine calibration is stable enough for the metrics used in the paper both at evaluation time and when the calibration is changed during RL training.
\Cref{sec:app_execution_lessons} summarizes the lessons we take from these checks for timing-based training and evaluation, and \cref{sec:app_execution_future_work} lists the measurement questions that remain open.

\subsection{What a code execution backend has to establish}
\label{sec:app_execution_claim}

For correctness-only RL, occasional timeout noise can remain secondary because the reward is dominated by pass/fail outcomes. For optimization RL, the measured duration itself can determine whether a rollout receives credit. This matters in exactly the regime created by DMC-Optim: even after large-input test generation, the pooled optimization-test $p95$ is $1.296\,\text{s}$, the $p99$ is $3.710\,\text{s}$, and only 6.87\% of optimization tests exceed $1\,\text{s}$ (\cref{tab:dataset_stats}). Sub-second perturbations can therefore change rankings, thresholds, and rewards. The measurement backend must isolate code timing from worker contention, expose failures explicitly, and support calibration when service state changes.

We considered two practical backends. The first was a local sandbox, which we can consider our base RLVR setup, running on the same training workers that host model inference, rollout orchestration, and other CPU and memory load from the RL job. The second is a dedicated remote sandboxed execution service, denoted CES below, that runs each code-test pair under a fixed resource envelope, with 1\,GB memory and a 10\,s hard limit, and returns both status and duration. CES has more overhead per call, but it separates timing from worker-side contention, lets us control retries and concurrency, and can be provisioned independently of the GPU job. The evidence that follows led us to use the calibrated CES setup for all optimization-RL runs in the paper.

Stable timing is needed not only to train with a speed reward, but also to compare ablations. Controlling execution noise makes differences between runs interpretable as changes in the dataset, reward, or training recipe, rather than uncontrolled changes in the backend.

\subsection{Local sandbox and why it is bad for timing measurements}
\label{sec:app_local_vs_ces}

Our base RLVR execution path is a local Python I/O-pair sandbox. A training worker sends the candidate program and a batch of input/output pairs to a sandboxed Python runner. The runner compiles the program once, replays the inputs sequentially by redirecting standard input and output, compares the produced output with the expected output, and returns one status and one elapsed duration per test case. The runner is launched through a fork server whose default backend is Bubblewrap \citep{bubblewrap}: the submitted code runs in a Linux namespace sandbox with dropped capabilities, a fresh temporary filesystem, resource limits, and process cleanup. This is a reasonable setup for correctness-only RLVR because it safely turns untrusted programs into pass/fail outcomes.

Three details matter for timing measurements. First, the local timer wraps the execution of each input/output pair after compilation, so it excludes sandbox startup and the one-time compilation step, but it still measures local worker wall-clock time. Second, the input/output pairs are replayed sequentially in one runner process; global state, imports, caches, or input/output redirection state can therefore carry across cases. Third, timeouts are detected by the parent process waiting for runner responses, so timeout behavior also depends on the local worker's ability to observe and clean up the sandboxed process on schedule. These details are acceptable for pass/fail RLVR, but they make the local path a weak timing instrument because process scheduling, shared worker load, and batch execution state can enter the measured duration.

BigO(Bench) further investigates how to make local sandbox timing more stable, including by refining the clock used to time programs and by trying to control CPU contention in its complexity-labeling sandbox \citep{chambon2025bigobench}.\footnote{\url{https://github.com/facebookresearch/BigOBench/tree/main/src/complexity/sandbox}} The main difference here is the workload around the sandbox. BigO(Bench) runs execution as an offline labeling job where execution can be isolated as the main workload. In optimization RL, execution runs concurrently with model inference, rollout orchestration, logging, and other worker-side load. Even if the sandbox itself is stabilized, the surrounding RL system can still perturb the durations used as rewards.

The local-timing failure has three layers: the local harness runs in a much faster regime, returns fewer usable timing cells, and gives unstable repeated measurements of the same code and tests. We then compare local and CES outputs directly: the mismatch is not just extra noise or a simple deterioration, because the two backends change which problems look duration-filterable and post-hoc fits cannot recover the CES ordering. Finally, we test the practical effect of local execution inside the RL loop, first with synthetic concurrent load and then with real local-execution RL runs.

\paragraph{Local sandbox speed pushes evaluation into the noisy sub-0.1\,s regime.}
As a post-hoc diagnostic, we can take four checkpoints from the result study in \cref{sec:main_results}: a 32B baseline, a 32B optimization-RL model, a 7B baseline, and a 7B optimization-RL model. We evaluated the same generated solutions with two harnesses, one backed by CES and one backed by the local sandbox. The local harness is much faster. At a 10\,s timeout the two environments are still close because almost all correct solutions have enough time to finish, but the difference grows rapidly as the timeout enters the sub-second range. At 0.1\,s, local execution inflates pass@1 by 21.3$\times$ on the 32B baseline, 32.8$\times$ on the 32B RL model, 50.3$\times$ on the 7B baseline, and 20.0$\times$ on the 7B RL model (\cref{tab:local_timeout_inflation}). This extra speed is not a useful measurement signal: it moves many executions into the regime where tiny wall-clock perturbations dominate. The table already shows the resulting instability. Under local execution at 0.1\,s, the 7B baseline scores higher than the 7B optimization-RL model, even though the calibrated DMC-Optim results show the optimization model is stronger at strict duration thresholds. Local execution therefore introduces a first problem: it places the evaluation in a much faster timing regime, where the instabilities studied next can only become more important.

\begin{table}[h!]
\centering
\small
\caption{\textbf{Tight-timeout pass@1 inflation under local execution.}
The same solution sets are evaluated under CES and local execution.
At a 0.1\,s timeout, local execution reports many more completions because the same programs receive much shorter measured runtimes under the local harness.
The table is used as a backend diagnostic, not as a model-quality result.
}
\label{tab:local_timeout_inflation}
\begin{tabular}{lrrr}
\toprule
{\bfseries Checkpoint} & {\bfseries CES pass@1 at 0.1\,s} & {\bfseries Local pass@1 at 0.1\,s} & {\bfseries Inflation} \\
\midrule
32B baseline & 0.84\% & 17.91\% & 21.3$\times$ \\
32B optimization RL & 0.57\% & 18.71\% & 32.8$\times$ \\
7B baseline & 0.28\% & 14.09\% & 50.3$\times$ \\
7B optimization RL & 0.56\% & 11.20\% & 20.0$\times$ \\
\bottomrule
\end{tabular}
\end{table}

\paragraph{Local sandbox distorts execution results and lowers coverage.}
Putting aside the timing values themselves, we look at whether the harness returns a usable timing matrix at all. We ran the same optimization-test family once on CES and once on the local sandbox. On a 50-problem slice of the raw outputs, local execution produced fewer usable timing measurements: success rate fell from 99.62\% to 97.41\%, non-success rate rose from 0.38\% to 2.59\%, and usable per-solution per-test timing coverage fell from 99.2\% to 93.3\%. Some problems deteriorated much more sharply, with timing-matrix coverage falling by up to 50\%. This is already a trustworthiness issue, independent of how the remaining durations are distributed. Missing cells are not harmless blanks: local out-of-memory events can be harder to isolate from the host worker, small runtime perturbations can push near-limit executions across a timeout boundary, and worker-side instability can change the execution status rather than only the measured duration.

\paragraph{Ask which code is fastest, get a different answer each time.}
We next tested whether the local sandbox can at least produce a stable timing signal relative to itself. In a dedicated local measurement campaign, we repeatedly re-evaluated unchanged candidate solutions on the same problems, asking only whether one rerun recovers the same within-problem timing ranking as another. For each optimization test, the candidate duration is ranked against the human-reference durations for that same test, with \(0\%\) denoting the fastest position and \(100\%\) the slowest. The mean percentile is the average of these per-test percentiles across the optimization tests. A mean percentile of \(20\%\) means the solution looks faster than most references; a mean percentile of \(70\%\) means it looks much slower. The diagnostic asks whether rerunning the same code on the same tests keeps that leaderboard position fixed.

Across 23 problems and 425 reruns, repeated local re-execution moved the mean-percentile score by 41.2 percentage points peak-to-peak on average: for each problem, we take the difference between the best and worst mean-percentile score obtained across reruns, then average that range across problems. Equivalently, the within-problem standard deviation of mean percentile averaged 10.961 percentage points and the within-problem range averaged 41.173 percentage points. On a ranked reward, this is not a cosmetic fluctuation: an 11-point standard deviation is large enough to move a solution across a top-30\% gate, and a 41-point range is the difference between looking like a clearly fast solution and a clearly mediocre one. In leaderboard terms, a solution can move from around the top 30\% to around the bottom 30\% across local reruns without any code change. The largest observed within-problem range, 73.743 percentage points, would be enough to move a candidate across almost the whole useful part of the leaderboard without changing the code.

We also check whether local reruns preserve the ordering of the tests themselves. For each rerun, we take each optimization test's median duration across re-executed human reference solutions, and fit a line from the stored median duration of that test to the newly measured local median duration. A stable timing backend should give a positive and similar slope across reruns: if the stored data says test A is a heavier workload than test B, rerunning locally should not make A look easier than B. A sign flip is the clearest failure case, because the fitted line changes direction; one rerun says larger stored workloads remain larger, whereas another says they become smaller. This happened on 18 of 22 problems with enough measurements. We also compute the fitted-slope range, the maximum fitted slope minus the minimum fitted slope across local reruns for the same problem. This range averaged 67.213, reached 208.401 at the 90th percentile, and reached 651.462 in the worst case. These are slopes in a linear map from stored seconds to local seconds after aggregating by test median, so their absolute scale is less important than the instability: the same problem can move from an almost flat fit to a very steep fit across reruns. The worst-problem \(R^2\) medians were near zero, meaning the stored workload ordering explains almost none of the local timing variation on those problems. Together with the mean-percentile movement above, the same code and test suite can therefore look substantially stronger or weaker across local reruns.

\begin{figure*}[t!]
\centering
\includegraphics[width=0.82\textwidth]{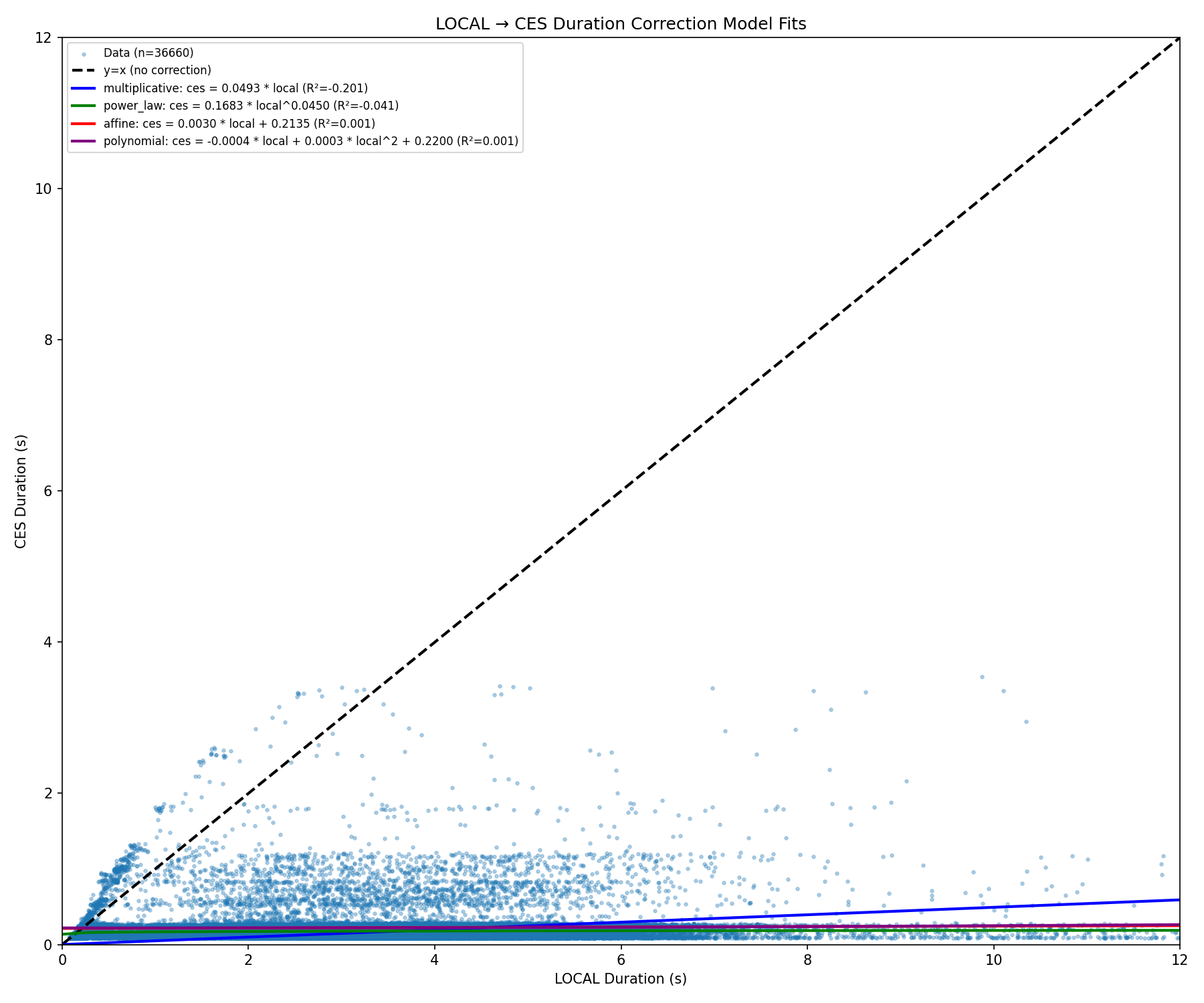}
\caption{\textbf{Attempting to fit local runtimes back to CES runtimes fails.}
The plot aggregates 36{,}660 joined local/CES duration pairs.
The identity line is shown only as a reference: the thin diagonal band above it is the benign case where local execution is faster than CES but still roughly ordered with it.
The larger pattern is different.
Many points with very small CES durations spread over a wide range of local durations, forming a long horizontal plume rather than a tight line.
This is the failure mode: local execution sometimes adds worker-side delay that is not explained by the code-test pair itself, so the same CES-fast execution can look locally fast, moderately slow, or extremely slow.
The fitted mappings therefore collapse toward nearly flat predictions, and all four have negative leave-one-problem-out \(R^2\).
Local timing is not missing only a scale factor; it contains extra variation that cannot be removed by a simple post-hoc calibration.
}
\label{fig:local_to_ces_fits}
\end{figure*}

\begin{figure*}[t!]
\centering
\includegraphics[width=0.92\textwidth]{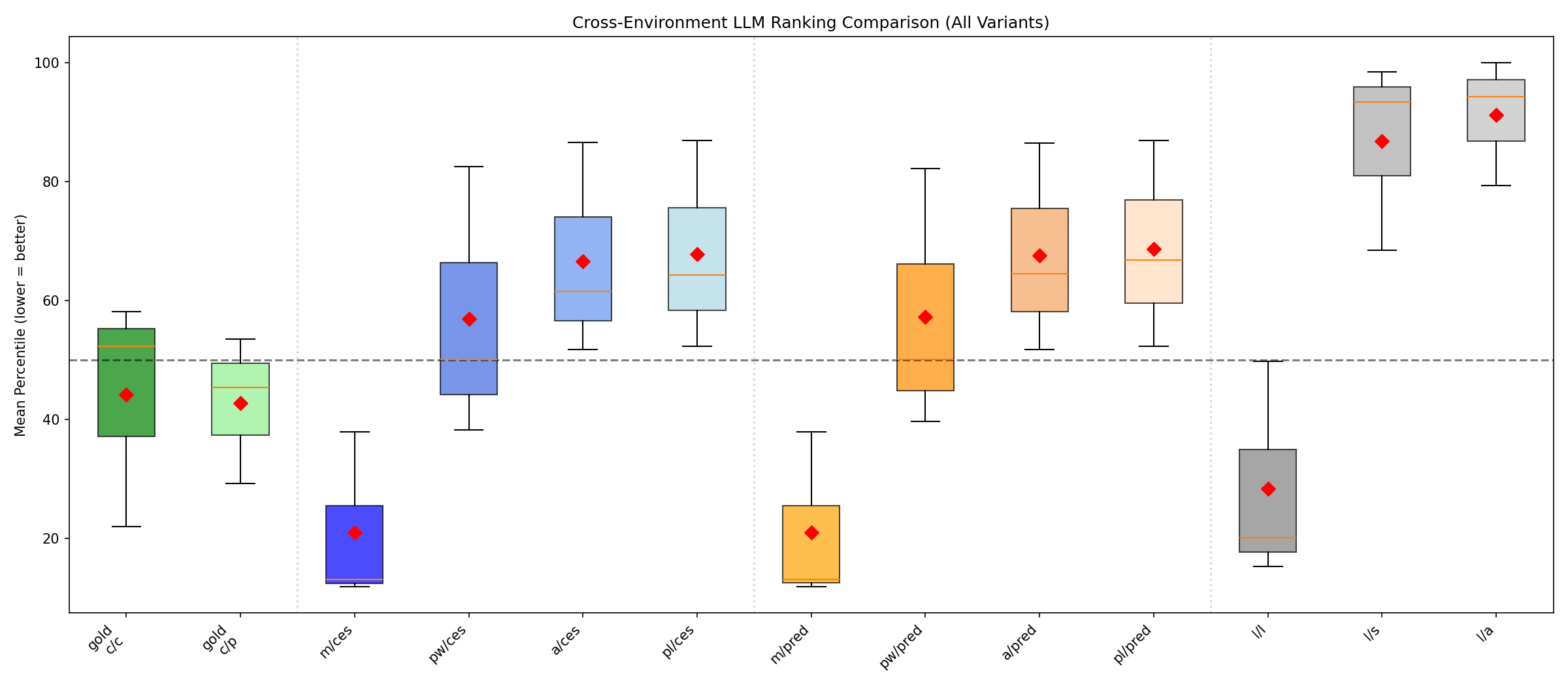}
\caption{\textbf{Local rankings remain misaligned even after cross-environment correction.}
The figure ranks model-generated solutions against distributions of human-reference solutions on the same problems and tests.
For one problem and one optimization test, we take the duration of the model solution under the protocol named on the x-axis, insert it into the duration distribution of human solutions for that same test, and record its percentile.
A percentile of \(0\%\) means the model solution is faster than all human references on that test; \(50\%\) means it is near the middle of the human distribution; \(100\%\) means it is slower than all human references.
The y-axis is the mean of these per-test percentiles over the optimization tests of the problem, so lower values mean that the model solution ranks faster relative to the human reference set.
Each box summarizes the three problem-level mean-percentile scores available in this paired local/CES diagnostic, and the red diamond marks their mean.
The dashed line at 50\% is the middle of the human reference distribution.
The x-axis labels are comparison protocols: \emph{gold c/c} ranks model durations measured on CES against human durations measured on CES, while \emph{gold c/p} keeps model CES durations but replaces human CES durations by affine-corrected stored human durations.
The blue and orange groups test whether local model durations can be repaired by local-to-CES maps: \(m\), \(pw\), \(a\), and \(pl\) denote multiplicative, power-law, affine, and polynomial corrections; the suffix \emph{/ces} ranks those corrected local model durations against human CES durations, and \emph{/pred} ranks them against affine-corrected stored human durations.
The gray diagnostic labels compare raw local model durations against raw local human durations (\emph{l/l}), stored human durations (\emph{l/s}), or affine-corrected stored human durations (\emph{l/a}).
The reference is therefore \emph{gold c/c}: a successful replacement for CES timing would put its box close to this one, problem by problem.
Lower is not automatically better in this diagnostic.
A box far below \emph{gold c/c} means local timing makes the same model solutions look spuriously fast; a box far above it means the correction makes them look spuriously slow.
\emph{Gold c/p} stays close to \emph{gold c/c}, which is the desired behavior of the stored-duration affine correction when model executions are still measured on CES.
Among local-to-CES repairs, the power-law variants are the least bad, but they still move the mean percentile from 44.1\% under \emph{gold c/c} to 56.9\%--57.3\%; the multiplicative variants move in the opposite direction to 20.9\%, and the affine/polynomial variants move to 66.6\%--68.6\%.
The gray raw-local diagnostics are worse as practical alternatives.
Even \emph{l/l}, the most favorable raw-local comparison because it re-executes both the model solution and all human reference solutions locally every time a model solution is ranked, is still far from the CES reference and makes the model look spuriously fast.
The mixed-backend raw-local comparisons are worse: \emph{l/s} and \emph{l/a} rank raw local model durations against stored or affine-corrected stored human durations, and push the mean percentile to 86.8\%--91.2\%.
This is the failure one would expect from comparing quantities produced by different timing backends.
This is why local timing cannot be used as a drop-in replacement for CES timing: even after correction, it does not recover the ranking used by the reward and evaluation.
}
\label{fig:local_cross_env_ranking}
\end{figure*}

\paragraph{Local and CES measurements are incompatible.}
After setting up a new remote execution backend that we call CES, we could compare the two backends on the same optimization tests. CES is not an absolute ground truth, but the two backends would need to be commensurate if they were ever mixed in the same reward path, for example by using local execution as a fallback when CES is overloaded. The comparison also checks whether our default local backend and the remote backend differ only by degree, with one being noisier but still agreeing on average, or whether they induce different timing orderings altogether. In the latter case, it probably means at least one backend cannot be trusted for timing measurements. They are not commensurate: local and CES measurements disagree by more than a small perturbation around the same signal. On the full dataset, the same optimization tests look much more duration-filterable locally: 81.1\% of problems are duration-filterable under the local sandbox versus 48.2\% on CES. Because the filterability criterion is based on a robust coefficient of variation, this gap is difficult to explain by a simple global speedup. A uniform rescaling would mostly preserve robust CV. The more plausible explanation is that local execution injects extra timing spread, making some otherwise non-filterable problems appear useful for timing. The 50-problem comparison points in the same direction: local execution reduces usable duration-cell coverage while making more problems appear duration-filterable. Finally, if local timing were only biased by a scale factor, we could fit a mapping from local durations to CES durations. We tried this on 36{,}660 joined local/CES duration pairs. Multiplicative, power-law, affine, and polynomial mappings all failed under leave-one-problem-out evaluation, with negative cross-validated \(R^2\) (\cref{tab:local_to_ces_fit_summary}). The fitted maps collapse toward nearly flat predictions rather than recovering CES durations, and ranking quality remains badly misaligned (\cref{fig:local_to_ces_fits,fig:local_cross_env_ranking}). Local timing is therefore not a quantity we can adjust or calibrate.

\begin{table}[h!]
\centering
\small
\caption{\textbf{Failed local-to-CES correction on 36{,}660 paired measurements.}
The mappings are fit under leave-one-problem-out cross-validation on the overlapping local/CES data.
If local timing were only a rescaled version of CES timing, one of these maps would predict held-out CES durations from local durations.
Instead, all cross-validated \(R^2\) values are negative, meaning the fitted maps are worse than ignoring the local duration and predicting an average held-out duration.
}
\label{tab:local_to_ces_fit_summary}
\begin{tabular}{lr}
\toprule
{\bfseries Mapping family} & {\bfseries Cross-validated \(R^2\)} \\
\midrule
Multiplicative & -0.360 \\
Power-law & -0.057 \\
Affine & -0.213 \\
Polynomial & -0.254 \\
\bottomrule
\end{tabular}
\end{table}

\paragraph{Local sandboxes collapse under CPU contention.}
The previous diagnostics already identify three failure points that make local timing hard to use for optimization RL. The remaining operational question is how local execution behaves when the worker is also under CPU pressure, as it would be during rollout collection. To measure sensitivity to contention, we reran the same 724 local timing jobs while adding background CPU load on the same node at 50\%, 100\%, and 200\% of node CPU capacity. Even with no added load, 294 of 724 jobs did not complete. This rose to 485/724 at 50\% load, 531/724 at 100\%, and 613/724 at 200\%. We observed very slow executions, local out-of-memory failures, and worker-side instability. These events cannot be isolated by a service-side retry because worker-side contention is the failure mechanism.

\clearpage
\paragraph{Local sandbox optimization RL fully fails.}
The standalone diagnostics above were obtained outside the full RL loop, which can probably only make things worse. We therefore tried the direct ablation: eight local-execution variants of existing CES-based Qwen 2.5 7B configurations, each targeting 5{,}000 RL steps and using direct local outcomes. These variants skipped CES during training, while the evaluation section still used CES, so the ablation isolates training-time execution rather than changing the reported evaluation backend as well. The correctness-only baseline, which runs no optimization tests, completed successfully. The seven optimization-test variants did not: each exhausted all six scheduler attempts (the original launch plus five restarts), accumulated 12--32 hours of wall-clock time, and reached only 160--1{,}620 training steps. In standalone measurements, local execution appeared faster per test than CES; inside the RL loop, the full runs were much slower. For configurations with optimization tests, local execution was 4--8$\times$ slower than CES end to end. The failures were out-of-memory kills on the training nodes, followed by distributed-communication failures across ranks, because local execution consumes memory on the same machines already holding the model and optimizer states. This is not a proof that no local-execution training configuration can ever run: lowering rollout concurrency, reducing the number of executed tests per step, or allocating more CPU and memory to the training workers would likely avoid some crashes. But those changes would directly reduce throughput, and the current local runs were already slow when they made progress: data-loading time reached 355\,s per step in the worst case, compared with single-digit seconds in the corresponding CES runs. This end-to-end ablation turns the timing argument into an operational one: for optimization-test RL at this model scale and throughput target, CES is not only a cleaner timing source but also what makes the training run feasible.

Local execution remains useful only for non-timing fallback. It can still help recover a verdict when a remote request fails. It cannot be used as the timing source for optimization rewards or evaluation, and it cannot be mixed with CES timings as if the two were commensurate. The two backends are fully misaligned for timing: local durations are faster, less complete, unstable across reruns, not calibrated back to CES, and coupled to the training workload itself. Mixing local and CES durations would therefore perturb any model trying to learn a serious timing-based reward.

\subsection{Remote execution service}
\label{sec:app_exec_architecture}

CES is a pre-existing remote execution backend that we use for timing-sensitive dataset construction, training, and evaluation; building the service itself is not a contribution of this paper. A training or evaluation worker sends candidate code and tests to the service; CES schedules the executions on service-managed virtual-machine capacity, runs them in remote sandboxes, and returns execution statuses and durations. The service property we rely on is isolation: each test runs under a 1\,GB memory limit and a 10\,s hard limit, and the execution capacity is provisioned so CPU cores are isolated and assigned to sandboxed jobs rather than implicitly shared with rollout processes. Sub-second reward thresholds are applied after CES returns durations rather than by changing the service limit. Building such a service at scale is an infrastructure problem in its own right and is outside the scope of this paper.

The main architectural property we use is separation from the rollout workers. Candidate code is no longer executed on the same machines that run model inference, rollout orchestration, data movement, and model or optimizer state. CES instead places execution behind a separately provisioned queue where concurrency can be controlled, retries are visible, and one submitted program is less likely to spill over into another through shared CPU or memory pressure. This is why a remote service can be preferable even if one isolated code-test call has more overhead than a local sandbox.

CES is not assumed to be noiseless: it has service overhead, infrastructure failures, and timing drift. Its role is to put timing under one monitored remote backend whose failures can be retried and whose stored durations can be calibrated against fresh executions. \Cref{sec:app_conservative_timing} defines the fallback rule, \cref{sec:app_fallback_impact} quantifies infrastructure failures, and \cref{sec:app_duration_correction} explains the stored-duration calibration.

\subsection{Fallback after remote execution failures}
\label{sec:app_conservative_timing}

\paragraph{Fallback path.}
When CES returns an inconclusive result, the system first retries on CES. If a code-test pair keeps failing to return a definitive remote result, local fallback may be used to recover a correctness verdict. Definitive CES results are kept as-is.

Early stopping interacts with fallback. When CES returns a definitive hard failure on any test, the system can skip local reruns for remaining inconclusive tests because the trajectory will receive a negative reward regardless. If the only definitive non-success status is timeout, local reruns may still proceed for inconclusive tests because an unresolved test could reveal a wrong output or runtime error, which can be more penalized than a timeout in reward variants that treat timeouts as a separate outcome.

\paragraph{Timing measurements for fallback executions.}
Local execution on a training node can be faster than CES. A solution that completes locally may not complete under the remote sandbox because of service overhead, different CPU allocation, or memory behavior. A local success therefore cannot be treated as evidence that CES would have reported success within the same time limit. This is even more true for the local duration itself, which has no reason to be close to the CES duration that would have been returned for the same code-test pair. For example, consider one problem with a large edge-case optimization test. Early in RL, a model rollout may solve the test but allocate an unnecessarily large data structure; CES returns a slow duration, and the timing reward can penalize it. After policy updates, a later rollout on the same problem may deteriorate on that edge case and allocate even more memory. If CES still returns a definitive result, that slow duration or failure remains the correct training signal. If instead the request crosses the remote execution envelope and ends without a usable CES duration, local fallback can rerun the same code-test pair under different CPU or memory pressure and report a successful shorter runtime. Importing that local runtime would reverse the signal: the model would receive better timing evidence for a generation that was worse under the remote execution backend.

To avoid trusting faster local timing, the system applies a conservative conversion on timing-sensitive optimization tests: a test that succeeds locally during fallback is reclassified as a timeout with duration set to the time limit. Correctness tests use local fallback only to recover pass/fail verdicts when CES remains inconclusive; the local runtime is not imported as timing evidence.

The two test categories are treated differently because a timeout has different consequences in each one. On correctness tests, converting a locally recovered success into a timeout would make an otherwise correct rollout fail the all-tests-must-pass gate and receive reward \(-1\). We therefore use local fallback there only to recover the pass/fail verdict, not to create a timing value. On optimization tests, timeouts are part of the timing signal. In the main experiments, the timeout tolerance is 1.0, so optimization timeouts are tolerated by the binary gate. A converted local success then contributes a 10\,s duration to the ranking computation and can lower the quality score, but it does not by itself make the solution incorrect. This is the intended tradeoff: conservative timing for optimization tests, without creating false correctness failures on correctness tests.

\subsection{Quantitative impact of infrastructure failures}
\label{sec:app_fallback_impact}

Fallback only enters the reward when CES does not return a definitive result. We therefore first ask how often those missing remote results can matter at the scale of a training run.

\paragraph{Rare failures add up over training.}
Let \(f\) be the per-test infrastructure-failure rate and \(n\) the number of correctness tests. Consider a correct solution that would pass whenever CES returns a definitive answer. If failures hit tests independently, the probability that at least one correctness test reaches fallback is
\begin{equation}
P(\text{at-risk fallback} \mid \text{correct solution}) = 1 - (1 - f)^n .
\label{eq:false_negative_prob}
\end{equation}
The independence assumption is a simplification, but it is enough to show why a uniform fallback rule can be dangerous on correctness tests. A per-test failure rate can look negligible while the probability of seeing at least one fallback in a rollout is not negligible. \Cref{tab:false_negative_rates} shows the sensitivity. At a 2\% per-test failure rate with 30 correctness tests, 45.5\% of correct rollouts would reach fallback at least once. Even at \(f=0.1\%\), the risk is 3.0\% with 30 tests, 9.5\% with 100 tests, and 13.9\% with 150 tests.

The scale becomes even larger once we multiply by the optimizer steps. A Qwen 2.5 7B run has 10{,}000 optimizer steps, 16 optimizer data-parallel replicas, and up to 32{,}768 packed tokens per replica per step. Across the run, the optimizer can process up to 5.24B packed training tokens. If each consumed code snippet contributes about 11k tokens to the trainer, filling those batches requires about \(4.8\times 10^5\) code snippets. Because the trainer packs whole snippets and does not split a long one, a stricter count with uniformly 11k-token snippets would fit two snippets per replica per step, or \(10{,}000 \times 16 \times 2 = 3.2\times 10^5\) consumed snippets. At \(f=0.1\%\) and \(n=150\), the 13.9\% per-rollout risk therefore corresponds to roughly \(4.5\times 10^4\)--\(6.6\times 10^4\) trainer-consumed snippets, or about 0.5--0.7B packed training tokens. The current training path avoids the specific false-negative failure mode by not converting locally recovered correctness successes into timeouts.

\begin{table}[h!]
\centering
\caption{\textbf{Per-rollout fallback exposure from rare per-test infrastructure failures.}
Values are computed with \cref{eq:false_negative_prob} as a function of per-test infrastructure failure rate \(f\) and number of correctness tests \(n\).
If locally recovered correctness successes were converted to timeouts, for example, any fallback exposure in this table would create a false-negative risk for an otherwise correct rollout.
}
\label{tab:false_negative_rates}
\begin{tabular}{@{}lccccc@{}}
\toprule
{\bfseries Failure rate \(f\)} & {\bfseries \(n = 10\)} & {\bfseries \(n = 30\)} & {\bfseries \(n = 50\)} & {\bfseries \(n = 100\)} & {\bfseries \(n = 150\)} \\
\midrule
2.0\% & 18.3\% & 45.5\% & 63.6\% & 86.7\% & 95.2\% \\
0.5\% & 4.9\% & 14.0\% & 22.2\% & 39.4\% & 52.9\% \\
0.1\% & 1.0\% & 3.0\% & 4.9\% & 9.5\% & 13.9\% \\
0.01\% & 0.1\% & 0.3\% & 0.5\% & 1.0\% & 1.5\% \\
0.001\% & 0.01\% & 0.03\% & 0.05\% & 0.10\% & 0.15\% \\
\bottomrule
\end{tabular}
\end{table}

\paragraph{CES reliability study.}
We next check how these rates look in actual RL runs. The case study covers 13 training runs: six CWM 32B runs and seven Qwen 2.5 7B runs, all trained for 10{,}000 optimizer steps with CES used for timing-sensitive execution. Metrics are logged every 10 optimizer steps. Within each model and reward family, the runs use the same dataset, environment, model initialization, seed, learning rate, GRPO settings, and duration calibration; the main uncontrolled variable is launch time, which changes the CES service state seen during rollout collection. The distributed asynchronous training system also introduces nondeterminism in rollout ordering and batch composition, so identical configurations are not expected to produce identical trajectories.

\Cref{tab:ces_postwarmup_runs} reports post-warmup statistics from all 13 runs, computed after step 1{,}000. The logged CES-only rate is a rollout-level quantity: it is high only when the model output parses, the code-test execution is correctly sent to CES, and CES resolves it without launching a local sandbox fallback. Local fallback is the complementary recovery path: it tracks cases where a local sandbox had to be launched after CES did not provide a definitive usable result. CES-only can therefore drop for two different reasons: parse failures before CES, or remote-execution failures after a parseable program is sent to CES. After step 1{,}000, parsing is above 97\% in these runs, so the remaining drops are mostly informative about CES and fallback behavior. In that post-warmup regime, Qwen 2.5 7B has 99.3\% CES-only execution on average, while CWM 32B has 98.1\%. The difference is expected because CWM 32B produces more correct programs, so more rollouts proceed to optimization-test execution and place concurrent load on CES. Local fallback averages 0.13\% for Qwen 2.5 7B and 0.21\% for CWM 32B; infrastructure and unknown failures average 0.10\% and 0.05\%, respectively. Here, an unknown result means that CES did not return a status that could be parsed into a normal program outcome. Since parse failures, wrong answers, exceptions, and timeouts are logged separately, these unknown results are treated as infrastructure-side inconclusive results in the analysis below.

\begin{table*}[t!]
\centering
\footnotesize
\setlength{\tabcolsep}{4pt}
\caption{\textbf{Post-warmup CES execution statistics across 13 RL runs.}
Statistics are computed over logged training metrics after step 1{,}000.
``CES-only'' is the logged fraction of rollout attempts that parse, are sent to CES, and are resolved by CES without launching a local sandbox fallback.
The infrastructure-failure column combines explicit infrastructure failures with unknown or inconclusive CES results; unknown results are unparsed CES outcomes after normal program errors and timeouts have been separated out.
The run-wide means are high, but the minimum column already shows that several runs contain short degraded intervals that are hidden by the mean.
}
\label{tab:ces_postwarmup_runs}
\begin{tabular}{llcccc}
\toprule
{\bfseries Run} & {\bfseries Reward} & {\bfseries CES-only mean} & {\bfseries CES-only min} & {\bfseries Local fallback} & {\bfseries Infra/unknown} \\
\midrule
CWM 32B p30 run 1 & p30 & 98.53\% & 95.09\% & 0.0003\% & 0.0001\% \\
CWM 32B p30 run 2 & p30 & 97.18\% & 54.33\% & 0.77\% & 0.16\% \\
CWM 32B p30 run 3 & p30 & 97.99\% & 91.59\% & \(<0.001\%\) & 0.0001\% \\
CWM 32B QP run 1 & QP & 98.40\% & 93.44\% & 0.0001\% & 0.0001\% \\
CWM 32B QP run 2 & QP & 98.04\% & 92.38\% & 0.49\% & 0.13\% \\
CWM 32B QP run 3 & QP & 98.52\% & 94.37\% & 0.0003\% & 0.002\% \\
\midrule
Qwen 2.5 7B p30 run 1 & p30 & 99.43\% & 97.98\% & 0.0006\% & 0.001\% \\
Qwen 2.5 7B p30 run 2 & p30 & 99.26\% & 79.94\% & 0.23\% & 0.18\% \\
Qwen 2.5 7B p30 run 3 & p30 & 98.68\% & 80.86\% & 0.71\% & 0.53\% \\
Qwen 2.5 7B p30 run 4 & p30 & 99.56\% & 97.70\% & 0.0004\% & 0.0007\% \\
Qwen 2.5 7B QP run 1 & QP & 99.42\% & 97.63\% & 0.0007\% & 0.001\% \\
Qwen 2.5 7B QP run 2 & QP & 99.44\% & 97.66\% & 0.0002\% & 0.0004\% \\
Qwen 2.5 7B QP run 3 & QP & 99.35\% & 97.12\% & 0.0008\% & 0.002\% \\
\bottomrule
\end{tabular}
\end{table*}

\begin{figure*}[t!]
\centering
\includegraphics[width=\textwidth]{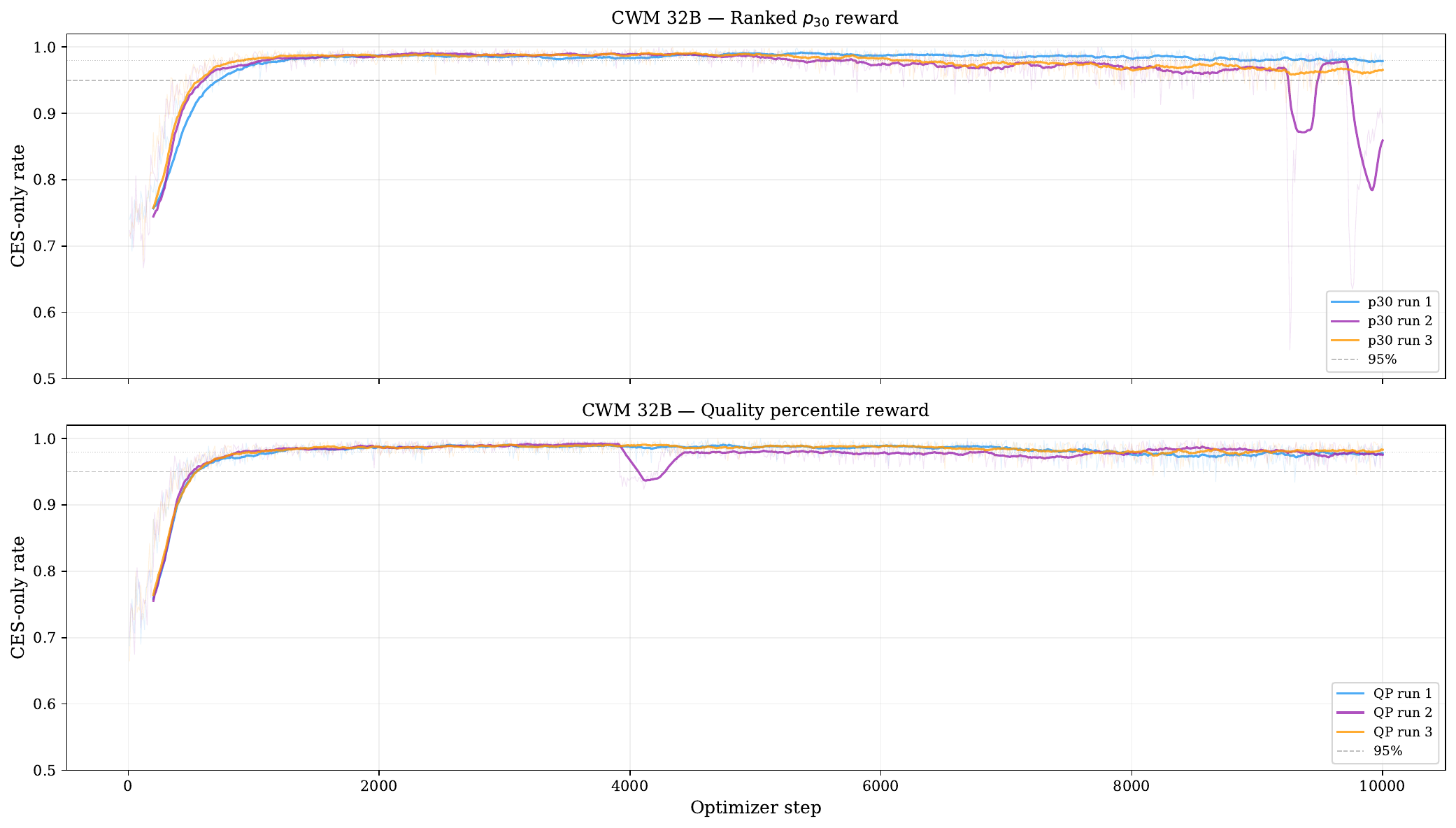}\\[0.75em]
\includegraphics[width=\textwidth]{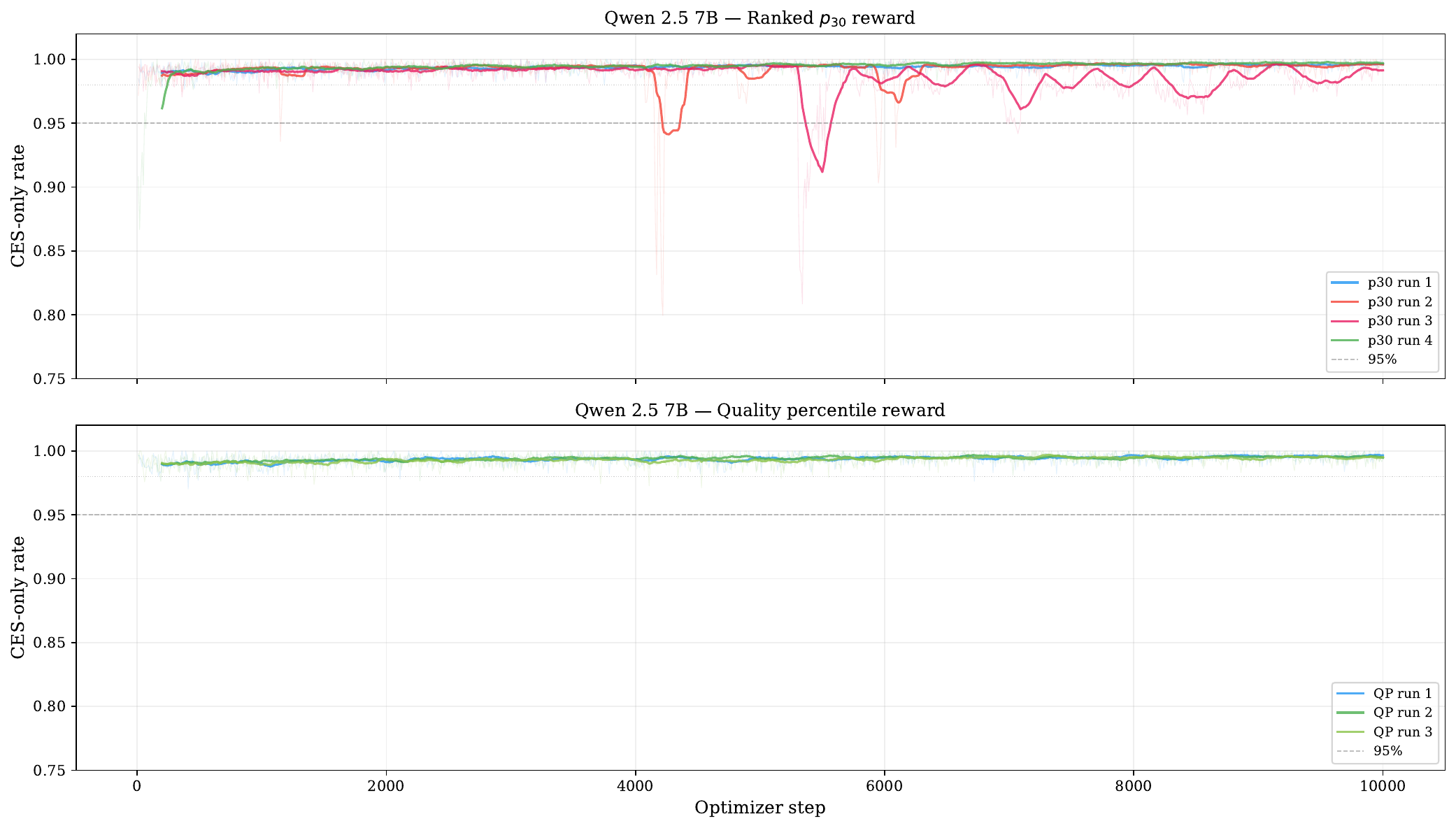}
\caption{\textbf{CES-only execution over 10{,}000 training steps.}
Top: six CWM 32B runs, split between ranked \(p_{30}\) and QP rewards.
Bottom: seven Qwen 2.5 7B runs, split between ranked \(p_{30}\) and QP rewards.
All runs show an early dip before step 1{,}000, then recover to the post-warmup operating regime.
This initial dip reflects parsing warmup rather than CES health: rollouts that fail to produce parseable code are never sent to CES and therefore do not count as CES-only.
The effect is largest for CWM 32B because the CWM recipe changes output format between SFT and RL~\citep{cwm2025}; at the start of RL, the model temporarily re-adapts to the required RL format.
Qwen 2.5 7B starts much closer to the final parse rate.
CWM 32B operates at slightly lower post-warmup CES-only rate because it produces more correct code and executes more tests.
After parsing has stabilized, the sharp drops in CWM 32B p30 run 2 and Qwen 2.5 7B p30 runs 2--3 are short remote-execution burst events rather than format-learning effects.
}
\label{fig:ces_case_ces_only}
\end{figure*}

\paragraph{Parsing warmups.}
\Cref{fig:ces_case_ces_only} also shows a universal early-training dip during the first roughly 500 optimizer steps, with recovery by step 1{,}000. This is not a CES infrastructure failure. The logged metrics show that this early dip tracks parse rate: rollouts that fail to produce parseable code never reach CES, while local rerun and infrastructure-failure rates remain near zero during the dip. For CWM 32B, the main early issue is the output-format switch between SFT and RL documented in the CWM report~\citep{cwm2025}: around step 10, roughly 26\% of rollouts fail to parse while the model re-adapts to the required RL format. Qwen 2.5 7B has a much smaller dip because its SFT checkpoint already produces well-formatted code for this environment. Once parsing stabilizes above 97\%, lower CES-only values are no longer a format-learning artifact and can be interpreted as remote-execution or fallback events. This is why the tables below report post-warmup statistics.

\paragraph{Random bursts of infra errors.}
After parsing warmup, the main visible failures are short bursts. \Cref{tab:ces_burst_events} lists every post-warmup interval where CES-only falls below 90\%. There are four events: two in CWM 32B p30 run 2, one in Qwen 2.5 7B p30 run 2, and one in Qwen 2.5 7B p30 run 3. The worst CWM 32B event drops CES-only to 54.3\%, pushes local rerun to 41.6\%, and raises timeout rate to 14.1\%. The Qwen 2.5 7B bursts are milder, with CES-only minima around 80\% and local rerun peaks around 19\%. Within the same model and reward family, other runs with the same configuration do not show these bursts, so these events are best treated as service-state events rather than properties of the reward.

\begin{table*}[t!]
\centering
\footnotesize
\setlength{\tabcolsep}{3.5pt}
\caption{\textbf{Post-warmup CES burst events.}
Each row is a contiguous window after step 1{,}000 where CES-only execution falls below 90\%.
The records column counts logged metric records; metrics are logged every 10 optimizer steps.
The return columns compare the mean trajectory return during the burst with the post-warmup run-wide mean.
Bursts are short but severe: the worst event lasts only 8 logged records but drops CES-only to 54.3\%.
}
\label{tab:ces_burst_events}
\begin{tabular}{lcccccc}
\toprule
{\bfseries Run} & {\bfseries Window} & {\bfseries Records} & {\bfseries CES-only min} & {\bfseries Peak fallback} & {\bfseries Peak timeout} & {\bfseries Return: burst / run} \\
\midrule
CWM 32B p30 run 2 & 9{,}240--9{,}310 & 8 & 54.3\% & 41.6\% & 14.1\% & \(+0.02 / +0.34\) \\
CWM 32B p30 run 2 & 9{,}720--10{,}000 & 29 & 63.5\% & 35.7\% & 10.1\% & \(+0.17 / +0.34\) \\
Qwen 2.5 7B p30 run 2 & 4{,}160--4{,}220 & 7 & 79.9\% & 19.0\% & 4.7\% & \(-0.49 / -0.31\) \\
Qwen 2.5 7B p30 run 3 & 5{,}310--5{,}390 & 9 & 80.9\% & 18.4\% & 5.2\% & \(-0.64 / -0.33\) \\
\bottomrule
\end{tabular}
\end{table*}

\begin{figure*}[t!]
\centering
\includegraphics[width=\textwidth]{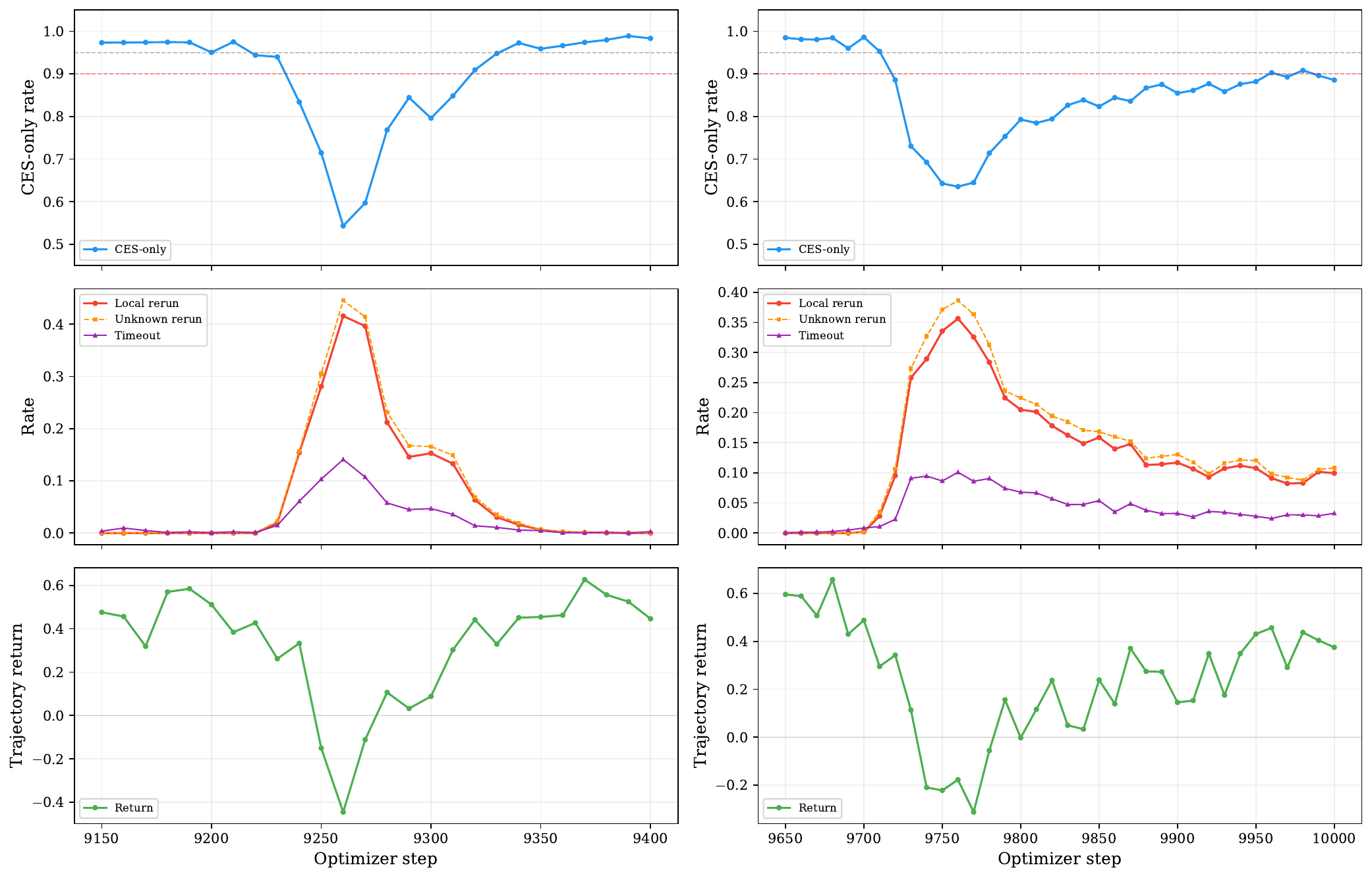}\\[0.75em]
\includegraphics[width=\textwidth]{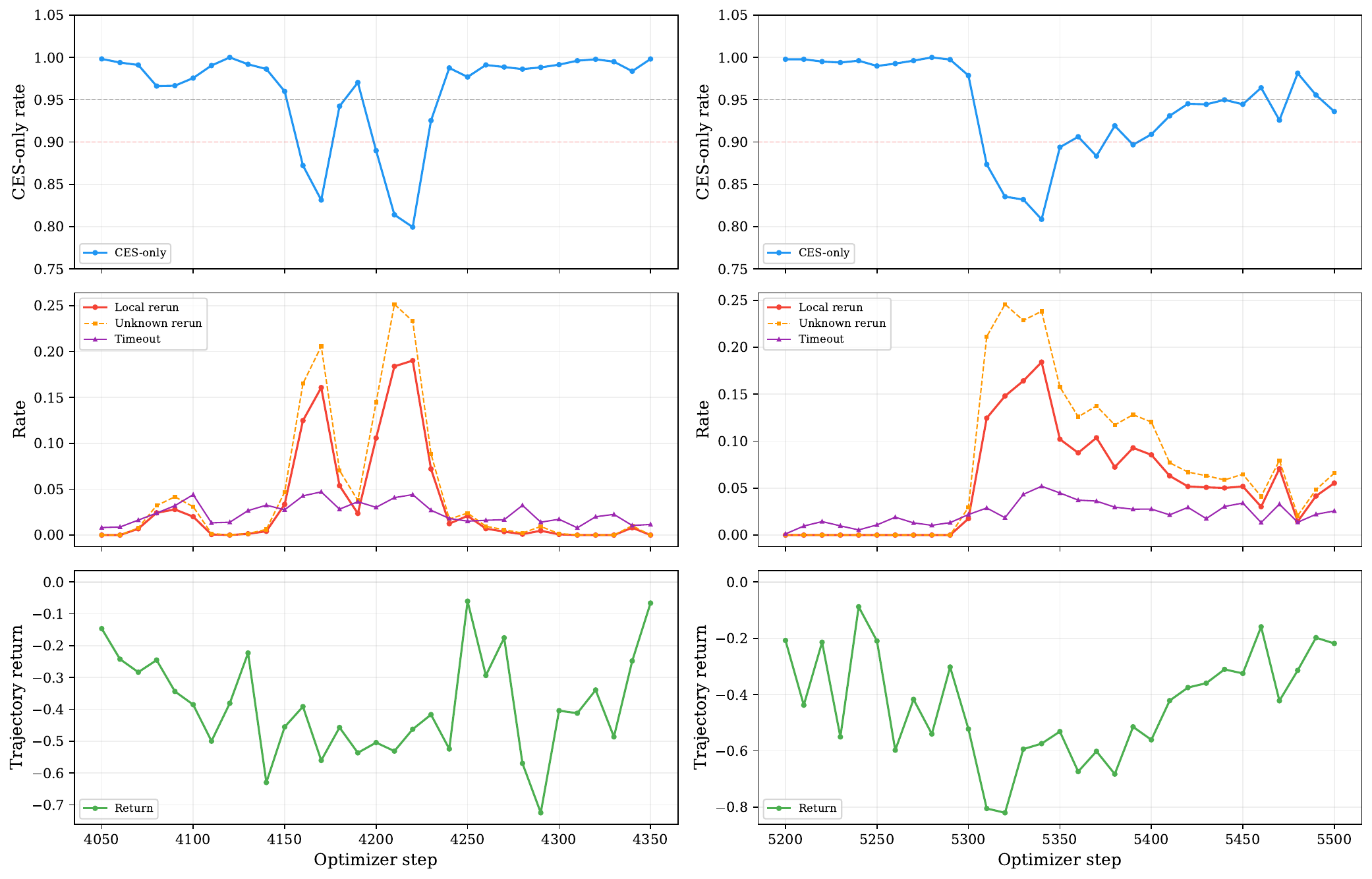}
\caption{\textbf{Anatomy of CES burst events.}
Top: the two CWM 32B p30 run 2 bursts.
Bottom: the two Qwen 2.5 7B p30 bursts.
Each panel shows CES-only execution, local rerun and timeout rates, and trajectory return around the degraded interval.
The mechanism is consistent across models: CES-only drops, inconclusive-result reruns rise, timeout rate rises, and the return falls during the burst.
For CWM 32B p30 run 2, CES-only drops from about 98\% to 54\% within two logged records and recovers within eight logged records in the first burst.
}
\label{fig:ces_case_burst_anatomy}
\end{figure*}

The burst mechanism is not a large spike in hard infrastructure errors. In the worst single logged step, CWM 32B p30 run 2 at step 9{,}260, hard infrastructure errors are 0.00\%, unknown or inconclusive CES results are 6.3\%, unknown-triggered local reruns reach 44.5\%, and timeouts reach 14.1\%. These unknown results are not model parse failures or ordinary test failures; those statuses are logged separately. They are CES outcomes that were not parsed into a normal verdict and are therefore grouped with infrastructure-side inconclusive results. A monitor that only tracks hard infrastructure errors would miss this degradation. The service is still returning responses, but too many of them are inconclusive, which pushes execution through fallback and changes the optimization-test durations seen by the reward.

\begin{figure*}[t!]
\centering
\includegraphics[width=0.95\textwidth]{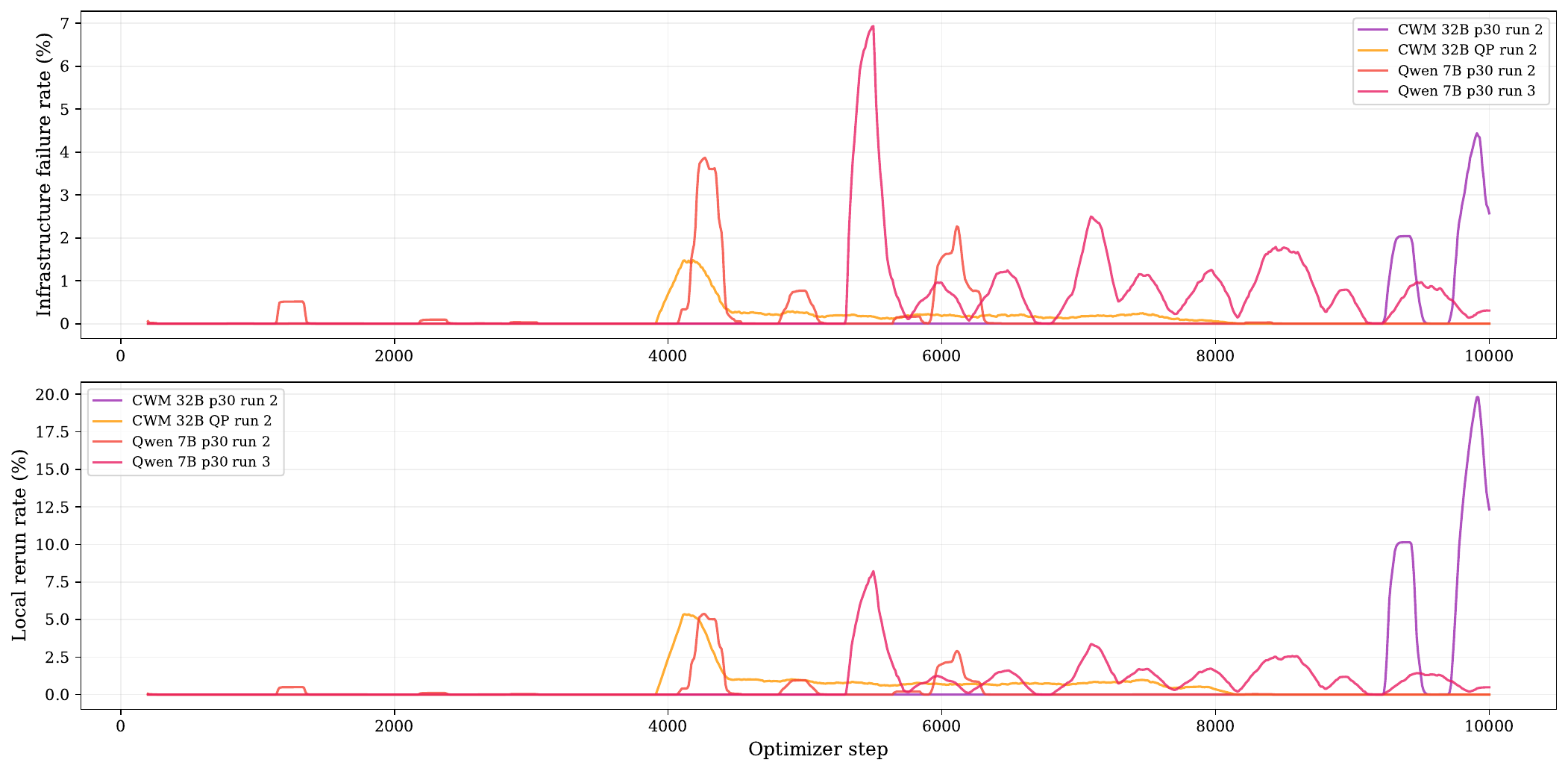}
\caption{\textbf{Inconclusive CES results and local fallback during bursty runs.}
The figure tracks infrastructure or unknown-result rates together with local fallback for the four runs that contain nonzero post-warmup events.
Unknown results are unparsed CES outcomes after normal test failures and timeouts have been logged separately, so they are treated as infrastructure-side inconclusive results here.
Failures are temporally concentrated: most steps have near-zero rates, then short intervals produce visible spikes.
The worst CWM 32B burst is dominated by unknown or inconclusive CES results that trigger local rerun, not by hard infrastructure-error responses.
}
\label{fig:ces_case_infra_failures}
\end{figure*}

\begin{figure*}[t!]
\centering
\includegraphics[width=0.95\textwidth]{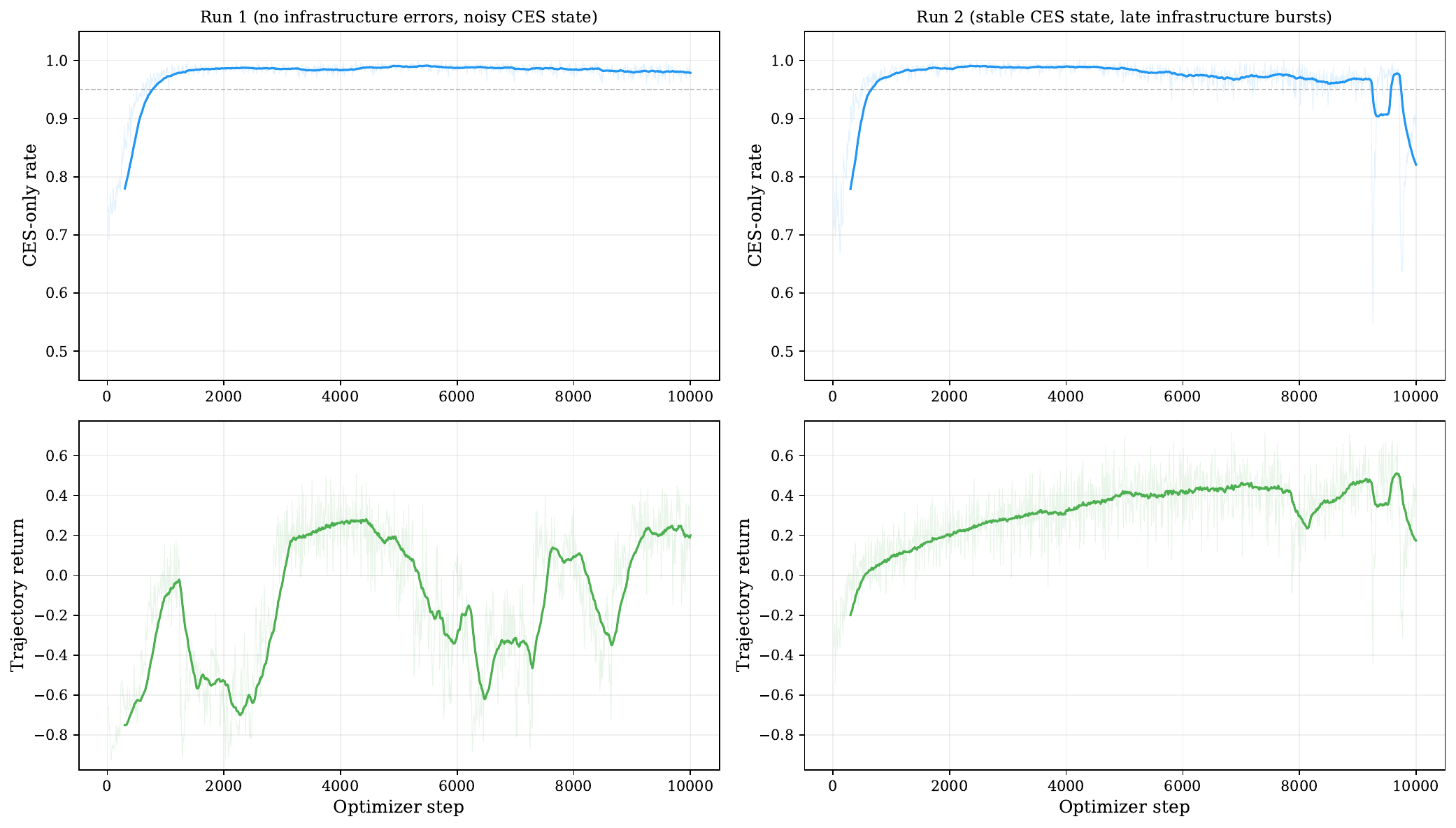}
\caption{\textbf{Silent CES degradation can affect timing rewards.}
The figure compares CWM 32B p30 run 1 and run 2.
Run 1 keeps CES-only above 95\% and has almost no local fallback, but its trajectory returns are volatile throughout training.
Run 2 has two late visible CES bursts, but between step 1{,}000 and step 9{,}000 its returns are smoother and higher.
This does not identify a single cause for run 1: training nondeterminism and early exploration can also change returns.
It shows why a service-state degradation can matter even without hard infrastructure errors: if CES still returns verdicts but records shifted or noisier durations, pass/fail outcomes may look normal while the timing reward changes.
}
\label{fig:ces_case_silent_bursty}
\end{figure*}

\paragraph{Beyond infra errors, silent service degradations.}
\Cref{fig:ces_case_silent_bursty} compares two CWM 32B p30 runs with the same configuration. Run 1 has almost no visible fallback or infrastructure events: 98.53\% mean CES-only, 95.09\% minimum CES-only, 0.0003\% local fallback, and 0.0001\% infrastructure or unknown failures. Run 2 has visible late bursts: 97.18\% mean CES-only, 54.33\% minimum CES-only, 0.77\% local fallback, and 0.16\% infrastructure or unknown failures. Yet run 1 has much noisier training returns between steps 1{,}000 and 9{,}000: mean return \(-0.14\), standard deviation 0.34, and range 1.43, compared with \(+0.34\), 0.15, and 0.85 for run 2. This comparison does not prove that the noisier run is caused by CES state alone; asynchronous rollout ordering and early exploration differences can also change the trajectory, and the pass-rate means differ from early training onward. It does show the failure mode we have to monitor: CES can keep returning verdicts, so pass/fail outcomes and fallback rates look healthy, while slower or noisier service state shifts the recorded durations used by the ranking reward.

\begin{table*}[t!]
\centering
\small
\caption{\textbf{Two CWM 32B p30 runs with different CES symptoms.}
Both runs use the same model, data, reward, and hyperparameters.
Run 1 has almost no visible fallback or infrastructure events but volatile returns; run 2 has late visible bursts but smoother returns before the bursts.
The comparison motivates monitoring the reward stream and duration distributions, not only hard failure rates.
}
\label{tab:ces_silent_vs_bursty}
\begin{tabular}{lcc}
\toprule
{\bfseries Metric} & {\bfseries Run 1: few visible failures, noisy return} & {\bfseries Run 2: late bursts} \\
\midrule
CES-only mean after step 1{,}000 & 98.53\% & 97.18\% \\
CES-only minimum after step 1{,}000 & 95.09\% & 54.33\% \\
Local fallback mean & 0.0003\% & 0.77\% \\
Infrastructure/unknown mean & 0.0001\% & 0.16\% \\
Return mean, steps 1{,}000--9{,}000 & \(-0.14\) & \(+0.34\) \\
Return std., steps 1{,}000--9{,}000 & 0.34 & 0.15 \\
Return range, steps 1{,}000--9{,}000 & 1.43 & 0.85 \\
Pass-rate mean after step 1{,}000 & 0.45 & 0.68 \\
\bottomrule
\end{tabular}
\end{table*}

For CWM 32B p30 run 2, 37 post-warmup logged records fall below 90\% CES-only and 73 fall below 95\%. With 16 optimizer data-parallel replicas and about two consumed snippets per replica per step, this corresponds to roughly 1{,}200 consumed code snippets below 90\% CES-only and 2{,}300 below 95\%. These are small fractions of the 10{,}000-step run, 0.4\% and 0.8\% under the logged-step count, but the per-step effect can be severe: during the worst burst, up to 41.6\% of executions are handled by local fallback. Fallback can be rare in the run-wide average, concentrated enough to affect only a small fraction of optimizer steps, and still large enough during a burst to change the reward stream for those steps.

\clearpage

\subsection{Alternative fallback designs}
\label{sec:app_fallback_tradeoffs}

\paragraph{Per-category conversion.}
In training, we handle correctness and optimization tests differently. For optimization tests, where the duration enters the reward, a locally recovered success is converted to a timeout. For correctness tests, where the goal is only to recover a pass/fail verdict, we keep the local verdict because the 10\,s limit makes a sub-second local success likely to remain a CES success. This removes the correctness false-negative mode while keeping conservative timing for optimization. If CES reliability degrades, the remaining refinement would be to tune category-specific fallback thresholds and decide when a trajectory should be discarded instead of recovered.

\paragraph{Immediate CES retry.}
Before local fallback is used, inconclusive CES results are retried immediately, before the rollout reward is finalized. This is useful for transient request failures: the reward can still be computed from a remote execution without mixing timing backends. It does not solve persistent bursts. If the same batch keeps returning infrastructure or unknown results, retrying only delays the step and the system still needs the per-category conversion above or a discard rule. The retry budget also has to stay short in asynchronous RL. If a rollout is retried for too long, it is eventually consumed by an optimizer whose policy has already moved on. The same issue appears inside a prompt group: if some rollouts for a problem are measured immediately while the inconclusive ones are re-executed much later, transient changes in CES speed can make the durations compared by the advantage computation less comparable.

\paragraph{Discarding or skipping affected data.}
One alternative is to discard any trajectory where at least one test falls back to local execution. This removes both false positives and false negatives from fallback, but it loses too much data. With \(n_{\text{total}}\) total tests, the discard rate is \(1-(1-f)^{n_{\text{total}}}\). At a 2\% per-test failure rate with 80 total tests, about 80\% of trajectories would be discarded. Even at 0.5\%, the discard rate reaches 33\%. The discard probability also correlates with the number of tests per problem, biasing training toward smaller suites. Another option is to skip failed tests and evaluate on the remaining tests. This avoids false negatives from converted timeouts, but it can create false positives: if the skipped test is the only one that catches a bug, a wrong solution receives positive reward. For correctness tests, where one missed bug can flip the reward sign, this trade-off is unfavorable over millions of trajectories.

\paragraph{Quarantining degraded windows.}
A stronger operational control would be to pause rollout collection, drop affected batches, or roll back to an earlier checkpoint when the service enters a bad state. This is plausible for both visible bursts and silent slowdowns if the degradation signal is reliable enough. It is also expensive: the policy may have already consumed noisy rewards before the degradation is diagnosed, discarding batches wastes GPU time, and rolling back a distributed RL run adds another source of operational complexity. We therefore do not implement such routine for the RL experiments of this paper.

\paragraph{Live calibration probes.}
Another natural idea is to send a fixed set of code snippets with known reference durations during training, use their measured slowdown to detect silent CES degradation, and rescale current rollout durations. This can work only if the probe set estimates the current service state accurately. Too few probes are noisy, especially in the sub-second regime where tens of milliseconds matter. Too many probes compete with the training workload for CES capacity; the monitoring traffic can slow the service and training progress enough to become part of the problem. This makes live probing more suitable as a health diagnostic than as a cheap per-step correction mechanism.

\paragraph{Reward choices.}
Reward design can reduce, but not remove, sensitivity to service drift. Same-prompt grouping and GRPO advantages compare rollouts collected close together, which helps when the service shift is common across a batch. A purely within-batch timing comparison would push this idea further, but we did not use it as the main reward because our ranked rewards also compare fresh model executions against stored human-reference durations. Those stored references still need the affine calibration studied next, and the reward ablations in \cref{app:training_support} show that data quality, fallback behavior, calibration, and reward design have to be handled together.

\subsection{CES noise, drift, and calibration}
\label{sec:app_duration_correction}

Switching to CES does not remove noise. It concentrates noise in a backend whose behavior can be monitored, retried, and recalibrated. CES timings affect four stages of the pipeline: dataset construction, where reference durations are stored; RL training, where each rollout is evaluated; evaluation, where timing-sensitive outcomes are measured again; and cross-run comparisons, where current results are compared either with stored human-reference durations used by ranked reward variants or with evaluation results from experiments run under an earlier CES state. The first three stages depend on live measurement noise. The fourth introduces service-state drift: stored durations or stored evaluation results may have been collected on a different CES fleet, cluster, load condition, or sandbox version than the current execution.

We therefore treat CES timing in two layers. Live CES noise asks whether repeated executions under the current service state are stable enough for ranking. Long-term CES drift asks whether historical measurements, either human-reference durations or evaluation outputs from earlier runs, remain comparable to current model executions. For stored reference durations, we apply the affine correction below; for cross-run evaluation comparisons, the same issue motivates re-execution or calibration-sensitivity checks.

\paragraph{Same-time grouping attenuates live drift.}
During RL, rollouts for the same prompt are executed and grouped contemporaneously, and the trainer does not use a replay buffer for these configurations. This helps because common-mode backend shifts within a prompt group are partially shared and are not mixed with trajectories from much older service regimes. It does not remove reward noise: ranked rewards still compare model executions against calibrated historical human-reference durations, so calibration and health monitoring remain necessary.

\paragraph{Calibration campaign setup.}
The first calibration campaign fits the correction on human-reference duration pairs. Each run also records one model solution for ranking diagnostics, but the fitted map itself is estimated from human references because the goal is to map stored reference durations into the current CES timing scale before a fresh model solution is inserted into the same leaderboard. We sample 33 problems from the filtered DMC-Optim pool. Across these problems, the human-reference pool contains 10{,}841 solutions, with 14--3{,}185 solutions per problem and median 158. The optimization-test suites contain 4{,}035 tests in total, with 30--150 tests per problem and median 135. Each problem is re-executed independently about 20 times through CES; 27 problems have all 20 runs and 6 problems have 19 runs because this calibration setup does not allow local fallback, so runs that remain too unstable without fallback are removed from the fit. This gives 654 problem-runs.

The raw scale is larger than the problem count suggests. A single problem-run executes all human references on the optimization tests, so the campaign yields 26{,}675{,}153 raw pairs \((d_{\text{stored}}, d_{\text{fresh}})\). Before fitting, we pool repeated measurements by the unique triple \((\text{problem}, \text{test}, \text{solution})\): the stored duration is fixed by the dataset, while the fresh CES duration is averaged across the 19--20 runs. This gives 1{,}369{,}381 pooled observations. The fit uses only optimization tests, because these are the tests whose duration enters the ranking reward; correctness tests can be useful for pass/fail recovery, but they are not the target distribution for timing calibration.

\paragraph{Short-run CES variability.}
Repeated CES executions show that short-run noise is present, but the noise scale is far below the local-sandbox instability in \cref{sec:app_local_vs_ces}. At the raw problem-test-solution level, the coefficient of variation across repeated CES measurements has mean 9.1\%, median 8.2\%, and a p5--p95 range of 3.5\%--18.0\%. In absolute terms, the standard deviation of a repeated duration measurement has mean 12.4\,ms and median 9.9\,ms, with p5--p95 range 6.2--26.7\,ms. These numbers are not zero: they are large enough to matter for the fastest tests, where a few tens of milliseconds can move a solution across a tight percentile boundary. They are also much smaller than the cross-solution timing signal the reward tries to exploit.

The same conclusion holds after converting durations into the leaderboard statistic used by the ranked rewards. For each optimization test, the candidate duration is ranked against human-reference durations for that same test, with 0\% denoting fastest and 100\% slowest. The mean percentile is the average of these per-test percentiles across optimization tests. Across the 20 repeated CES runs, the standard deviation of mean percentile averages 2.1 percentage points, with median 2.0 and range 1.1--4.5 percentage points. A two-point percentile noise scale is still visible in strict metrics, but it is far from the 10.961-point local standard deviation and 41.173-point local range reported above.

\begin{table*}[t!]
\centering
\small
\caption{\textbf{Short-run CES variability in the first calibration campaign.}
The top rows summarize repeated duration measurements for each pooled problem-test-solution triple.
The bottom row summarizes the induced variability after those durations are converted into the mean-percentile leaderboard statistic used by ranked rewards.
}
\label{tab:ces_short_run_noise}
\begin{tabular}{lccc}
\toprule
{\bfseries Statistic} & {\bfseries Mean} & {\bfseries Median} & {\bfseries Range / p5--p95} \\
\midrule
Duration CV across repeated CES runs & 9.1\% & 8.2\% & 3.5--18.0\% \\
Duration std. across repeated CES runs & 12.4\,ms & 9.9\,ms & 6.2--26.7\,ms \\
Mean duration of pooled observations & 0.160\,s & 0.108\,s & 0.080--6.38\,s \\
\midrule
Mean-percentile std. across runs & 2.1 pp & 2.0 pp & 1.1--4.5 pp \\
\bottomrule
\end{tabular}
\end{table*}

\paragraph{Variance decomposition.}
The raw-duration decomposition is favorable. The between-triple variance is 0.02906, while the within-triple variance from repeated CES measurement is 0.00027, against total variance 0.02994. Equivalently, about 97.1\% of the variation is explained by real differences between problem-test-solution triples, while the repeated-measurement component is about 0.9\% of the total scale. This is the reason averaging fresh measurements across runs gives a stable fit: the calibration is fitting the systematic relationship between stored and fresh durations, not mostly CES jitter. For duration-ranked rewards, the remaining 2.1-point mean-percentile noise corresponds to a reward perturbation on the order of \(0.01\)--\(0.02\) per episode, small relative to the correctness swing from \(-1\) to a positive reward and partly averaged out across the roughly 120 optimization tests per problem in the final training split.

\paragraph{Long-term CES drift.}
The larger issue is comparing current CES measurements against previously stored durations, either from human references or from model generations evaluated under an earlier CES state. An early 20-problem pilot showed fresh-to-stored duration ratios ranging from 0.96 to 1.34, with mean 1.15, too heterogeneous for one global multiplier. The full 33-problem campaign makes the shape of the drift clearer. Taking a fixed set of previously stored human-reference durations, fresh CES measurements from the new calibration campaign are often below the stored values on longer tests, so a slope below one is needed. For the fastest tests, however, a fixed overhead appears: a stored duration near zero does not map to a fresh duration near zero. This is exactly the regime used by strict timing percentiles, so ignoring the intercept can distort the ranking even if a multiplicative fit looks plausible on medium and long tests. In the larger calibration study, uncorrected stored rankings have only 0.54 Spearman correlation with fresh CES rankings, and 29 of 33 problems are biased in the same direction by an average of 19.6 percentage points. Even if live CES calls were noiseless, frozen stored durations would distort reference rankings unless recalibrated.

\paragraph{Affine correction.}
We correct long-term drift by fitting an affine map from stored to current CES durations:
\begin{equation}
\label{eq:affine_correction}
d_{\text{corrected}} = \alpha \cdot d_{\text{stored}} + \beta,
\qquad
\alpha = 0.6306,
\quad
\beta = 0.0529\,\text{s},
\quad
R^2_{\text{CV}} = 0.993 .
\end{equation}
The fit is performed with leave-one-problem-out cross-validation: each fold holds out one complete problem, fits on the other 32, and predicts fresh durations on the held-out problem. This is stricter than randomly holding out individual duration pairs, because all tests and solutions from the held-out problem are unseen during fitting. \Cref{tab:duration_correction_models} reports the four families considered. The intercept is needed. A purely multiplicative correction reaches only \(R^2_{\text{CV}}=0.900\), while the affine model captures a fixed overhead of about 53\,ms on top of proportional scaling. The polynomial has a slightly higher \(R^2_{\text{CV}}\), 0.9936 versus 0.9933, and a slightly lower RMSE, 13.7\,ms versus 13.9\,ms, but this is not a meaningful gain for the reward path and adds an extra parameter. Applying the affine correction raises the Spearman correlation between stored and fresh CES rankings from 0.54 to 0.96. At data-load time, invalid stored durations are replaced by a conservative 10\,s fallback before correction, and corrected durations are clamped to the valid \([0,10]\) second execution range before use.

\begin{table*}[t!]
\centering
\small
\caption{\textbf{Stored-to-fresh duration correction families fit on the first CES calibration campaign.}
All models are fit on 1{,}369{,}381 pooled problem-test-solution observations.
Cross-validation leaves out an entire problem at a time.
The affine model is the chosen calibration because it captures the fixed overhead with essentially the same held-out error as the polynomial, while the multiplicative and power-law families leave structured residuals.
}
\label{tab:duration_correction_models}
\begin{tabular}{llccc}
\toprule
{\bfseries Model} & {\bfseries Form} & {\bfseries Train \(R^2\)} & {\bfseries CV \(R^2\)} & {\bfseries CV RMSE} \\
\midrule
Multiplicative & \(d_{\text{fresh}} = 0.7190\,d_{\text{stored}}\) & 0.9254 & 0.8998 & 54.0\,ms \\
Power law & \(d_{\text{fresh}} = 0.5551\,d_{\text{stored}}^{0.6661}\) & 0.7896 & 0.7798 & 80.0\,ms \\
Affine (selected) & \(d_{\text{fresh}} = 0.6306\,d_{\text{stored}} + 0.0529\) & 0.9945 & 0.9933 & 13.9\,ms \\
Polynomial & \(d_{\text{fresh}} = 0.6491\,d_{\text{stored}} - 0.0040\,d_{\text{stored}}^2 + 0.0502\) & 0.9948 & 0.9936 & 13.7\,ms \\
\bottomrule
\end{tabular}
\end{table*}

\begin{figure*}[t!]
\centering
\includegraphics[width=0.88\textwidth]{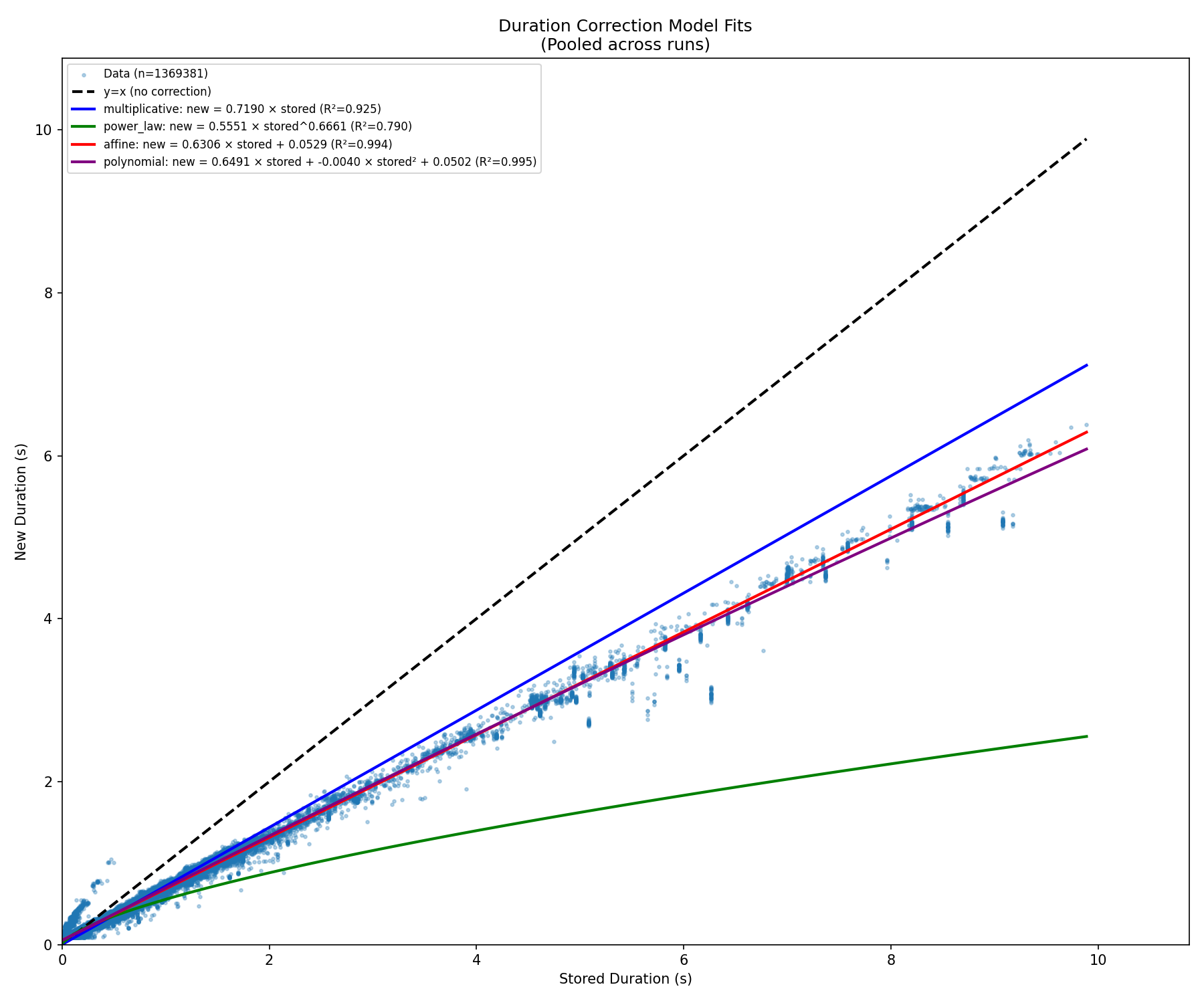}
\caption{\textbf{Stored-versus-fresh CES durations and the correction families considered for calibration.}
Each point is one pooled problem-test-solution observation from the first 33-problem calibration campaign, after averaging fresh CES measurements across repeated runs.
The dashed line is the identity relation.
Fresh CES durations are not obtained by applying a single global speed multiplier to stored durations: long tests need a slope below one, while fast tests show a visible additive floor.
The affine and polynomial fits nearly overlap across the observed range, which is why the calibration uses the affine map in \cref{eq:affine_correction}.
}
\label{fig:duration_correction_fits}
\end{figure*}

\begin{figure*}[t!]
\centering
\includegraphics[width=0.9\textwidth]{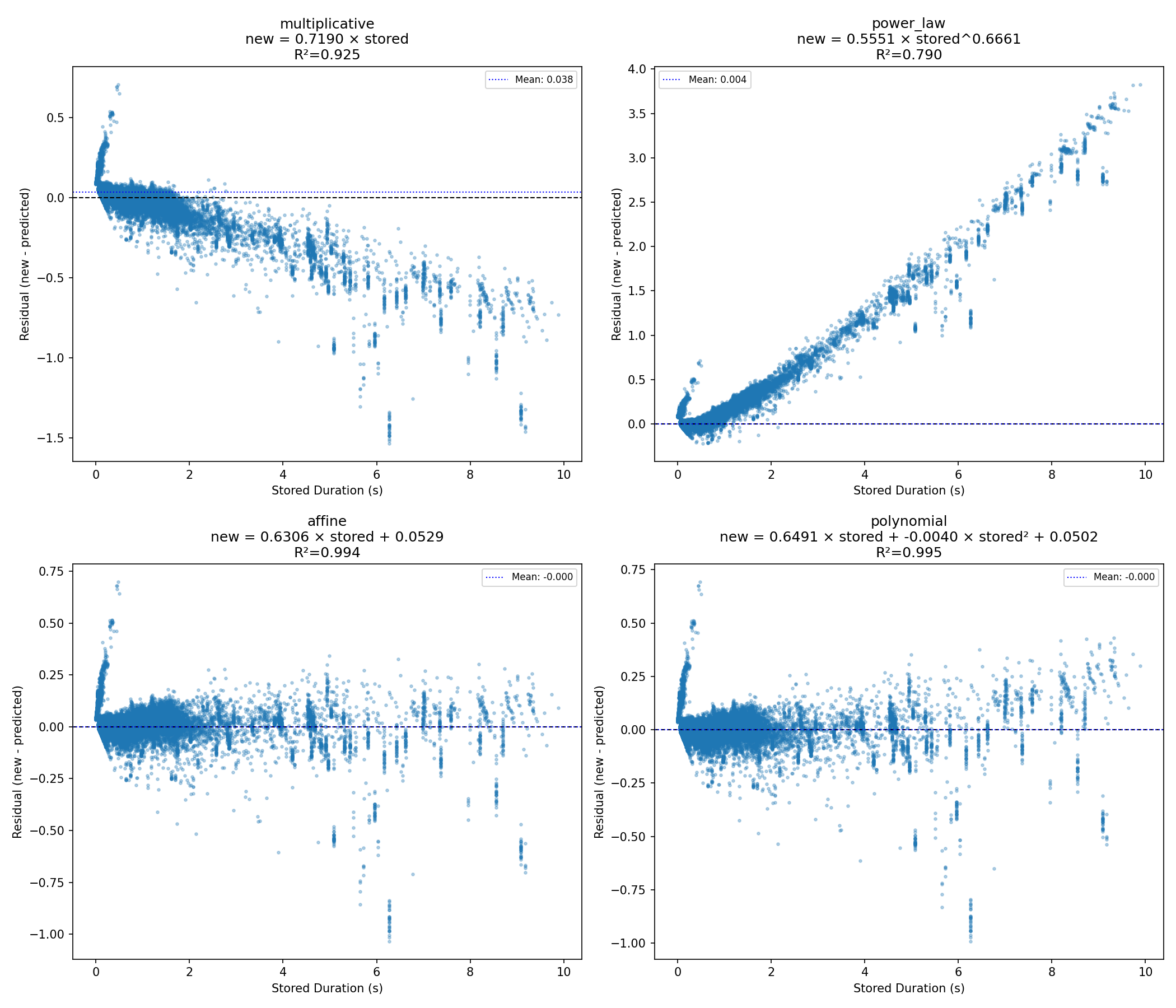}
\caption{\textbf{Residual structure for the candidate duration-correction families.}
The residual is \(d_{\text{fresh}}-\hat{d}_{\text{fresh}}\), plotted against stored duration.
The multiplicative model leaves a duration-dependent bias because it cannot represent the fixed overhead on fast tests and the proportional speed change on longer tests at the same time.
The power-law fit bends in the wrong way in the original duration scale.
Affine and polynomial correction leave residuals centered near zero across most of the observed range; the polynomial reduces error only marginally relative to affine, so the simpler affine correction is used for training and evaluation.
}
\label{fig:duration_correction_residuals}
\end{figure*}

\begin{figure*}[t!]
\centering
\includegraphics[width=0.9\textwidth]{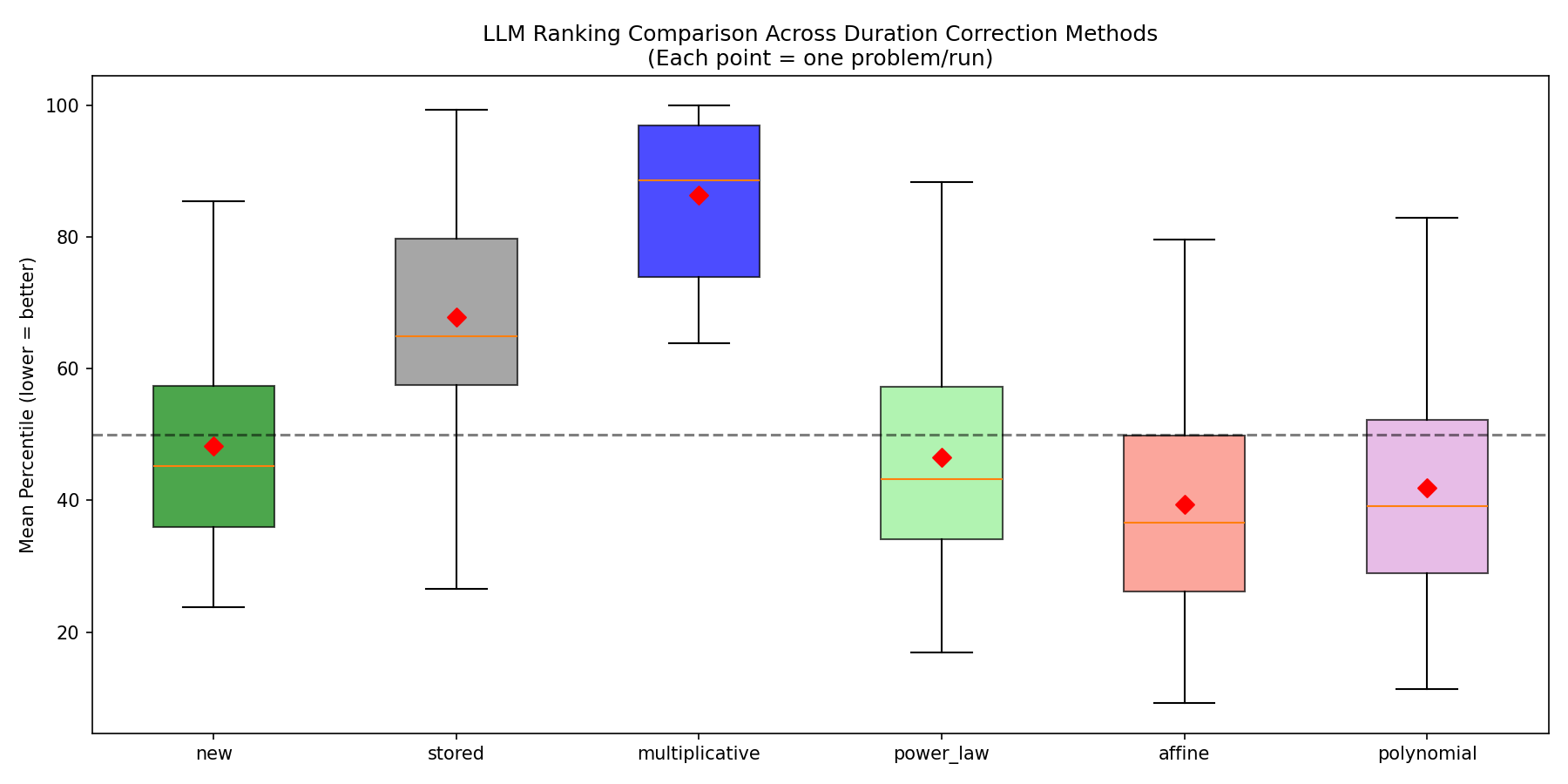}
\caption{\textbf{Impact of duration correction on solution rankings.}
Each box summarizes 654 problem-runs: 33 selected problems, each re-executed about 20 times.
For each selected problem, the diagnostic uses one model-generated candidate solution from a saved rollout and compares it to the human-reference solutions for that same problem.
On each optimization test, the candidate's fresh CES duration is ranked against the human-reference durations for that test; the plotted value is the mean of these per-test percentiles across the optimization suite.
Lower is better because the candidate is faster than more human references.
The boxes differ only in how the human-reference durations are represented: the green box uses freshly measured CES human-reference durations and is treated as the target distribution, while the other boxes use uncorrected stored durations or stored durations transformed by one correction family.
Uncorrected stored durations shift the mean percentile from 48.3 to 67.9, a 19.6-point bias.
The multiplicative correction is worse, with mean 86.4, because it removes proportional scale without correcting the fast-test offset.
The affine correction moves the mean to 39.4 and the polynomial to 42.0; both are much closer to the fresh-CES reference than uncorrected stored durations, and the affine family is kept because its residual and cross-validation behavior are nearly identical to the polynomial with one fewer parameter.
}
\label{fig:ranking_comparison}
\end{figure*}

\FloatBarrier

\subsection{Temporal stability of calibration}
\label{sec:app_temporal_stability}

The preceding subsection used a first calibration campaign to map fixed stored dataset durations into a newer CES timing scale. That is the high-shift case, especially relevant after execution-service infrastructure changes such as moving to a new CPU fleet. There is also a more routine question: when the service infrastructure is nominally fixed, how much does the calibration move across day-to-day reruns? During the course of our experiments, we repeated the calibration several times on the same 33 problems with the same repeated-execution design. In this subsection, we report one such rerun on top of the results already reported above. Each campaign produces approximately 27M raw duration pairs, pooled to about 1.37M unique observations. Both campaigns fit the same correction families using leave-one-problem-out cross-validation.

\Cref{tab:calibration_stability} compares the affine parameters. The slope changes by only 1.0\% (\(0.6306 \to 0.6243\)), while the intercept shifts from 53\,ms to 38\,ms. For a 1\,s stored duration, this changes the predicted execution time by roughly 3\%; for a 5\,s duration, by roughly 1.5\%. Cross-validated \(R^2\) remains above 0.98 in both campaigns.

\begin{table}[h!]
\centering
\small
\caption{\textbf{Affine calibration parameters and fit quality across two independent campaigns.}
The slope \(\alpha\) is stable to within 1\%.
The intercept \(\beta\) shifts by 15\,ms, which has small impact on predicted durations for typical stored values.
Both campaigns achieve cross-validated \(R^2 > 0.98\), confirming that the affine model captures the stored-to-measured relationship reliably.
}
\label{tab:calibration_stability}
\begin{tabular}{lcccc}
\toprule
{\bfseries Parameter} & {\bfseries Campaign 1} & {\bfseries Campaign 2} & {\bfseries \(\Delta\)} & {\bfseries \(\Delta\) (\%)} \\
\midrule
\(\alpha\) (slope) & 0.6306 & 0.6243 & -0.006 & -1.0\% \\
\(\beta\) (intercept, s) & 0.053 & 0.038 & -0.015 & -28\% \\
\midrule
Train \(R^2\) & 0.9945 & 0.9858 & -0.009 & --- \\
CV \(R^2\) & 0.9933 & 0.9820 & -0.011 & --- \\
CV RMSE (s) & 0.014 & 0.023 & +0.009 & --- \\
\bottomrule
\end{tabular}
\end{table}

The model-performance profile across problems is also stable between campaigns. For each of the 33 problems, we compute the model candidate's fresh-CES mean percentile against the human references for that problem: lower means that the candidate is faster relative to more human solutions on that problem. We then ask whether the same problems remain easy or hard for the model across the two calibration campaigns. The Spearman correlation is \(\rho=0.99\), meaning that problems where the model candidate looks relatively strong in the first campaign are mostly the same problems where it looks relatively strong in the rerun campaign, and likewise for weak problems. This supports using the same affine family for day-to-day recalibration, while still allowing the numerical \((\alpha,\beta)\) values to move with the service state.

\subsection{Evaluation-time sensitivity to affine calibration}
\label{sec:app_calibration_sensitivity}

\begin{figure*}[t!]
\centering
\begin{minipage}[t]{0.49\textwidth}
\centering
\includegraphics[width=0.88\linewidth]{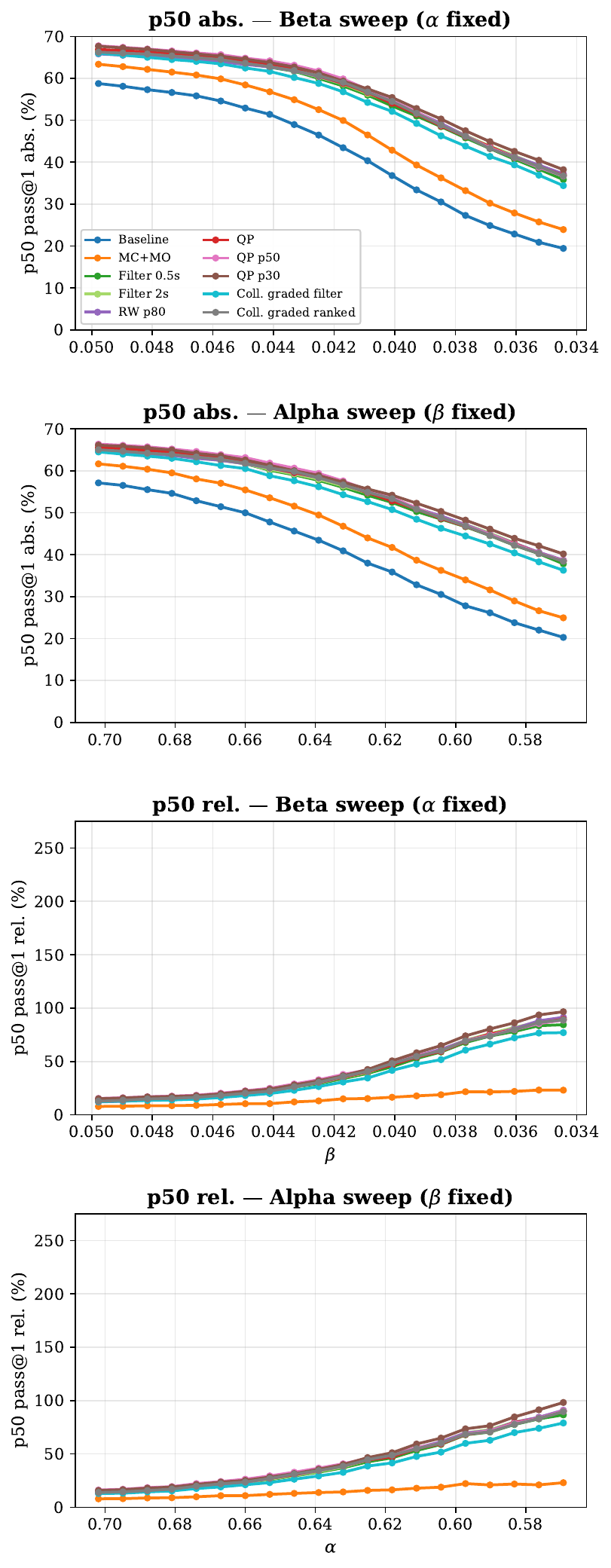}
\end{minipage}\hfill
\begin{minipage}[t]{0.49\textwidth}
\centering
\includegraphics[width=0.88\linewidth]{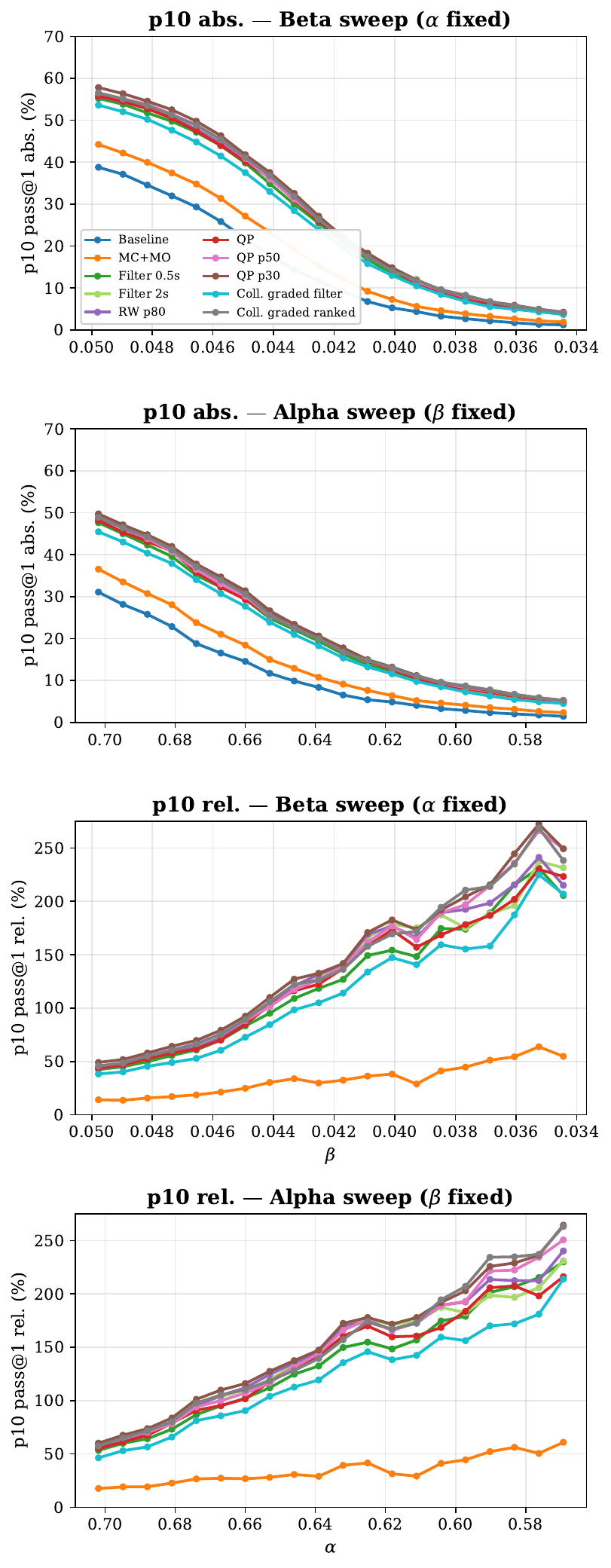}
\end{minipage}
\caption{\textbf{Sensitivity of \(p_{50}\) and strict \(p_{10}\) pass@1 to affine calibration parameters.}
\emph{Left:} Absolute CWM 32B DMC-Optim pass@1 and relative improvement at \(p_{50}\) as \(\beta\) is swept with \(\alpha\) fixed or \(\alpha\) is swept with \(\beta\) fixed.
\emph{Right:} The same evaluation-time sensitivity check at \(p_{10}\).
The additive offset matters because it is a large fraction of the corrected duration for short tests. The relative ordering of optimization-trained configurations remains stable across the displayed range.
As \(\beta\) becomes more stringent, the relative gains of Filter 2s, QP (quality percentile), and QP train \(p_{30}\) grow from 45\%, 44\%, and 49\% at the lenient endpoint to 141\%, 137\%, and 141\% at the intermediate \(\beta\) value, and up to 233\%, 225\%, and 250\% at the most stringent endpoint, though there the baseline is so low that any nonzero optimization-trained score looks much larger in relative terms.
}
\label{fig:app_alphabeta_sweep}
\end{figure*}

Timing calibration keeps freshly recorded model-generated solution durations comparable to the fixed stored human-solution durations used by the ranking metric.
We now ask whether the calibration itself affects the evaluation results: do different plausible calibrations lead to different optimization scores, or do they mainly move the absolute score scale while preserving comparisons between models?
We therefore sweep a current affine fit, holding model executions fixed and changing only the corrected human-reference durations used by the \(p_{50}\) and \(p_{10}\) ranking metrics.
This is an evaluation-time sensitivity check: decreasing \(\alpha\) or \(\beta\) makes corrected references shorter, so the same model executions are judged against a stricter speed threshold.
\Cref{fig:app_alphabeta_sweep} shows the result for CWM 32B.
In relative terms, the sweep moves \(\beta\) by about 31\% and \(\alpha\) by about 19\% from their starting values; the \(\beta\) movement also matches more or less the drift pattern in \cref{sec:app_temporal_stability}, where \(\beta\) moved by 28\% across calibration campaigns while \(\alpha\) moved by 1\%, and for \(\alpha\) we choose to do a wider cautious sweep.
Perturbing the additive offset \(\beta\) remains at least as important as perturbing the multiplicative slope \(\alpha\): reducing \(\beta\) from 0.0498 to about 0.0344\,s drops the baseline \(p_{10}\) pass@1 from 38.8\% to 1.2\%, while reducing \(\alpha\) from 0.7019 to 0.5692 with fixed \(\beta\) drops the baseline from 31.1\% to 1.4\%.
The scale explains why this effect appears on strict percentiles.
The full \(\beta\) sweep changes every corrected reference by about 15.3\,ms, whereas the \(\alpha\) sweep changes a stored duration \(d\) by \(0.1326d\): 13.3\,ms at \(d=100\)ms, 132.6\,ms at \(d=1\)s, and 663.2\,ms at \(d=5\)s.
Thus the intercept is the larger perturbation in the fast-reference regime, while the slope is larger on the multi-second tail.
\Cref{tab:dataset_stats} shows that optimization tests create a much heavier tail than the original DMC tests, but most duration records are still sub-second: the pooled median is 0.099\,s, \(p90\) is 0.736\,s, and only 6.87\% exceed 1\,s.
Moreover, a \(p_{10}\) decision compares the model execution against the fastest human-reference durations for each test, not against the median human duration.
The relevant cutoff is therefore often in the part of the distribution where a 15\,ms additive shift is comparable to, or larger than, the effect of the tested slope perturbation.

The bottom panels give the more important training conclusion.
Making the calibration stricter lowers all absolute \(p_{10}\) scores, but it lowers the correctness-only baseline faster than the optimization-trained models.
At the lenient beta endpoint, Filter 2s, QP (quality percentile), and QP train \(p_{30}\) score 56.3\%, 55.8\%, and 57.8\%, compared with 38.8\% for the baseline; their relative gains are 45\%, 44\%, and 49\%.
At a more stringent beta point, \(\beta\simeq0.042\), the baseline falls to 9.2\%, while the same three optimization-trained models remain at 22.2\%, 21.8\%, and 22.2\%; the relative gains grow to 141\%, 137\%, and 141\%.
At the most stringent beta point, the baseline falls to 1.2\%, while the same three optimization-trained models remain at 4.0\%, 3.9\%, and 4.2\%; the relative gains grow to 233\%, 225\%, and 250\%.
This endpoint is less interesting because the baseline is already so low that any nonzero optimization-trained score looks much larger in relative terms.
The alpha sweep shows the same pattern: at \(\alpha=0.6321\), the baseline is 6.5\%, while the three optimization-trained models reach 17.5\%, 17.0\%, and 17.8\%, for gains of 169\%, 162\%, and 174\%.
MC+MO also stays above the baseline, but with smaller gains: 14\% at the lenient beta endpoint and 58\% at the stringent beta point.
This means that stricter calibration does not erase the optimization-RL effect; on the contrary, it makes the gap more visible, because the baseline loses a larger share of its top-decile solutions.
The optimization-RL setups used here therefore appear particularly beneficial under more demanding evaluation settings.

\FloatBarrier

\subsection{Joint training-time and evaluation-time calibration sweeps}
\label{sec:app_training_eval_calibration_sensitivity}

\Cref{sec:app_calibration_sensitivity} held the trained models fixed and changed only the affine correction used for post-training scoring. The next question is whether calibration also matters while learning. This is reward-dependent: rewards based only on pass/fail outcomes or raw timeouts do not need a calibrated human-reference distribution, but the post-execution ranked rewards do, because they compare each rollout to a fixed distribution of stored human-reference durations. If such rewards were very sensitive to the exact \((\alpha,\beta)\) used while training, then a small calibration mismatch could change the learned policy even if final scoring rankings were stable. We therefore run a two-dimensional sensitivity study on Qwen 2.5 7B: one axis changes the affine correction used during RL training, and the other axis changes the affine correction used during post-training resimulation.

\paragraph{Training-time calibration sweep.}
We train 16 Qwen 2.5 7B models across two post-execution ranking environments, both defined in \cref{sec:app_env_qwen7b_definitions}. The first is ranked \(p_{30}\): after execution, the candidate is inserted into the calibrated human-reference leaderboard and receives the binary optimization gate only if it falls inside the top 30\% of the reference-quality distribution. The second is QP (quality percentile): the same leaderboard construction is used, but each correct candidate receives a graded reward according to its quality percentile over the full reference pool. We train eight models with each environment. Thus the two families use the same underlying calibrated duration distribution but expose it to RL differently: \(p_{30}\) turns the ranking into a sharper pass/fail optimization gate, while QP gives a smoother graded signal to all correct candidates.

Within each family, the only intended training change is the affine correction applied to stored reference durations before duration filters and reference rankings are computed. We index the eight reward-calibration points as ab1--ab8. They linearly sweep from \(\alpha=0.594939,\beta=0.054482\) to \(\alpha=0.719699,\beta=0.048993\), a range chosen from the calibration variation observed across repeated CES recalibration campaigns. The ab3 point, \(\alpha=0.630585,\beta=0.052914\), matches the original affine fit from \cref{sec:app_duration_correction} up to rounding. The ablation runs are launched as one concurrent sweep under the same CES setup, so the live execution backend is held fixed while the stored-reference correction used by the reward changes. Two replication references trained with the ab3 coefficients under older CES states are included to check how much changes when the reward calibration is fixed but the service state differs.

\paragraph{Evaluation-time calibration sweep.}
All swept training models and the two replication references are re-executed concurrently on CES to obtain fresh candidate timings under the same service state. We then resimulate each model under 10 scoring calibrations, denoted s0--s9, linearly interpolating from \((\alpha=0.701876,\beta=0.049777)\) to \((\alpha=0.606033,\beta=0.038306)\). These endpoints are two plausible affine corrections from repeated recalibration runs and are used here to test whether the choice of correction matters once the model outputs are fixed. Lowering \(\alpha\) and \(\beta\) makes the corrected human references shorter, so the same model execution is judged against a stricter speed target. We evaluate \(p_{100}\), \(p_{20}\), and \(p_{10}\) pass@1, plus \(p_{10}\) pass@10. \Cref{tab:training_eval_ab_grid} records the two calibration grids.

\begin{table}[h!]
\centering
\scriptsize
\caption{\textbf{Reward and post-training scoring calibration grids for the two-dimensional alpha/beta sensitivity study.}
Reward-calibration points ab1--ab8 change the affine correction used inside RL. Scoring points s0--s9 change the affine correction used after re-executing the trained models together on CES. The range is chosen to cover the variation observed across different calibration campaigns during this work.}
\label{tab:training_eval_ab_grid}
\begin{tabular}{lrrlrr}
\toprule
{\bfseries Reward} & \(\boldsymbol{\alpha}\) & \(\boldsymbol{\beta}\) &
{\bfseries Score} & \(\boldsymbol{\alpha}\) & \(\boldsymbol{\beta}\) \\
\midrule
ab1 & 0.594939 & 0.054482 & s0 & 0.701876 & 0.049777 \\
ab2 & 0.612762 & 0.053698 & s1 & 0.691227 & 0.048502 \\
ab3 & 0.630585 & 0.052914 & s2 & 0.680578 & 0.047228 \\
ab4 & 0.648407 & 0.052130 & s3 & 0.669928 & 0.045953 \\
ab5 & 0.666230 & 0.051345 & s4 & 0.659279 & 0.044679 \\
ab6 & 0.684053 & 0.050561 & s5 & 0.648630 & 0.043404 \\
ab7 & 0.701876 & 0.049777 & s6 & 0.637981 & 0.042130 \\
ab8 & 0.719699 & 0.048993 & s7 & 0.627331 & 0.040855 \\
--- & --- & --- & s8 & 0.616682 & 0.039581 \\
--- & --- & --- & s9 & 0.606033 & 0.038306 \\
\bottomrule
\end{tabular}
\end{table}

\begin{figure*}[p]
\centering
\includegraphics[width=0.90\textwidth]{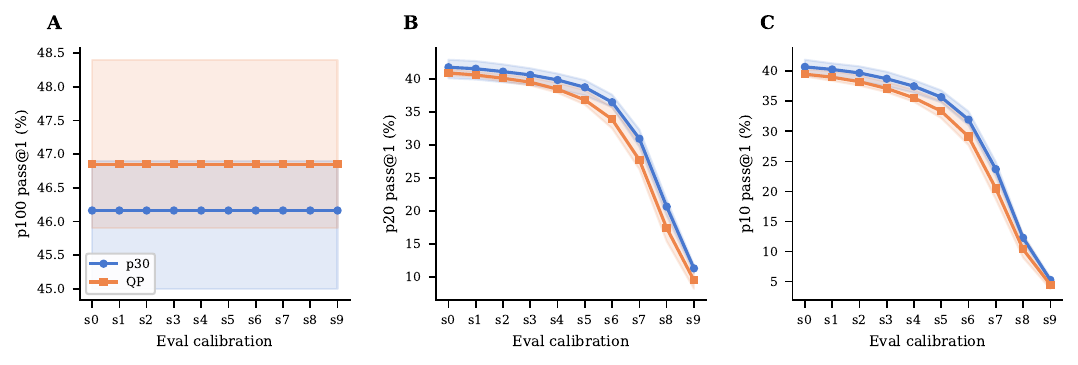}
\caption{\textbf{Post-training scoring calibration drives the mean score, while reward calibration mostly appears as a narrow band.}
Each panel plots score against scoring calibration s0--s9.
Colors distinguish the ranked \(p_{30}\) and QP reward families; shaded bands are the min--max range over the eight reward calibrations ab1--ab8.
The \(p_{100}\) panel is flat because correctness does not depend on reference-duration calibration.
At \(p_{20}\) and especially \(p_{10}\), lowering the scoring calibration from s0 to s9 makes corrected human references shorter and sharply reduces absolute scores, while the within-band spread from changing the reward calibration remains small.
}
\label{fig:training_eval_ab_layoutA}
\end{figure*}

\begin{figure*}[p]
\centering
\includegraphics[width=0.90\textwidth]{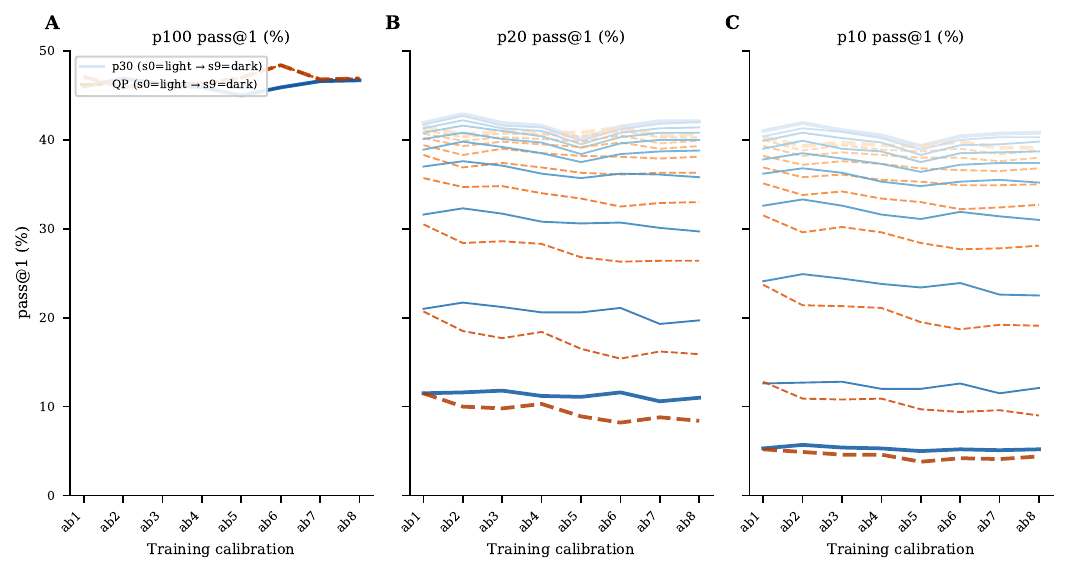}
\caption{\textbf{Reward calibration has weak effect at fixed scoring calibration.}
The x-axis is the reward calibration ab1--ab8.
Each curve is one scoring calibration s0--s9; blue solid curves are ranked \(p_{30}\) and orange dashed curves are QP.
Within a fixed scoring curve, the lines are nearly flat compared with the vertical separation between scoring calibrations.
This is the main evidence that changing the reward-side \((\alpha,\beta)\) during training does not strongly change the learned model quality over the tested range.
}
\label{fig:training_eval_ab_layoutE1}
\end{figure*}

\begin{figure*}[p]
\centering
\includegraphics[width=0.76\textwidth]{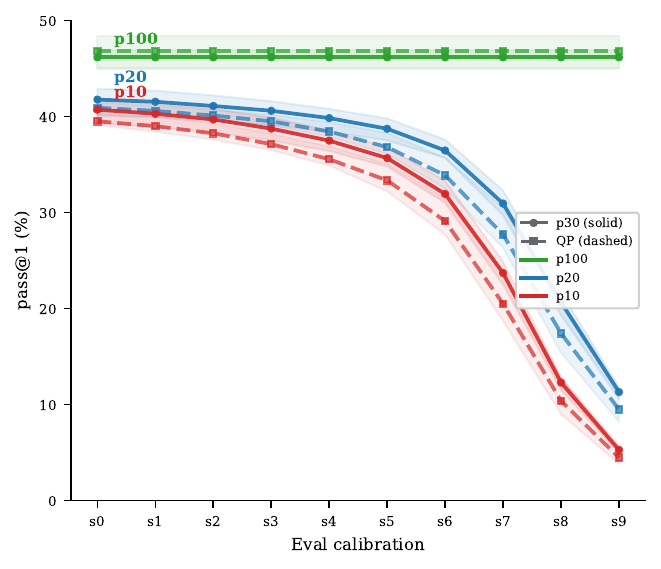}
\caption{\textbf{Stricter optimization percentiles amplify scoring-calibration sensitivity.}
All three pass@1 percentiles are plotted against scoring calibration in one view, with min--max bands over reward calibrations.
\(p_{100}\) remains invariant because it only asks whether the solution is correct.
\(p_{20}\) drops substantially as the scoring calibration moves from s0 to s9, and \(p_{10}\) drops further because top-decile membership depends on the shortest corrected reference durations.
The figure separates correctness effects from timing-threshold effects: reward calibration barely changes the bands, while scoring calibration changes the strict optimization metrics.
}
\label{fig:training_eval_ab_layoutF2}
\end{figure*}

\begin{figure*}[p]
\centering
\includegraphics[width=0.90\textwidth]{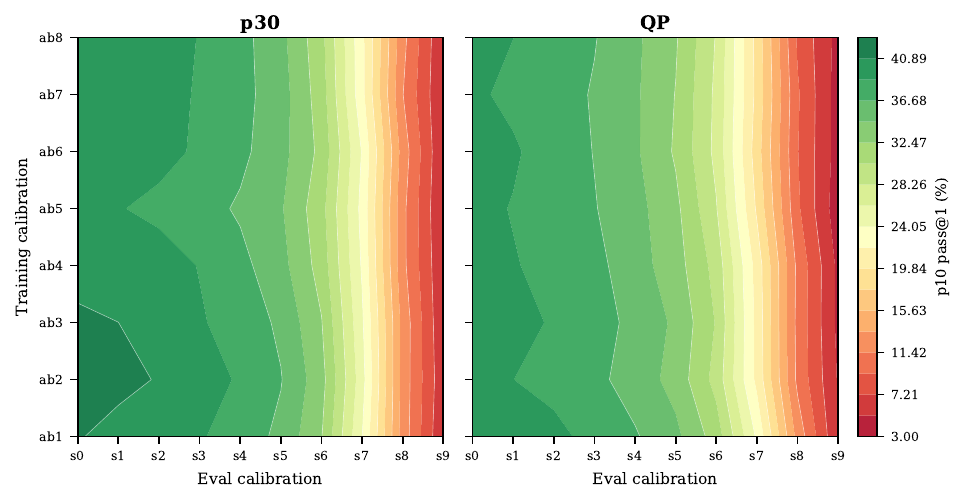}
\caption{\textbf{Strict \(p_{10}\) score surface over reward and scoring calibration.}
The color shows \(p_{10}\) pass@1 for each reward calibration ab1--ab8 and scoring calibration s0--s9.
The contours are nearly vertical in both the ranked \(p_{30}\) and QP panels, meaning that the score changes mainly when the scoring calibration changes, not when the reward calibration changes.
There is no visible diagonal ridge suggesting that training with a reward calibration close to the later scoring calibration is necessary.
}
\label{fig:training_eval_ab_layoutH1}
\end{figure*}

\begin{figure*}[p]
\centering
\includegraphics[width=0.90\textwidth]{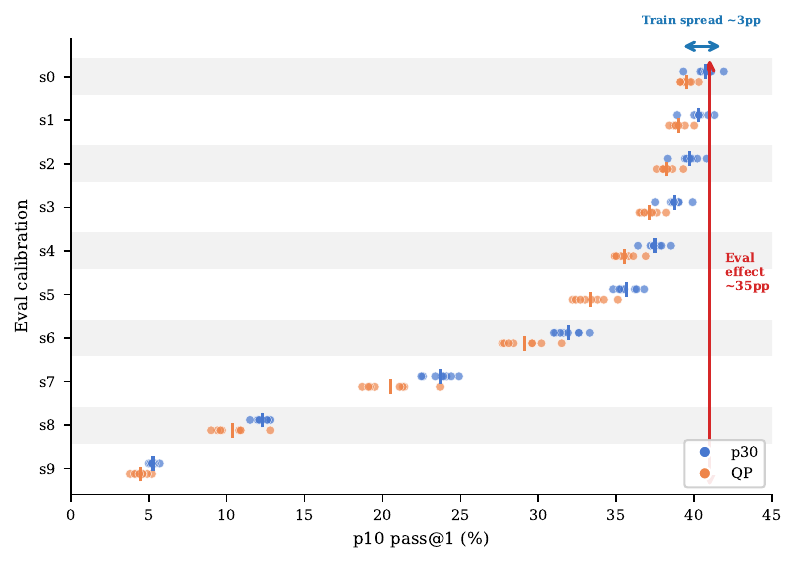}
\caption{\textbf{Scale separation between reward-calibration spread and scoring-calibration movement.}
Each row fixes a scoring calibration and shows the eight reward calibrations as dots.
The dots cluster tightly within a row, while the row centers move by roughly 35 percentage points from s0 to s9 at strict \(p_{10}\).
This visualizes the main quantitative conclusion: reward-side alpha/beta choices move the final score by only a few points, while post-training scoring calibration determines the absolute strict-\(p_{10}\) scale.
}
\label{fig:training_eval_ab_layoutH2}
\end{figure*}

\paragraph{Results.}
\Cref{fig:training_eval_ab_layoutA,fig:training_eval_ab_layoutE1,fig:training_eval_ab_layoutF2,fig:training_eval_ab_layoutH1,fig:training_eval_ab_layoutH2} show complementary projections of the same 16-by-10 scoring matrix. \Cref{fig:training_eval_ab_layoutA} averages over reward calibrations and shows that the post-training scoring calibration is the dominant axis. \Cref{fig:training_eval_ab_layoutE1} fixes the reward-calibration axis on the x-axis and shows that, at a fixed scoring point, changing reward calibration produces nearly flat curves. \Cref{fig:training_eval_ab_layoutF2} overlays percentiles and shows why strict metrics are more sensitive. \Cref{fig:training_eval_ab_layoutH1} focuses on strict \(p_{10}\), where most variation follows the scoring axis rather than the reward-calibration axis. \Cref{fig:training_eval_ab_layoutH2} shows the same scale separation as a dot strip: reward calibrations cluster tightly within each scoring calibration, while row centers move substantially across scoring calibrations. \Cref{tab:training_eval_ab_p10_raw} gives the full strict \(p_{10}\) pass@1 matrix, and \cref{tab:training_eval_ab_anchor_raw} reports \(p_{100}\), \(p_{20}\), and \(p_{10}\) pass@10 anchors at s0, s5, and s9.

\begin{table}[p]
\centering
\scriptsize
\caption{\textbf{Raw strict \(p_{10}\) pass@1 scores for the reward and scoring alpha-beta sweep.}
Rows are trained Qwen 2.5 7B models. Columns are scoring calibrations s0--s9, with s0 and s9 denoting the two endpoints in \cref{tab:training_eval_ab_grid}. All values are percentages at step 10k after shared CES re-execution.}
\label{tab:training_eval_ab_p10_raw}
\resizebox{\textwidth}{!}{%
\begin{tabular}{llrrrrrrrrrr}
\toprule
{\bfseries Family} & {\bfseries Reward cal.} & {\bfseries s0} & {\bfseries s1} & {\bfseries s2} & {\bfseries s3} & {\bfseries s4} & {\bfseries s5} & {\bfseries s6} & {\bfseries s7} & {\bfseries s8} & {\bfseries s9} \\
\midrule
\(p_{30}\) & ref & 41.1 & 40.6 & 40.0 & 39.0 & 37.7 & 36.3 & 32.5 & 24.8 & 13.3 & 5.8 \\
\(p_{30}\) & ab1 & 41.0 & 40.4 & 39.9 & 39.0 & 37.8 & 36.2 & 32.6 & 24.1 & 12.6 & 5.3 \\
\(p_{30}\) & ab2 & 41.9 & 41.3 & 40.8 & 39.9 & 38.5 & 36.8 & 33.3 & 24.9 & 12.7 & 5.7 \\
\(p_{30}\) & ab3 & 41.1 & 40.9 & 40.2 & 39.0 & 37.9 & 36.3 & 32.6 & 24.4 & 12.8 & 5.4 \\
\(p_{30}\) & ab4 & 40.5 & 40.2 & 39.7 & 38.7 & 37.3 & 35.3 & 31.6 & 23.8 & 12.0 & 5.3 \\
\(p_{30}\) & ab5 & 39.3 & 38.9 & 38.3 & 37.5 & 36.4 & 34.8 & 31.1 & 23.4 & 12.0 & 5.0 \\
\(p_{30}\) & ab6 & 40.4 & 40.0 & 39.4 & 38.5 & 37.2 & 35.3 & 31.9 & 23.9 & 12.6 & 5.2 \\
\(p_{30}\) & ab7 & 40.7 & 40.3 & 39.5 & 38.6 & 37.4 & 35.5 & 31.4 & 22.6 & 11.5 & 5.1 \\
\(p_{30}\) & ab8 & 40.8 & 40.3 & 39.8 & 38.7 & 37.4 & 35.2 & 31.0 & 22.5 & 12.1 & 5.2 \\
\midrule
QP & ref & 40.1 & 39.6 & 38.8 & 37.8 & 36.2 & 34.2 & 29.7 & 21.2 & 10.5 & 4.6 \\
QP & ab1 & 40.3 & 40.0 & 39.3 & 38.2 & 36.9 & 35.1 & 31.5 & 23.7 & 12.8 & 5.2 \\
QP & ab2 & 39.3 & 38.8 & 38.2 & 37.2 & 35.8 & 33.8 & 29.6 & 21.4 & 10.9 & 4.9 \\
QP & ab3 & 39.7 & 39.4 & 38.6 & 37.6 & 36.1 & 34.2 & 30.2 & 21.3 & 10.8 & 4.6 \\
QP & ab4 & 39.4 & 38.9 & 38.3 & 37.3 & 35.5 & 33.4 & 29.6 & 21.1 & 10.9 & 4.6 \\
QP & ab5 & 39.3 & 38.7 & 38.0 & 36.8 & 35.3 & 33.0 & 28.4 & 19.5 & 9.7 & 3.8 \\
QP & ab6 & 39.8 & 39.0 & 38.0 & 36.6 & 34.9 & 32.2 & 27.7 & 18.7 & 9.4 & 4.2 \\
QP & ab7 & 39.1 & 38.4 & 37.6 & 36.5 & 34.9 & 32.4 & 27.8 & 19.2 & 9.6 & 4.1 \\
QP & ab8 & 39.1 & 38.8 & 38.0 & 36.8 & 35.0 & 32.7 & 28.1 & 19.1 & 9.0 & 4.4 \\
\bottomrule
\end{tabular}}
\end{table}

\begin{table}[p]
\centering
\scriptsize
\caption{\textbf{Raw anchor scores for correctness, \(p_{20}\), and pass@10 under the reward and scoring alpha-beta sweep.}
The table reports the scoring endpoints s0 and s9 plus the midpoint s5. \(p_{100}\) is invariant to scoring calibration and is shown once per trained model. All values are percentages at step 10k after shared CES re-execution.}
\label{tab:training_eval_ab_anchor_raw}
\resizebox{\textwidth}{!}{%
\begin{tabular}{llrrrrrrrrrr}
\toprule
 & & & \multicolumn{3}{c}{{\bfseries \(p_{20}\) pass@1}} & \multicolumn{3}{c}{{\bfseries \(p_{10}\) pass@1}} & \multicolumn{3}{c}{{\bfseries \(p_{10}\) pass@10}} \\
\cmidrule(lr){4-6}\cmidrule(lr){7-9}\cmidrule(lr){10-12}
{\bfseries Family} & {\bfseries Reward cal.} & {\bfseries \(p_{100}\)} & {\bfseries s0} & {\bfseries s5} & {\bfseries s9} & {\bfseries s0} & {\bfseries s5} & {\bfseries s9} & {\bfseries s0} & {\bfseries s5} & {\bfseries s9} \\
\midrule
\(p_{30}\) & ref & 46.2 & 42.1 & 39.1 & 12.0 & 41.1 & 36.3 & 5.8 & 61.6 & 57.1 & 17.8 \\
\(p_{30}\) & ab1 & 46.0 & 41.9 & 38.9 & 11.5 & 41.0 & 36.2 & 5.3 & 61.4 & 57.1 & 17.0 \\
\(p_{30}\) & ab2 & 46.9 & 42.9 & 39.8 & 11.6 & 41.9 & 36.8 & 5.7 & 60.8 & 56.7 & 16.8 \\
\(p_{30}\) & ab3 & 46.3 & 41.9 & 39.2 & 11.8 & 41.1 & 36.3 & 5.4 & 60.8 & 56.9 & 16.8 \\
\(p_{30}\) & ab4 & 45.9 & 41.6 & 38.5 & 11.2 & 40.5 & 35.3 & 5.3 & 60.2 & 56.0 & 15.7 \\
\(p_{30}\) & ab5 & 45.0 & 40.1 & 37.5 & 11.1 & 39.3 & 34.8 & 5.0 & 60.5 & 56.7 & 14.7 \\
\(p_{30}\) & ab6 & 45.9 & 41.5 & 38.4 & 11.6 & 40.4 & 35.3 & 5.2 & 62.3 & 58.1 & 16.6 \\
\(p_{30}\) & ab7 & 46.6 & 42.1 & 38.7 & 10.6 & 40.7 & 35.5 & 5.1 & 61.8 & 57.3 & 15.8 \\
\(p_{30}\) & ab8 & 46.7 & 42.1 & 38.8 & 11.0 & 40.8 & 35.2 & 5.2 & 62.0 & 56.5 & 15.6 \\
\midrule
QP & ref & 47.4 & 41.7 & 37.6 & 9.9 & 40.1 & 34.2 & 4.6 & 59.4 & 54.2 & 13.7 \\
QP & ab1 & 47.1 & 41.5 & 38.3 & 11.5 & 40.3 & 35.1 & 5.2 & 62.9 & 58.4 & 15.9 \\
QP & ab2 & 45.9 & 40.6 & 36.9 & 10.0 & 39.3 & 33.8 & 4.9 & 59.7 & 55.8 & 15.1 \\
QP & ab3 & 46.4 & 41.0 & 37.4 & 9.8 & 39.7 & 34.2 & 4.6 & 59.8 & 55.1 & 13.7 \\
QP & ab4 & 46.3 & 40.8 & 36.9 & 10.3 & 39.4 & 33.4 & 4.6 & 61.3 & 56.5 & 14.6 \\
QP & ab5 & 47.0 & 40.8 & 36.3 & 8.9 & 39.3 & 33.0 & 3.8 & 60.1 & 54.8 & 12.7 \\
QP & ab6 & 48.4 & 41.4 & 36.1 & 8.2 & 39.8 & 32.2 & 4.2 & 59.4 & 53.8 & 12.9 \\
QP & ab7 & 46.8 & 40.5 & 36.3 & 8.8 & 39.1 & 32.4 & 4.1 & 59.3 & 54.0 & 12.4 \\
QP & ab8 & 46.9 & 40.7 & 36.3 & 8.4 & 39.1 & 32.7 & 4.4 & 61.2 & 56.0 & 14.2 \\
\bottomrule
\end{tabular}}
\end{table}

\paragraph{Reward calibration has small effect.}
At any fixed scoring calibration, the spread across the eight reward calibrations is small compared with the absolute score movement across scoring calibrations. For \(p_{100}\), which does not depend on scoring alpha/beta, the spread is 1.9 percentage points for ranked \(p_{30}\) and 2.5 points for QP. This is the purest training-time readout: changing the calibration used inside the reward does not materially change correctness. The strict \(p_{10}\) readout is similar. At s0, ranked \(p_{30}\) spans 39.3--41.9\% across ab1--ab8, a 2.6 point range; QP spans 39.1--40.3\%, a 1.2 point range. At s5, the spreads are 2.0 and 2.9 points. These ranges are comparable to the 1--2.5 point run-to-run variation observed in the rerun-robustness analysis, so we should not interpret small differences between adjacent reward-calibration points as meaningful model-quality changes.

\Cref{fig:training_eval_ab_layoutH1} and \cref{fig:training_eval_ab_layoutH2} test whether reward and scoring calibrations need to be matched. If they did, \cref{fig:training_eval_ab_layoutH1} would show a diagonal ridge: models trained under lower reward-calibration points would do best under lower scoring calibrations, and models trained under higher reward-calibration points would do best under higher scoring calibrations. We do not see that pattern. The contours are nearly vertical, and \cref{fig:training_eval_ab_layoutH2} shows tight within-row clusters. The practical conclusion is that the tested calibration range changes the reward scale but does not select a different kind of solution strongly enough to dominate training noise. This is useful operationally because optimization-RL runs can span two days or more: within the range tested here, the training run does not need continuous reward-side recalibration as long as the service-state variation stays comparable to the range we study. A likely reason is that GRPO computes advantages within the on-policy batch, so common shifts in absolute reward values are partly absorbed before the parameter update. The post-training scoring side is different. The ranking of runs remains stable, but absolute strict-percentile scores can move substantially with the scoring calibration, and stricter scoring conditions emphasize optimization-trained models that preserve more top-decile solutions.

\paragraph{Scoring calibration dominates absolute strict metrics.}
The scoring axis has much larger effect. For the ranked \(p_{30}\) replication reference, \(p_{10}\) pass@1 falls from 41.1\% at s0 to 5.8\% at s9, a 35.3 point drop. For QP, it falls from 40.1\% to 4.6\%, a 35.5 point drop. This is more than an order of magnitude larger than the reward-calibration spread at a fixed scoring point. The same pattern holds at \(p_{20}\), though the drop is less severe: ranked \(p_{30}\) moves from 42.1\% to 12.0\% and QP from 41.7\% to 9.9\%. Scoring calibration therefore controls the absolute strict-percentile scale. This does not mean the models change; the same shared re-executed model outputs are being rescored against shorter corrected references.

The percentile dependence is useful. \(p_{100}\) is invariant because it measures correctness under the timing-sensitive reexecution, not membership in a speed percentile. \(p_{20}\) retains 28.5\% of the s0 ranked-\(p_{30}\) score at s9 and 23.7\% of the QP score. Strict \(p_{10}\) retains only 14.1\% and 11.5\%, respectively. \(p_{10}\) pass@10 is less brittle than pass@1 in relative terms, retaining 28.9\% for ranked \(p_{30}\) and 23.1\% for QP, but its absolute drop is even larger because the starting pass@10 scores are higher. This is the expected behavior of a strict percentile metric: as references are shortened, the boundary for being in the top decile moves faster than the boundary for being correct or merely within the top fifth.

\paragraph{The two reward families respond similarly, but \(p_{30}\) is consistently stronger on strict optimization.}
The ranked \(p_{30}\) and QP families are not two calibration methods; they are two ways of turning the same calibrated human-reference leaderboard into reward. QP supplies a graded reward across the full leaderboard, while ranked \(p_{30}\) makes the top-30\% event a binary optimization gate. Across this sweep, ranked \(p_{30}\) is consistently ahead on strict \(p_{10}\) pass@1. Averaging over ab1--ab8, the ranked-\(p_{30}\) advantage over QP is 1.2 points at s0, 2.3 points at s5, and 0.8 points at s9. The gap is modest compared with the scoring-calibration drop, but it persists across the calibration axis. This supports the reward-design result from \cref{sec:app_env_qwen7b_definitions}: for this Qwen 7B setting, the sharper \(p_{30}\) gate produces a stronger strict-optimization signal than the smoother QP reward, even though both use the same calibrated duration reference.

The QP family sometimes has slightly higher \(p_{100}\), for example 48.4\% for QP ab6 versus a ranked-\(p_{30}\) maximum of 46.9\%. That does not contradict the strict optimization result. It suggests that the smoother QP reward can preserve or improve broad correctness while being less selective for top-decile speed. Conversely, ranked \(p_{30}\) is more aligned with the strict \(p_{10}\) target, but does not dominate every correctness-side number.

\paragraph{Cross-service-state replication.}
For the two reward families, we also include reference runs trained with the same affine coefficients as ab3 but collected under a different CES service state. After shared re-execution, these references can be compared with ab3 to ask whether changing the service state during training, while keeping the reward calibration fixed, visibly changes the learned model. The answer is mostly no. For ranked \(p_{30}\), the reference and ab3 rows are identical or within 0.4 points across the strict \(p_{10}\) scoring grid. For QP, they are also closely aligned. This is the same effect we tried to simulate when we keep CES executions fixed and vary the calibrated reference durations: it changes the relative position of model timings and human-reference timings. The replication check confirms that, within this range, changing CES conditions under fixed reference calibration behaves similarly to changing reference calibration under fixed CES executions.

The absolute scores in this subsection are diagnostic and should be compared only within this shared re-execution campaign. In the main result tables, we rerun a calibration campaign at evaluation time and use the resulting fit to approximate, as closely as possible, the score we would obtain in an ideal no-shift, no-noise timing system: running the same snippet on the same test would always return the same duration, so stored human-reference durations and freshly executed model durations would already live on the same timing scale. That correction moves stored human-reference durations toward the timing scale of the CES state used by the evaluation. For example, the ranked \(p_{30}\) replication reference scores \(p_{10}=41.1\%\) here under s0, while the corresponding main-result table reports 6.0\%. This difference is not a model improvement; it comes from using a different CES state and calibration. The point of this subsection is therefore the within-campaign comparison across reward and scoring calibrations, not an update of the headline scores.

\paragraph{Takeaway.}
The two-dimensional sweep separates three facts that are easy to conflate. First, reward-time affine calibration over the tested range has weak effect on the learned Qwen 7B models. Second, post-training scoring calibration has large effect on absolute strict-percentile scores, especially \(p_{10}\), because it changes the corrected reference durations against which fixed model executions are ranked, though it preserves the relative ordering of the runs in this sweep, which supports the robustness of the comparisons. Third, the relative comparison between the two reward families is stable enough to interpret: ranked \(p_{30}\) remains the stronger strict-optimization training signal, while QP can preserve a bit more correctness.

\clearpage

\subsection{Lessons learned for timing-based RL}
\label{sec:app_execution_lessons}

From the analyses above, we retain that:
\begin{enumerate}
    \item Timing-sensitive dataset construction, training, and evaluation should use a controlled, confined service to measure runtimes of generated code snippets, rather than local sandboxes running concurrently with the RL run: the issue is not only an extra degree of timing noise; it can completely corrupt the timing signal.
    \item Local sandbox execution on the RL workers can recover inconclusive correctness verdicts, as long as timeouts are generous and not constraining, but its durations should not enter timing rewards or timing-sensitive metrics.
    \item Aggregate service health can hide short harmful windows: rare infrastructure bursts can affect a large fraction of rollouts in a few optimizer steps, and silent service slowdowns can shift measured durations even when pass/fail counters look normal.
    \item Model comparisons should use shared and concurrent re-execution in the same controlled setting rather than mixing timings collected under different service states.
    \item A calibration campaign should be rerun for timing-sensitive evaluation, and the affine stored-duration correction should be refit when moving to a new execution-service fleet, after a sandbox upgrade, or after a sustained service-state change.
    \item GRPO within-batch advantage computation can partially absorb common live-service shifts during training; post-training scoring is more sensitive to calibration in absolute strict scores, although the relative model ranking is stable and optimization-trained models stand out more under stricter scoring settings.
\end{enumerate}

\subsection{Improvements and experiments we did not have the time to conduct}
\label{sec:app_execution_future_work}

\paragraph{Controlled timing-noise injection.}
The studies above use noise that happened in real runs: repeated CES reruns, bursts, slowdowns, and calibration changes. A next study could add known noise after the sandbox returns a result. For example, we could add 5--50\,ms to each duration, recompute whether each test should now count as a timeout, and update the stored duration. We could also multiply durations, add larger noise on short tests, or simulate bursts that last for several optimizer steps. Because the injected noise is known, we could compare reward environments directly and see which ones break first, and which ones stay more stable.

\paragraph{Replaying service states on fixed rollouts.}
The RL-run analysis is hard to isolate because the policy, sampled problems, sampled rollouts, and CES state all change at the same time. A cleaner study would keep the rollout logs fixed. Given saved rollouts with code, tests, statuses, and durations, we could recompute rewards under different service traces: healthy execution, real burst windows, synthetic slowdowns, different fallback policies, discard, or quarantine. This could help measure whether the RL environment can correctly smooth part of the execution noise after sandbox execution. It could also estimate how much reward variance comes from the service state. GRPO could then use this estimate to normalize advantages, downweight noisy batches, or trigger extra re-execution. Replaying fixed rollout batches under controlled service states could also measure how the policy update would drift under each service state.

\paragraph{Storing standard deviation estimates.}
Several environments use human-reference durations. The repeated-CES campaign could let us store not only one corrected duration, but also an estimate of its standard deviation. This could be stored for each problem-test-solution duration, or more coarsely for each test-level reference distribution. These estimates could then be used in several places. Filtering could avoid tests whose duration spread is mostly measurement noise. Timeout choices could include a safety margin for noisy tests. Ranking rewards could soften comparisons when the generated snippet and the reference boundary are too close to distinguish reliably. The same estimate could also trigger selective re-execution, or give GRPO a reward-standard-deviation estimate when it computes or normalizes advantages.

\paragraph{Adaptive monitoring and recalibration.}
Live calibration probes could help detect silent service drift, but they also use execution capacity. Too few probes are noisy; too many slow down rollout execution. A sentinel suite could be run from time to time during training and used to detect changes in slope, offset, or tail behavior. When the shift is large enough, the system could refit the affine correction, quarantine recent batches, or pause rollout collection.

\clearpage
\section{Designing Environments and Rewards for Optimization RL}
\label{app:reward}
Optimization tests by themselves do not define an RL reward. They produce execution records, and the reward design decides which parts of those records should affect learning. This appendix gives the detailed version of the argument compressed in \cref{sec:reward}: timing constraints can enter the environment before execution, during execution, or after execution; all settings are reduced to a common reward-facing interface that separates correctness, hard optimization gate, and duration quality into three numerical signals that the reward can then be built upon.

\subsection{Problem taxonomy}
\label{sec:app_env_intervention_points}

For this paper, an RL environment \(\mathcal E\) is the full routine around the model to prepare a rollout that the optimizer will then take a step upon. Given a programming problem \(x\), the environment prepares the prompt shown to the model, receives a generated answer \(y\), decides which tests to run and under which constraints, aggregates the resulting statuses and durations into outcomes, and returns a reward \(R_{\mathcal E}(x,y)\). We define the taxonomy so that each RL environment exposes a common set of signals, from which different reward functions map the rollout to the reward range \([-1,1]\).

Each test \(t\) is an input-output evaluation item with input \(u_t\), expected output \(v_t\), and a default execution cap \(\tau_t\); optimization environments may later replace this cap by an effective limit \(\lambda_t\). Running a candidate program \(y\) on \(t\) returns a measured duration \(\delta_t(y)\) and a status
\begin{equation}
s_t(y)\in\mathcal S=\{\mathrm{success},\mathrm{timeout},\mathrm{failure}\}.
\end{equation}
Here \(\mathrm{success}\) means that the output matches \(v_t\) within the effective time limit, \(\mathrm{timeout}\) means that the run did not finish within that limit and therefore produced no output to compare with \(v_t\), and \(\mathrm{failure}\) means that the run finishes within the effective time limit but its produced output does not match \(v_t\). The executor records more detailed statuses, but the reward notation below uses only this three-way status. For each problem we also have a calibrated human reference pool \(H(x)\): accepted reference solutions with stored duration traces on optimization tests. For \(h\in H(x)\), \(\tilde d_{t,h}\) denotes the calibrated duration of reference solution \(h\) on test \(t\); this pool is used to construct reference-derived limits and percentile rankings, not as a model output.

We split the tests attached to \(x\) into correctness tests \(T^{\mathrm C}(x)\) and a raw optimization-test pool \(T^{\mathrm O}(x)\). The correctness tests are fixed for a given environment and produce the base correctness signal; executed optimization tests can then refine this signal by rejecting failures. The design choices studied in \cref{sec:taxonomy} act on the optimization-test pool, the timing constraints, or the duration record. At this level, the optimization-side routine can be written as three parameterized operators:
\begin{align}
T^{\mathrm O}_{\mathrm{used}} &= F_{\mathrm{pre}}(T^{\mathrm O}(x);\theta_{\mathrm{pre}}),\\
\mathcal Z^{\mathrm O}(x,y) &= E_{\mathrm{intra}}(y,T^{\mathrm O}_{\mathrm{used}};H(x),\theta_{\mathrm{intra}}),\\
\bigl(g_{\mathrm{post}}(x,y),q_{\mathrm{post}}(x,y)\bigr) &= Q_{\mathrm{post}}(\mathcal Z^{\mathrm O}(x,y);H(x),\theta_{\mathrm{post}}),
\end{align}
where \(\mathcal Z^{\mathrm O}(x,y)\) is the optimization-side execution record, including the executed tests, statuses, durations, and intended limits. These operators are meta-classes of environments. A pre-execution environment is defined by the routine used to instantiate \(F_{\mathrm{pre}}\), such as an absolute-duration filter, a relative-duration filter, or a character-length filter. An intra-execution environment is defined by the constraints in \(E_{\mathrm{intra}}\), such as absolute timeouts, relative timeouts, or ranked-worst time limits. A post-execution environment is defined by whether \(Q_{\mathrm{post}}\) turns the observed durations into a ranked gate, a graded ranked scalar, or both. After these routines run, the reward-facing interface contains a correctness signal \(c_{\mathrm{cor}}\), its strict optimization-aware refinement \(\tilde c_{\mathrm{cor}}\), a hard optimization gate \(g\), and a graded duration quality signal \(q\).

\begin{table}[h]
\centering
\small
\caption{\textbf{Notation for optimization-aware RL environments.}
The same symbols apply to pre-execution filters, intra-execution time limits, and post-execution ranking rewards.}
\label{tab:app_env_notation}
\begin{tabular}{p{0.18\linewidth}p{0.72\linewidth}}
\toprule
\bfseries Symbol & \bfseries Meaning \\
\midrule
\(\mathcal E,x,y\) & RL environment, programming problem, and generated candidate program \\
\(u_t,v_t,\tau_t\) & input, expected output, and default execution cap for test \(t\) \\
\(T_0(x)\) & original public, private, and generated correctness tests \\
\(T_+(x)\) & additional correctness tests used in more-correctness settings \\
\(T^{\mathrm C}(x)\) & correctness tests used to judge semantic correctness \\
\(T^{\mathrm O}(x)\) & raw optimization-test pool \\
\(T^{\mathrm O}_{\mathrm{used}}(x)\) & optimization tests retained by pre-execution filtering \\
\(\mathcal Z^{\mathrm O}(x,y)\) & optimization-side execution record for candidate \(y\) \\
\(H(x)\) & calibrated human/reference solutions with stored duration traces \\
\(\tilde d_{t,h}\) & calibrated duration of reference solution \(h\) on test \(t\) \\
\(D_t\) & calibrated duration distribution for test \(t\) over \(H(x)\) \\
\(A,\bar d_t\) & aggregation operator, either mean or median, and aggregate duration \(\bar d_t=A(D_t)\) used by duration filters \\
\(\mathcal S\) & execution statuses: success, timeout, or failure \\
\(s_t(y),\delta_t(y)\) & execution status and measured duration for candidate \(y\) on test \(t\) \\
\(\lambda_t\) & intended optimization time limit for test \(t\) \\
\(\phi,\rho\) & optimization timeout fraction and timeout-tolerance threshold \\
\(p_t(y),a(y)\) & normalized per-test reference percentile and mean per-test percentile for candidate \(y\), with lower values faster \\
\(c_{\mathrm{cor}},\tilde c_{\mathrm{cor}}\) & correctness-test pass indicator and strict optimization-aware correctness prerequisite used before efficiency credit \\
\(g,g_{\mathrm{to}},g_{\mathrm{dur}}\) & final hard optimization gate, timeout-ratio gate, and duration-ranking gate \\
\(q,q_{\mathrm{to}},q_{\mathrm{QP}},q_{\mathrm{QAR}}\) & final graded optimization scalar, timeout-rate scalar, leaderboard percentile, and per-test percentile, all normalized so \(0\) is best or neutral and \(1\) is worst \\
\bottomrule
\end{tabular}
\end{table}

\subsection{Classes of Optimization RL Environments}
\label{sec:app_env_general_operators}
\label{sec:app_env_qwen7b_definitions}
\label{sec:app_reward_scalarizers}

The construction and timing stack in \cref{app:dataset,app:ces_fallback} give us optimization tests, calibrated human-reference durations, and a sandbox that can record per-test timing, but these ingredients are not yet a good optimization RL environment. A good environment should induce a useful learning curve: early rollouts should still receive a correctness signal, correct but slow snippets should not be indistinguishable from wrong code, and progressively faster correct snippets should see progressively better optimization feedback. Left unshaped, raw optimization tests can make the rollout mostly a harder correctness check, make the reward too sparse because only near-top solutions ever receive efficiency credit, or make the scalar too focused on speed and give credit to degenerate fast-but-wrong programs. The sparsity regime also matters: a constraint that gives efficiency credit only to the top \(1\%\) of isolated candidate snippets may never reward early GRPO rollouts, while a loose constraint can produce dense rewards with little optimization pressure. The raw-duration controls in \cref{sec:app_naive_duration_rewards} and the baseline discussion in \cref{sec:training} show this limitation empirically: even when optimization tests are included in the rollout, directly adding duration to the scalar reward gives only small strict-threshold gains relative to reference-normalized environment/reward choices.

We study optimization RL as a one-shot generation-from-scratch problem. The model produces one program from the problem statement, not an iterative refinement trajectory around a known correct solution, and the environment assigns \(R_{\mathcal E}(x,y)\) to that one generated program \(y\) for one problem \(x\). Calibrated human references may set filters, time limits, or ranks, but any relative comparison among model snippets appears later, when GRPO turns the scalar rewards of a sampled batch into advantages; it is not part of the environment score itself. The goal of the taxonomy is therefore to be exhaustive with respect to this rollout pipeline: a timing constraint can enter before sandbox execution, during sandbox execution, or after execution. Within each stage, an environment family fixes the type of constraint, and a concrete environment instance fixes its parameterization.

Any environment starts from the loaded sample \(x\), before anything is queried to the model. The sample provides the problem statement, the correctness tests \(T_0(x)\) and \(T_+(x)\), the raw optimization-test pool \(T^{\mathrm O}(x)\), and the reference duration traces from \(H(x)\). At this stage, the environment fixes which correctness tests will judge the answer, with \(T^{\mathrm C}=T_0\) for our baseline standard RLVR and \(T^{\mathrm C}=T_0\cup T_+\) for any of our optimization-aware environments. In the case of any optimization environment, it starts with the full pool \(T^{\mathrm O}(x)\) of optimization tests before any pre-execution filter or time-limit rule is applied. Stored reference human durations on these optimization tests pass through
\begin{equation}
\psi(d)=
\begin{cases}
d, & d>0,\\
10, & \text{otherwise},
\end{cases}
\qquad
\tilde d=\min\{10,\max\{0,\Gamma(\psi(d))\}\},
\end{equation}
where \(\Gamma\) is either the identity or an affine correction fitted from repeated CES measurements; \cref{sec:app_duration_correction} gives the calibration evidence. Candidate durations \(\delta_t(y)\) are not transformed by \(\Gamma\): they are measured in the live execution-service scale and compared to calibrated reference distributions only after execution. For each optimization test \(t\), let \(D_t=\{\tilde d_{t,h}:h\in H(x)\}\) be the sanitized reference-duration distribution when available. The aggregate duration used by duration filters is \(\bar d_t=A(D_t)\), where \(A\) is either the mean or the median.

The model is then queried with the problem statement and returns a candidate program \(y\). The environment must now judge the quality of \(y\), beginning with the subset of optimization tests that will be run.

Pre-execution filtering is the part of the environment that acts before the generated program is run. The no-op filter keeps the entire optimization pool,
\begin{equation}
F_{\mathrm{id}}(T)=T.
\end{equation}
Active pre-execution families, in contrast, are parameterized filters that select a subset of these optimization tests. We explore the following subclasses of pre-execution environments:
\begin{enumerate}
    \item \textbf{Absolute-duration filtering} uses the calibrated human reference durations to keep only optimization tests whose aggregate duration is below a threshold \(a\):
    \begin{equation}
    F_{\mathrm{abs}}^{a}(T)=\{t\in T:\ D_t\ \mathrm{usable},\ \bar d_t<a\}.
    \end{equation}
    \textbf{More concretely, what it does is:}
    \begin{enumerate}
        \item[(a)] For each optimization test, look at the recorded durations of the references on that test and compute the test-level mean duration \(\bar d_t\).
        \item[(b)] Keep only tests with \(\bar d_t<a\). For example, with \(a=2\,\text{s}\), keep only optimization tests whose mean reference duration is below \(2\,\text{s}\).
        \item[(c)] Run only these retained tests with a uniform optimization timeout, for instance \(\ell=0.5\,\text{s}\). With timeout tolerance \(\rho=0.1\), the optimization side passes if at most \(10\%\) of retained tests time out.
    \end{enumerate}
    \item \textbf{Character-length filtering} uses the size of the input-output pair as a proxy for workload and keeps only tests below a cutoff \(L\):
    \begin{equation}
    F_{\mathrm{len}}^{L}(T)=\{t\in T:\ |u_t|+|v_t|<L\}.
    \end{equation}
    \textbf{More concretely, what it does is:}
    \begin{enumerate}
        \item[(a)] For each optimization test, compute the serialized input-output size \(|u_t|+|v_t|\).
        \item[(b)] Keep only tests with \(|u_t|+|v_t|<L\). For example, with \(L=10^3\), keep tests whose input plus output has fewer than \(1{,}000\) characters.
        \item[(c)] Run only these retained tests with a uniform optimization timeout, for instance \(\ell=0.5\,\text{s}\). With timeout tolerance \(\rho=0.1\), the optimization side passes if at most \(10\%\) of retained tests time out.
    \end{enumerate}
    \item \textbf{Relative-duration filtering} removes the slowest calibrated fraction of the optimization pool. Let \(T_D=\{t\in T:D_t\ \mathrm{usable}\}\) and let \(S_r(T_D)\) be the \(\lfloor r|T_D|/100\rfloor\) slowest tests in \(T_D\) under \(\bar d_t\). Then
    \begin{equation}
    F_{\mathrm{rel}}^{r}(T)=T\setminus S_r(T_D).
    \end{equation}
    \textbf{More concretely, what it does is:}
    \begin{enumerate}
        \item[(a)] For each optimization test, look at the recorded durations of the references on that test, compute the test-level mean duration \(\bar d_t\), and sort the rankable optimization tests by \(\bar d_t\), slowest first.
        \item[(b)] Drop the slowest \(r\%\) of rankable optimization tests. For example, with \(r=80\), drop the slowest \(80\%\) and keep roughly the fastest \(20\%\).
        \item[(c)] Run only these retained tests with a uniform optimization timeout, for instance \(\ell=0.5\,\text{s}\). With timeout tolerance \(\rho=0.1\), the optimization side passes if at most \(10\%\) of retained tests time out.
    \end{enumerate}
\end{enumerate}
Thus \(F_{\mathrm{abs}}^{a}\) retains only tests whose calibrated aggregate is defined and below the cutoff, \(F_{\mathrm{len}}^{L}\) depends only on input-output character count, and \(F_{\mathrm{rel}}^{r}\) retains tests without usable aggregate durations while dropping the slowest calibrated fraction among the tests that can be ranked. These are parameterized families: the environment sweep varies \(a\), \(L\), or \(r\), and otherwise leaves the rest of the rollout routine fixed.

These filters do not by themselves reward speed; they choose which workloads are allowed to contribute to the later timeout or ranking signal. Absolute-duration filtering makes the workload scale explicit. With \(a=2\,\text{s}\), for example, the retained optimization tests are those whose calibrated aggregate human runtime is below two seconds, so choosing a comparable optimization cap has an interpretable scale. The same global threshold can nevertheless mean different things across problems: some problems already live entirely below two seconds and are effectively unfiltered, while others have a duration profile that the threshold cuts into two very different subsets. Character-length filtering connects to the duration/length filterability analysis in \cref{sec:data_filterability,sec:app_filterability}. It is cheap, deterministic, and can be applied even without stored reference durations, but it assumes serialized input-output length is a usable proxy for computational workload; this assumption can fail when runtime is driven by input structure, values, or control flow rather than character count. Relative-duration filtering avoids a global cutoff by removing a per-problem fraction of slow calibrated tests, which can look better matched to heterogeneous tasks. Its risk is that the removed tail may be exactly where the optimization pressure lives: even a small removal fraction can delete hard cases on some problems, and because the ordering comes from the human reference pool, it may overfit to references by discarding tests that are slow on average across human solutions but could have produced a broad duration spread across model rollouts during RL. Tuning \(a\), \(L\), and \(r\) is therefore part of defining the learning curve, trading broad coverage against reward sparsity and strict optimization pressure.

Once the optimization-test pool is fixed---either as the full pool under \(F_{\mathrm{id}}\) or as a subset selected by a pre-execution environment---the candidate program \(y\) and the retained tests are sent to the execution sandbox. The sandbox is controlled by the timeout assigned to each test. In the default execution setting, the intended time limit is the sample's default cap, \(\lambda_t=\tau_t\). Active intra-execution environments act on the sandbox by replacing this default with a parameterized optimization timeout:
\begin{enumerate}
    \item \textbf{Absolute-timeout environments} use the same intended optimization limit \(\ell\) for every retained optimization test:
    \begin{equation}
    \lambda_t^{\mathrm{abs}}(\ell)=\ell.
    \end{equation}
    \textbf{More concretely, what it does is:}
    \begin{enumerate}
        \item[(a)] Start from the retained optimization-test pool.
        \item[(b)] Set the same intended optimization timeout for every retained test. For example, with \(\ell=0.5\,\text{s}\), every retained optimization test has deadline \(0.5\,\text{s}\).
        \item[(c)] Run the retained tests. If a successful execution has \(\delta_t(y)\geq\ell\), record that test as a timeout. With timeout tolerance \(\rho=0.1\), the optimization side passes if at most \(10\%\) of retained tests time out.
    \end{enumerate}
    \item \textbf{Relative-timeout environments} set a per-test limit from the calibrated reference distribution for that test:
    \begin{equation}
    \lambda_t^{\mathrm{rel}}(p)=\max\{10^{-3},\operatorname{Percentile}_{p}(D_t)\}.
    \end{equation}
    \textbf{More concretely, what it does is:}
    \begin{enumerate}
        \item[(a)] Start from the retained optimization-test pool, and for each retained test read the recorded reference durations \(D_t\).
        \item[(b)] Set the intended timeout of each test to a percentile of its own reference durations. For example, with \(p=30\), test \(t\) receives the 30th-percentile duration of \(D_t\) as \(\lambda_t\).
        \item[(c)] Run the retained tests. If a successful execution has \(\delta_t(y)\geq\lambda_t\), record that test as a timeout. With timeout tolerance \(\rho=0.1\), the optimization side passes if at most \(10\%\) of retained tests time out.
    \end{enumerate}
    \item \textbf{Ranked-reference timeout environments} first select a top reference set \(H_p(x)\), then set each test limit from the selected reference durations:
    \begin{equation}
    \lambda_t^{\mathrm{ranked}}(p,B)=B\bigl(\{\tilde d_{t,h}:h\in H_p(x)\}\bigr).
    \end{equation}
    \textbf{More concretely, what it does is:}
    \begin{enumerate}
        \item[(a)] Rank the reference solutions on the problem by their average per-test rank on the optimization tests, and keep the best \(p\%\) as \(H_p(x)\).
        \item[(b)] For each optimization test \(t\), set \(\lambda_t\) from the selected reference durations \(\{\tilde d_{t,h}:h\in H_p(x)\}\). For example, ranked-worst \(p=80\) with \(B=\max\) keeps the top \(80\%\) of references and sets \(\lambda_t\) to their slowest duration on each test.
        \item[(c)] Run the retained tests. If a successful execution has \(\delta_t(y)\geq\lambda_t\), record that test as a timeout. With timeout tolerance \(\rho=0.1\), the optimization side passes if at most \(10\%\) of retained tests time out.
    \end{enumerate}
\end{enumerate}
Here \(\ell\) is an absolute time limit in seconds, \(p\) is a reference percentile, \(B\) is either the maximum or median aggregation, and \(H_p(x)\) is the best \(p\%\) of reference solutions after ranking references by their average per-test rank on \(T^{\mathrm O}_{\mathrm{used}}(x)\). The retained ranked-worst environments use \(B=\max\). When a reference-derived limit needs \(D_t\), tests without usable reference durations can be omitted before execution; in that case \(T^{\mathrm O}_{\mathrm{used}}\) denotes the remaining executed optimization tests. Whenever the CES path is given an optimization time-limit override, execution runs under the 10s service cap and then applies the intended limit by post-filtering: a successful test is changed to timeout whenever \(\delta_t(y)\geq\lambda_t\).

Intra-execution environments introduce optimization pressure through the sandbox budget itself. This is different from pre-execution filtering: a pre-execution filter chooses which tests form the workload and then lets execution unfold blindly, whereas an intra-execution environment keeps the retained workload fixed and changes how much time the candidate is allocated to complete it. These choices are therefore not equivalent. Removing a test removes its evidence from both the hard gate and the graded signal, while tightening \(\lambda_t\) leaves the test in the rollout but changes whether the run becomes a timeout. Absolute timeouts are the simplest version of this idea and make the deadline easy to read, but a single \(\ell\) can be too coarse when the retained tests span very different workloads. Relative timeouts adapt test by test by extracting a quantile from \(D_t\), but they can assemble an unrealistic simultaneous constraint when different tests reward different algorithmic trade-offs: the \(20\%\) fastest duration on each test need not come from the same kind of solution. Ranked-reference timeouts try to avoid that per-test mixture by first selecting a coherent pool of strong reference solutions and then deriving deadlines from their traces. They can fail in the opposite direction: if each selected reference is weak on a different subset of tests, the resulting deadlines may allow a candidate to match the worst selected reference on every test without matching any reference's strengths. Both reference-derived variants also inherit the biases of the human reference pool, so their parameterization controls not only strictness but also whether the environment potentially overfits the observed reference pool.

After execution, the optimization-side record is
\begin{equation}
\mathcal Z^{\mathrm O}(x,y)=\{(t,s_t(y),\delta_t(y),\lambda_t):t\in T^{\mathrm O}_{\mathrm{used}}(x)\}.
\end{equation}

Once execution finishes, statuses and durations are summarized separately. Execution statuses produce the correctness signal in the same way for all environment families:
\begin{equation}
c_{\mathrm{cor}}(x,y)=\mathrm{1}\{s_t(y)=\mathrm{success}\ \forall t\in T^{\mathrm C}(x)\},
\end{equation}
where \(T^{\mathrm C}=T_0\) for standard RLVR and \(T^{\mathrm C}=T_0\cup T_+\) for the ``more-correctness'' setting and for all optimization RL environment classes considered below. Optimization-aware rewards use the stricter prerequisite
\begin{equation}
\tilde c_{\mathrm{cor}}(x,y)=c_{\mathrm{cor}}(x,y)\,
\mathrm{1}\{s_t(y)\in\{\mathrm{success},\mathrm{timeout}\}\ \forall t\in T^{\mathrm O}_{\mathrm{used}}(x)\}.
\end{equation}
Thus executed optimization tests may be correct or may time out, but any failure on an executed optimization test makes \(\tilde c_{\mathrm{cor}}=0\). At this point, the environment has produced its first reward-facing output signal, \(\tilde c_{\mathrm{cor}}\): a strict correctness prerequisite that can be fed to a reward function.

The same status summary also records the optimization timeout fraction
\begin{equation}
\phi(x,y)=\frac{1}{|T^{\mathrm O}_{\mathrm{used}}(x)|}\sum_{t\in T^{\mathrm O}_{\mathrm{used}}(x)} \mathrm{1}\{s_t(y)=\mathrm{timeout}\},
\end{equation}
with \(\phi=0\) when no optimization tests are used. We use the same lower-is-better convention as the percentile quantities below: \(q=0\) is best and \(q=1\) is worst, with reward functions later converting \(q\) into higher-is-better credit. A timeout tolerance \(\rho\) defines the status-side optimization hard gate, and the same timeout fraction can be exposed as a graded optimization scalar:
\begin{equation}
g_{\mathrm{to}}^{\rho}(x,y)=\mathrm{1}\{\phi(x,y)\leq\rho\},
\qquad
q_{\mathrm{to}}(x,y)=\phi(x,y).
\end{equation}
The setting \(\rho=1\) makes the timeout gate a no-op, while smaller values turn the timeout fraction into a hard optimization constraint. This tolerance parameterization is not a class of its own but an attached parameter of the different classes of optimization RL environments, because it directly controls how timeout statuses are transferred into downstream reward impact. Set to \(\rho=1\), it shuts down any optimization pressure through the hard optimization gate for the workload selected by pre-execution filters and constrained by intra-execution time limits.

Active post-execution environments process the measured per-test durations \(\{\delta_t(y):t\in T^{\mathrm O}_{\mathrm{used}}\}\) after the sandbox has finished. A raw-duration reward is the simplest version of this idea: it averages durations over a chosen test set and normalizes that average into the reward range, without using the calibrated reference pool. The reference-normalized variants go further by comparing the candidate to \(H(x)\). For each test in \(T^{\mathrm O}_{\mathrm{used}}\) with usable \(D_t\), the candidate duration is inserted into the reference durations and scored as a normalized percentile \(p_t(y)\in[0,1]\), with \(0\) assigned to the fastest participant and \(1\) to the slowest; exact ties receive the previous participant's percentile. The mean-percentile statistic averages only over retained tests with usable reference durations:
\begin{equation}
a(y)=\frac{1}{|\{t\in T^{\mathrm O}_{\mathrm{used}}:D_t\ \mathrm{usable}\}|}
\sum_{\substack{t\in T^{\mathrm O}_{\mathrm{used}}\\D_t\ \mathrm{usable}}}p_t(y).
\end{equation}
This gives the per-test percentile (QAR) family: in the retained ranking variants, the per-test rank is normalized as the percentile \(p_t(y)\), and QAR uses the aggregate \(a(y)\) directly,
\begin{equation}
q_{\mathrm{QAR}}(y)=a(y).
\end{equation}
The leaderboard percentile (QP) family uses the same aggregate only as an intermediate score: it sorts the candidate and references by \(a(\cdot)\), then converts the candidate's solution-level position into a leaderboard percentile. Let \(b(y)\) be the zero-indexed position of \(y\) in that ordering, and let \(N_{\mathrm{rank}}\) be the number of ranked participants. Its graded scalar is
\begin{equation}
q_{\mathrm{QP}}(y)=\frac{b(y)}{\max\{1,N_{\mathrm{rank}}-1\}}.
\end{equation}
\textbf{Post-execution ranking} is parameterized by a duration threshold \(p\in[0,1]\). This turns into a hard optimization gate on the selected ranked scalar:
\begin{align}
g_{\mathrm{dur,QAR}}^{p}(y)&=\mathrm{1}\{q_{\mathrm{QAR}}(y)\leq p\},\\
g_{\mathrm{dur,QP}}^{p}(y)&=\mathrm{1}\{q_{\mathrm{QP}}(y)\leq p\},\\
g_{\mathrm{dur}}^{p}(y)&\in\{g_{\mathrm{dur,QAR}}^{p}(y),g_{\mathrm{dur,QP}}^{p}(y)\},\\
q_{\mathrm{dur}}(y)&\in\{q_{\mathrm{QAR}}(y),q_{\mathrm{QP}}(y)\}.
\end{align}
Equivalently, if the ranked outputs are written as higher-is-better quality scores \(\bar q_{\mathrm{QAR}}=1-q_{\mathrm{QAR}}\) and \(\bar q_{\mathrm{QP}}=1-q_{\mathrm{QP}}\), the same lower-is-better top-\(p\) threshold becomes \(\bar q_{\mathrm{QAR}}(y)\geq1-p\) and \(\bar q_{\mathrm{QP}}(y)\geq1-p\), respectively.
When no duration post-processing is configured, \(g_{\mathrm{dur}}\equiv1\) and the final scalar below falls back to \(q=q_{\mathrm{to}}\) by setting \(q_{\mathrm{dur}}\equiv0\). Conversely, when an environment relies only on duration-based post-execution ranking, we set \(\rho=1\), so \(g_{\mathrm{to}}^{\rho}\equiv1\), and use infinite intended optimization timeouts so that \(q_{\mathrm{to}}\equiv0\) and \(q=q_{\mathrm{dur}}\). In practice, each test execution is still capped at 10s for systems reasons, and timeout durations are recorded at that cap before being ranked.
\textbf{More concretely, what post-execution ranking do is:}
\begin{enumerate}
    \item[(a)] Run the retained optimization tests under the 10s service cap and record one duration \(\delta_t(y)\) for each executed test; if the run reaches the cap, record \(\delta_t(y)=10\,\text{s}\).
    \item[(b)] For each usable test, insert \(\delta_t(y)\) into the recorded reference durations \(D_t\) and compute the per-test percentile \(p_t(y)\in[0,1]\).
    \item[(c)] Example: QAR averages these percentiles into \(q_{\mathrm{QAR}}(y)=a(y)\). QP ranks the same \(a(y)\) against the reference aggregates to compute \(q_{\mathrm{QP}}(y)\); with duration threshold \(p=0.3\), the duration gate passes candidates that rank in the top \(30\%\) under the selected ranked scalar.
\end{enumerate}

Post-execution ranking applies optimization pressure by making the candidate compete against a reference leaderboard after execution, instead of choosing a workload cutoff before execution or changing the sandbox budget during execution. Relative to pre- and intra-execution families, this moves the strictness choice out of workload construction: the environment does not need to tune \(a\), \(L\), \(r\), or a per-test deadline \(\lambda_t\) to make speed visible, although a downstream threshold \(p\) can still turn the ranked scalar into a hard gate. In practice, the model is rewarded for moving upward in the reference ordering on the same problem, which is close to how many code-optimization benchmarks are evaluated as ranked speed comparisons. The price is that the reference distribution becomes part of the environment. If \(H(x)\) is weak, too narrow, or biased toward one implementation family, then a generated solution near the reference pack and a much faster solution may not be separated enough, or the model may overfit to the observed reference trade-offs. Note that this does not make this type of environment less generalizable, or stuck in this potential pitfall: the reference distribution could be generated, learned, refreshed adversarially against the current model, or augmented with past rollout positions. We keep it fixed here for the sake of simplicity, as stated in the limits of our study at the beginning of this subsection.

The QAR/QP split is the main scalarization choice once post-execution ranking is selected. QAR keeps the average per-test percentile \(a(y)\) itself, so it rewards broad movement across the retained tests and makes the per-test-to-problem aggregation choice explicit; the displayed definition uses a mean, while \cref{sec:app_ranking_metrics} compares this family to median, trimmed, and filtered variants. QP takes the same aggregate \(a(y)\), sorts the candidate among the references, and converts that position into a problem-level leaderboard percentile. The tradeoff is dynamic range. If different reference solutions are fast on different tests and slow on others, the raw average \(a(y)\) can compress many candidates into a narrow interval; QP spreads this aggregate back into leaderboard positions, making more relative positions visible to the reward. The cost is saturation at the extreme: once a candidate is best on the existing ladder, QP remains at the best position, while QAR can still decrease if the candidate keeps improving across tests beyond any pre-existing reference. \Cref{sec:app_ranking_metrics} therefore studies which ranked scalar has low repeated-CES noise, enough within-problem discrimination, and stable ordering after calibration, before the reward-composition subsection decides how to consume \(g\) and \(q\).

The final optimization interface multiplies the status-side and duration-side gate factors,
\begin{equation}
\begin{aligned}
g(x,y)&=g_{\mathrm{to}}^{\rho}(x,y)\,g_{\mathrm{dur}}^{p}(x,y),\\
q(x,y)&=1-\bigl(1-q_{\mathrm{to}}(x,y)\bigr)\bigl(1-q_{\mathrm{dur}}(x,y)\bigr)\\
&=
\begin{cases}
q_{\mathrm{dur}}(x,y), & \text{for active post-execution ranking},\\
q_{\mathrm{to}}(x,y), & \text{otherwise}.
\end{cases}
\end{aligned}
\end{equation}
where inactive gate factors are set to \(1\) and inactive graded scalars are set to \(0\). Thus the environment exposes one hard optimization gate \(g\) and one graded optimization scalar \(q\), after separating the correctness prerequisite \(\tilde c_{\mathrm{cor}}\).

\Cref{tab:app_qwen7b_env_definitions} lists the screened environment families and the signals they expose. The parameter ranges are the offline-simulator grids; Qwen 7B and CWM 32B online runs promote selected points from these grids rather than rerunning every setting at scale. For compactness, let \(\mathcal L=\{0.1,0.2,0.5,1,2,5,10\}\) seconds, \(\mathcal C=\{10^2,10^3,\ldots,10^8\}\) characters, \(\mathcal P=\{10,20,\ldots,90\}\), \(\mathcal R=\{0,0.01,0.02,0.05,0.10,0.20,0.50,0.90\}\), and \(\mathcal G=\{0.10,0.20,\ldots,0.90\}\).

\begin{table}[h!]
\centering
\scriptsize
\setlength{\tabcolsep}{3pt}
\renewcommand{\arraystretch}{1.08}
\caption{\textbf{Environment definitions and screened parameter ranges.}
The table covers the families in \cref{tab:qwen7b_environments}. It lists the test transformation or timing constraint, then the signals exposed to reward functions. Optimization environments use \(T^{\mathrm C}=T_0\cup T_+\) with a 10s correctness cap and include the optimization tests.}
\label{tab:app_qwen7b_env_definitions}
\begin{tabular}{p{0.19\linewidth}p{0.46\linewidth}p{0.27\linewidth}}
\toprule
\bfseries Environment class & \bfseries Environment definition & \bfseries Exposed signals \\
\midrule
Standard RLVR & \(T^{\mathrm C}=T_0\), \(T^{\mathrm O}_{\mathrm{used}}=\emptyset\). & \(\tilde c_{\mathrm{cor}}=c_{\mathrm{cor}}\), \(g\equiv1\), \(q\equiv0\). \\
MC & \(T^{\mathrm C}=T_0\cup T_+\), \(T^{\mathrm O}_{\mathrm{used}}=\emptyset\). & \(\tilde c_{\mathrm{cor}}=c_{\mathrm{cor}}\), \(g\equiv1\), \(q\equiv0\). \\
MC + TLoc & Same as MC, with a 10s cap on correctness tests. & \(\tilde c_{\mathrm{cor}}=c_{\mathrm{cor}}\), \(g\equiv1\), \(q\equiv0\). \\
MC + MO & \(T^{\mathrm O}_{\mathrm{used}}=T^{\mathrm O}\), 10s optimization cap. & \(g=g_{\mathrm{to}}^{0}\); \(q=q_{\mathrm{to}}\). \\
\midrule
Absolute duration filters & \(T^{\mathrm O}_{\mathrm{used}}=F_{\mathrm{abs}}^{a}(T^{\mathrm O})\), \(a\in\mathcal L\); sandbox cap \(\lambda_t=\ell\), \(\ell\in\mathcal L\). & \(g=g_{\mathrm{to}}^{\rho}\), \(\rho\in\mathcal R\); \(q=q_{\mathrm{to}}\). \\
Character-length filters & \(T^{\mathrm O}_{\mathrm{used}}=F_{\mathrm{len}}^{L}(T^{\mathrm O})\), \(L\in\mathcal C\); sandbox cap \(\lambda_t=\ell\), \(\ell\in\mathcal L\). & \(g=g_{\mathrm{to}}^{\rho}\), \(\rho\in\mathcal R\); \(q=q_{\mathrm{to}}\). \\
Relative duration filters & \(T^{\mathrm O}_{\mathrm{used}}=F_{\mathrm{rel}}^{r}(T^{\mathrm O})\), \(r\in\mathcal P\); sandbox cap \(\lambda_t=\ell\), \(\ell\in\mathcal L\). & \(g=g_{\mathrm{to}}^{\rho}\), \(\rho\in\mathcal R\); \(q=q_{\mathrm{to}}\). \\
\midrule
Absolute timeout & No pre-filter; \(\lambda_t^{\mathrm{abs}}(\ell)=\ell\), \(\ell\in\mathcal L\). & \(g=g_{\mathrm{to}}^{\rho}\), \(\rho\in\mathcal R\); \(q=q_{\mathrm{to}}\). \\
Relative timeout & No pre-filter; \(\lambda_t=\lambda_t^{\mathrm{rel}}(p_{\lambda})\), \(p_{\lambda}\in\mathcal P\). & \(g=g_{\mathrm{to}}^{\rho}\), \(\rho\in\mathcal R\); \(q=q_{\mathrm{to}}\). \\
Ranked-reference timeout & No pre-filter; \(\lambda_t=\lambda_t^{\mathrm{ranked}}(p_{\lambda},B)\), \(p_{\lambda}\in\mathcal P\), \(B\in\{\max,\mathrm{median}\}\). The online ranked-worst family uses \(B=\max\). & \(g=g_{\mathrm{to}}^{\rho}\), \(\rho\in\mathcal R\); \(q=q_{\mathrm{to}}\). \\
\midrule
QAR ranked quality & No pre-filter; 10s optimization cap; post-execution QAR ranking against \(D_t\), with \(\rho=1\) and duration threshold \(p\in\mathcal G\). & \(g=g_{\mathrm{dur,QAR}}^{p}\); \(q=q_{\mathrm{QAR}}\). \\
QP ranked quality & No pre-filter; 10s optimization cap; post-execution QP ranking against \(D_t\), with \(\rho=1\) and duration threshold \(p\in\mathcal G\). & \(g=g_{\mathrm{dur,QP}}^{p}\); \(q=q_{\mathrm{QP}}\). \\
\bottomrule
\end{tabular}
\end{table}
\FloatBarrier

\subsection{How to create a good ranking score based on recorded durations}
\label{sec:app_ranking_metrics}

In \cref{sec:app_env_general_operators}, post-execution environments instantiate \(Q_{\mathrm{post}}\): after the sandbox has run the candidate, they turn the recorded durations into a ranked optimization signal. A ranking score can be generous or harsh, depending on whether it frequently places candidates in good tiers or in the worst tiers; it can use a wide effective range or collapse most candidates into a small interval; and it can be more or less sensitive to timing measurements, and therefore more or less stable across reruns. The formulas in the previous subsection use normalized \([0,1]\) quantities that we here rather map to the \(0\)--\(100\) percentile interval, with lower values better.

Fix a problem \(x\), a candidate program \(y\), and the optimization tests \(\mathcal T\) on which \(y\) has a usable duration and at least one reference duration is usable. For each \(t\in\mathcal T\), let \(D_t=\{d_{t,h}:h\in H_t\}\) be the usable reference durations on that test, and let \(n_t=|D_t|+1\) be the number of ranked participants after inserting \(y\). The basic object is the per-test percentile
\begin{equation}
p_t(y)=100\cdot\frac{\bar r_t(y)-1}{n_t-1},
\end{equation}
where \(\bar r_t(y)\) is the one-indexed rank of \(\delta_t(y)\) after sorting \(D_t\cup\{\delta_t(y)\}\) from fastest to slowest. Tied durations are assigned the average tied rank in this metric sweep. Thus \(p_t(y)=0\) means fastest on test \(t\), \(p_t(y)=100\) means slowest, and metrics can differ in how they aggregate or transform the vector \(\{p_t(y):t\in\mathcal T\}\). We write \(\Pi(z,A)\) for this operation of ranking a scalar value \(z\) among a reference set \(A\), after adding \(z\) to the set.

The most direct aggregation keeps the per-test percentile and averages it:
\begin{equation}
M_{\mathrm{mean}}(y)=\frac{1}{|\mathcal T|}\sum_{t\in\mathcal T}p_t(y),
\end{equation}
while median percentile replaces the average by the median. Trimmed mean percentile sorts the per-test percentiles, removes the lowest and highest 10\% before averaging, and is only defined when enough tests remain after trimming; this drops \(m=\max\{1,\lfloor0.1|\mathcal T|\rfloor\}\) values at each end when \(|\mathcal T|\geq5\). The motivation is that the lowest and highest per-test percentiles may come from timing-measurement outliers, so removing them can make the ranking function less sensitive to timing noise. Log-space percentile first replaces each positive duration by its logarithm before computing \(p_t\), then averages across tests; this mainly tries to be more robust towards outliers that concentrate near the slowest percentiles rather than the fastest percentiles. And also, near the slowest percentiles, small variations are probably less meaningful and can be flattened by the log.

Another approach is to avoid feeding the scalar percentiles directly, since they can be noisy, and instead threshold them to filter out part of the timing noise. For \(k\in\{25,50\}\), the top-\(k\) share is
\begin{equation}
S_k(y)=\frac{100}{|\mathcal T|}\sum_{t\in\mathcal T}\mathbb{I}\{p_t(y)<k\}.
\end{equation}
This is a higher-is-better candidate ranking score: \(S_{25}=60\) means that the candidate lands in the fastest quartile on 60\% of the retained tests.

Another possibility is to aggregate raw durations across tests before ranking. Total-duration percentile ranks the candidate total duration against reference totals,
\begin{equation}
M_{\mathrm{tot}}(y)=
\Pi\!\left(\sum_{t\in\mathcal T}\delta_t(y),\,
\left\{\sum_{\substack{t\in\mathcal T\\d_{t,h}\ \mathrm{usable}}}d_{t,h}:h\in H_{\mathcal T}\right\}\right),
\end{equation}
where \(H_{\mathcal T}\) denotes references with at least one usable duration on the tests being compared. The idea is that aggregating across tests gives more measurements, so individual timing outliers can cancel out. It can also let a small number of long tests dominate the problem-level value; this can be useful if these tests are the ones that contain the real optimization signal, but harmful if the other tests also matter for scoring whether a candidate is good according to the reference pool.

We also tested metrics that compare the shape of the candidate runtime curve to the reference pool. At a high level, these metrics ask whether the candidate slows down more or less than the references as tests become harder. In practice, it is the median stored human-reference duration on that test,
\begin{equation}
w_t=\operatorname{median}\{d^{\mathrm{stored}}_{t,h}:h\in H(x),\ d^{\mathrm{stored}}_{t,h}\ \mathrm{usable}\}.
\end{equation}
If human references usually take longer on a test, that test is treated as harder. The slope-with-intercept metric asks whether candidate runtime grows slowly or quickly as this human-derived difficulty increases, and fits
\begin{equation}
(\alpha_y,\beta_y)=\arg\min_{\alpha,\beta}
\sum_{t\in\mathcal T}\left(\delta_t(y)-\alpha-\beta w_t\right)^2,
\end{equation}
then ranks \(\beta_y\) against the corresponding reference slopes. The zero-intercept variant fits
\begin{equation}
\beta_y^{0}=\arg\min_{\beta}
\sum_{t\in\mathcal T}\left(\delta_t(y)-\beta w_t\right)^2,
\end{equation}
then ranks \(\beta_y^{0}\). Lower slopes correspond to candidates whose duration grows less quickly on tests that are slower for the reference pool. These scores therefore add a second modeling choice on top of the ranking itself: the scalar no longer depends only on where the candidate stands on each test, but also on how well a single linear scaling model summarizes its duration curve.

Finally, some metrics discard most of the magnitude information. Mean win rate and median win rate turn each test into a binary comparison against a reference summary,
\begin{equation}
W_{\mathrm{mean}}(y)=\frac{1}{|\mathcal T|}\sum_{t\in\mathcal T}\mathbb{I}\{\delta_t(y)<\mu_t\},
\qquad
W_{\mathrm{median}}(y)=\frac{1}{|\mathcal T|}\sum_{t\in\mathcal T}\mathbb{I}\{\delta_t(y)<m_t\},
\end{equation}
where \(\mu_t\) and \(m_t\) are the mean and median of \(D_t\). These are higher-is-better. The filtered percentile variants sort tests by reference difficulty, using the mean reference duration on that test for the duration source being evaluated, then recompute either the mean or median percentile after dropping the easiest 10\% of tests, the hardest 25\%, or the hardest 50\%. These six variants ask whether the easiest or hardest tests should be ignored before aggregation. As with trimming, at least one test is removed when the filtered score is defined, so we only score these variants when enough tests remain.

So in total we have 17 metrics to be screened: four direct percentile aggregations (mean, median, trimmed mean, log-space mean), two top-\(k\) shares, one total-duration percentile, two slope percentiles, two win rates, and six filtered percentile variants. We screened them along three axes. First, the score should not be too easy or too harsh: if almost every candidate lands in the worst tier, it creates a sparse reward, while a score that gives many candidates good values can remove optimization pressure. Second, it should use a meaningful dynamic range within each problem, otherwise correct solutions that differ in speed still look similar to the reward. Third, it should be stable under repeated CES measurements and after duration calibration, because calibration should improve agreement with fresh timings rather than introduce deformations of the ranking scale.

\begin{figure*}[p]
\centering
\includegraphics[width=\textwidth,height=0.94\textheight,keepaspectratio]{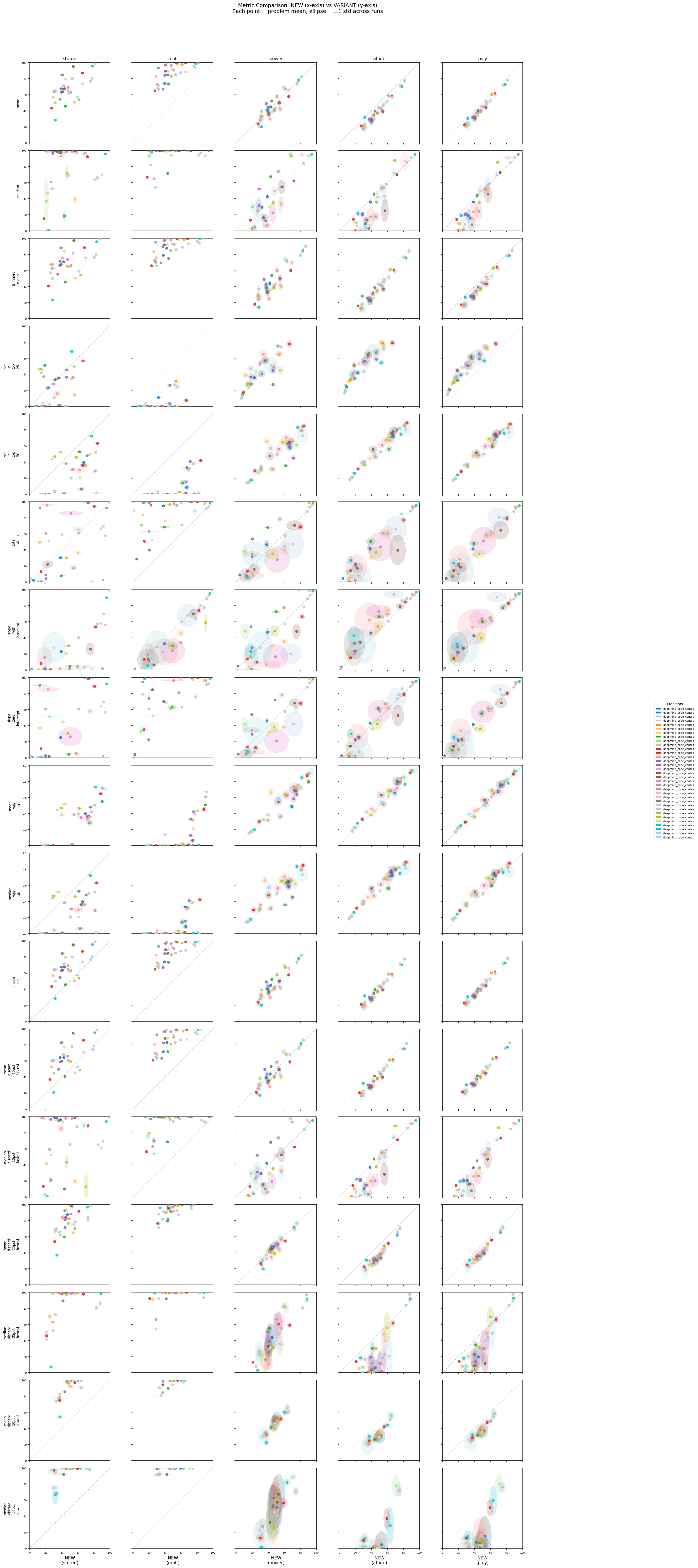}
\caption{\textbf{Cross-validation recap of ranking metrics.}
Each dot is one problem averaged over repeated runs; ellipses show one standard deviation on each axis. We apologize for the small labels: the figure is kept tiny so the full grid fits on one page, and most explanations are left in the text.}
\label{fig:ranking_metric_cv_recap}
\end{figure*}

The full-page grid in \cref{fig:ranking_metric_cv_recap} visualizes this screening. Rows are candidate metrics and columns are duration sources or fitted calibration variants, evaluated under leave-one-problem-out cross-validation. In one panel, a dot is one problem: its \(x\)-coordinate is the mean, over repeated runs, of the problem-level metric computed from fresh CES durations, and its \(y\)-coordinate is the same metric computed from the stored or calibrated durations for the column. The ellipse around the dot shows one standard deviation across repeated runs on the \(x\)- and \(y\)-axes. A good panel therefore has dots close to the diagonal, small ellipses, and a spread that uses a meaningful part of the axis range; this means the metric agrees with fresh timings, is stable across reruns, and does not collapse to a narrow score band. A bad panel bends away from the diagonal, has large ellipses, or collapses most dots into a narrow band or an extreme corner. The grid makes these failure modes visible at once: some metrics are unstable across reruns, while others are stable but occupy only a narrow part of the score range.

Reading \cref{fig:ranking_metric_cv_recap} with this rule, the candidate metrics fail for different reasons. Win-rate metrics occupy a useful range of values under the affine calibration that we retain, but some problems have wide ellipses, so repeated CES runs can move the score substantially. Median percentile and median-filtered variants show even more rerun noise, and the calibrated score tends to over-reward candidates relative to fresh timings. Top-\(k\) shares also have substantial noise, but tend to under-reward candidates under calibration. The slope-with-intercept metric has very large rerun variability, making the fitted slope too sensitive to individual timing measurements. Total-duration percentile also varies strongly across reruns, although its calibrated values are better aligned with fresh-duration values than the slope scores. These patterns point to the mean-based percentile family, which keeps a useful dynamic range while reducing rerun variability and avoiding the visible over- or under-rewarding patterns of other families. We also inspect the calibration variants that are not retained, because the selected affine correction may not be the exact underlying recalibration model; a metric is more attractive when it behaves similarly under these alternative duration-correction scenarios.

\Cref{tab:ranking_metric_finalists} zooms into the mean-based ranking metrics. It compares plain mean percentile to trimmed and filtered variants, which test whether dropping extreme or difficult tests improves stability enough to justify an extra design choice. The cross-validation columns measure alignment to fresh CES timings after fitting the duration correction on other problems; the within-problem column measures rerun noise for the same problem.

\begin{table}[h]
\centering
\small
\caption{\textbf{Stability of mean-based percentile variants.}
The two cross-validation columns report, under leave-one-problem-out fitting of the affine or polynomial duration correction, the standard deviation of the difference between the metric computed from corrected stored durations and the same metric computed from fresh CES durations. The within-problem column reports rerun noise: for each problem we compute the standard deviation of the metric across repeated CES reruns, then report the mean and standard deviation of that quantity across problems. Values are percentile points; lower is better.}
\label{tab:ranking_metric_finalists}
\begin{tabular}{lccc}
\toprule
\bfseries Metric & \bfseries CV diff std (affine) & \bfseries CV diff std (poly) & \bfseries Within-problem std \\
\midrule
Mean percentile & \textbf{3.8} & \textbf{2.9} & \(3.0 \pm 1.0\) \\
Trimmed mean percentile & 5.1 & 4.0 & \(3.5 \pm 1.2\) \\
Mean percentile, discard fastest 10\% & 4.2 & 3.3 & \(2.8 \pm 1.0\) \\
Mean percentile, discard slowest 25\% & 4.2 & 3.3 & \(3.7 \pm 1.1\) \\
Mean percentile, discard slowest 50\% & 4.8 & 4.0 & \(5.3 \pm 1.7\) \\
\bottomrule
\end{tabular}
\end{table}

\begin{figure}[h]
\centering
\input{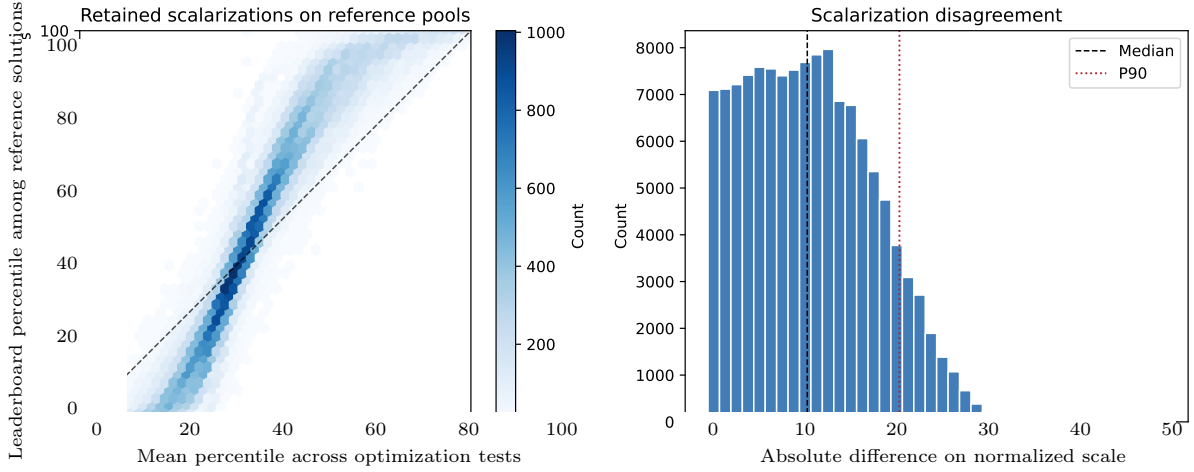}
\caption{\textbf{Comparison of the two post-execution scalarizations selected for RL on the training split.}
Left: comparison between the mean percentile across optimization tests \(a(y)\) and the leaderboard percentile \(q_{\mathrm{QP}}\).
Right: histogram of \(|a(y)-q_{\mathrm{QP}}|\).
The full training split contains 119.4 optimization tests per problem on average (median 135), a median of 66 ranked references, and a median leaderboard-percentile resolution of 1.54 points.
Across 127{,}841 reference solutions from 1{,}000 training problems, the two signals have Spearman \(\rho=0.9696\) and Pearson \(r=0.9407\), but their mean absolute difference is 12.7 points, with 90th and 95th percentile gaps of 23.6 and 26.6 points.
The two scalarizations preserve nearly the same ordering but can move solutions across reward buckets, so we try both for RL.}
\label{fig:quality_signal_comparison}
\end{figure}

Mean percentile is retained because it has the lowest cross-validation dispersion, comparable rerun noise to filtered variants, and a 47-point spread between strong and weak solutions without adding trimming hyperparameters. We then keep two ways of turning this aggregate into the post-execution signal used by the RL ablations. QAR ranked quality uses the mean per-test percentile itself, so \(q_{\mathrm{QAR}}\) keeps the average location of the candidate across tests. QP ranked quality first sorts references and the candidate by that same mean-percentile statistic, then reports the candidate's problem-level leaderboard percentile. \Cref{fig:quality_signal_comparison} shows this is mostly a rotation of the same ranking information: we keep both because one may have too much or too little dynamic range, or saturate too early at the extreme, and the online RL sweep can retain the scalarization that works best.

\subsection{Composing rewards that best balance correctness and optimization objectives}
\label{sec:app_reward_composition}

Once an environment is reduced to \((\tilde c_{\mathrm{cor}},g,q)\), the reward ablation is a composition rule. The design question is how to balance the correctness objective and the optimization objective. A strict outer correctness gate makes efficiency credit conditional on a valid program, while softer blends or multitask mixtures can expose some optimization signal even when strict correctness is not yet satisfied. This can be harmful if the model learns fast but wrong programs, or useful if it gives early optimization feedback before the model reaches high pass rates. We use the normalized interface \(q\in[0,1]\), where \(q=0\) is best. The signal \(c_{\mathrm{cor}}\) is the correctness-test pass indicator, while \(\tilde c_{\mathrm{cor}}\) is the strict prerequisite from \cref{sec:app_env_qwen7b_definitions}: correctness tests must pass, and executed optimization tests may be successes or timeouts but not failures. The binary optimization signal is \(g\in\{0,1\}\), either a timeout-tolerance gate or a ranked threshold. For the rewards that use \(q\), we first apply one of the following two transformations:
\begin{align}
m_{\mathrm{bucket}}(q)&=
\begin{cases}
1, & q=0,\\
\max\{0,\,0.9-0.1\lfloor 10(q-10^{-6})\rfloor\}, & q>0,
\end{cases}
&
m_{\mathrm{linear}}(q)&=1-q,
\end{align}
where \(m_{\mathrm{bucket}}\) is a ten-bin map and \(m_{\mathrm{linear}}\) is the continuous variant. The idea behind bucketing is that discretizing nearby quality values may smooth out part of the timing noise before it reaches the reward.
We also use range-compressed maps that keep the same ordering but change the reward gap to potentially give more weight to the \(q\) signal:
\begin{equation}
m_{[\ell,1]}(q)=\ell+(1-\ell)(1-q),
\qquad \ell\in\{1/3,1/2\}.
\end{equation}
All signals in \([0,1]\), whether \(c_{\mathrm{cor}}\), \(\tilde c_{\mathrm{cor}}\), \(g\), or \(q\), are converted to signed reward units with
\begin{equation}
m_{\mathrm{signed}}(z)=2z-1.
\end{equation}
Thus \(m_{\mathrm{signed}}(c_{\mathrm{cor}})=2c_{\mathrm{cor}}-1\), and analogously for \(\tilde c_{\mathrm{cor}}\), \(g\), and \(m(q)\). For simplicity, the rest of this subsection, including \cref{tab:reward_composition_formulas}, writes \(\tilde c_{\mathrm{cor}}\), \(c_{\mathrm{cor}}\), \(g\), and \(q\) for these signed reward components; for \(q\), this means we write \(q\) instead of repeatedly writing \(m(q)\) after the selected quality map and signed conversion have been applied.
\Cref{tab:reward_composition_formulas} separates the six ways these signals are mixed: correctness-only, optimization-only, collapsed, two-gate, additive blending, and multitask. Collapsed rewards make correctness and optimization one conjunction. Two-gate rewards first check \(\tilde c_{\mathrm{cor}}\), then expose either the binary optimization gate or the graded quality score. The reward we call quality percentile is a graded two-gate reward in which \(q\) is computed from the mean-percentile aggregation of recorded durations described in \cref{sec:app_ranking_metrics}; it is not a separate composition family. Blended rewards add signed component rewards, and multitask variants alternate separate correctness and optimization rollouts instead of forming a single scalar inside one rollout. We investigate these reward forms on top of two main optimization RL environments. In the duration-filtered optimization environment, \(T^{\mathrm O}_{\mathrm{used}}=F_{\mathrm{abs}}^{a}(T^{\mathrm O})\) with \(a=2\,\mathrm{s}\), and the optimization timeout is 0.5s; binary versions use \(g=g_{\mathrm{to}}^{0.1}\), while graded versions use the timeout fraction \(q=q_{\mathrm{to}}\), with graded standing for continuous or bucketed rewards. In the ranked optimization environment, optimization tests are all kept and executed under a 10s time limit; binary versions use \(g=g_{\mathrm{dur}}^{0.30}\), while graded versions use the selected ranked scalar \(q\in\{q_{\mathrm{QAR}},q_{\mathrm{QP}}\}\).

\begin{table}[h]
\centering
\small
\caption{\textbf{Properties of the graded reward functions.}
The passing range is the image of the quality map \(m\) on \(q\in[0,1]\).
The correctness--failure gap is \(m(1)-(-1)\), and the final column compares that gap with the optimization range \(m(0)-m(1)\).}
\label{tab:reward_mapping_properties}
\begin{tabular}{lcccc}
\toprule
\bfseries Mapping & \bfseries Passing range & \bfseries \(m(0.5)\) & \bfseries Corr.--fail.\ gap & \bfseries Gap : range \\
\midrule
Bucketed & \([0,1]\) & 0.5 & 1.0 & \(1{:}1\) \\
Linear & \([0,1]\) & 0.5 & 1.0 & \(1{:}1\) \\
Range \([1/3,1]\) & \([1/3,1]\) & 0.67 & 1.33 & \(2{:}1\) \\
Range \([1/2,1]\) & \([1/2,1]\) & 0.75 & 1.50 & \(3{:}1\) \\
\bottomrule
\end{tabular}
\end{table}

\begin{table*}[t!]
\centering
\scriptsize
\setlength{\tabcolsep}{2.5pt}
\renewcommand{\arraystretch}{1.32}
\newcommand{\rcell}[1]{\begin{minipage}[t]{\linewidth}\vspace*{0.16em}\raggedright #1\par\vspace*{0.24em}\end{minipage}}
\newcommand{\fcell}[1]{\begin{minipage}[t]{\linewidth}\vspace*{0.28em}#1\par\vspace*{0.18em}\end{minipage}}
\caption{\textbf{Reward-composition families used in the ablation grid.}
All reward families consume the same environment outputs \((\tilde c_{\mathrm{cor}},g,q)\), but they differ in where correctness enters the scalar reward and whether optimization-side credit can appear when strict correctness is not satisfied. The table writes \(\tilde c_{\mathrm{cor}}\), \(c_{\mathrm{cor}}\), \(g\), and \(q\) as signed reward components in \([-1,1]\); in graded rows, \(q\) is the shorthand for the selected mapped-and-signed quality component. Controls isolate one side of the signal, collapsed and two-gate rewards enforce correctness before efficiency credit, while blended and multitask rewards test softer ways of balancing the two objectives.}
\label{tab:reward_composition_formulas}
\begin{tabular}{@{}p{0.11\textwidth}@{\hspace{2pt}{\color{black!25}\vrule width 0.2pt}\hspace{2pt}}p{0.35\textwidth}@{\hspace{2pt}{\color{black!25}\vrule width 0.2pt}\hspace{2pt}}p{0.09\textwidth}@{\hspace{2pt}{\color{black!25}\vrule width 0.2pt}\hspace{2pt}}p{0.11\textwidth}@{\hspace{2pt}{\color{black!25}\vrule width 0.2pt}\hspace{2pt}}p{0.22\textwidth}@{}}
\toprule
\rcell{{\bfseries Family}} & \rcell{{\bfseries Reward form}} & \rcell{{\bfseries Correctness role}} & \rcell{{\bfseries Speed credit if wrong?}} & \rcell{{\bfseries Role in the sweep}} \\
\midrule
\rcell{Correctness only} &
\fcell{\(R_{\mathrm{corr}}=c_{\mathrm{cor}}\)} &
\rcell{\(c_{\mathrm{cor}}\) only} &
\rcell{No speed term} &
\rcell{Standard RLVR control: optimization tests may be present as correctness tests, but no timing quality is exposed.} \\
\addlinespace[0.08em]\midrule[0.15pt]\addlinespace[0.08em]
\rcell{Optimization only} &
\fcell{\(\begin{aligned}[t]
R_{\mathrm{opt-bin}}&=g\\[0.18em]
R_{\mathrm{opt-gr}}&=q
\end{aligned}\)} &
\rcell{None} &
\rcell{Yes} &
\rcell{Control for pure duration-side learning; it tests whether optimization pressure alone can sustain code generation.} \\
\addlinespace[0.08em]\midrule[0.15pt]\addlinespace[0.08em]
\rcell{Collapsed} &
\fcell{\(\begin{aligned}[t]
R_{\mathrm{coll-bin}}&=\dfrac{(1+\tilde c_{\mathrm{cor}})(1+g)}{2}-1\\[0.32em]
R_{\mathrm{coll-gr}}&=\dfrac{(1+\tilde c_{\mathrm{cor}})(1+q)}{2}-1
\end{aligned}\)} &
\rcell{Product gate} &
\rcell{No} &
\rcell{Hard conjunction of correctness and efficiency. It withholds optimization credit unless strict correctness holds, but a correct slow program can collapse to the failure reward.} \\
\addlinespace[0.08em]\midrule[0.15pt]\addlinespace[0.08em]
\rcell{Two-gate} &
\fcell{\(\begin{aligned}[t]
R_{\mathrm{2g-bin}}&=\dfrac{(1+\tilde c_{\mathrm{cor}})(1+g)}{4}-\dfrac{1-\tilde c_{\mathrm{cor}}}{2}\\[0.32em]
R_{\mathrm{2g-gr}}&=\dfrac{(1+\tilde c_{\mathrm{cor}})(1+q)}{4}-\dfrac{1-\tilde c_{\mathrm{cor}}}{2}
\end{aligned}\)} &
\rcell{Outer gate} &
\rcell{No} &
\rcell{Correctness is checked first, then binary or graded optimization quality separates correct solutions. Quality percentile is the graded instance using mean-percentile aggregation for \(q\).} \\
\addlinespace[0.08em]\midrule[0.15pt]\addlinespace[0.08em]
\rcell{Additive blend} &
\fcell{\(\begin{aligned}[t]
R_{\mathrm{blend-bin}}&=\lambda c_{\mathrm{cor}}+(1-\lambda)g\\[0.18em]
R_{\mathrm{blend-gr}}&=\lambda c_{\mathrm{cor}}+(1-\lambda)q
\end{aligned}\)} &
\rcell{Soft sum} &
\rcell{Yes} &
\rcell{Tests whether weighted sums provide a better balance. The primary \(1{:}1\) cells use \(\lambda=1/2\); \(2{:}1\) and \(1{:}2\) variants use \(2/3\) and \(1/3\).} \\
\addlinespace[0.08em]\midrule[0.15pt]\addlinespace[0.08em]
\rcell{Multitask} &
\rcell{Sample \(R_{\mathrm{corr}}\) or \(R_{\mathrm{opt-bin/gr}}\) from separate task streams} &
\rcell{Stream-separated} &
\rcell{Yes in opt.-only rollouts} &
\rcell{Tests whether separating the two objectives across rollouts is easier than combining both signals inside one scalar. Primary variants use a \(1{:}1\) task mixture.} \\
\bottomrule
\end{tabular}
\end{table*}

Depending on the reward form, the correctness gate consumed by the scalar can be either \(c_{\mathrm{cor}}\) or \(\tilde c_{\mathrm{cor}}\). Collapsed and two-gate rewards, including quality percentile, require correctness tests to pass and executed optimization tests to avoid failures before giving any efficiency credit. Timeouts may be tolerated only through the environment's explicit timeout-tolerance rule, while failures stay at \(-1\). Optimization-only, blend, and multitask variants relax this coupling in different ways, testing whether some optimization-side feedback before strict correctness improves or hurts online learning.

\begin{figure*}[t!]
    \centering
    \includegraphics[width=\textwidth]{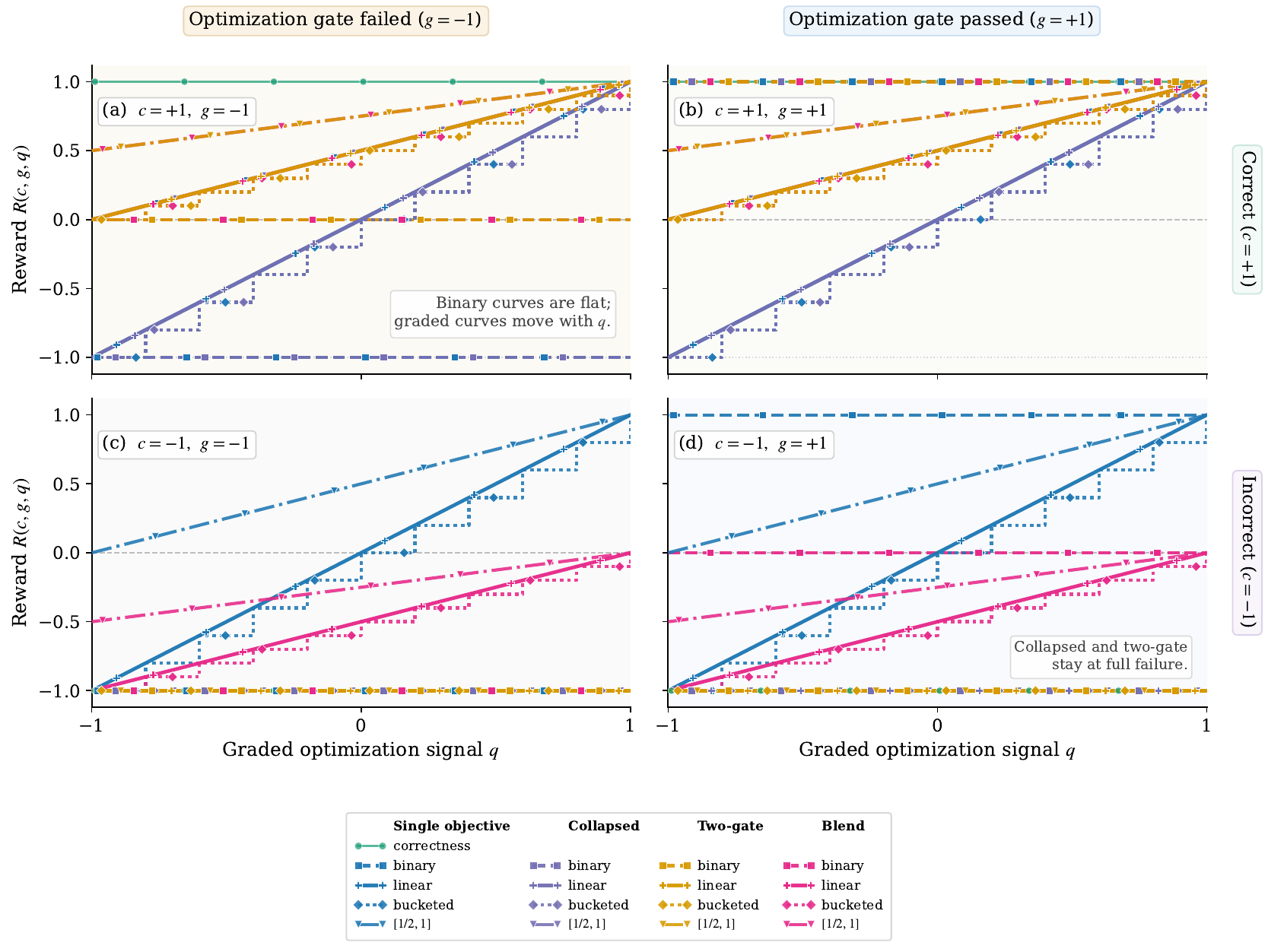}
    \caption{\textbf{Reward composition families in the shared signed \((c,g,q)\) interface.}
    For this visualization only, we set \(c=c_{\mathrm{cor}}=\tilde c_{\mathrm{cor}}\) and use signed components \(c,g,q\in[-1,1]\).
    Rows fix correctness \(c\), columns fix the hard optimization gate \(g\), and the horizontal axis varies the graded optimization signal \(q\) on the signed reward scale, from worst (\(-1\)) to best (\(+1\)).
    The legend groups the curves into four columns: single-objective controls, collapsed rewards, two-gate rewards, and additive blending.
    The single-objective column contains the correctness-only control and the optimization-only binary/graded controls; in the other columns, dashed lines are binary gates, solid lines are continuous graded maps, dotted step lines are bucketed maps, and dash-dotted lines use the range-compressed \(m_{[1/2,1]}\) quality map.
    Multitask is omitted because it mixes separate rollout streams rather than defining one scalar curve.
    Collapsed and two-gate rewards stay at \(-1\) when \(c=-1\), while optimization-only and blended rewards show where optimization-side signal can remain outside a strict correctness gate.}
    \label{fig:reward_composition}
\end{figure*}

All graded families keep the global reward range in \([-1,1]\) after the signed conversion used by the corresponding composition rule. The linear map sends passing solutions to \(1-q\), and the bucketed map discretizes the same interval into ten bins. The range-compressed maps \(m_{[1/3,1]}\) and \(m_{[1/2,1]}\) keep the same ordering but raise the reward assigned to the worst passing solution, so that the failure-to-passing gap is larger than the optimization range among passing solutions. These mappings ablate the relative scale of correctness and optimization for a fixed composition rule, after the reward has already defined how the correctness and optimization signals interact.

\FloatBarrier

\subsection{Raw-durations (aka naive) reward baselines}
\label{sec:app_naive_duration_rewards}

The naive-duration baselines test the most direct way we could think of as a reward for code optimization: add raw seconds directly to the correctness reward. Let \(\bar d_{\mathrm{base}}\) be the mean duration on the original public, private, and generated correctness tests, excluding the additional correctness tests, and let \(\bar d_{\mathrm{opt}}\) be the mean duration on optimization tests. Timeouts, missing durations, and non-positive durations are replaced by the effective time limit for that test source before averaging, most of the time 10s; if a source has no tests, the same effective limit is used. The four baselines choose \(s\in\{\mathrm{base},\mathrm{opt}\}\) and one of two maps,
\begin{align}
m_{\mathrm{lin}}(\bar d_s)&=\operatorname{clip}\left(1-\frac{\bar d_s}{10},0,1\right),\\
m_{\log}(\bar d_s)&=\operatorname{clip}\left(\frac{1-\log_{10}(\max\{\bar d_s,10^{-6}\})}{3},0,1\right),
\end{align}
then set
\begin{equation}
R_{\mathrm{naive}}(x,y)=c_{\mathrm{cor}}(x,y)+m(\bar d_s(x,y))-1.
\end{equation}
Thus a correct 10s average solution receives \(0\), a correct 1s linear solution receives \(0.9\), and an incorrect but very fast solution can receive a reward close to \(0\) rather than \(-1\). This is the behavior the reference-normalized families avoid. The idea behind trying a log map is that, for many problems, the dynamic range of possible average durations may be concentrated near the left of the \([0\,\mathrm{s},10\,\mathrm{s}]\) range.

\begin{table}[h!]
\centering
\small
\caption{\textbf{Naive-duration baselines.}
The variants differ only in which tests provide the raw average duration and how that duration is mapped to \([0,1]\). None compares the candidate to per-problem human references.}
\label{tab:naive_duration_definitions}
\begin{tabular}{lll}
\toprule
\bfseries Variant & \bfseries Duration source & \bfseries Duration map \\
\midrule
Base-linear & \(\bar d_{\mathrm{base}}\) & \(m_{\mathrm{lin}}\) \\
Base-log & \(\bar d_{\mathrm{base}}\) & \(m_{\log}\) \\
Optimization-linear & \(\bar d_{\mathrm{opt}}\) & \(m_{\mathrm{lin}}\) \\
Optimization-log & \(\bar d_{\mathrm{opt}}\) & \(m_{\log}\) \\
\bottomrule
\end{tabular}
\end{table}

These baselines obviously suffer from the fact that they attribute absolute timing rewards whatever the underlying problem is. A 0.1s program receives nearly the same raw-duration credit whether the problem's reference solutions run in 0.05s or 5s, so the reward is not normalized by problem difficulty or by the attainable speed distribution. A big advantage remains that such a reward does not need any reference distribution or extra data to be calibrated: it can be computed directly from the executed input-output tests. The corresponding training results are summarized in \cref{sec:training}.

\subsection{Taxonomy of optimization/efficiency definitions in previously published papers}
\label{sec:app_related_efficiency_taxonomy}

The taxonomy built above tries to cover the different works already published around code optimization: they differ by how they define the optimization problem, and whether they train for it, evaluate on it, or both. A benchmark can construct hard inputs before execution, enforce time limits during execution, or rank a candidate after execution, though this does not necessarily make it an online RL environment. Similarly, an inference-time search system may use execution feedback as a fitness score without updating the model policy. \Cref{tab:app_adjacent_efficiency_taxonomy_benchmarks,tab:app_adjacent_efficiency_taxonomy_training_search} therefore go over previously published studies on topics close to our work, and describe what they mainly do in light of the environment taxonomy and reward families defined above. The comparison remains high-level. Some systems use memory, instruction counts, complexity classes, or task-specific evaluator scores rather than wall-clock time, and small implementation choices can matter a lot.

\newcommand{\taxcell}[1]{\begin{minipage}[t]{\linewidth}\vspace*{0.12em}\raggedright #1\par\vspace*{0.16em}\end{minipage}}
\newcommand{\envtag}[1]{{\bfseries\underline{#1}}}

\begin{table*}[p]
\centering
\fontsize{6.5}{7.5}\selectfont
\setlength{\tabcolsep}{2.3pt}
\renewcommand{\arraystretch}{1.0}
\caption{\textbf{Benchmark papers adjacent to our environment taxonomy.}
Here we classify benchmark work by where efficiency enters the evaluation routine. \envtag{Pre-execution} refers to workload or test construction before the candidate is run; \envtag{intra-execution} refers to time limits or right-censoring that are part of the scoring rule; and \envtag{post-execution} refers to ranking or scoring a candidate after running it.}
\label{tab:app_adjacent_efficiency_taxonomy_benchmarks}
\begin{tabular}{p{0.115\textwidth}p{0.095\textwidth}p{0.255\textwidth}p{0.255\textwidth}p{0.235\textwidth}}
\toprule
\taxcell{{\bfseries Paper}} & \taxcell{{\bfseries Kind}} & \taxcell{{\bfseries What it studies}} & \taxcell{{\bfseries Fit to our environment taxonomy}} & \taxcell{{\bfseries Training or reward analogue}} \\
\midrule
\taxcell{\textsc{Mercury}\\ \citep{du2024mercury}} &
\taxcell{Benchmark + SFT/DPO} &
\taxcell{Python efficiency benchmark with human/runtime reference distributions and the \textsc{Beyond} percentile metric.} &
\taxcell{Mostly \envtag{post-execution} reference-runtime percentile scoring; timeout is used as an execution guard rather than a learned timing environment.} &
\taxcell{SFT on fast solutions and DPO over faster/slower correct solutions. The benchmark metric is correctness-gated, but the training signal is pairwise preference rather than one of our scalar reward compositions.} \\
\addlinespace[0.12em]
\taxcell{\textsc{EffiBench}\\ \citep{huang2024effibench}} &
\taxcell{Benchmark} &
\taxcell{LeetCode-style benchmark measuring execution time and memory of generated Python code against canonical efficient solutions.} &
\taxcell{\envtag{Post-execution} normalized runtime and memory scoring; generated tests are benchmark construction, not train-time \envtag{pre-execution} filtering.} &
\taxcell{No training reward.} \\
\addlinespace[0.12em]
\taxcell{\textsc{EffiBench-X}\\ \citep{qing2025effibenchx}} &
\taxcell{Benchmark} &
\taxcell{Multi-language extension of \textsc{EffiBench} with expert references and relative execution-time and memory metrics.} &
\taxcell{\envtag{Post-execution} expert-reference-normalized time and memory scoring; translation and test generation happen before evaluation.} &
\taxcell{No training reward.} \\
\addlinespace[0.12em]
\taxcell{\textsc{EvalPerf}\\ \citep{liu2024evalperf}} &
\taxcell{Benchmark} &
\taxcell{Creates performance-exercising inputs and reference-solution clusters, then evaluates candidates with profiled execution time.} &
\taxcell{\envtag{Pre-execution} benchmark construction for hard inputs, followed by \envtag{post-execution} runtime ranking against reference clusters.} &
\taxcell{No training reward; the benchmark improves input selection and runtime comparison, but remains an offline evaluation protocol rather than an online reward.} \\
\addlinespace[0.12em]
\taxcell{\textsc{ENAMEL}\\ \citep{qiu2025enamel}} &
\taxcell{Benchmark} &
\taxcell{HumanEval-derived benchmark with expert algorithms, level-based strong inputs, right-censored execution, and \(\mathrm{eff}@k\).} &
\taxcell{Mixed \envtag{intra-execution} and \envtag{post-execution} design: reference-derived time limits and right-censoring are part of scoring, then candidates receive normalized efficiency scores.} &
\taxcell{No training reward.} \\
\addlinespace[0.12em]
\taxcell{\textsc{COFFE}\\ \citep{peng2025coffe}} &
\taxcell{Benchmark} &
\taxcell{Function- and file-level code-efficiency benchmark using stress tests and stable CPU instruction-count measurements.} &
\taxcell{\envtag{Pre-execution} stress-test generation and selection, then \envtag{post-execution} correctness-gated instruction-count scoring.} &
\taxcell{No training reward.} \\
\addlinespace[0.12em]
\taxcell{\textsc{SWE-Perf}\\ \citep{he2025sweperf}} &
\taxcell{Benchmark} &
\taxcell{Repository-level benchmark from real performance-improving pull requests, with performance tests and expert patches.} &
\taxcell{\envtag{Post-execution} repository workload timing with a correctness gate and statistical performance-gain check; target functions and tests are benchmark curation.} &
\taxcell{No training reward.} \\
\addlinespace[0.12em]
\taxcell{\textsc{SWE-fficiency}\\ \citep{ma2025swefficiency}} &
\taxcell{Benchmark} &
\taxcell{Repository-level workload benchmark where models patch real Python projects and are scored by captured expert speedup.} &
\taxcell{\envtag{Post-execution} workload timing normalized to expert speedup, with correctness and anti-cheating checks.} &
\taxcell{No training reward.} \\
\addlinespace[0.12em]
\taxcell{\textsc{BigO(Bench)}\\ \citep{chambon2025bigobench}} &
\taxcell{Benchmark} &
\taxcell{Evaluates whether models can generate code with controlled time and space complexity through scaled execution and curve fitting.} &
\taxcell{\envtag{Post-execution} empirical complexity inference and ranking; complexity constraints may be requested in the prompt, but the evidence comes from measured growth curves.} &
\taxcell{No training reward.} \\
\bottomrule
\end{tabular}
\end{table*}

\begin{table*}[p]
\centering
\fontsize{6.5}{7.5}\selectfont
\setlength{\tabcolsep}{2.3pt}
\renewcommand{\arraystretch}{1.0}
\caption{\textbf{Training and inference-time search papers.}
The table separates online policy training from supervised training, preference optimization, and inference-time search. The taxonomy column highlights whether the execution feedback is mainly \envtag{pre-execution}, \envtag{intra-execution}, or \envtag{post-execution}; the final column describes the closest reward analogue.}
\label{tab:app_adjacent_efficiency_taxonomy_training_search}
\begin{tabular}{p{0.115\textwidth}p{0.095\textwidth}p{0.255\textwidth}p{0.255\textwidth}p{0.235\textwidth}}
\toprule
\taxcell{{\bfseries Paper}} & \taxcell{{\bfseries Kind}} & \taxcell{{\bfseries What it studies}} & \taxcell{{\bfseries Fit to our environment taxonomy}} & \taxcell{{\bfseries Training or reward analogue}} \\
\midrule
\taxcell{Performance-Aligned LLMs\\ \citep{nichols2024performancealigned}} &
\taxcell{Training} &
\taxcell{Uses supervised data, preference data, reward modeling, and PPO/RLPF-style optimization to generate faster code.} &
\taxcell{\envtag{Post-execution} runtime measurement on executable coding tasks; learned reward modeling is used when direct execution is not available.} &
\taxcell{For executable tasks, a correctness-gated continuous speedup reward: incorrect code gets a failure reward, correct faster code receives larger reward. This is close to gated continuous \(q\), but without a separate hard optimization gate \(g\).} \\
\addlinespace[0.12em]
\taxcell{\textsc{SuperCoder}\\ \citep{wei2025supercoder}} &
\taxcell{Benchmark + RL} &
\taxcell{Assembly superoptimization against \texttt{gcc -O3}, with RL improving both correctness and speedup.} &
\taxcell{\envtag{Post-execution} compilation, testing, and speedup measurement for generated assembly.} &
\taxcell{Online RL uses a correctness-gated continuous reward: zero unless all tests pass, then speedup. This is closest to a two-gate-style reward with \(q\) as speedup, but no explicit ranked gate.} \\
\addlinespace[0.12em]
\taxcell{HPC online RL\\ \citep{mikasa2026hpc}} &
\taxcell{Training} &
\taxcell{Online GRPO for matrix multiplication on real machines, using measured GFLOPS as performance feedback.} &
\taxcell{\envtag{Post-execution} real-machine performance reward; staged restrictions on allowed optimization techniques are a curriculum-like \envtag{pre-execution} constraint.} &
\taxcell{Status-penalty ladder for format, compile, timeout, runtime, and verification failures, with continuous GFLOPS on success. This is not a strict two-gate or collapsed reward; it is closer to ordinal failure penalties plus continuous success quality.} \\
\addlinespace[0.12em]
\taxcell{\textsc{Afterburner}\\ \citep{du2025afterburner}} &
\taxcell{Training / refinement} &
\taxcell{Iterative code-efficiency refinement with SFT, DPO, and GRPO using runtime, memory, and integral feedback.} &
\taxcell{\envtag{Post-execution} feedback in a refinement environment: the model edits an existing solution after observing prior metrics, rather than solving one-shot from scratch.} &
\taxcell{GRPO uses an additive blend of format, correctness, and efficiency rewards. The efficiency term is gated by output correctness, but the final scalar is still a weighted blend rather than a hard correctness-preserving two-gate reward.} \\
\addlinespace[0.12em]
\taxcell{\textsc{EffiCoder}\\ \citep{huang2025efficoder}} &
\taxcell{SFT + preference ablations} &
\taxcell{Builds efficient-code fine-tuning data by selecting candidates with good execution time and memory.} &
\taxcell{Offline \envtag{post-execution} selection and curation of efficient correct code.} &
\taxcell{No online RL reward; the closest analogues are supervised learning on fastest or lowest-memory correct samples and offline preference objectives over efficient candidates.} \\
\addlinespace[0.12em]
\taxcell{\textsc{PIE}\\ \citep{shypula2024pie}} &
\taxcell{Dataset + SFT} &
\taxcell{Slow-fast C++ edit dataset with measured performance-improving code edits.} &
\taxcell{\envtag{Post-execution} measured edit improvement in a code-to-code setting.} &
\taxcell{Supervised and edit-pair training rather than scalar RL. Evaluation is correctness-gated speedup-like, but the model is not trained with our reward families.} \\
\addlinespace[0.12em]
\taxcell{\textsc{PerfCodeGen}\\ \citep{peng2025perfcodegen}} &
\taxcell{Search} &
\taxcell{Inference-time iterative refinement using execution and profiling feedback to find faster generated code.} &
\taxcell{\envtag{Post-execution} profiling, repair, and fastest-correct selection inside an inference-time loop.} &
\taxcell{No policy-update reward; the search objective resembles correctness-gated runtime feedback.} \\
\addlinespace[0.12em]
\taxcell{\textsc{SBLLM}\\ \citep{gao2025sbllm}} &
\taxcell{Search} &
\taxcell{Search-based LLM optimization with iterative candidate scoring, retrieval, and revision.} &
\taxcell{\envtag{Post-execution} execution-based ranking inside search; retrieval and prompting are search machinery rather than an RL environment operator.} &
\taxcell{No train-time scalar reward for the model policy; the search fitness is closest to correctness-gated speedup or ranking.} \\
\addlinespace[0.12em]
\taxcell{\textsc{MaxCode}\\ \citep{ou2025maxcode}} &
\taxcell{Search + value model} &
\taxcell{Frames code optimization as max-reward inference-time search with execution feedback, critique, and a reward-to-go model.} &
\taxcell{\envtag{Post-execution} iterative search over execution feedback with a frozen code-proposal policy.} &
\taxcell{Uses bucketed speedup labels for search/value guidance. This is adjacent to a correctness-gated bucketed quality signal, but it is not online RL on the code-generation policy with our scalar reward.} \\
\addlinespace[0.12em]
\taxcell{\textsc{FunSearch}\\ \citep{romeraparedes2023funsearch}} &
\taxcell{Search / evolution} &
\taxcell{LLM-guided evolutionary program search for mathematical and combinatorial objectives.} &
\taxcell{Mostly outside runtime-efficiency training: candidates receive \envtag{post-execution} task-specific evaluator scores, and timeouts are guardrails.} &
\taxcell{No policy-training reward; selection uses task-specific evaluator scores rather than code-runtime reward composition.} \\
\addlinespace[0.12em]
\taxcell{\textsc{AlphaEvolve}\\ \citep{novikov2025alphaevolve}} &
\taxcell{Search / evolution} &
\taxcell{General evolutionary coding agent for scientific and algorithmic discovery, sometimes optimizing runtime or resource objectives.} &
\taxcell{Broad \envtag{post-execution} black-box evaluator. Runtime/resource cases fit \envtag{post-execution} scoring, but the paper is not limited to code-efficiency RL.} &
\taxcell{No policy-training reward; evolutionary selection may use multiple evaluator scores rather than collapsed, two-gate, or blended RL rewards.} \\
\bottomrule
\end{tabular}
\end{table*}

We notice two things: first, benchmark papers often solve measurement but stop before the online reward question: they can use stress tests, hardware counters, right-censoring, or curve fitting that are carefully refined for a specific test set, including with heavy computations to make scores as reliable as possible, but can then become too expensive or too indirect to be evaluated on every GRPO rollout or provide usable feedback to update the policy in the right direction. Second, training and search papers that do use execution feedback often work in an easier regime than one-shot generation, either by editing an existing program or selecting among candidates at inference time. Our environment definitions cover the main places where timing can enter such systems, and our retained training setting tries to avoid any form of simplification: one-shot generation from scratch, based on the problem description only, with no execution feedback, evaluated with a live execution sandbox that gives reward based on execution results.

\clearpage
\section{Doing offline RL simulations to spare compute of online RL runs}
\label{app:rl_simulator}
The environment space in \cref{sec:reward} is too large to search only with online GRPO runs. This appendix explains the offline simulator used to prune that space before training. The simulator keeps the same environment-side filtering, timeout, ranking, and outcome computation as online RL, but replaces model generation and live execution with sampled human solutions and stored calibrated durations. We use the term ``simulator'' for convenience, but it remains a lightweight computation of RL-environment outcomes on reference samples, not a model of online learning itself.
\suppressfloats[t]

\subsection{Same environment, cheaper executions}
\label{sec:app_simulator_mechanics}

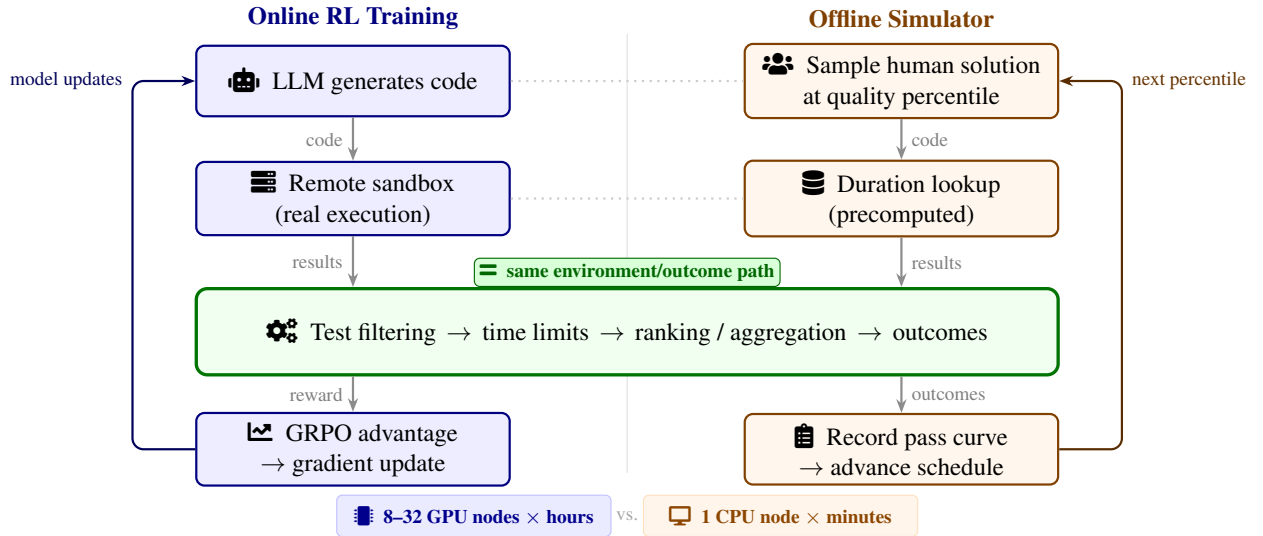
\begin{figure}[h!]
\centering
\begin{neuripsfigurefonts}
\resizebox{\textwidth}{!}{%
\begin{tikzpicture}[
  stage/.style={rectangle, rounded corners=3pt, draw=#1!50!black, fill=#1!7, thick,
    minimum height=0.9cm, minimum width=4.0cm, align=center, font=\small},
  shared/.style={rectangle, rounded corners=4pt, draw=green!45!black, fill=green!6,
    line width=1.2pt, align=center, font=\small},
  lbl/.style={font=\scriptsize, text=black!50},
  costbox/.style={rectangle, rounded corners=2pt, inner sep=4pt, font=\scriptsize\bfseries,
    minimum width=3.5cm, align=center},
  header/.style={font=\small\bfseries, text=#1!50!black},
  arr/.style={-{Stealth[length=5pt]}, thick, black!45},
  feedarr/.style={-{Stealth[length=5pt]}, thick, #1!35!black},
  replaces/.style={dotted, thick, black!20},
]

\def\lcol{-3.5}
\def\rcol{3.5}
\def\rowA{0}
\def\rowB{-1.5}
\def\rowC{-3.2}
\def\rowD{-4.7}

\node[header=blue] at (\lcol, 0.8) {Online RL Training};
\node[header=orange] at (\rcol, 0.8) {Offline Simulator};
\draw[black!10, line width=0.5pt] (0, 0.6) -- (0, \rowD-0.3);

\node[stage=blue] (llm) at (\lcol, \rowA) {\faIcon{robot}\; LLM generates code};
\node[stage=orange] (sampler) at (\rcol, \rowA) {\faIcon{users}\; Sample human solution\\at quality percentile};
\node[stage=blue] (sandbox) at (\lcol, \rowB) {\faIcon{server}\; Remote sandbox\\(real execution)};
\node[stage=orange] (lookup) at (\rcol, \rowB) {\faIcon{database}\; Duration lookup\\(precomputed)};
\node[shared, minimum width=11.0cm, minimum height=1.1cm] (envreward) at (0, \rowC)
  {\faIcon{cogs}\; Test filtering \,$\rightarrow$\, time limits \,$\rightarrow$\, ranking / aggregation \,$\rightarrow$\, outcomes};
\node[fill=green!15, draw=green!40!black, rounded corners=2pt, inner sep=2pt,
  font=\scriptsize\bfseries, text=green!40!black, anchor=south] at (envreward.north) {\faIcon{equals}\; same environment/outcome path};
\node[stage=blue] (gradient) at (\lcol, \rowD) {\faIcon{chart-line}\; GRPO advantage\\$\rightarrow$ gradient update};
\node[stage=orange] (record) at (\rcol, \rowD) {\faIcon{clipboard-list}\; Record pass curve\\$\rightarrow$ advance schedule};

\draw[arr] (llm) -- node[left, lbl] {code} (sandbox);
\draw[arr] (sandbox) -- node[left, lbl] {results} (envreward.north -| sandbox);
\draw[arr] (envreward.south -| gradient) -- node[left, lbl] {reward} (gradient);
\draw[arr] (sampler) -- node[right, lbl] {code} (lookup);
\draw[arr] (lookup) -- node[right, lbl] {results} (envreward.north -| lookup);
\draw[arr] (envreward.south -| record) -- node[right, lbl] {outcomes} (record);

\draw[feedarr=blue, rounded corners=4pt] (gradient.west) -- ++(-0.8, 0) |- (llm.west)
  node[pos=0.5, left, font=\scriptsize, text=blue!35!black] {model updates};
\draw[feedarr=orange, rounded corners=4pt] (record.east) -- ++(0.8, 0) |- (sampler.east)
  node[pos=0.5, right, font=\scriptsize, text=orange!40!black] {next percentile};
\draw[replaces] (llm.east) -- (sampler.west);
\draw[replaces] (sandbox.east) -- (lookup.west);

\node[costbox, fill=blue!8, draw=blue!25, text=blue!55!black, anchor=east]
  at (-0.2, \rowD-0.85) {\faIcon{microchip}\; 8--32 GPU nodes $\times$ hours};
\node[costbox, fill=orange!8, draw=orange!25, text=orange!55!black, anchor=west]
  at (0.2, \rowD-0.85) {\faIcon{desktop}\; 1 CPU node $\times$ minutes};
\node[font=\scriptsize, text=black!35] at (0, \rowD-0.85) {vs.};
\node[anchor=east, font=\scriptsize, opacity=0] at (-6.3, {(\rowA+\rowD)/2}) {next percentile};
\end{tikzpicture}%
}
\end{neuripsfigurefonts}
\caption{\textbf{Online RL training loop vs.\ offline simulator, on the worker side.}
The simulator evaluates candidate environment configurations by replacing LLM inference and sandbox execution with human solution sampling and precomputed duration lookups.
The environment-side computation stays the same: tests are filtered, time limits are enforced, and, in ranked environments, solutions are ranked against human references identically.
Online training then maps outcomes to rewards for GRPO updates; whereas the simulator records diagnostics from those outcomes and then samples better candidates for the next round, according to precomputed solution scores.}
\label{fig:simulator_vs_online}
\end{figure}

\Cref{fig:simulator_vs_online} shows the equivalence. Online training has two expensive components on the worker side that are not needed for a first screening pass: the current model must generate code, and the sandbox must execute every generated program (on the trainer side, the full RL run also includes gradient computations; across trainers and workers, it also includes communication costs, including weight broadcasting). The simulator replaces those two components while leaving the environment/outcome path fixed. In the simulator, a human solution is sampled from a quality band (built using the pre-recorded durations of that solution, which assigns it a quality score for how optimized it is beyond correctness), stored durations replace sandbox calls, and the same environment/outcome path produces pass-rate and quality observations. Moving the quality band from weaker to stronger reference solutions yields one simulated pass-rate curve per environment configuration, in the same manner that an online-RL-trained policy would generate stronger and stronger samples. This simulates only one side of the RL run: how the environment reacts to and rewards stronger generations, not how the policy reacts to those assigned rewards.

\subsection{Quantifying what a good RL environment is}
\label{sec:app_simulator_metrics}

A useful environment is not one that passes everything or fails everything, nor one that rewards an intermediate fraction of sampled solutions without correlating with better generated samples. Let \(c_i\in[0,1]\) be the simulated pass rate at normalized step \(x_i\) (so between 0 and 1), and let \(\tilde c_i\) be a moving-average smoothed version of the \(c_i\) curve. We then measure the following quantities on a full simulated run of a certain RL environment configuration:
\begin{enumerate}
\item \textbf{AUC.}
\[
\mathrm{AUC}_{\mathrm{raw}} = \frac{1}{n}\sum_{i=0}^{n-1} c_i,
\qquad
\mathrm{AUC}_{\mathrm{smooth}} = \frac{1}{m}\sum_{i=0}^{m-1}\tilde c_i,
\qquad m=n-w+1 .
\]
AUC measures pass-outcome density across the simulated run. It flags environments that reward almost everything, or regimes that are too sparse to generate useful online updates. It quantifies the amount of activation that goes through the environment, without telling how it activates or the shape of that activation.

\item \textbf{Steepness.}
\[
\hat c(x)=\operatorname{clip}(ax+b,0,1),
\qquad
S = \frac{a}{1+|a|}.
\]
Steepness is the bounded slope of a clipped linear fit to the activation curve. It measures whether the environment moves smoothly from mostly failing sampled solutions to mostly passing sampled solutions, or whether the signal turns on abruptly in a narrow quality band. Whereas AUC measures the total amount of activation that goes through, steepness measures how it activates, whether abruptly or smoothly.

\item \textbf{Variance around the smoothed curve.}
\[
V = \frac{1}{m}\sum_{i=0}^{m-1}(c_{i+\Delta}-\tilde c_i)^2,
\qquad \Delta=w-1 .
\]
Variance measures local wobble around the moving-average trend. This metric does not measure how the environment behaves and rewards samples, but how stable or noisy this reward is.

\item \textbf{Deviation from the diagonal.}
\[
D_{y=x} = \frac{1}{n}\sum_i |c_i-x_i| .
\]
Deviation from the diagonal measures how far the activation curve is from a gradual progression over the full quality range. This diagnostic mixes several effects: it overlaps with AUC on how much activation goes through and with steepness on how abruptly activation changes, but also captures where the environment activates along the quality sweep. A curve that smoothly activates near the middle of the sampled-quality range stays close to the diagonal; a curve with the same smooth shape that activates only later, when samples become stronger, has larger deviation.

\item \textbf{Quality-correlation.}
\[
Q_{\mathrm{raw}} = \rho_{\mathrm{Spearman}}(\{x_i\},\{c_i\}),
\qquad
Q_{\mathrm{smooth}} = \rho_{\mathrm{Spearman}}(\{x_i\},\{\tilde c_i\}).
\]
Quality-correlation measures whether pass outcomes tend to increase with normalized sample quality. This catches environments that have a reasonable average density but do not actually order sampled solutions in a useful way. The smoothed version tries to remove individual-sample noise to check whether the environment in general has a useful direction.
\end{enumerate}

\subsection{Do these metrics correlate with downstream online RL performance?}
\label{sec:app_simulator_online_comparison}

The simulator is useful only if its offline diagnostics say something about the online runs we would otherwise have to launch. We compare it to 20 RL-optimization environments launched with Qwen 2.5 7B, across pre-execution, intra-execution, and post-execution families. Each run is evaluated on DMC-Optim test at \(p_{100}\), \(p_{80}\), \(p_{50}\), and \(p_{30}\). The main statistic is Spearman correlation between a simulator diagnostic and the online RL pass@$1$ reached by that run.

\begin{table*}[t!]
\centering
\small
\caption{\textbf{Spearman correlation between simulator diagnostics and online test pass@1 on a set of Qwen 2.5 7B RL runs.}
The simulator is weak on pure correctness (\(p_{100}\)) but becomes more predictive as the optimization constraint gets stronger in the evaluation.
Lower deviation from \(y=x\) is better, hence the negative correlations; the \(p\) columns report the two-sided Spearman \(p\)-values.}
\label{tab:app_simulator_correlation_matrix}
\begin{tabular}{lcccccccc}
\toprule
& \multicolumn{2}{c}{\bfseries \(p_{100}\)} & \multicolumn{2}{c}{\bfseries \(p_{80}\)} & \multicolumn{2}{c}{\bfseries \(p_{50}\)} & \multicolumn{2}{c}{\bfseries \(p_{30}\)} \\
\cmidrule(lr){2-3}\cmidrule(lr){4-5}\cmidrule(lr){6-7}\cmidrule(lr){8-9}
\bfseries Metric & \bfseries \(r_s\) & \bfseries \(p\) & \bfseries \(r_s\) & \bfseries \(p\) & \bfseries \(r_s\) & \bfseries \(p\) & \bfseries \(r_s\) & \bfseries \(p\) \\
\midrule
AUC (smoothed) & 0.176 & 0.486 & -0.197 & 0.433 & -0.389 & 0.110 & -0.448 & 0.062 \\
AUC (raw) & 0.190 & 0.450 & -0.193 & 0.443 & -0.383 & 0.117 & -0.448 & 0.062 \\
Quality-correlation (smoothed) & 0.262 & 0.293 & 0.625 & 0.006 & 0.716 & 0.001 & 0.752 & 0.0003 \\
Quality-correlation (raw) & 0.353 & 0.151 & 0.660 & 0.003 & 0.763 & 0.0002 & 0.787 & 0.0001 \\
Variance around trend & -0.337 & 0.172 & -0.179 & 0.478 & -0.059 & 0.817 & -0.002 & 0.994 \\
Deviation from \(y=x\) & -0.291 & 0.241 & -0.652 & 0.003 & -0.788 & 0.0001 & -0.832 & 0.00002 \\
Steepness \(S\) & 0.382 & 0.118 & 0.633 & 0.005 & 0.720 & 0.0008 & 0.723 & 0.0007 \\
\bottomrule
\end{tabular}
\end{table*}

According to \cref{tab:app_simulator_correlation_matrix}, the metrics computed above do not predict correctness performance: at \(p_{100}\), all correlations are weak and none is significant. As the evaluation threshold becomes stricter, some become predictive. By \(p_{30}\), the strongest signals are deviation from \(y=x\) (\(r_s=-0.832\), \(p=2\times10^{-5}\)), raw quality-correlation (\(r_s=0.787\), \(p=10^{-4}\)), and steepness (\(r_s=0.723\), \(p=7\times10^{-4}\)), while curve noise remains uninformative (\(r_s=-0.002\), \(p=0.994\)).

When correlating these metrics with earlier checkpoint evaluations in the RL runs, the correlations are weaker, consistent with the run variability observed at the end of \cref{sec:training}. By the end of the RL runs, however, some metrics show stronger signal.

\begin{figure}[p]
\centering
\includegraphics[height=0.82\textheight]{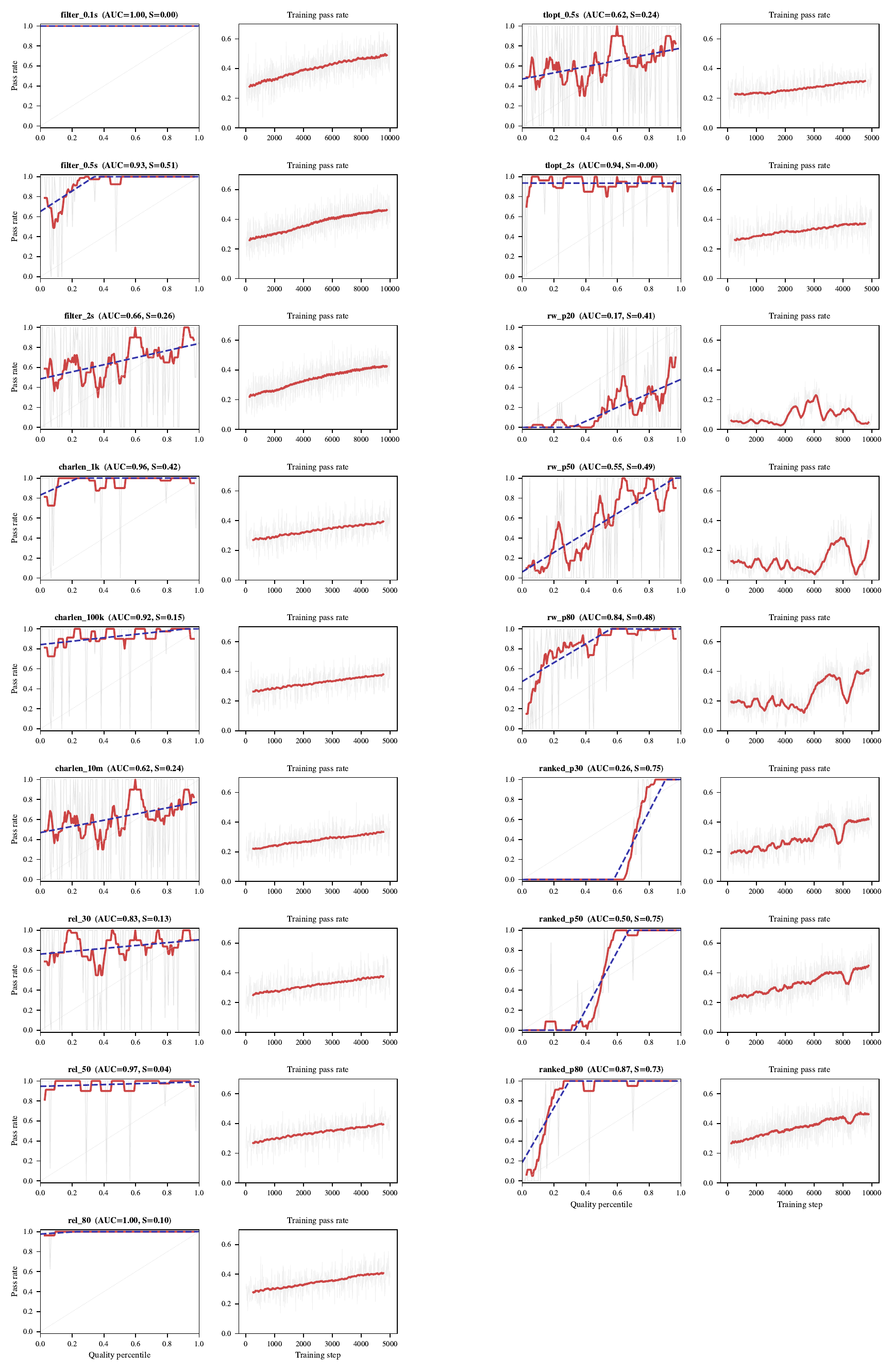}
\caption{\textbf{Offline simulator curves compared with online training pass rates for matched Qwen 2.5 7B environments.}
Each row pairs one environment configuration with its simulator curve on the left and the online training pass rate recorded on the same training environment on the right.
Simulator panels show the raw pass-rate curve in gray, a smoothed curve in red, the clamped-linear fit in blue, and the \(y=x\) reference in light gray; titles report simulator AUC and steepness \(S\).
Training panels show raw logged pass rates in gray and a smoothed curve in red.
These online curves are not the held-out resimulation pass@1 values used in \cref{tab:app_simulator_correlation_matrix}; they provide a qualitative view of the pass-rate signal seen during training.
}
\label{fig:app_simulator_training_curves}
\end{figure}

A useful follow-up would study the simulator at finer granularity, both across RL-optimization environment families and within each family as its parameters vary. We do not have enough downstream RL runs to conduct that analysis. In the current study, the simulator remains most useful as a pruning tool: it separates highly degenerate environments from configurations that can support online learning, but it is still unclear whether offline diagnostics can reliably rank the remaining viable configurations. The diagnostics in \cref{tab:app_simulator_correlation_matrix} suggest that some metrics carry signal about end-of-training test pass@1 under a strict optimization constraint, but they do not establish that the simulator predicts the shape of the online training curve. \Cref{fig:app_simulator_training_curves} compares the offline simulation curve induced by a fixed quality-sampling schedule with the online RL training curve for the Qwen 2.5 7B configurations reported in \cref{tab:cross_model}; most of these were successful candidates in the first place, and the main failures in the broader matched-curve comparison are RW \(p_{20}\) and RW \(p_{50}\). The same panels show the clamped-linear fit used to compute steepness.

\subsection{Using the offline simulator to select the most promising parametrization of each environment}
\label{sec:app_simulator_sweep_surfaces}

\begin{figure*}[t!]
\centering
\includegraphics[width=0.92\textwidth,height=0.24\textheight,keepaspectratio]{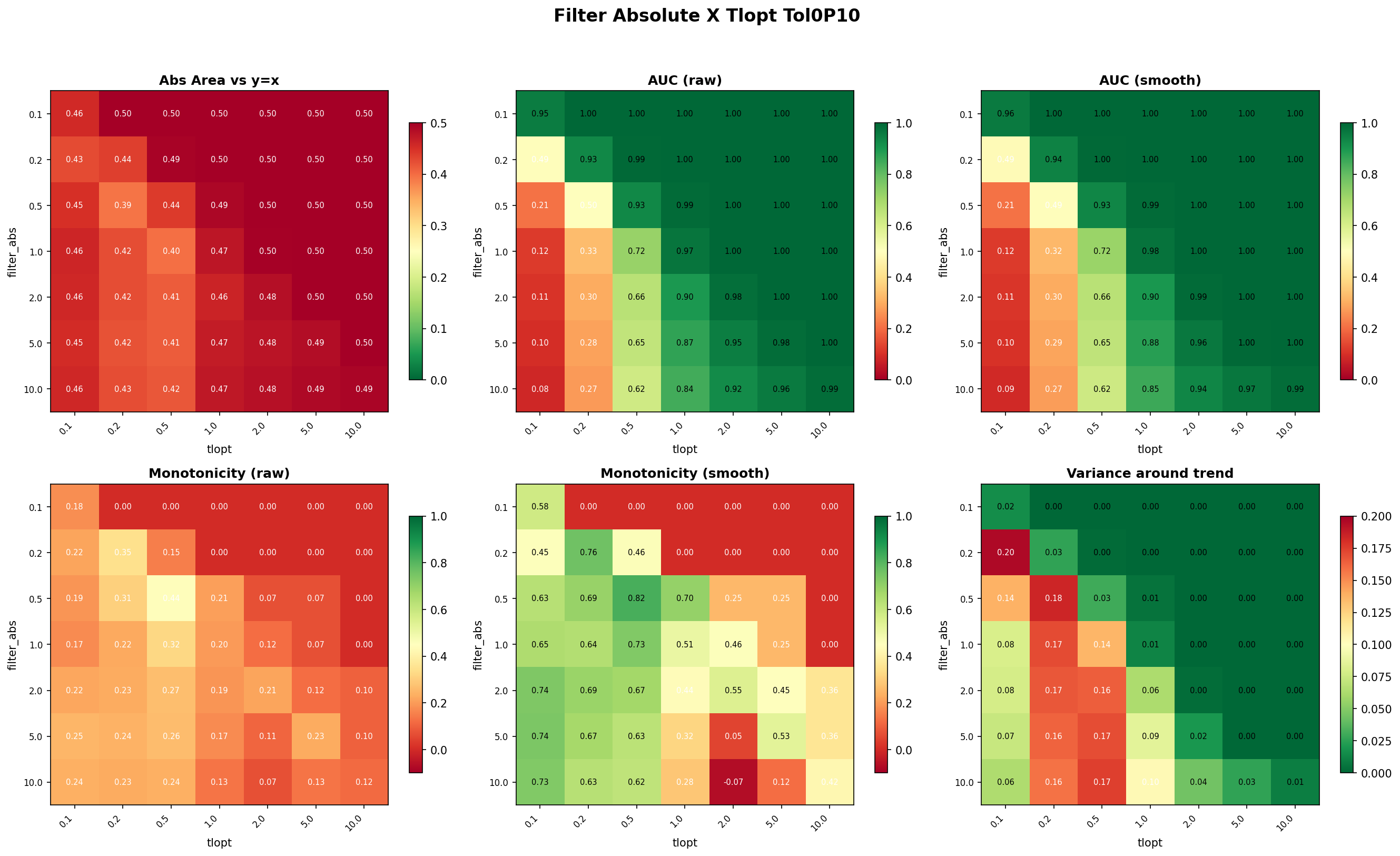}
\vspace{0.15em}
\includegraphics[width=0.92\textwidth,height=0.24\textheight,keepaspectratio]{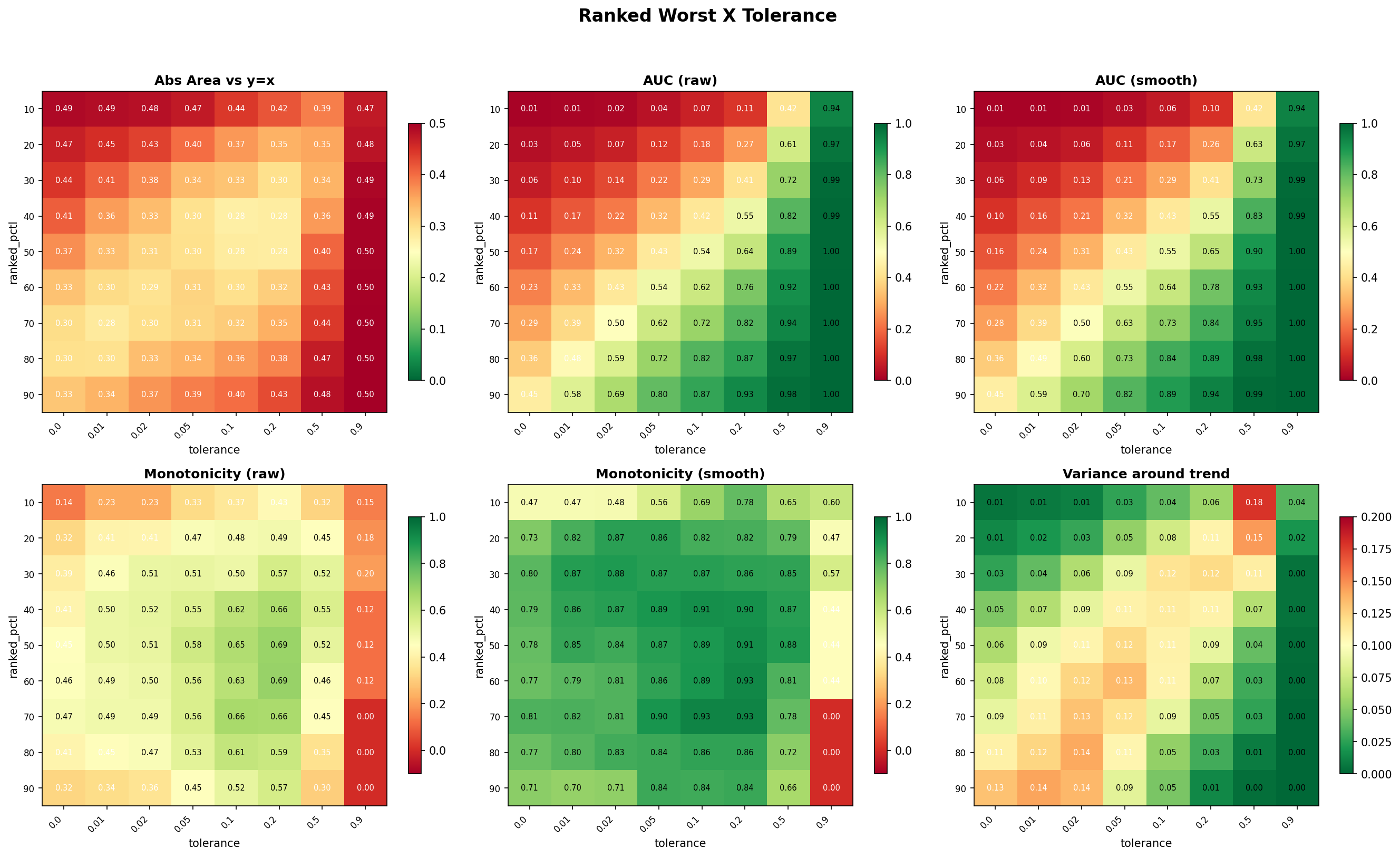}
\vspace{0.15em}
\includegraphics[width=0.92\textwidth,height=0.24\textheight,keepaspectratio]{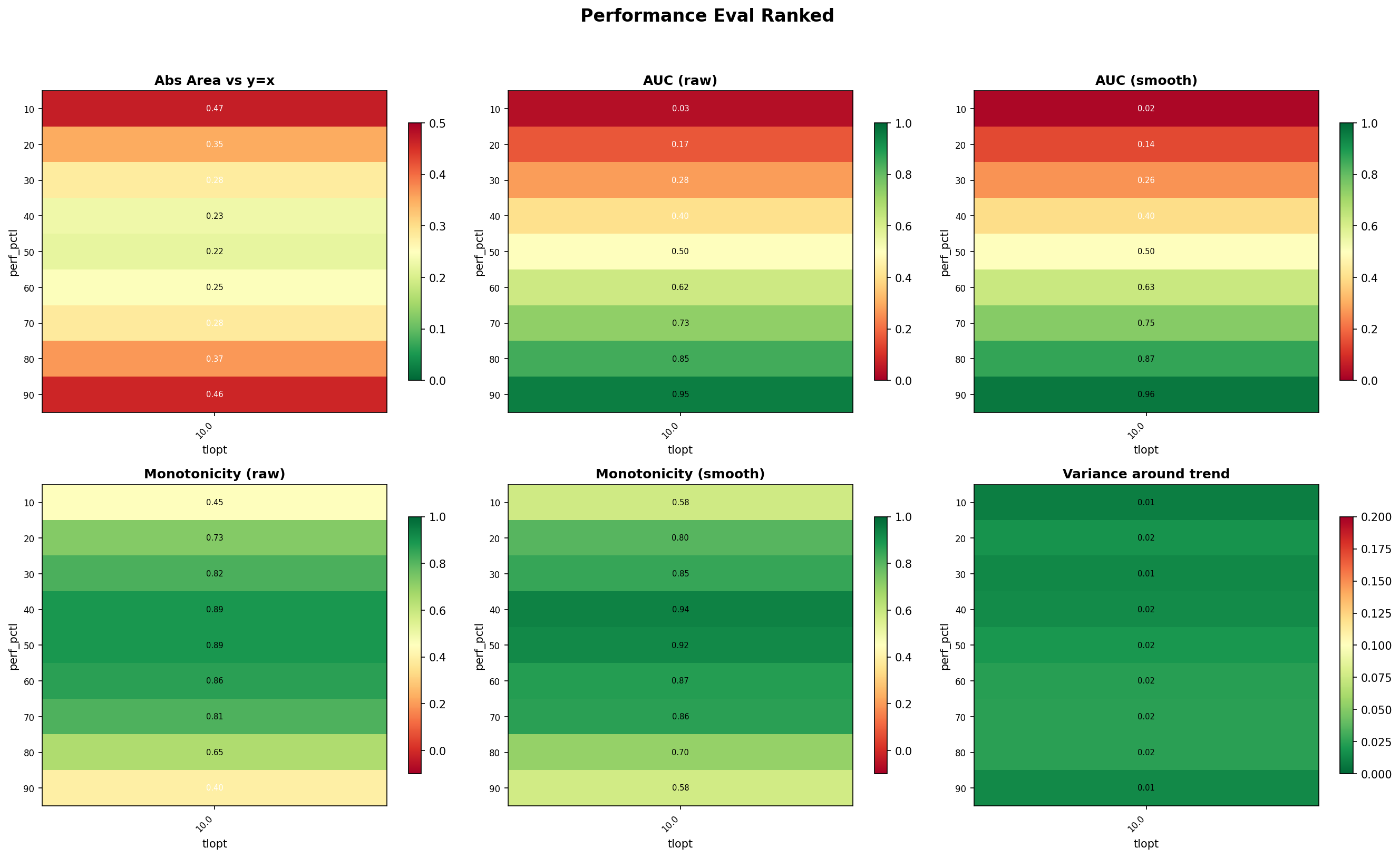}
\caption{\textbf{Simulator heatmap examples for three optimization-aware environment families.}
\emph{Top:} pre-execution absolute-duration filtering; rows sweep the retained-test duration filter from \(10\,\text{s}\) to \(0.1\,\text{s}\), columns sweep the absolute optimization time limit over the same range, and timeout tolerance is fixed to \(10\%\). Other simulator sweeps vary this tolerance.
\emph{Middle:} intra-execution ranked-worst time limits; rows sweep the reference percentile used to set the per-test timeout and columns sweep timeout tolerance.
\emph{Bottom:} post-execution leaderboard-percentile ranking; the sweep varies only the leaderboard percentile threshold.
All panels report the diagnostics defined earlier in this appendix: deviation from \(y=x\), raw and smoothed AUC, raw and smoothed quality-correlation (shown as monotonicity in the figure labels), and variance around trend. The red--yellow--green scales show relative metric values within each panel; they do not imply that one color is always better, since \cref{tab:app_simulator_correlation_matrix} shows that some diagnostics correlate positively with downstream test pass@1, some correlate negatively, and others correlate only weakly.}
\label{fig:app_simulator_heatmap_family_examples}
\end{figure*}

In practice, we use the offline simulator mainly to sweep the large parameterization space within each environment family, rather than to eliminate whole families at once. For example, pre-execution environments can have three continuous-valued controls: the filter threshold, the optimization time limit, and the timeout tolerance. \Cref{fig:app_simulator_heatmap_family_examples} shows the type of heatmaps produced by these sweeps. The absolute-filtering pre-execution sweep shows that many filter/time-limit pairs have high AUC, meaning they likely reward too many samples and therefore provide little pressure toward optimized generations; their low quality-correlation points in the same direction. The ranked-worst intra-execution family shows a broader range of behaviors, from degenerate settings that are simply too hard, such as \(p_{10}\) with low timeout tolerance, to middle-ground settings at higher percentiles and timeout tolerance around \(0.1\). Finally, post-execution leaderboard ranking is lower-dimensional: once the ranking-score aggregation is fixed, as studied in \cref{sec:app_ranking_metrics}, this environment mainly depends on the leaderboard percentile threshold. The sweep suggests that \(p_{10}\) is too strict while \(p_{100}\) and \(p_{90}\) are too permissive. Although \cref{tab:app_simulator_correlation_matrix} shows that variance around trend does not correlate much with downstream online RL performance, it stays low across this post-execution family.

\clearpage
\section{Async-RL and training objective}
\label{app:training_support}
\paragraph{GRPO objective.}
For a prompt \(x\), workers generate a same-prompt group of \(G\) rollouts \(\{y_i\}_{i=1}^{G}\).
The trainer centers each return against a prompt-level token-weighted mean:
\begin{equation}
\label{eq:advantage}
\hat{A}_i = R_i - \mu_x,
\qquad
\mu_x = \frac{\sum_{j=1}^{G} L_j R_j}{\sum_{j=1}^{G} L_j},
\end{equation}
where \(L_j\) is the number of model-generated tokens in rollout \(j\).
We then optimize the clipped surrogate objective
\begin{equation}
\label{eq:grpo_loss}
\mathcal{J}(\theta)=
\frac{1}{N}
\sum_{y_i \in \mathcal{B}}
\sum_{t=1}^{|y_i|}
M_{i,t}
\min\!\Bigl[
\rho_{i,t}(\theta)\hat{A}_i,\;
\operatorname{clip}\!\bigl(\rho_{i,t}(\theta), 1-\varepsilon_{\mathrm{low}}, 1+\varepsilon_{\mathrm{high}}\bigr)\hat{A}_i
\Bigr],
\end{equation}
with a fixed token horizon \(N\) rather than the realized response length.

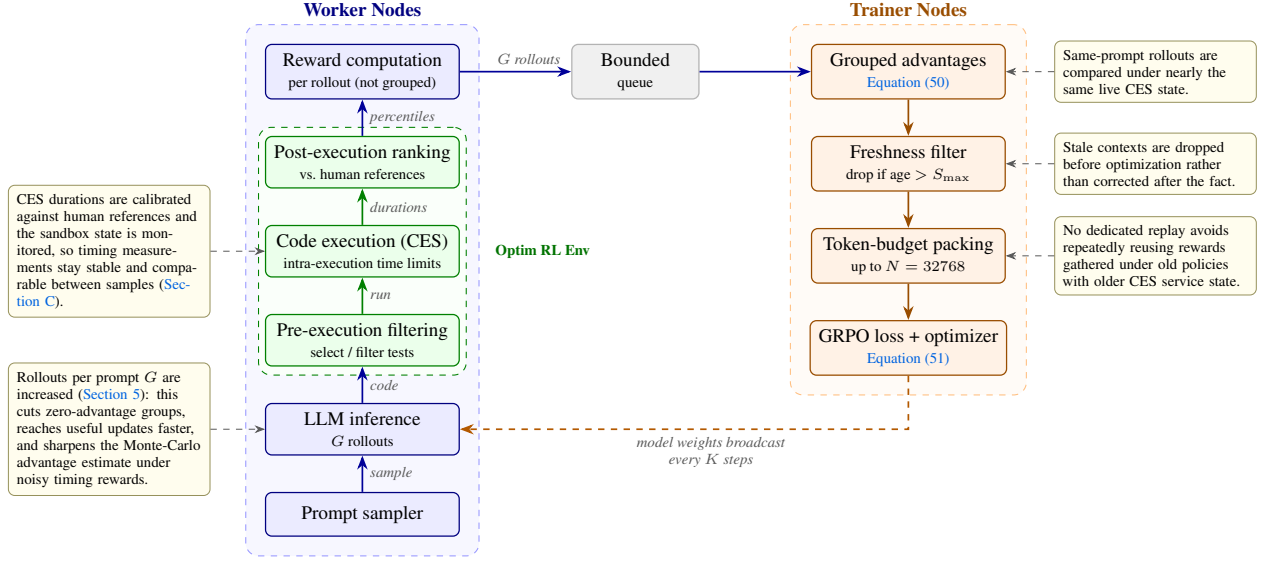
\begin{figure*}[t!]
\centering
\begin{neuripsfigurefonts}
\resizebox{\textwidth}{!}{%
\begin{tikzpicture}[
    >=Stealth,
    node distance=0.6cm and 1.0cm,
    mainbox/.style={draw, rounded corners=3pt, minimum height=0.72cm, align=center, font=\small, inner sep=4pt, line width=0.6pt},
    worker/.style={mainbox, fill=blue!8, draw=blue!50!black, minimum width=3.2cm},
    trainer/.style={mainbox, fill=orange!10, draw=orange!60!black, minimum width=3.2cm},
    env/.style={mainbox, fill=green!8, draw=green!50!black, minimum width=3.2cm},
    queue/.style={mainbox, fill=gray!12, draw=gray!60, minimum width=2.1cm, minimum height=0.7cm},
    note/.style={draw, rounded corners=3pt, fill=yellow!8, draw=yellow!50!black, font=\scriptsize, align=left, text width=3.05cm, inner sep=4pt},
    group/.style={draw, rounded corners=6pt, dashed, inner sep=9pt, line width=0.5pt},
    dataarrow/.style={->, thick, color=blue!60!black},
    trainerarrow/.style={->, thick, color=orange!60!black},
    modelarrow/.style={->, thick, color=orange!70!black, dashed},
    envarrow/.style={->, thick, color=green!50!black},
    notearrow/.style={->, dashed, color=black!65},
    lbl/.style={font=\scriptsize\itshape, text=gray!70!black},
]

\node[worker] (reward) {Reward computation\\[-1pt]{\scriptsize per rollout (not grouped)}};
\node[env, below=0.6cm of reward] (postrank) {Post-execution ranking\\[-1pt]{\scriptsize vs.\ human references}};
\node[env, below=0.6cm of postrank] (exec) {Code execution (CES)\\[-1pt]{\scriptsize intra-execution time limits}};
\node[env, below=0.6cm of exec] (prefilter) {Pre-execution filtering\\[-1pt]{\scriptsize select / filter tests}};
\node[worker, below=0.6cm of prefilter] (fastgen) {LLM inference\\[-1pt]{\scriptsize \(G\) rollouts}};
\node[worker, below=0.6cm of fastgen] (prompts) {Prompt sampler};

\begin{scope}[on background layer]
\node[group, draw=blue!40, fill=blue!3,
      fit=(prompts)(fastgen)(prefilter)(exec)(postrank)(reward),
      label={[font=\small\bfseries, text=blue!50!black]above:Worker Nodes}] (wbox) {};
\node[group, draw=green!50!black, fill=green!2, inner xsep=3pt, inner ysep=4pt,
      fit=(prefilter)(exec)(postrank),
      label={[font=\scriptsize\bfseries, text=green!45!black, label distance=10pt]right:Optim RL Env}] (envbox) {};
\end{scope}

\node[trainer, right=5.8cm of reward] (adv) {Grouped advantages\\[-1pt]{\scriptsize \cref{eq:advantage}}};
\node[trainer, below=0.6cm of adv] (stale) {Freshness filter\\[-1pt]{\scriptsize drop if age \(> S_{\max}\)}};
\node[trainer, below=0.6cm of stale] (pack) {Token-budget packing\\[-1pt]{\scriptsize up to \(N=32768\)}};
\node[trainer, below=0.6cm of pack] (loss) {GRPO loss + optimizer\\[-1pt]{\scriptsize \cref{eq:grpo_loss}}};

\begin{scope}[on background layer]
\node[group, draw=orange!50, fill=orange!3,
      fit=(adv)(stale)(pack)(loss),
      label={[font=\small\bfseries, text=orange!60!black]above:Trainer Nodes}] (tbox) {};
\end{scope}

\node[queue] at ($(reward.east)!0.5!(adv.west |- reward.east)$) (dataqueue) {Bounded\\[-1pt]{\scriptsize queue}};

\draw[dataarrow] (prompts) -- node[right, lbl] {sample} (fastgen);
\draw[dataarrow] (fastgen) -- node[right, lbl] {code} (prefilter);
\draw[envarrow] (prefilter) -- node[right, lbl] {run} (exec);
\draw[envarrow] (exec) -- node[right, lbl] {durations} (postrank);
\draw[dataarrow] (postrank) -- node[right, lbl] {percentiles} (reward);
\draw[dataarrow] (reward.east) -- node[above, lbl, pos=0.62] {\(G\) rollouts} (dataqueue.west);
\draw[dataarrow] (dataqueue.east) -- ++(0.45,0) |- (adv.west);

\draw[trainerarrow] (adv) -- (stale);
\draw[trainerarrow] (stale) -- (pack);
\draw[trainerarrow] (pack) -- (loss);

\draw[modelarrow] (loss.south) |- (fastgen.east)
    node[pos=0.72, below, lbl, align=center] {model weights broadcast\\every \(K\) steps};

\node[note, left=0.9cm of fastgen] (noteG)
    {Rollouts per prompt \(G\) are increased (\cref{sec:grpo}): this cuts zero-advantage groups, reaches useful updates faster, and sharpens the Monte-Carlo advantage estimate under noisy timing rewards.};
\node[note, left=0.9cm of exec] (noteCES)
    {CES durations are calibrated against human references and the sandbox state is monitored, so timing measurements stay stable and comparable between samples (\cref{app:ces_fallback}).};

\draw[notearrow] (noteG.east) -- (fastgen.west);
\draw[notearrow] (noteCES.east) -- (exec.west);

\node[note, right=0.8cm of adv] (note1) {Same-prompt rollouts are compared under nearly the same live CES state.};
\node[note, right=0.8cm of stale] (note2) {Stale contexts are dropped before optimization rather than corrected after the fact.};
\node[note, right=0.8cm of pack] (note3) {No dedicated replay avoids repeatedly reusing rewards gathered under old policies with older CES service state.};

\draw[notearrow] (note1.west) -- (adv.east);
\draw[notearrow] (note2.west) -- (stale.east);
\draw[notearrow] (note3.west) -- (pack.east);
\end{tikzpicture}
}
\end{neuripsfigurefonts}
\caption{\textbf{Asynchronous trainer-worker architecture for efficiency RL.}
Each worker samples one prompt, generates \(G\) rollouts, and runs them through the optimization RL environment: pre-execution test filtering, code execution on the calibrated remote sandbox (CES) under intra-execution time limits, and post-execution ranking against human references, followed by a per-rollout reward computation (rewards are \emph{not} grouped on the worker side).
The resulting group of \(G\) rollouts is sent to the trainers through a bounded queue.
The trainers compute grouped advantages (\cref{eq:advantage}), discard contexts older than \(S_{\max}=30\) optimizer steps, pack the remaining contexts up to \(N=32768\) tokens, and apply the clipped GRPO loss (\cref{eq:grpo_loss}).
Updated model weights are broadcast to the workers every \(K\) optimizer steps.
As explained in the GRPO paragraph of \cref{sec:grpo}, the number of rollouts \(G\) is deliberately increased to reduce zero-advantage groups and stabilize the advantage estimate under sparse, noisy timing rewards.}
\label{fig:rl_pipeline}
\end{figure*}

\clearpage
\section{Examples of Optimization-RL improvements}
\label{app:judge_case_studies}
\subsection{Judge prompt used for the pairwise analysis}
\label{sec:app_judge_prompt}

The following user message is sent to GPT-OSS 120B for each blinded pair.

\begingroup
\scriptsize
\begin{verbatim}
Here is a competitive programming problem and two solutions. Analyze them carefully.

## Problem

{problem_description}

## Solution 1

```python
{solution_1_code}
```

## Solution 2

```python
{solution_2_code}
```

## Instructions

1. Analyze both solutions in detail. Compare their algorithms, data structures,
   and implementation choices. Reference specific lines of code.

2. Then output a single JSON object with the following fields:

{
  "predicted_faster": "Solution 1" or "Solution 2" or "Same",
  "algorithm_analysis": "Your detailed analysis as a single string. Explain what
    each solution does, how they differ, and why one is faster. Reference
    specific code patterns.",
  "complexity_comparison": {
    "solution_1": {"best": "O(...)", "average": "O(...)", "worst": "O(...)"},
    "solution_2": {"best": "O(...)", "average": "O(...)", "worst": "O(...)"}
  },
  "faster_solution_complexity_improves": {
    "best": true or false or null,
    "average": true or false or null,
    "worst": true or false or null
  },
  "primary_category": one of ["ALGORITHM_CHANGE", "DATA_STRUCTURE_CHANGE",
    "COMPLEXITY_REDUCTION", "IO_OPTIMIZATION", "CACHING_MEMOIZATION",
    "CONSTANT_FACTOR", "IMPLEMENTATION_STYLE", "MATHEMATICAL_SHORTCUT"],
  "secondary_categories": [...],
  "classification": "ALGORITHMIC" or "SUPERFICIAL",
  "classification_reason": "One-line summary of why",
  "key_differences": ["specific difference 1", "specific difference 2", ...]
}

Category definitions:
- ALGORITHM_CHANGE: Different algorithmic paradigm (brute force -> DP,
  BFS -> Dijkstra, recursion -> iterative)
- DATA_STRUCTURE_CHANGE: Different core data structure (list -> heap, dict ->
  sorted set, array -> segment tree)
- COMPLEXITY_REDUCTION: Same paradigm but tighter complexity (removing nested
  loop, binary search instead of linear scan)
- IO_OPTIMIZATION: Faster input/output (sys.stdin vs input(), printf vs cout)
- CACHING_MEMOIZATION: Adding lru_cache, precomputation tables, memoization
- CONSTANT_FACTOR: Same algorithm, same complexity, fewer operations per
  iteration (pruning, sparse representation, early termination)
- IMPLEMENTATION_STYLE: Code style differences with negligible performance impact
- MATHEMATICAL_SHORTCUT: Closed-form formula, modular arithmetic trick,
  combinatorial identity

Classification definitions:
- ALGORITHMIC: The faster solution uses a fundamentally different algorithm,
  paradigm, or approach to solve the problem (e.g., brute force vs DP, BFS vs
  Dijkstra, completely different mathematical formulation). The two solutions
  reflect different problem-solving strategies.
- SUPERFICIAL: The faster solution uses the same core algorithm/approach but
  with implementation-level improvements -- tighter loops, pruning, better data
  structures within the same framework, I/O tricks, or constant-factor
  optimizations. This includes cases where optimizations within the same
  paradigm happen to improve asymptotic complexity (e.g., adding a binary search
  to skip a linear scan in an otherwise identical algorithm).

For "faster_solution_complexity_improves": compare the predicted-faster
solution's complexity against the other. true = faster solution has strictly
better complexity at that level, false = same or worse, null = cannot determine.
\end{verbatim}
\endgroup

\subsection{Cases where optimization-RL beats standard RLVR}
\label{sec:app_optimrl_vs_rlvr_cases}

In these two problems, both CWM 32B submissions are correct, but optimization-RL recovers the known intended solution while standard RLVR produces a correct but more generic program.

\paragraph{Codeforces 1466F, ``Euclid's Nightmare''.}
\textbf{Problem.} Given \(n\) vectors over \(\mathbb{Z}_2^m\), each with at most two non-zero coordinates, compute the size of their span modulo \(10^9+7\) and output the lexicographically smallest minimum subset that spans the same space. Here \(n,m\leq 5\cdot 10^5\). \textbf{Measured:} RLVR \(\,46.4\,s\rightarrow\) optimization-RL \(28.5\,s\) (\(\times1.63\)); judge label: algorithm change / algorithmic.

\paragraph{RLVR solution.}
\begingroup
\scriptsize
\begin{verbatim}
def main():
    import sys
    input = sys.stdin.read().split()
    ptr = 0
    n = int(input[ptr])
    ptr += 1
    m = int(input[ptr])
    ptr += 1

    vectors = []
    for i in range(n):
        k = int(input[ptr])
        ptr += 1
        coords = list(map(int, input[ptr:ptr + k]))
        ptr += k
        coords.sort()
        if k == 1:
            vec = (coords[0],)
        else:
            vec = (coords[0], coords[1])
        vectors.append((vec, i + 1))

    basis = {}
    selected_indices = []

    for vec, idx in vectors:
        current = vec
        while True:
            if not current:
                break
            pivot = current[0]
            if pivot not in basis:
                break
            basis_vec, basis_idx = basis[pivot]
            combined = list(current) + list(basis_vec)
            count = {}
            for x in combined:
                count[x] = count.get(x, 0) + 1
            new_current = []
            for x in sorted(count):
                if count[x] %
                    new_current.append(x)
            new_current = tuple(new_current)
            current = new_current

        if current:
            pivot = current[0]
            basis[pivot] = (current, idx)
            selected_indices.append(idx)

    mod = 10**9 + 7
    k = len(selected_indices)
    size_T = pow(2, k, mod)
    print(size_T, k)
    if k > 0:
        print(' '.join(map(str, sorted(selected_indices))))
    else:
        print()

if __name__ == "__main__":
    main()
\end{verbatim}
\endgroup

\paragraph{Optimization-RL solution.}
\begingroup
\scriptsize
\begin{verbatim}
import sys
MOD = 10**9 + 7

def main():
    import sys
    sys.setrecursionlimit(1 << 25)
    n, m = map(int, sys.stdin.readline().split())
    parent = list(range(m + 1))
    size = [1] * (m + 1)
    has_single = [False] * (m + 1)
    basis = []

    def find(u):
        while parent[u] != u:
            parent[u] = parent[parent[u]]
            u = parent[u]
        return u

    def merge(a, b):
        if a == b:
            return
        if size[a] < size[b]:
            a, b = b, a
        parent[b] = a
        size[a] += size[b]
        has_single[a] = has_single[a] or has_single[b]

    for idx in range(1, n + 1):
        parts = list(map(int, sys.stdin.readline().split()))
        k = parts[0]
        if k == 1:
            x = parts[1]
            rx = find(x)
            if not has_single[rx]:
                basis.append(idx)
                has_single[rx] = True
        else:
            x, y = parts[1], parts[2]
            rx = find(x)
            ry = find(y)
            if rx == ry:
                continue
            s_rx = has_single[rx]
            s_ry = has_single[ry]
            if s_rx and s_ry:
                merge(rx, ry)
            elif s_rx or s_ry:
                basis.append(idx)
                merge(rx, ry)
            else:
                basis.append(idx)
                merge(rx, ry)

    basis.sort()
    t_size = pow(2, len(basis), MOD)
    print(t_size, len(basis))
    print(' '.join(map(str, basis)))

if __name__ == "__main__":
    main()
\end{verbatim}
\endgroup

\paragraph{What changed.}
Both programs compute a basis of the given GF(2) vectors and report the span size \(2^{\mathrm{rank}}\) together with a minimum generating subset. RLVR builds a generic linear basis, keyed by the smallest non-zero coordinate of each vector, and repeatedly reduces incoming vectors by symmetric-differencing coordinate sets. This is valid Gaussian elimination over GF(2), but it ignores the special structure of the input. Optimization-RL uses the intended graph reformulation: a two-coordinate vector \((x,y)\) is an edge between coordinates, a one-coordinate vector is a component-level single, and the rank can be maintained with a disjoint-set union carrying one \texttt{has\_single} flag per component. Generic elimination is replaced by near-linear union--find.

\paragraph{AtCoder ARC085, ``ABS''.}
\textbf{Problem.} A deck has \(N\) visible card values. Players \(X\) and \(Y\), starting with cards \(Z\) and \(W\), alternately draw a non-empty prefix of the remaining deck and keep the last drawn card. When the deck is exhausted, \(X\) maximizes and \(Y\) minimizes the absolute difference of their final held cards. Here \(N\leq 2000\). \textbf{Measured:} RLVR \(55.1\,s\rightarrow\) optimization-RL \(9.2\,s\) (\(\times5.96\)); judge label: mathematical shortcut / algorithmic.

\paragraph{RLVR solution.}
\begingroup
\scriptsize
\begin{verbatim}
import sys
from functools import lru_cache

def main():
    sys.setrecursionlimit(1 << 25)
    N, Z, W = map(int, sys.stdin.readline().split())
    a = list(map(int, sys.stdin.readline().split()))

    @lru_cache(maxsize=None)
    def dp(i, last_opponent, is_X_turn):
        if i >= N:
            return 0
        if is_X_turn:
            max_score = -float('inf')
            for j in range(i, N):
                current_card = a[j]
                if j == N - 1:
                    score = abs(current_card - last_opponent)
                else:
                    score = dp(j + 1, current_card, False)
                if score > max_score:
                    max_score = score
            return max_score
        else:
            min_score = float('inf')
            for j in range(i, N):
                current_card = a[j]
                if j == N - 1:
                    score = abs(current_card - last_opponent)
                else:
                    score = dp(j + 1, current_card, True)
                if score < min_score:
                    min_score = score
            return min_score

    print(dp(0, W, True))

if __name__ == '__main__':
    main()
\end{verbatim}
\endgroup

\paragraph{Optimization-RL solution.}
\begingroup
\scriptsize
\begin{verbatim}
import sys

def main():
    N, Z, W = map(int, sys.stdin.readline().split())
    a = list(map(int, sys.stdin.readline().split()))
    if N == 1:
        print(abs(a[0] - W))
    else:
        option1 = abs(a[-1] - W)
        option2 = abs(a[-1] - a[-2])
        print(max(option1, option2))

if __name__ == "__main__":
    main()
\end{verbatim}
\endgroup

\paragraph{What changed.}
RLVR implements the game literally: a memoized minimax recursion over the next deck position, the opponent's last card, and the player to move, trying every possible prefix length at every state. Optimization-RL collapses the game to the known closed form. If \(N=1\), the answer is \(|a_0-W|\); otherwise it is \(\max(|a_{N-1}-W|, |a_{N-1}-a_{N-2}|)\). The entire minimax disappears because optimal play only depends on the last one or two cards.

\subsection{Cases where optimization-RL beats the fastest available human}
\label{sec:app_optimrl_vs_human_cases}

Here ``best human'' means the fastest correct human solution available in the source dataset for the problem, not necessarily the globally best Codeforces or AtCoder submission. The examples below are still real competitive-programming solutions; the point is to inspect when optimization-RL writes a cleaner or more optimized variant.

\paragraph{Codeforces 571A, ``Lengthening Sticks''.}
\textbf{Problem.} Given stick lengths \(a,b,c\) and a budget \(\ell\), count triples of non-negative integer increments with total at most \(\ell\) such that the resulting sticks form a non-degenerate triangle. Here \(a,b,c,\ell\leq 3\cdot 10^5\). \textbf{Measured:} best human \(17.2\,s\rightarrow\) optimization-RL \(13.5\,s\) (\(\times1.28\)); judge label: mathematical shortcut / algorithmic.

\paragraph{Best human solution.}
\begingroup
\scriptsize
\begin{verbatim}
def solve(a, b, c):
    ans = 0
    for da in range(max(0, b + c - a), l + 1):
        x = min(a - b - c + da, l - da)
        ans += (x + 1) * (x + 2) // 2
    return ans


a, b, c, l = map(int, input().split())
print((l + 1) * (l + 2) * (l + 3) // 6 - solve(a, b, c) - solve(b, a, c) - solve(c, a, b))
\end{verbatim}
\endgroup

\paragraph{Optimization-RL solution.}
\begingroup
\scriptsize
\begin{verbatim}
def count_invalid(s1, s2, s3, l):
    d = s1 + s2 - s3
    if d > 0:
        if l < d:
            return 0
        m = (l - d) // 2
        C = l - d + 1
        sum1 = (m + 1) * (m + 2) // 2
        sum2 = m * (m + 1) * (m + 2) // 3
        return C * sum1 - 2 * sum2
    else:
        k = -d
        m1 = min(k, l)
        part1 = 0
        if m1 >= 0:
            sum1_part1 = (m1 + 1) * (m1 + 2) // 2
            sum2_part1 = m1 * (m1 + 1) * (m1 + 2) // 3
            part1 = (l + 1) * sum1_part1 - sum2_part1
        s_start = k + 1
        s_end = (l + k) // 2
        part2 = 0
        if s_start <= s_end:
            C = l + k + 1
            a = s_start
            b = s_end
            sum1_part2 = ((b + 1) * (b + 2) // 2) - (a * (a + 1) // 2)
            sum2_part2 = (b * (b + 1) * (b + 2) - (a - 1) * a * (a + 1)) // 3
            part2 = C * sum1_part2 - 2 * sum2_part2
        return part1 + part2

def main():
    import sys
    a, b, c, l = map(int, sys.stdin.readline().split())
    total = (l + 1) * (l + 2) * (l + 3) // 6 if l >= 0 else 0
    countA = count_invalid(a, b, c, l)
    countB = count_invalid(a, c, b, l)
    countC = count_invalid(b, c, a, l)
    ans = total - (countA + countB + countC)
    print(max(ans, 0))

if __name__ == "__main__":
    main()
\end{verbatim}
\endgroup

\paragraph{What changed.}
The count is \(\binom{\ell+3}{3}\) minus the three invalid cases where one side is at least the sum of the other two; at most one such inequality can hold at a time. The human solution is legitimate work: it uses inclusion--exclusion and computes each invalid count with an \(O(\ell)\) triangular-number sweep, which is exactly the kind of solution many strong contestants would submit because it passes comfortably. Optimization-RL goes one step further by recognizing that the sweep sums a quadratic polynomial and evaluating it in closed form. The measured \(\times1.28\) speedup is modest because \(\ell\) is bounded, but the asymptotic change from \(O(\ell)\) to \(O(1)\) per invalid side is real.

\paragraph{Codeforces 1237B, ``Balanced Tunnel''.}
\textbf{Problem.} Given the entry and exit permutations of \(n\) cars in a one-way tunnel, count the cars that must have overtaken at least one other car. Here \(n\leq 10^5\). \textbf{Measured:} best human \(17.7\,s\rightarrow\) optimization-RL \(17.4\,s\) (\(\times1.02\)); judge label: mathematical shortcut / algorithmic.

\paragraph{Best human solution.}
\begingroup
\scriptsize
\begin{verbatim}
from sys import stdin, setrecursionlimit
setrecursionlimit(10**7)

def iin(): return int(stdin.readline())

def lin(): return list(map(int, stdin.readline().split()))

def main():
    n=iin()
    e=lin()
    ex=lin()
    fine=set()
    i,j=0,0
    while i<n and j<n:
        if e[i]==ex[j]:
            i+=1
            j+=1
        else:
            while i<n and  (e[i] in fine):
                i+=1
            while j <n and  ex[j] !=e[i]:
                fine.add(ex[j])
                j+=1
    print(len(fine))

main()
\end{verbatim}
\endgroup

\paragraph{Optimization-RL solution.}
\begingroup
\scriptsize
\begin{verbatim}
import sys

def main():
    n = int(sys.stdin.readline())
    a = list(map(int, sys.stdin.readline().split()))
    b = list(map(int, sys.stdin.readline().split()))
    exit_pos = [0] * (n + 1)
    for idx, car in enumerate(b):
        exit_pos[car] = idx
    max_exit = -1
    count = 0
    for car in a:
        current_exit = exit_pos[car]
        if current_exit < max_exit:
            count += 1
        else:
            max_exit = current_exit
    print(count)

if __name__ == "__main__":
    main()
\end{verbatim}
\endgroup

\paragraph{What changed.}
A car is fined iff, scanning cars in entry order, its exit index is below the maximum exit index seen so far. The human solution is a correct two-pointer implementation with a \texttt{set} of already-fined cars, but it is the fiddly version of the problem. Optimization-RL writes the canonical one-pass running-maximum solution with an array of exit positions. This is not an algorithmic leap in the asymptotic sense---both are \(O(n)\)---but it is the cleaner implementation with less hashing and less pointer bookkeeping.

\paragraph{Codeforces 534B, ``Covered Path''.}
\textbf{Problem.} Choose a length-\(t\) speed sequence starting at \(v_1\) and ending at \(v_2\), where adjacent speeds differ by at most \(d\), to maximize total distance. Here \(t\leq 100\) and \(d\leq 10\). \textbf{Measured:} best human \(13.1\,s\rightarrow\) optimization-RL \(12.7\,s\) (\(\times1.04\)); judge label: mathematical shortcut / algorithmic.

\paragraph{Best human solution.}
\begingroup
\scriptsize
\begin{verbatim}
inp = input()
inp = inp.split()
v1 = int(inp[0])
v2 = int(inp[1])
inp = input()
inp = inp.split()
t = int(inp[0])
d = int(inp[1])
ans = v1
t -= 1
while(t > 0):
    for i in range(-d,d+1):
        v = v1 - i
        if d * (t - 1) >= abs(v2 - v):
            ans += v
            v1 = v
            t -= 1
            break
print(ans)
\end{verbatim}
\endgroup

\paragraph{Optimization-RL solution.}
\begingroup
\scriptsize
\begin{verbatim}
import sys

def main():
    v1, v2 = map(int, sys.stdin.readline().split())
    t, d = map(int, sys.stdin.readline().split())
    total = 0
    for i in range(1, t+1):
        a = v1 + (i-1) * d
        b = v2 + (t - i) * d
        total += min(a, b)
    print(total)

if __name__ == "__main__":
    main()
\end{verbatim}
\endgroup

\paragraph{What changed.}
At time \(i\), the speed is upper-bounded both by ramping up from the start, \(v_1+(i-1)d\), and by still being able to ramp down to the end, \(v_2+(t-i)d\). Summing the minimum of those two bounds gives the optimal distance directly in \(O(t)\). The human solution is a correct greedy simulation with a per-step scan over feasible speed changes, so it is \(O(td)\) and still passes easily under the original constraints. Optimization-RL writes the intended two-ramp observation directly; this is a cleaner solution more than a surprising one.

\clearpage
\begin{table}[h!]
\refstepcounter{table}
\label{tab:asset_licenses}
\normalsize
\noindent\textbf{Table~\thetable: Licenses of data/models used in this work.}
External datasets, benchmarks, and model checkpoints used in this work, together with their role in the paper and their license or usage terms.
\par\vspace{0.5em}

\centering
\small
\setlength{\tabcolsep}{5pt}
\renewcommand{\arraystretch}{1.3}
\begin{tabular}{@{}p{0.37\textwidth} p{0.39\textwidth} p{0.19\textwidth}@{}}
\toprule
{\bfseries Asset} & {\bfseries Use in paper} & {\bfseries License / terms} \\
\midrule
DeepMind Code Contests (CodeContests) / Codeforces human references & Source corpus for DMC-Optim and human-speed reference solutions & Apache 2.0 \href{https://www.apache.org/licenses/LICENSE-2.0}{[link]} \\
LiveCodeBench (LCB, including LCB I/O and LCB native) & Out-of-distribution evaluation benchmark & MIT \href{https://opensource.org/licenses/MIT}{[link]} \\
OpenCodeReasoning-2 & Part of the Qwen 2.5 reasoning-only SFT mix & CC BY 4.0 \href{https://creativecommons.org/licenses/by/4.0/}{[link]} \\
OpenMathReasoning & Part of the Qwen 2.5 reasoning-only SFT mix & CC BY 4.0 \href{https://creativecommons.org/licenses/by/4.0/}{[link]} \\
Qwen 2.5 7B / 32B base checkpoints & Base models for the Qwen optimization-RL runs & Qwen License \href{https://huggingface.co/Qwen/Qwen2.5-72B/blob/main/LICENSE}{[link]} \\
CWM 32B & Released SFT checkpoint used for optimization RL & FAIR Noncommercial Research License \href{https://huggingface.co/facebook/cwm-sft}{[link]} \\
GPT-OSS 120B & LLM-as-judge model for the code-pair analysis in \cref{sec:analysis} & Apache 2.0 \href{https://www.apache.org/licenses/LICENSE-2.0}{[link]} \\
BigO(Bench) & External benchmark used for decontamination and motivation & CC BY-NC 4.0 \href{https://creativecommons.org/licenses/by-nc/4.0/}{[link]} \\
\bottomrule
\end{tabular}
\end{table}

\end{document}